\documentclass[12pt,letterpaper]{report}          % Single-sided printing for the library
\usepackage[intlimits]{amsmath}
\usepackage{amsfonts,amssymb}
\DeclareSymbolFontAlphabet{\mathbb}{AMSb}
\usepackage{apalike}
\usepackage{float}
\usepackage[bf]{caption}       
\setcaptionmargin{0.5in}
\usepackage{fancyhdr}
\usepackage{fancybox}
\usepackage{ifthen}
\usepackage{bu_ece_thesis}
\usepackage{lscape,afterpage}
\usepackage{xspace}
\usepackage{epstopdf} 
\usepackage{graphicx}
\usepackage{appendix}

\usepackage[linesnumbered,ruled,vlined]{algorithm2e}
\usepackage{algpseudocode}
\usepackage{listings}
\usepackage{subcaption}
\usepackage{tikz}
\usepackage{colortbl}
\usepackage{multirow}
\usepackage{longtable}

\PassOptionsToPackage{hyphens}{url}
\usepackage[hidelinks]{hyperref}
\hypersetup{breaklinks=true}
\usepackage{url}

\usepackage{array}
\usepackage{enumitem}
\usepackage{ragged2e}

\usepackage{todonotes}
\newcommand\eg{\emph{e.g.,}\xspace}
\newcommand\ie{\emph{i.e.,}\xspace}

\newcommand\gymfcGeneric{GymFC\xspace}
\newcommand\gym{GymFCv1\xspace}
\newcommand\gymfc{GymFCv1\xspace}
\newcommand\newgym{GymFCv1.5\xspace}
\newcommand\fc{Neuroflight\xspace}
\newcommand\aircraft{NF1\xspace}
\newcommand\nn{NN\xspace}
\newcommand\nns{NNs\xspace}
\newcommand\nf{Neuroflight\xspace}

\newcommand\gymfcOne{GymFCv1\xspace}
\newcommand\gymfcTwo{GymFCv2\xspace}
\newcommand\uav{uncrewed aerial vehicle\xspace}

\newcommand{\new}[1]{{#1}}	  % don't highlight

\newcommand*{\circled}[2][]{\tikz[baseline=(C.base)]{
    \node[inner sep=0pt] (C) {\vphantom{1g}#2};
	\node[draw, circle, inner sep=2pt, yshift=1pt] at (C.center) 
		{\vphantom{1g}};}}

\def\addauthnote#1#2{%
    \expandafter\def\csname#1\endcsname##1{\todo[inline,color=#2]{#1: 
        ##1}\xspace}
    \expandafter\def\csname#1m\endcsname##1{\todo[color=#2]{#1: ##1}\xspace} 
}

\addauthnote{wil}{green!25} 

\begin{document}

\title{Flight Controller Synthesis via Deep Reinforcement Learning}

\author{William Frederick Koch III}

\degree=2

\prevdegrees{B.S., University of Rhode Island, 2008\\
	M.S., Stevens Institute of Technology, 2013}

\department{Department of Computer Science}

\defenseyear{2019}
\degreeyear{2019}

\reader{First}{Azer Bestavros, PhD}{Professor of Computer Science}
\reader{Second}{Renato Mancuso, PhD}{Assistant Professor of Computer Science}
\reader{Third}{Richard West, PhD}{Professor of Computer Science}

\numadvisors=1
\majorprof{Azer Bestavros, PhD}{{Professor of Computer Science}}
\maketitle
\cleardoublepage

\copyrightpage
\cleardoublepage

\approvalpage
\cleardoublepage

\newpage
\phantom{.}
\vspace{4in}

\begin{singlespace}
\begin{quote}
Just flow with the chaos...
\end{quote}
\end{singlespace}

\cleardoublepage

\newpage
\section*{\centerline{Acknowledgments}}
What an adventure this has been. The past five years have been some of the best years
of my life. I have been fortunate enough to have the opportunities to  
work on projects and research that are dear to me, 
form life long relationships and travel around the world.  
Its hard to imagine going through my PhD  without the 
love and  support of my family, friends, and colleagues who I would like to
thank.% have had on my life. 

I would like to start off by thanking members of my committee Azer Bestavros,
Rich West and Renato Mancuso. 
Azer, you have been
there for me since the beginning. Your wisdom and guidance has
helped shape my perspective on the world and how to step back and see the
bigger picture. I appreciate your support over the years and the partnerships and relationships you
have helped me form. In the context of research we have been on quite a roller coaster ride,  from cyber
security to flight control. 
Rich, thank you for always making me feel welcome in your lab. I will always cherish our
conversations and shared interests in racing. Your energy has helped me
pursue an area of research that
was intimidating and unknown. 
Renato, you could not have joined BU at any more perfect time. This research
would not have been possible without your support and involvement. Your
expertise in the field of real-time systems and flight control has provided
invaluable insight. Working together has been a pleasure and will not be forgotten.   
Additionally I would like to thank Manuel Egele who I worked with for years conducting
research in cyber security before pursing my current research area in flight control
systems. I have learned a great deal from you and you have helped shaped me to
become a better
researcher. 

My current research all began with drone racing. I would like to thank my friends and
classmates Ethan
Heilman, William Blair and Craig Einstein for the countless flying sessions and
races over the years, especially Ethan for first introducing the rest of us to the
hobby. These gatherings are what eventually led to the formation of Boston
Drone Racing~(BDR), and it has been incredible to
see where it has evolved to today. With that I would like all the members of
BDR, it truly has been a blast and 
 it is amazing to see everyone's  progression. On behalf of Boston Drone
Racing we are grateful to the BU CS department staff who have always helped and supported us
and Renato Mancuso for allowing us to store racing equipment in the lab.

Additionally I would like to thank my other classmates and friends Aanchal Malhotra, Thomas Unger,
Nikolaj Volgushev and Sophia Yakoubov. No matter what we faced during our time
at BU, we were going through it together. Our awesome times living in Allston will never
be forgotten.  Although we are now  scattered
across the globe, the relationships we forged will always remain close.   
I would like to thank my friends Zack, Melissa, Dave, Kat, Matt, Sydney, Drew and the URI crew 
for their support over these years. You have always been
there for me, we have experienced countless adventures, you are family.

Dad, thank you for your support over the years. I will treasure our
conversations we had
throughout my research about aeronautics. Flight definitely runs through our blood.
Mom, you have had unconditional love for me my entire life.  
Thank you for the scarifies
you have made for me over the years, and the opportunities you have given me. 
To my brothers Cole,
Spence and Carter, I am so proud of you all, always follow you dreams and passions in
life. I will always be there for you. Randy and Ellen, I cannot begin to
thank you for your generosity, kindness and hospitality over the years. Mark,
Alissa, Shannon, Nick, my nieces and nephew, I am so fortunate to have you in my
life.

To my wife Kristen, thank you for your kindness, encouragement, patience and love. You are my
soul mate, best friend and rock in my life. You have helped me maintain a balance in life through this 
chaotic journey. No matter what is happening in life, you and Liam  make me
smile. I love the two of you with all of my heart.

\cleardoublepage

\begin{abstractpage}
Traditional control methods are inadequate in many deployment settings involving autonomous control of Cyber-Physical Systems (CPS). In such settings, CPS controllers must operate and respond to unpredictable interactions, conditions, or failure modes. Dealing with such unpredictability requires the use of executive and cognitive control functions that allow for planning and reasoning. Motivated by the sport of drone racing, this dissertation addresses these concerns for state-of-the-art flight control by investigating the use of deep artificial neural networks to bring essential elements of higher-level cognition to bear on the design, implementation, deployment, and evaluation of low level (attitude) flight controllers.

First, this thesis presents a feasibility analyses and results which confirm
that neural networks, trained via reinforcement learning, are more accurate
than traditional control methods used by commercial uncrewed aerial vehicles
(UAVs) for attitude control. Second, armed with these results, this thesis
reports on the development and release of an open source, full solution stack
for building neuro-flight controllers. This stack consists of a tuning
framework for implementing training environments (GymFC) and firmware for the
world's first neural network supported flight controller (Neuroflight). GymFC's novel approach fuses together the digital twinning paradigm with flight control training to provide seamless transfer to hardware. Third, to transfer models synthesized by GymFC to hardware, this thesis reports on the toolchain that has been released for compiling neural networks into Neuroflight, which can be flashed to off-the-shelf microcontrollers. This toolchain includes detailed procedures for constructing a multicopter digital twin to allow the research and development community to synthesize flight controllers unique to their own aircraft. Finally, this thesis examines alternative reward system functions as well as changes to the software environment to bridge the gap between simulation and real world deployment environments.

The design, evaluation, and experimental work summarized in this thesis
demonstrates that deep reinforcement learning is able to be leveraged for the
design and implementation of neural network controllers capable not only of
maintaining stable flight, but also precision aerobatic maneuvers in real world settings. As such, this work provides a foundation for developing the next generation of flight control systems.

\end{abstractpage}
\cleardoublepage

\tableofcontents
\cleardoublepage

\newpage
\listoftables
\cleardoublepage

\newpage
\listoffigures
\cleardoublepage

\chapter*{List of Abbreviations}
\begin{center}
  \begin{tabular}{lll}
    \hspace*{2em} & \hspace*{1in} & \hspace*{4.5in} \\
    API  & \dotfill & application programming interface \\
    DDPG & \dotfill & Deep Deterministic Policy Gradient \\
    DOF & \dotfill & degrees of freedom \\
    ESC & \dotfill & electronic speed controller \\
	FC  & \dotfill & flight controller \\
	FPV & \dotfill & first person view \\
    IMU & \dotfill & inertial measurement unit \\
    HITL & \dotfill & hardware in the loop \\
    NF  & \dotfill & Neuroflight \\
    NN  & \dotfill & neural network \\
    PPO & \dotfill & Proximal Policy Optimization \\
    PWM & \dotfill & pulse width modulation \\
    RL  & \dotfill & reinforcement learning \\
	RX  & \dotfill & receiver \\
	SITL & \dotfill & software in the loop \\
    TRPO & \dotfill & Trust Region Policy Optimization \\
    UAV & \dotfill & \uav \\
	VTX & \dotfill & video transmitter 
  \end{tabular}
\end{center}
\cleardoublepage

\chapter*{List of Symbols}
\begin{center}
  \begin{longtable}{lll}
    \hspace*{2em} & \hspace*{1in} & \hspace*{4.5in} \\
    $a$        && agent action \\
    $b$         && number of propeller blades \\
    $B$         && thrust factor \\
    $C_T,C_Q$           && thrust and torque coefficient \\
    $D$              && degrees of freedom\\
    $e$             && angular velocity error \\
    $e_\phi, e_\theta, e_\psi$             && angular velocity error elements \\
    $F$         && force \\
    $F_\text{min},F_\text{max}$  && min and max change in rotor force \\
    $H$             && rotor velocity transfer function \\
    $J$         && advance ratio \\
    $K_T,K_Q$       && thrust  and torque constant \\
    $K_P, K_I, K_D$       && PID gains \\
    $K_v$       && motor constant \\
    $l$     && multicopter arm length \\
    $M$                 && aircraft actuator count\\
    $r$                 && reinforcement learning reward\\
    $S$                 && aircraft state\\
    $t$          && time in seconds \\
    $T$             && thrust \\
    $\mathbf{T}$         && desired throttle \\
$\widehat{\mathbf{T}}$ && actual throttle \\
    $u$                 && control signal\\
$U_T, U_\phi, U_\theta, U_\psi$ && aerodynamic affect for thrust, roll, pitch
and yaw \\
    $x$         && neural network input \\
    $y$         && neural network output \\
    $\Omega$            && angular velocity \\
    $\Omega_\phi,\Omega_\theta,\Omega_\psi$       && angular velocity axis elements 
    \\
    $\Omega^*$          && desired angular velocity \\
    $\eta_{(\text{ax}, \mu)}$ && mean gyro noise for axis ax\\
    $\eta_{(\text{ax}, \sigma)}$ && variance of gyro noise for axis ax\\
    $\phi,\theta,\psi$              && roll, pitch and yaw axis\\
    $\tau$          && torque \\
    $\rho$     && air mass density \\
    $\omega$            && angular velocity array for each rotor\\ 
    $\omega_i$            && angular velocity of rotor $i$\\ 
    $\pi$             && policy \\
$\gamma$     && PPO discount \\
$\lambda$  && GAE parameter \\
$\delta$    && simulation stability metric \\
  \end{longtable}
\end{center}

\cleardoublepage

\newpage
\endofprelim
        
\cleardoublepage

\setlength{\tabcolsep}{0.2em}
\renewcommand{\arraystretch}{1.15}

\graphicspath{ {1_Intro/figures/} }
\chapter{Introduction}
\label{chapter:Introduction}
\thispagestyle{myheadings}

Recent advances in science and engineering, coupled with affordable processors
and sensors, has led to an explosive growth in Cyber-Physical Systems~(CPS).
Software components in a CPS are tightly intertwined with their physical operating environment. 
This software reacts to changes in its environment in order to control  physical elements in the real world.
Typically a CPS incorporates a control algorithm to reach a desired state, % in the real world,
for example to control the movement of a robotic arm, navigate an autonomous
automobile or to stabilize an \uav~(UAV) during flight.

A CPS's environment is inherently complex and dynamic, from the  degradation of the physical elements over the life time of the system, to its operating environment~(weather, external disturbances, electrical noise, etc.). 
To achieve optimal control in these environments, that is to derive a control law that has been optimized for a particular objective function, one requires sophisticated control strategies. 
Although control theory has a rich history dating back to the 19th century~\cite{maxwell1868governors}, traditional control methods have their limitations.  
Primarily they lack executive functions and cognitive control that allow for memory, learning and planning. 
Such functionality in a controller is fundamental for the safety,
reliability and performance of next generation CPS's
that will be closely integrated into our lives. 
For example, these controllers %for autonomous CPS's, especially in the case for safety critical applications, 
must have the intellectual capacity to instantaneously react to
catastrophes as well as being able to predict and mitigate future failures.

Over the last decade artificial neural network~(\nn) based controllers~(neuro-controllers), for use in a CPS, have become practical for continuous control tasks in the real world.
A \nn  is a mathematical model mimicking a biological brain capable of approximating any continuous function~\cite{cybenko1989approximation}.
Unlike traditional control methods, they provide the essential components for achieving high order cognitive functionality. 
Each neuron (node) connection of the \nn is associated with a numerical weight that emulates the strength of the neuron. 
To achieve the desired performance, these weights are tuned through a process called training.% to each connection to provide the desired control. 

Part of the success of \nn based controllers for continuous tasks can be attributed to exponential progress in the field of deep reinforcement learning~(RL). 
Deep RL is a machine learning paradigm for training deep \nns. The term deep
refers to the width of the \nn's architecture. As control problems increase in
complexity typically the width must also increase.
RL allows the \nn to interact with their operating environment (typically done in a simulation) to iteratively learn a  task. The \nn (commonly referred to as the agent) receives a numerical reward indicating how well they performed the task. 
Reward engineering is 
		the process of designing a reward system in order 
        reinforce the desired behavior of the agent~\cite{dewey2014reinforcement}.
The RL training algorithm's objective is to maximize these rewards over time.
Once the \nn has been trained, it can be transferred to execute on hardware in the real world.
This has become practical in recent years due to 
advancements in size, weight, power and cost (SWaP-C) optimized electronics.

\section{Challenges Synthesizing Neuro-controllers}
Although neuro-controllers trained in simulation via RL have  enormous
potential for the future CPS, there are still a number of challenges that must
be addressed. Particularly, how do we reach a desired level of performance
during training in simulation and successfully transfer the trained model into
hardware to achieve similar performance in the real world. 

\textbf{Performance.}
A controller is designed with a specific number of performance goals in mind depending on the application. The primary goal is to accurately control the physical system within some predefined level of tolerance that is usually governed by the underlying system.
For a robotic arm this may refer to the precision of the movements, or for a UAV attitude controller how well the angular velocity is able to be controlled.

However there are typically other sub-goals the controller should be optimized
for such as reducing energy consumption, and minimizing control output oscillations. 
Because of a \nns black box nature, which can consist of thousands if not
millions of connections, achieving the desired level of performance is not as straight forward as developing a transfer function for a traditional control system for which the step response characteristics can be calculated. 
A number of factors affect the controllers performance such as the \nn
architecture, RL training algorithm, hyperparameters, and the reward function. 

The reward function is specific to the CPS control task, and the desired
performance goals. 
		The rewards must encode the desired performance we wish the agent to
        obtain. To reach a desired level of control accuracy the reward system must include a representation of the error, that is the difference between the current state and the desired state.   However as the performance goals increase in complexity, it becomes increasingly more difficult to balance these goals to obtain the desired level of performance. 

\textbf{Transferability.} The ultimate goal is to be able to synthesize a
neuro-controller in simulation and transfer it seamlessly to hardware to be
used in the real world. Although in simulation we may be able to achieve a
desired level of performance, it is difficult to obtain the same level of
performance in the real world. This is due to the difference between the two
environments commonly referred to as the \textit{reality gap}. In simulation the fidelity of the environment and the CPS model both have an impact on the transferability.
 The world is a complex place, increasing simulation fidelity and modelling all of the dynamics in simulation is challenging and computationally expensive. Thus prioritizing modelling parameters and deriving strategies to aid in the transferability is required.
It is critical to address the reality gap  in order to  provide
seamless transfer of the controller from simulation to hardware while still
gaining the desired level of performance.

\begin{figure}
	\centering
	{\includegraphics[width=0.9\textwidth]{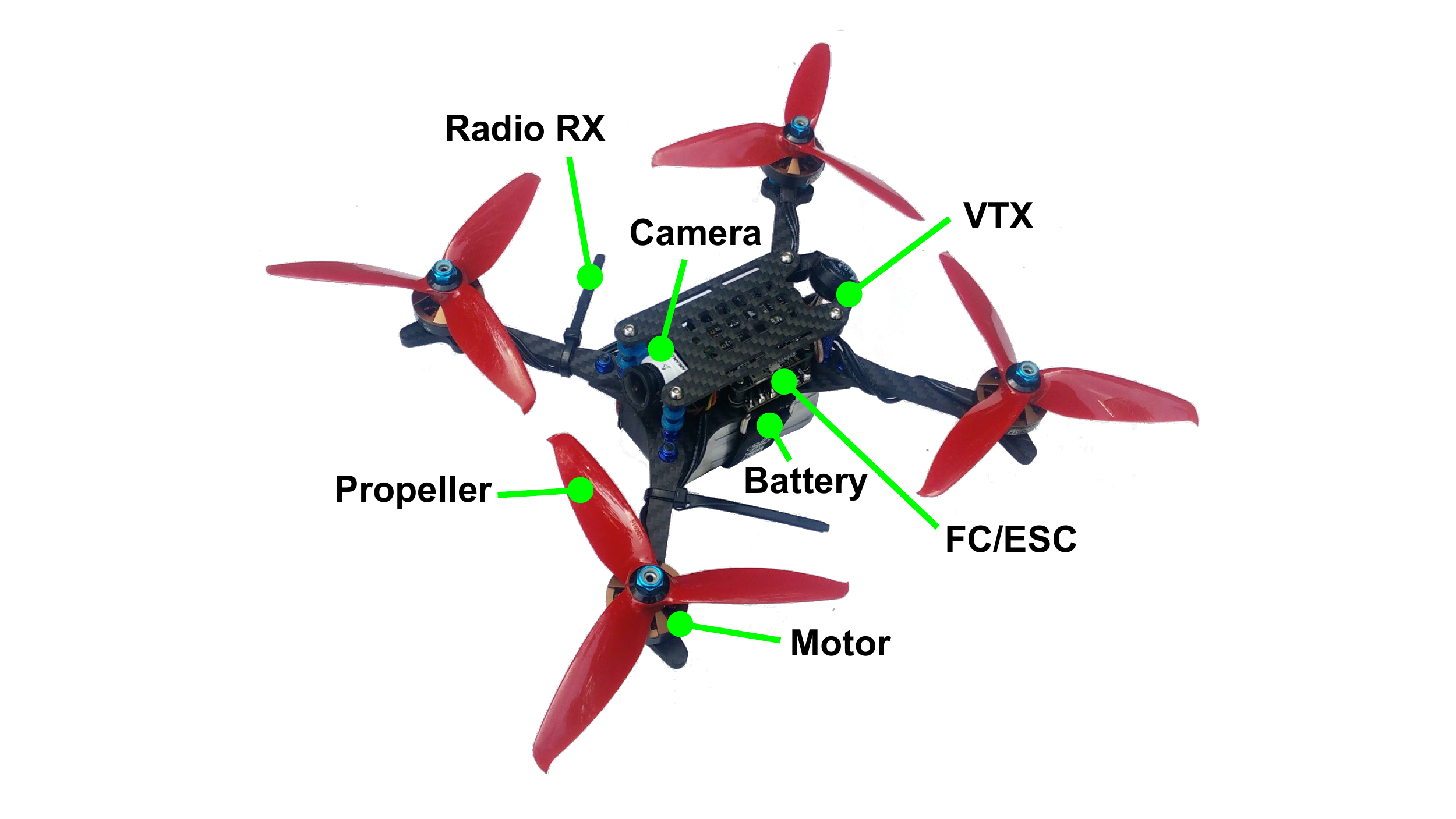}}
	\caption{FPV racing drone.}
	\label{fig:fpv}
\end{figure}

\section{Scope and Contributions}
Motivation for this work has been driven by  drone
racing. The sport of drone racing demands 
the highest level of flight performance to maintain a competitive edge. In drone
racing, a UAV is remotely piloted by first-person-view~(FPV). FPV
provides an immersed flying experience  allowing the UAV to be piloted from the perspective as if you were onboard the aircraft. 
This is accomplished by transmitting the video feed of an onboard camera 
to goggles with an embedded monitor worn by the pilot.
The pilot manually controls the angular velocity (attitude) of the aircraft and mixes in throttle to achieve translational movements. 
A typical FPV equipped racing drone is pictured in Fig.~\ref{fig:fpv}.
A racing drone is an interesting CPS for studying control as they are capable of high speeds and aggressive maneuvers.
Furthermore the controller is exposed to a number of nonlinear dynamics.

\begin{figure}
	\centering
	{\includegraphics[trim=50 40 50
      40,clip,width=\textwidth]{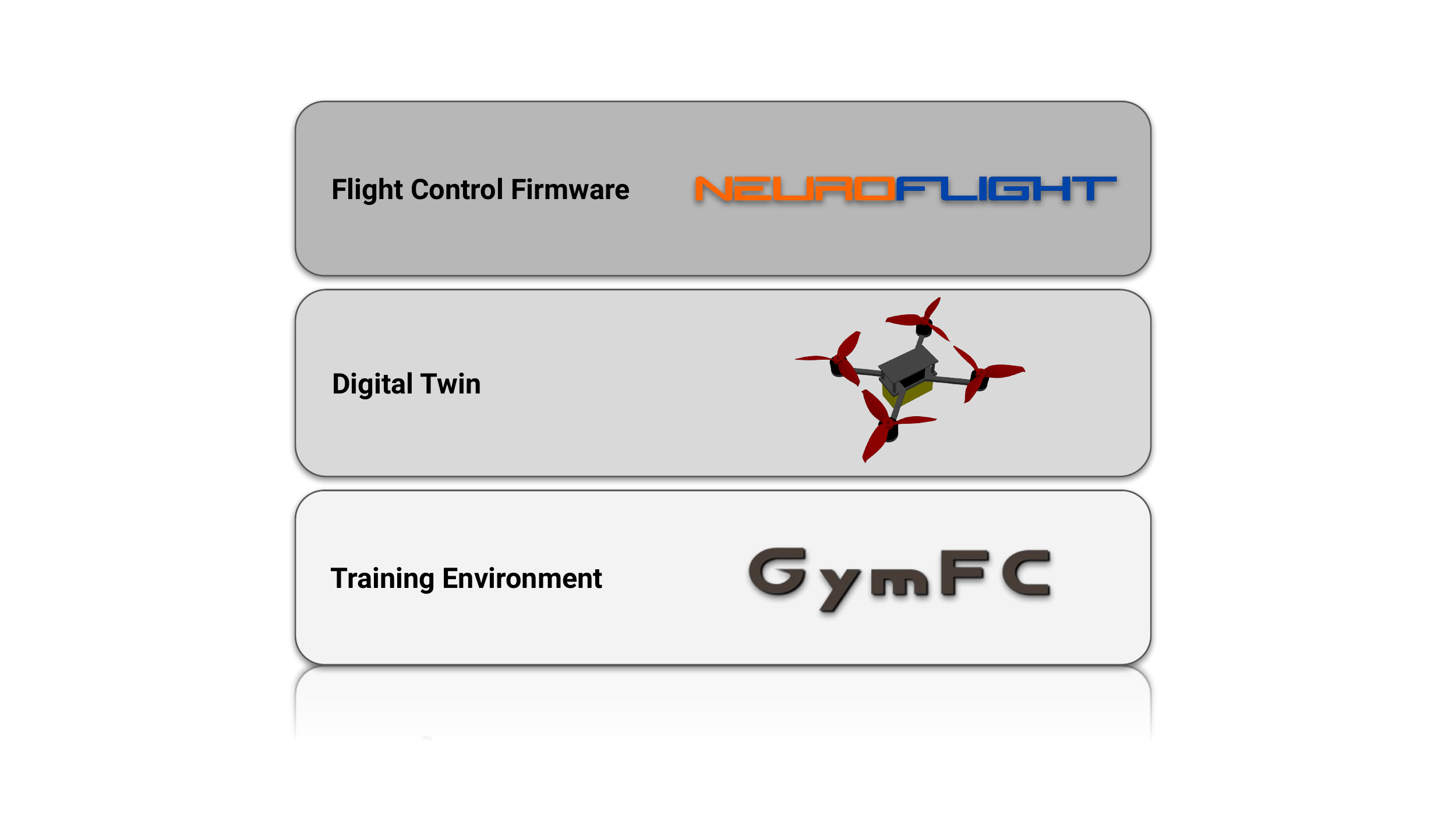}}
	\caption{Neuro-flight controller solution stack.}
	\label{fig:stack}
\end{figure}

Using a racing drone as our experimental platform we study the aforementioned
challenges for synthesizing neuro-controllers. In response to the study, the main contribution of this dissertation
is a full solution stack depicted in Fig.~\ref{fig:stack} for 
synthesizing neuro-flight controllers.
This stack includes  a simulation training environment,
digital twin modelling methodology, and  flight control firmware. 

Throughout this dissertation we synthesize neuro-controllers for the quadcopter aircraft,  
however the training methods described in this work are generic to
most space and aircraft.
Specifically our contributions are in training low level attitude controllers.
Previous work~\cite{kim2004autonomous,abbeel2007application,hwangbo2017control,dos2012experimental,palossi201964mw} has
focused on high level navigation and guidance tasks, while it has remained
unknown how well these type of controllers perform for low level control. 

This dissertation is scoped to synthesizing neuro-controllers
offline in simulation. This is a precursor for practical deployment %this is a necessary 
 as the controller must have initial knowledge of how to achieve
stable flight. 
We provide an initial study of these type of controllers and
publish open source software and frameworks for researchers to progress their
performance. % of these type of controllers.
For neuro-controllers to be adopted in the future
we believe  a hybrid solutions that
incorporates online learning methods 
to compensate for unmodelled dynamics in the simulation environment will be
required.
However as the saying
goes, one must learn to walk before one can run.

Given the capacity and potential of \nns, we believe 
they are the
future for developing high performance, reliable flight control systems. Our contributions and impact are predominately in the development and
release of open source software allowing others to build off of our work to advance the progression in intelligent flight controller design.
We will now briefly summarize the contributions of each item in the solution stack.

\subsection{Tuning Framework and Training Environment}
Most control algorithms  are associated with a set of adjustable parameters that must be tuned for their specific application. 
Tuning a flight controller in the real world is a time consuming task and few systematic approaches are openly available. 
Simulated environments, on the other hand, are an attractive option for developing automated systematic methods for tuning. 
They are cost effective, run faster than real time, and easily allow software to automate tasks.

The benefits of a simulated environment for tuning flight controllers is 
not unique to RL-based controllers, but also applies to traditional controllers as well.
In the context of neuro-controllers, training is just the process of tuning the \nns weights.
In summary this dissertation makes the following contributions in controller
tuning and RL training environments.

\textbf{GymFC:} The first item in our solution stack is an open source tuning framework for synthesizing neuro-flight controllers in simulation called \gymfcGeneric.
\gymfcGeneric was originally developed as an RL training environment for synthesizing
attitude flight controllers. The initial environment architecture is introduced
in Chapter~\ref{chapter:gymfc} and has been published in \cite{gymfc}.
Since the projects release \gymfcGeneric has matured into a generic universal tuning framework
based on feedback received from the community. 
Revisions to \gymfc, discussed in Chapter~\ref{chapter:twin}, increase user
flexibility providing a framework to provide 
custom 
reward systems and aircraft models. 
Additionally \gymfcGeneric is no longer tied to an
RL environment but now opens up the possibilities for other optimization
algorithms to tune traditional controllers.
In Chapter~\ref{chapter:twin} we demonstrate the modular design of the
framework by implementing
a dynamometer for validating motor models in simulation, and a PID controller
tuning system.  
 %Various optimization algorithms such as genetic algoriths \cite{fatan2013adaptive} and partical swarm optimization have been  
%
Our goal with \gymfcGeneric is to provide  the research community a standardized way for tuning flight controllers in simulation. The source code is available at \cite{gymfccode}.  

\textbf{Flight control reward system:} In the context of RL-based flight controllers
the training environment must provide the agent with a reward they are doing the right thing. 
This dissertation shows the progression of our reward system development 
to synthesize accurate controllers and address challenges transferring
controllers to the real world. 
In Chapter~\ref{chapter:gymfc} we introduce rewards to minimize
error which has also been published in \cite{gymfc}.
From experimentation we find in Chapter~\ref{chapter:nf} that additional
rewards are necessary in order to transfer the trained policy into hardware which also appear in \cite{koch2019neuroflight}.
As the accuracy of our aircraft model continued to increase we fine tuned the
reward system in Chapter~\ref{chapter:twin} to decrease error.

\textbf{RL evaluation:} The field of RL is progressing rapidly and
 a number of algorithms have been proposed for continuous control tasks.  
The RL algorithm can be thought of as the \nn tuner. It determines how
the \nn weights are updated depending on the agents current, and past interactions with the environment and rewards
received. This dissertation does not introduce new RL algorithms but
instead uses off-the-shelf implementations for the purpose of synthesizing flight
controllers. Specifically this dissertation makes its contribution in the  
performance evaluation of several  state-of-the-art RL algorithms, including Deep 
	  Deterministic Policy Gradient~(DDPG)~\cite{lillicrap2015continuous}, Trust Region Policy 
	  Optimization~(TRPO)~\cite{schulman2015trust},
	  and Proximal Policy Optimization~(PPO)~\cite{schulman2017proximal}.
    These results were first  published in  \cite{gymfc}.

\subsection{Digital Twin Development } 
Every aircraft is unique in its own way. Off the assembly line, accumulation
of tolerances of each individual part from the manufacturing process results in a
slightly different aircraft. In some cases performance between the same parts, such as
sensors, can
vary greatly \cite{miglino1995evolving}. Once an aircraft is put into service,
they continue to diverge from their initial state as they age. 

To  maximize performance, a controller would ideally be synthesized
uniquely for each individual aircraft, at least in the scope of offline
training strategies. To synthesize this controller in simulation, what we need
is a digital replica, or \textit{digital twin} of the aircraft. A digital twin
is a relatively new paradigm, generic to digitizing any CPS which resides in an ultra high fidelity
simulator.  Once the
CPS is put into service, it is kept in synchronization with its digital twin 
through the collection of state information from its senors. Typical use cases
for the digital twin are for analytics, design and forecasting 
failures. 

This work is the first to fuse together digital twinning concepts for 
neuro-flight controller training. In contrast, previous work has primarily
used a
mathematical model of the UAV
\cite{hwangbo2017control,waslander2005multi,kim2004autonomous,abbeel2007application}
rather than a physics simulator.
In summary we make the following contributions in digital twinning.

\textbf{Multicopter Digital Twin Development Processes:}
Most flight control research performed in simulation use prebuilt aircraft models from 
Gazebo \cite{koenig2004design} or PX4 \cite{meier2015px4} as they are readily
available. In Chapter~\ref{chapter:gymfc} for our initial feasibility analysis,
we also took this approach using the Iris quadcopter~\cite{iris} model provided
by Gazebo. We improved the motor models to more accurately reflect the
motors used by our real quadcopter in Chapter~\ref{chapter:nf}.
Lastly in Chapter~\ref{chapter:twin} we provide our methodology for
creating a
digital twin from the ground up and apply these processes to create a digital twin of our custom built
racing quadcopter. 

Our novel dynamometer for identifying parameters of our  propulsion system 
repurposes the avionics to capture the electronic dynamics that would be  experienced 
during flight which cannot otherwise be captured from commercial 
dynamometers. This results in a higher fidelity motor model which encodes
dynamics such as power delivery from the electronic speed controller~(ESC) and control
signal latency.

Our contributions are in the initial construction of the digital twin, we do not
maintain synchronization with the twin after the aircraft is deployed in this
work. Although our development is specific our quadcopter, these processes are
applicable to any multicopter.

\textbf{Propulsion System Models:} The performance capabilities of a multicopters
propulsion system (motor and propeller pair) have a large influence in the
overall performance of the aircraft. This work builds upon the software in the
loop~(SITL) motor models developed by the PX4 firmware project~\cite{px4sitl}.
These models have been ported to \gymfcGeneric and we have introduced additional dynamics to
increase realism such as motor response and throttle curve mapping. 
These models have been made open source available from \cite{gymfcplugins}.

\textbf{Simulation Stability Analysis:}
Multicopters (particular those found in racing) are capable of achieving high angular velocities, which
induce large centripetal forces. Under certain circumstances this can result in
the digital twin becoming  unstable in  
simulation. In this work we discuss the conditions in which instabilities can
occur. We also propose an algorithm for measuring simulation stability and have
included an implementation with \gymfcGeneric \cite{gymfccode}.
Using this software we perform an analysis of our digital twin.

\subsection{Flight Control Firmware} 
Common approaches for deploying  a neuro-controller to a UAV is to use a
companion computer and run the \nn in user space. However this is usually only
suitable for slower than real-time applications that do not have strict
deadlines and the UAV can permit the size, and weight of the additional hardware. Companion computers are typically used for high level control tasks such as navigation and guidance  in flight control systems which need the additional computational resources but have a slower control loop in comparison to the low level stability control. 

To meet control loop timing requirements, UAVs currently use microcontrollers to execute the real-time task of low level flight control. 
However there previously did not exist solutions for deploying neuro-controllers to microcontrollers let alone a flight control firmware that supported neuro-controllers. 
   
To evaluate our neuro-controllers trained in simulation in the real world it was first necessary for us to develop methods for compiling a \nn to run on a microcontroller. 
With these methods established we developed the flight control firmware Neuroflight to support neuro-attitude flight controllers. 
The results from this work first appeared  in \cite{koch2019neuroflight}.
In summary, this dissertation makes the following contributions in the area of
flight control firmware.

\textbf{Neuroflight:} Prior to this work, every open source flight control firmware available used PID control~\cite{ebeid2018survey}.
In this work we have created the world's first open source \nn supported flight
control firmware for UAVs, Neuroflight. The firmware provides the community
with a platform to experiment with their own trained policies and further progress advancements in field of flight control.
The source code is available from \cite{neuroflight}.

\textbf{Toolchain:}  The target hardware for most UAV flight control firmware is significantly resource constrained. 
The off-the-shelf microcontrollers supported by the family of high performance drone racing firmwares only consists of 1MB of flash memory, 320KB
of SRAM and an ARM Cortex-M7
processor with a clock speed of 216MHz~\cite{STM32F745VG}. 
This dissertation proposes a toolchain  to allow \nns to be compiled  to run on
off-the-shelf microcontrollers with hard floating point arithmetic. 
The impact of this toolchain reaches beyond flight control for UAVs and opens up the
possibilities of using neuro-control for other CPS's in resource
constrained environments.

\textbf{Flight Performance Evaluation:}  In the context of low level attitude
control, this work provides the first evaluation of a neuro-controller trained
in simulation and transferred to hardware to fly in the real world. Our timing
analysis reveals the \nn-based attitude control task is able to execute at over
2kHz on an Arm Cortex-M microcontroller. We demonstrate our training environment, and reward functions are capable
of synthesizing controllers with remarkable performance in the real world. Our
real world flight evaluations validate these controllers are capable of stable flight and the execution of
aerobatic maneuvers. 

\section{Structure}
In summary, the remainder of this dissertation is organized as follows. In
Chapter~\ref{chapter:bg} we discuss important background information and
related work pertinent to synthesizing neuro-based flight controllers.
In Chapter~\ref{chapter:gymfc} we present our flight control training
environment \gymfcGeneric and provide a feasibility analysis on whether neuro-flight controllers can accurately provide attitude control in simulation. 
To identify if the synthesized controllers can achieve stable flight in the
real world we present our firmware, \fc and its accompanying toolchain in
Chapter~\ref{chapter:nf}.  
We propose our digital twin development methodology in
Chapter~\ref{chapter:twin} and introduce our revisions to \gymfcGeneric to support
training of arbitrary aircraft models. Finally in
Chapter~\ref{chapter:conclusion} we conclude with our final remarks and future
work.

\cleardoublepage

\graphicspath{ {1_Related/figures/} {2_GymFC/figures/} }

\chapter{Background and Related Work}
\label{chapter:bg}
\thispagestyle{myheadings}

In this chapter we discuss background concepts and related work. We begin in
Section~\ref{sec:bg:history} with
the history and evolution of flight control for fixed wing aircraft
leading up to the rise of the quadcopter. In Section~\ref{sec:bg:dynamics} we provide an overview
of quadcopter flight dynamics and review flight control systems found in
commercial UAVs in Section~\ref{sec:bg:pid}. In Section~\ref{sec:bg:academia} we
discuss flight control
research being conducted in academia and the trend towards intelligent control
systems. In Section~\ref{sec:bg:rl} we emphasize the academic research related
to deep reinforcement learning in the
context of flight control. To successfully transfer models from simulation to
hardware a number of strategies have been proposed which we review in
Section~\ref{sec:bg:transfer}. Lastly we provide an overview of digital
twinning in Section~\ref{sec:bg:twin} particularity in the context of flight control. 

\section{History of Flight Control}
\label{sec:bg:history}
Aviation has a rich history in flight control dating back to the 1960s. During 
this time supersonic aircraft were being developed which demanded more 
sophisticated dynamic flight control than what a linear controller could 
provide.  Gain scheduling~\cite{leith2000survey} was developed allowing multiple 
linear controllers of different configurations to be used in designated 
operating regions. This however was inflexible and insufficient for handling the 
nonlinear dynamics at high speeds but paved way for adaptive control.

During the 1950s there was a period 
know as the \textit{brave era} in which various adaptive control techniques were 
tested with little time between conception and implementation.  The lack of 
theoretical analysis and guarantees resulted in  fatalities most notably in the 
X-15 crash~\cite{hovakimyan2011mathcal}.
Eventually this led to the development of Model Reference Adaptive 
Control~(MRAC)~\cite{whitaker1958design} which introduced a reference model 
specifying the desired performance of the controller during adaptation. A 
reference model usually consists of the transient response characteristics such as 
rise time, setting time and steady state error.  However early developments of 
MRAC did not have stability guarantees during adaptation. It was not until later that MRAC used 
the Lyapunov function for stability~\cite{aastrom2013adaptive}.
To improve upon  tuning challenges found in MRAC, $L_1$  was proposed which 
includes a lowpass filter to decouple the rate of adaptation and robustness.  An 
$L_1$ control system was tested in the U.S. Air Force's VISTA 
F-16 aircraft~\cite{f16}. However there has been considerable debate in the 
control community due to two rebuttal papers questioning the true benefits of 
$L_1$ adaptive control~\cite{black2014adaptive}.

There has been a trend towards using artificial intelligence for adaptive
control in fixed wing crewed aircraft to compensate for the nonlinear aircraft
dynamics, and uncertainties. Specifically the use of artificial \nns 
which provide capabilities that are beyond that of traditional control such as
their ability to learn and approximate any function. For an introduction to \nns
with applications to control we refer to \cite{hagan1999neural}. 

Work provided by \cite{kim1993nonlinear}
sought to create a single controller valid throughout the entire flight
envelope to remove the need for gain scheduling. The use of nonlinear
controllers such as feedback linearization are an
attractive option as they are able to transform the nonlinear system into an
equivalent linear representation. Once in a linear representation a
linear controller, such as PID or linear quadratic Gaussian (LQG) can be used. 
However feedback linearization requires a model of the aircraft which can
contain errors. To develop an
aircraft model, the authors utilized a \nn which is first trained offline using
mathematical models, and then fine tuned, online using a second \nn to compensate for any model
errors. Another interesting contribution to this work was the use of the circle
theorem~\cite{zames1966input} as a way to bound the stability of this
controller even in the presence of the \nns.

The Intelligent Flight Control System (IFCS) project lead by NASA was created
to investigate the capabilities of \nns for adaptive control, with a focus in
providing stability during failure~\cite{williams2005flight}. 
Failure in this work is scoped to malfunctioning of the control surfaces. 
The
project's
test aircraft is a highly modified F-15; however this work only reports
simulation results. 
Simulation results  demonstrate the \nn is able to restore the aircraft to a stable
state after the occurrence of failure, in less time and smoother than without
the presence of the \nn.  
Starting in 2006 real flight tests began~\cite{smith2010design}.
During these test flights, two failures were emulated, locking of the left stabilator
and change to the baseline angle of attack of the canard~(a small forward
wing). Overall the test pilots reported improved handling with the \nn enabled
during failure. These results show a promising  future for  these type of
controllers.

As a result of the significant cost reduction for sensors and small-scale embedded computing platforms  over the course of the last couple decades, UAVs, 
particularity quadcopters, have surged
in popularity.   Due to their unique complex dynamics 
quadcopters have their own set of challenges related to flight control.
However we are seeing similar patterns in the progress of flight control
 for UAVs as we have seen for fixed wing crewed 
aircraft. Although this dissertation's focus is in the development of flight controllers
for quadcopters, nonetheless the majority of what is discussed is applicable to
most multicopter configurations and fixed wing aircraft as well.

\section{Quadcopter Flight Dynamics}
\label{sec:bg:dynamics}
Before we can discuss the specifics of flight control pertaining to the
quadcopter aircraft it is necessary to  
understand some basics of their dynamics. 

\begin{figure}
    \centering
    %trim l b r t
    {\includegraphics[trim=35 405 20 100, clip, width=0.75\textwidth]{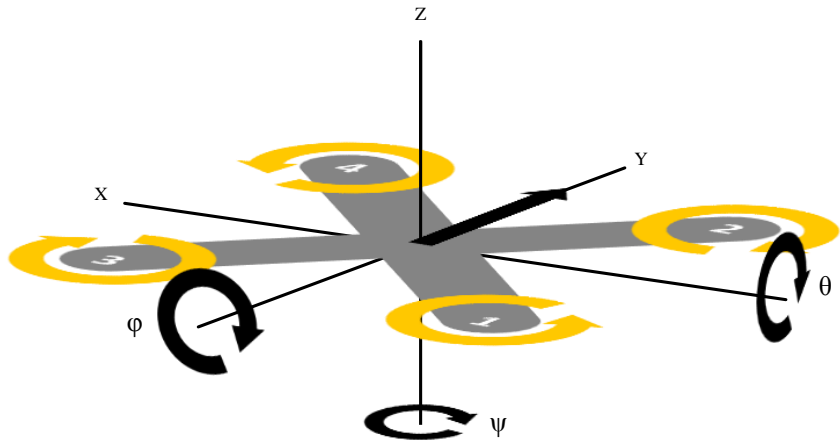}}
    \caption{Axis of rotation}
    \label{fig:axis}
\end{figure}

A quadcopter is an aircraft with four (quad) motors using a propeller
propulsion system. It has six degrees of freedom (DOF), three
rotational and three translational as depicted in Fig.~\ref{fig:axis}. Throughout this dissertation we will use 
the motor ID and order referenced in this figure, starting at index one, to be consistent
with the ordering used
to configure our flight control firmware, while the subscript used in the
mathematical  notation begins with
zero. 
We
indicate with $\omega_i, i \in {0, \dots, M-1}$ the rotation speed of
each rotor where $M=4$ is the total number of motors for a quadcopter. These 
have a direct impact on the resulting Euler angles
$\phi, \theta, \psi$, \ie roll, pitch, yaw respectively and translation in each $x$, $y$, and $z$ direction.  

The aerodynamic effect that each $\omega_i$ produces depends upon the
configuration of the motors. 
The motor configuration (\ie location of each motor) can have a significant affect in flight performance 
depending on the distance the motors are from each axis of rotation.  Intuitively 
the greater the distance the motor is from the axis of rotation the more
torque 
will be required to travel along this arc compared to when a motor is mounted 
closer to the axis.  
In the context of classical mechanics, torque is defined as $\tau = l \times F$ 
where $l$ is the length of the lever and $F$ is the applied force.  Translated to 
a quadcopter, each motor and propeller pair generates a force $F$ at some distance 
$l$  from the axis
of rotation.  

The most popular configuration is
an \textbf{X} configuration, depicted in Fig.~\ref{fig:axis} which has the motors 
mounted in an \textbf{X} formation relative to what is considered the front of the 
aircraft.  This
configuration provides more stability compared to a \textbf{+} configuration which in 
contrast has its motor configuration rotated an additional $45^{\circ}$ along 
the z-axis. This is due to the differences in torque generated along each axis 
of rotation in respect to the distance of the motor from the axis.  
Additionally the \textbf{X} configuration is a more practical arrangement for
mounting cameras used for  navigation.

For a \textbf{+} configuration the distance, in relation to pitch,  
is equivalent to the distance of the arm $l$.  
An \textbf{X} configuration with the same arm length $l$ has a distance from the 
axis $l \times cos(\pi/4)$ resulting in less torque required. A decrease in the 
arm length provides increased responsiveness. Furthermore the motor rotation in a 
\textbf{+} configuration is in the same direction along an axis of rotation 
leading to less stability than an \textbf{X} configuration. 
Based on these dynamics, frames are optimized depending on their application. For
example racing frames are often \textit{stretched} such that the distance
between motors 3 and 4, and motors 1 and 2, are at a greater distance than 
between motors 1 and 3, and motors 2 and 4.
This results in less torque along the roll axis
providing  a more responsive aircraft for performing turns.

The aerodynamic affect $U$ that 
each rotor speed $\omega_i$ has on
thrust and Euler angles, is given by:
\begin{align}
U_T &= B ( \omega_0^2 + \omega_1^2 + \omega_2^2 +
\omega_3^2)\label{eq:qc_dynamics_thrust}\\
U_\phi &= B ( \omega_0^2 + \omega_1^2 - \omega_2^2 - \omega_3^2)\label{eq:qc_dynamics_roll}\\
U_\theta &= B ( \omega_0^2 - \omega_1^2 + \omega_2^2 - \omega_3^2)\label{eq:qc_dynamics_pitch}\\
U_\psi &= B ( \omega_0^2 - \omega_1^2 - \omega_2^2 + \omega_3^2)\label{eq:qc_dynamics_yaw}
\end{align}
where $U_T, U_\phi, U_\theta, U_\psi$ is the thrust, roll, pitch, and
yaw effect respectively, while $B$ is a thrust factor that captures
propeller geometry and the motor configuration. 
The torque $\tau_B$ applied to the aircraft is the torque applied to each axis 
$\phi, \theta, \psi$ for roll, pitch, yaw respectively. The model developed by 
\cite{luukkonen2011modelling,bouabdallah2004design} modified for \textbf{X} 
configuration as,

\begin{equation}
	\tau_B = \begin{bmatrix}
		\tau_\phi \\
		\tau_\theta \\
		\tau_\psi \\
	\end{bmatrix} = \begin{bmatrix}
		l cos(\pi/4) B ( \omega_0^2 + \omega_1^2 - \omega_2^2 - \omega_3^2)\\
		l cos(\pi/4) B (\omega_0^2 - \omega_1^2 + \omega_2^2 - \omega_3^2)\\
		\sum_{i=0}^{M-1} \tau_{M_i}
	\end{bmatrix}
\end{equation}
where  $\tau_{M_i}$ is the 
torque of each motor.

\begin{figure*}
	\centering
	\begin{subfigure}{.23\textwidth}
	%trim l b r t
		\includegraphics[trim=0 650 460 0, clip, width=\textwidth]{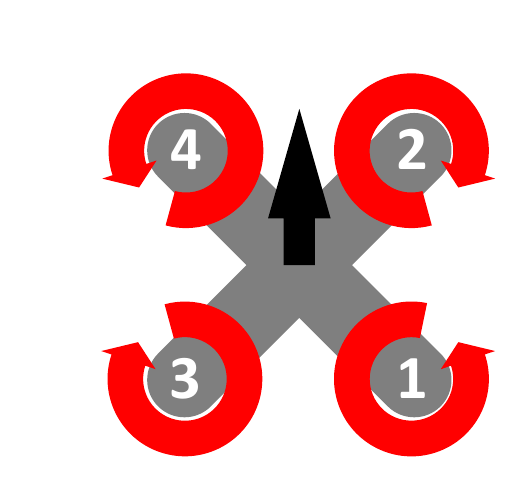}
		\caption{Accel}
        \label{fig:fd:accel}
	\end{subfigure}
~
	\begin{subfigure}{.23\textwidth}
		\includegraphics[trim=0 650 460 0, clip, width=\textwidth]{yawcw}
		\caption{Yaw CW}
        \label{fig:fd:yawcw}
	\end{subfigure}
~
	\begin{subfigure}{.23\textwidth}
		\includegraphics[trim=0 650 460 0, clip, width=\textwidth]{pitchpos}
		\caption{Pitch forward}
        \label{fig:fd:pitchforward}
	\end{subfigure}
~
	\begin{subfigure}{.23\textwidth}
		\includegraphics[trim=0 650 460 0, clip, width=\textwidth]{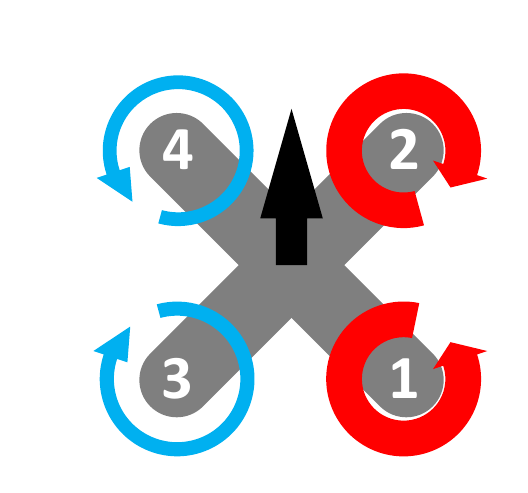}
		\caption{Roll left}
        \label{fig:fd:rollleft}
	\end{subfigure}
\\
	\begin{subfigure}{.23\textwidth}
	%trim l b r t
		\includegraphics[trim=0 650 460 0, clip, width=\textwidth]{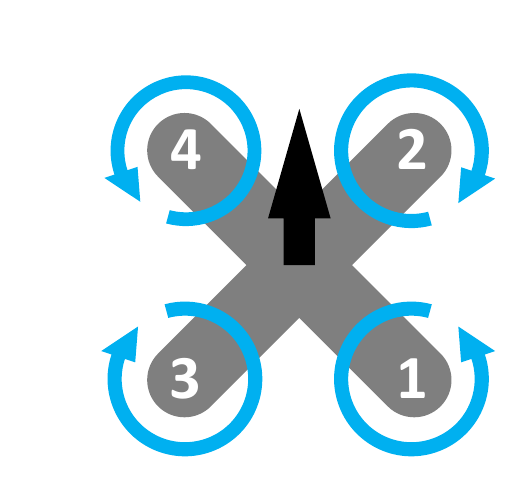}
		\caption{Decel}
        \label{fig:fd:decel}
	\end{subfigure}
~
	\begin{subfigure}{.23\textwidth}
		\includegraphics[trim=0 650 460 0, clip, width=\textwidth]{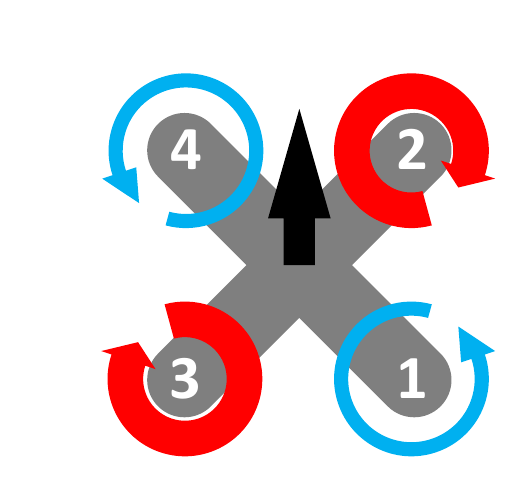}
		\caption{Yaw CCW}
        \label{fig:fd:yawccw}
	\end{subfigure}
~
	\begin{subfigure}{.23\textwidth}
		\includegraphics[trim=0 650 460 0, clip, width=\textwidth]{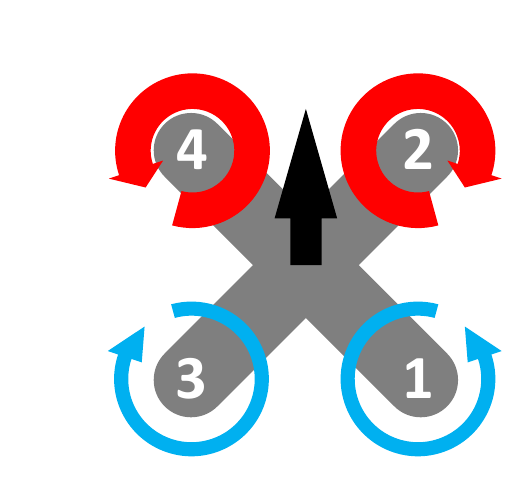}
		\caption{Pitch backward}
        \label{fig:fd:pitchbackward}
	\end{subfigure}
~
	\begin{subfigure}{.23\textwidth}
		\includegraphics[trim=0 650 460 0, clip, width=\textwidth]{rollright}
		\caption{Roll right}
        \label{fig:fd:rollright}
	\end{subfigure}
	\caption{Commands of a quadcopter. Red wide arrows represent faster  angular 
velocity, while blue narrow arrows represent slower angular velocity.  Faster 
and slower velocities are relative to when its net force is zero. }
\label{fig:motion}
\end{figure*}

To perform rotational movement the velocity of each rotor is manipulated 
according to the relationship expressed in Eq.~\ref{eq:qc_dynamics_roll}
Eq.~\ref{eq:qc_dynamics_pitch}, Eq.~\ref{eq:qc_dynamics_yaw} and as illustrated in 
Fig.~\ref{fig:motion}.  For example, to 
roll right (Fig.~\ref{fig:fd:rollright}) more thrust is delivered to motor~3
and~4 (\ie $\omega_2 >\omega_0$ and  $\omega_3 >\omega_1$).
However yaw is not achieved directly through difference in 
thrust generated by the rotor as roll and pitch are, but instead through a
difference in torque generated by the velocity of the rotors.
For example, as shown in Fig.~\ref{fig:fd:yawcw}, higher rotational speed for rotors~1 and~4 
allow the aircraft to yaw clockwise. A net positive torque of the rotors in the
counter-clockwise direction causes the 
aircraft to rotate clockwise in the opposite direction due to Newton's second law of motion. 

Attitude, in respect to the orientation of the aircraft, can be expressed as
the   
angular velocities of each axis $\Omega=[\Omega_\phi, \Omega_\theta, 
\Omega_\psi]$. The objective of attitude control is to compute the required 
motor control signals $u=[u_0, \dots, u_{M-1}]$  to achieve some desired attitude $\Omega^*$.
In autopilot systems attitude control is typically executed as an inner control loop  and 
is time-sensitive.  Once the desired attitude is achieved, translational 
movements (in the X, Y, Z direction) are accomplished by applying thrust 
proportional to each motor.
For further details
about the mathematical models of quadcopter dynamics please refer
to~\cite{bouabdallah2004design}.

\section{Flight Control for Commercial UAVs}
\label{sec:bg:pid}
Of the commercially available flight control systems and open source
flight control firmwares currently available every single one uses a static
linear controller called 
proportional, integral, and derivative~(PID)  control~\cite{ebeid2018survey}.

A PID 
controller is a linear feedback controller expressed mathematically as,
\begin{equation}
	y(t) = K_P e(t) + K_I \int_0^t e(\tau) d \tau + K_D \frac{d e(t)}{dt}
\end{equation}
where $K_P, K_I, K_D$ are configurable constant gains and $y(t)$ is the output. The effect of each term can be thought of as the P term considers the 
current error, the I term considers the history of errors and the D term 
estimates the future error. In the context of attitude control there 
is a PID controller to control  each roll, pitch and yaw axis. The attitude
controller controls the orientation of the aircraft, typically by its angular
velocity. A PID attitude controller results in a total of 9
gains that must be collectively tuned for each aircraft.

Every time a PID attitude controller is evaluated,
the PID for each axis is computed. The output of each of the PIDs 
must be combined together to form the control signal for each motor. 
This process is 
called \textit{mixing}. 
Mixing uses a table consisting of constants 
to compensate for the motor configuration described in
Section~\ref{sec:bg:dynamics}.
The control signal for each motor  
$u_i$ is loosely defined as,
\begin{equation}
u_i=\mathbf{T} + m_{(i,\phi)} y_\phi + m_{(i,\theta)} y_\theta +m_{(i,\psi)} y_\psi 
\label{eq:mix}
\end{equation} where $m_{(i,\phi)},m_{(i,\theta)},m_{(i,\psi)}$ are the mixer 
values for motor $i$ and $\mathbf{T}$ is the throttle. 

To adapt to nonlinear dynamics experienced during flight, 
the firmware of some flight controllers~(\eg Betaflight~\cite{betaflight}) use gain
scheduling. This gain scheduler  adjusts the PID gains for certain operating
regions such as the throttle value and  battery voltage levels.

\section{Flight Control Research in Academia}
\label{sec:bg:academia}
As flight control methods continue to develop for fixed wing crewed aircraft,
accelerated growth in multicopters have forged new areas of research for this new bread of
aircraft. This has been beneficial for flight control development in general as
the low cost of a quadcopter has made it practical for anyone to engage in this
research. 

Quadcopters are naturally unstable and underactuated, meaning
each of the six degrees of freedom cannot be controlled directly. These complex 
dynamics present an interesting control problem. In order to maintain
stability, a quadcopter requires a control algorithm to calculate the power to
apply to each motor.

In academia there has been extensive research in flight control systems for 
quadcopters~\cite{zulu2014review,li2012survey}.   
Optimal  control algorithms have been applied using linear quadratic
Gaussian~(LQG)~\cite{minh2010modeling}, 
and $H_\infty$ which minimize a specific cost function until an optimally 
defined criteria is achieved. However these algorithms tend to lack 
robustness~\cite{zulu2014review,li2012survey}. Adaptive control using feedback 
linearization~\cite{palunko2011adaptive} have also been applied which allows for 
the system control parameters to adapt to change over time however these 
algorithms typically rely on mathematical models of the aircraft.  

Similar to flight control for crewed aircraft, there has also been a shift
towards intelligent control 
methods for UAVs to address limitations of traditional control methods.
Intelligent control is a control system that uses various artificial intelligent 
algorithms~\cite{santoso2017state}. 
These algorithms are broadly categorized 
into three different classes for what they provide: knowledge, learning and 
global search.  Knowledge algorithms consist of fuzzy and expert systems, 
learning algorithms encompass \nns, and global search contains search 
and optimization algorithms such as genetic algorithms and swarm intelligence.  
Each of these algorithms have their own advantages and disadvantages when it
comes to developing fight 
control systems. However  knowledge 
and global search algorithms do not have the functionality and capabilities to
provide  direct control of
the aircraft actuators. 
Knowledge-based algorithms are unable to adapt to new unseen events and lack
robustness, qualities that are undesirable for control tasks with noisy sensors
and complex nonlinear dynamics.
While global search algorithms are far to time consuming for real-time
control of an aircraft. \nns, on the other hand, have a number of characteristics that are
attractive for control.  They are universal 
approximators, resistant to noise~\cite{miglino1995evolving}, and provide predictive 
control~\cite{hunt1992neural}.

Intelligent PID flight control~\cite{fatan2013adaptive} methods have been 
proposed in which PID gains are dynamically updated online providing adaptive 
control as the environment changes. However these solutions still  
inherit disadvantages associated with PID control, such as integral windup,  
need for mixing, and most significantly, they are feedback controllers and 
therefore inherently \textit{reactive}.  On the other hand feedforward control (or 
predictive control) is \textit{proactive}, and allows the controller to output 
control signals before an error occurs. For feedforward control, a model of the system must exist.  
Learning-based intelligent control has been proposed to develop models of the 
aircraft for predictive control using artificial \nns.

Notable work by \cite{dierks2010output} proposes an intelligent 
flight control system constructed with \nns to learn the quadcopter 
dynamics, online, to navigate along a specified path. This method allows the 
aircraft to adapt in real-time to external disturbances and unmodelled dynamics.  
Matlab simulations demonstrate that their approach outperforms a PID controller in the 
presence of unknown dynamics, specifically in regards to control effort 
required to track the desired trajectory. Nonetheless the proposed approach  
requires prior knowledge of the aircraft mass and moments of inertia to estimate 
velocities.  While online learning is an essential component to construct a 
complete intelligent flight control system, nonetheless it is fundamental  to develop 
accurate offline models to establish an initial stable controller. Offline
learning can also teach the \nn how to respond to rare occurring events ahead
of time before
encountering them in
the real world~\cite{santoso2017state}.

To build offline models, previous work has used supervised learning to train 
intelligent flight control systems using a variety of data sources such as test 
trajectories~\cite{bobtsov2016hybrid}, and PID step responses 
\cite{shepherd2010robust}.
The limitation of this approach is that training data may not accurately reflect 
the underlying dynamics. In general, supervised learning on its own is not ideal 
for interactive problems such as control~\cite{sutton1998reinforcement}.

There is, however, an alternative learning paradigm for building offline models that is ideal for continuous control tasks, does
not make assumptions about the aircraft dynamics and is capable of creating
optimal control policies. This learning paradigm is known as reinforcement learning~(RL).

\subsection{Flight Control via  Reinforcement Learning}
\label{sec:bg:rl}

\begin{figure}
    \centering
    {\includegraphics[trim=5 290 250 10, clip, 
	width=0.9\textwidth]{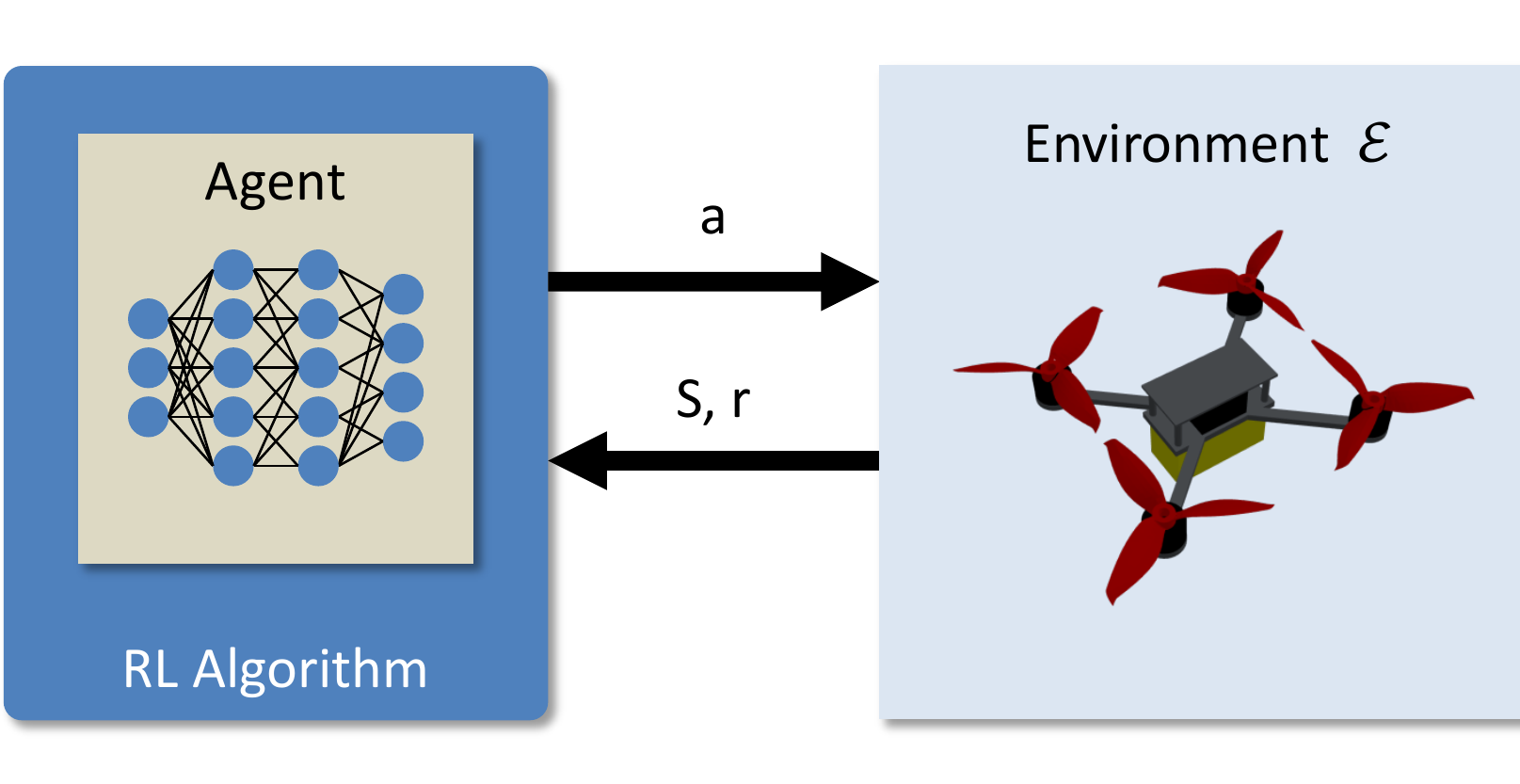}}
	\caption{{Deep RL architecture.}}
	\label{fig:rl}
\end{figure}

RL is a machine learning paradigm in which an agent
interacts with its environment in order to learn a task over time. 
Deep RL refers to the use of a \nn as the agent that contains two or more
hidden layers.
In this work we consider a deep RL architecture as depicted in Fig.~\ref{fig:rl}.
We will now describe the agents interaction with the environment in the context of
neuro-flight controller training.

At each discrete time-step $t$, the agent~(\ie \nn) receives an observation $S_t$ from the 
environment $\mathcal{E}$.  The environment 
consists of the aircraft and also the simulation world while observations are
obtained through various sensors onboard the aircraft. 
Because the agent is only receiving  sensor data, it is unaware of the 
entire physical environment and aircraft dynamics and therefore $\mathcal{E}$ is 
only partially observed by the agent. 
These observations are 
in the continuous observation spaces $S_t \in \mathbb{R}$. 
The 
observations are used as input to evaluate the agent to produce the action
$a_{t}$.
The action values are  also in 
the continuous range $a_t \in  \mathbb{R}^M$ and corresponds to the $M$ control
signals
to send to the ESC.
This action is applied to the environment and in return the agent receives a
single numerical reward $r_{t+1}$ 
indicating the performance of this action along with the updated state of the
environment $S_{t+1}$. 

In reality, during training, an RL algorithm acts as a shim between the agent and
the environment. The RL algorithm uses the action, state, and reward history in
order to adjust the weights of the \nn.

The interaction between the agent and $\mathcal{E}$ is formally defined as a 
Markov decision processes~(MDP) where the state transitions are defined as the 
probability of transitioning to state $s'$ given the current state and action 
are $s$ and $a$ respectively, $Pr\{s_{t+1} = s' | s_t = s, a_t = a\}$.  The 
behavior of the agent is defined by its policy $\pi$ which is essentially a 
mapping of what action should be taken for a particular state.  The objective of 
the agent is to maximize the returned reward overtime to develop an optimal 
policy.  We invite the reader to refer to \cite{sutton1998reinforcement} for 
further details on \new{RL}.

\new{RL} has similar goals to adaptive control in which a policy improves 
overtime interacting with its environment.  
RL has been applied to 
autonomous helicopters to learn how to track trajectories, 
specifically how to hover in place and perform various maneuvers
~\cite{bagnell2001autonomous,kim2004autonomous,abbeel2007application}.
Work by \cite{kim2004autonomous,abbeel2007application} validated their trained helicopter's 
capabilities in helicopter competitions requiring the aircraft to perform 
advanced acrobatic  maneuvers. Performance was compared to trained pilots, 
nevertheless it is unknown how their controllers compare to PID control for 
tracking trajectories.

The first use of RL in  quadcopter control was presented by  
\cite{waslander2005multi} for altitude control. The authors developed a 
model-based RL algorithm to search for an optimal control policy. The controller 
was rewarded for accurate tracking and damping. Their design provided 
significant improvements in stabilization in comparison to linear control
methods.

Up until recently control in continuous action spaces was considered  difficult 
for RL. Significant progress has been made combining the power of \nns with RL. 
State-of-the-art algorithms such as 
Deep Deterministic Policy 
Gradient~(DDPG)~\cite{lillicrap2015continuous}, Trust Region Policy 
Optimization~(TRPO)~\cite{schulman2015trust} and Proximal Policy
Optimization~(PPO)~\cite{schulman2017proximal}
have shown to be effective methods of training deep
\nns~\cite{duan2016benchmarking,gymfc}.
DDPG provides improvement to Deep Q-Network~(DQN)~\cite{mnih2013playing}
for the continuous action domain. It employs an actor-critic architecture using 
two \nns for each actor and critic.  It is also a model-free algorithm 
meaning it can learn the policy without having to first generate a model. TRPO 
is similar to natural gradient policy methods however this method guarantees 
monotonic improvements. PPO~\cite{schulman2017proximal}
 is known to out perform other state-of-the-art  methods in challenging 
environments. PPO is also a policy gradient method and has similarities to
TRPO. 
Its  novel objective function allows for a
Trust Region update to the policy at each training iteration. 
Many RL algorithms can be very sensitive
to hyperparameter tuning in order to obtain good results. Part of the reason PPO is so
widely adopted is due to it being easier to tune than other RL algorithms.

More recently \cite{hwangbo2017control} has used \new{deep RL} for 
quadcopter control, particularly for navigation control. They developed a novel 
deterministic on-policy learning algorithm that outperformed 
TRPO~\cite{schulman2015trust} and DDPG~\cite{lillicrap2015continuous} in regards 
to training time. Furthermore the authors validated their results in the real 
world, transferring their policy trained in simulation to a physical quadcopter.  Path 
tracking turned out to be adequate.  However the authors discovered major 
differences transferring from simulation to the real world.  

The vast majority of prior work has focused on performance of navigation and 
guidance. There is limited and insufficient data justifying the accuracy and 
precision of \nn-based intelligent attitude flight control and none 
previously for controllers trained via RL. 

\section{Transfer learning}
\label{sec:bg:transfer}
The desire  to train and evaluate intelligent control systems in simulation dates back to 
the 1990s as discussed in \cite{husbands1992evolution}. It is simply not practical 
to accomplish most training tasks in the real world as it would take far to long 
and be costly. However the fidelity and accuracy of the simulator drastically 
determines the controllers performance in the real world, in fact in some cases 
robots trained in simulated environments completely fail when transferred to a 
robot in the real world  \cite{brooks1992artificial}.  To address these issues 
several studies have proposed methods to reduce the reality gap.

In \cite{miglino1995evolving} the authors developed a simulator to train a 
neuro-controller for a two wheeled Khepera robot using evolutionary
algorithms.
The inputs of the \nn was connected directly to eight infrared sensors, and
the output was connected directly to two motors.
During their research they found the accuracy of the infrared sensors  
varied drastically from one another.  To adjust for these discrepancies in 
simulation the robot sensors were randomly sampled in the real world. To 
compensate for changes in light conditions noise was introduced in the simulated 
environment. Models of the robots motors were constructed in a similar way 
introducing noise to account for uncertainties in the environment (\eg 
imperfections on the floor).  
Individuals were evaluated based on how fast they were able to  
travel in a straight line while still avoiding obstacles. Results show the robot 
had  decreased in performance when transferred to a real robot, however 
continued training in the real world for a small number generations can revert 
and actually improve performance. The major contribution of this paper
demonstrates the 
reality gap can be greatly reduced by introducing noise into the training data.  
Noise accounts for uncertainties found in the real world, as \nns are 
noise resistant the \nn is able to learn the underlying dynamics 
despite the additional noise. 

Around the same time, work by \cite{jakobi1995noise} explored three 
claims made by  \cite{husbands1992evolution} to reduce the reality gap. 
First, a large amount of empirical data should be collected from the robots
    sensors, actuators and operating environment to be used to build
    accurate 
    simulation environments. The authors discuss what is
    now referred to as hardware in the loop (HITL) as a method to further
    increase the accuracy by using the actual hardware of the robot.
	Second, noise should be injected at all inputs to blur the two running environments together.  
    Third, adaptive noise tolerant elements should be used to absorb the 
	discrepancies in the simulated environment from the real world.

The authors also performed their evaluations with the Khepera robot.
First, mathematical models for each 
sensor and actuator in the system was defined based on elementary physics and 
control theory.  Several experiments were conducted to collect empirical data on 
these devices and then mapping techniques were created to map the calculated 
value to the sampled value. 
To identify the ideal amount 
of noise to introduced into the simulation  the \nn 
was trained on three noise levels: zero, observed, and double observed. Observed 
noise is created from a Gaussian distribution with the standard deviation equal 
to that of the collected empirical data. Results verify previous work claims 
that the addition of noise in the simulator provides improved performance in the 
real world. Furthermore it was found that the normal observed noise level 
provided the best performance of the three. However there is a fine line in the amount 
of noise that is best, in some cases injecting double the observed noise 
performed worse than no noise at all.

If neuro-controllers synthesized in simulation via RL
are to be adopted for use in real CPS,
it is critical to reduce the reality gap.
There have been several studies addressing the reality gap in the context of
RL.

In \cite{tobin2017domain}, the authors explore a method called domain
randomization for reducing the reality gap. Domain
randomization randomizes parts of the simulation environment with the idea
being if the simulation has enough variety, the real world will just appear as
another variation to the agent. In relation to the use of noise, domain
randomization is a generalized method for adding variation to the environment which
consists of the use of noise.  
The authors particular application is in computer vision in which a \nn is trained to
detect the location of an object. They randomize the location, number and shape
of the objects. Additionally textures of object and environment were
randomized. Similar to \cite{miglino1995evolving} noise and also lighting conditions were
also randomized. Their evaluation shows that domain randomization can provide
high enough accuracy to locate and grasp an object from clutter.

In more recent work by \cite{andrychowicz2018learning}
the authors applied deep RL to learn dexterous in-hand manipulation, a task that is beyond the capabilities of traditional control methods.
The intention of this work is to show 
transferability of the learned policy to a real robot.
To overcome the reality gap, the authors randomized most aspects of the
simulation environment. In addition to applying noise to the observations, and
randomizing visual properties  they also
randomized physical parameters such as friction and introduce delays and noise
to the actions. Although domain randomization did narrow the reality gap, the
real robot performed worse than in simulation. Transferability was most
successful when the entire training environment state was randomized but they
did point out that the affects of observation randomization had the least
impact which they attribute to the accuracy  of their motion capture system.   
Another interesting observation was the fact that training on a randomized
environment converged significantly slower, than when trained without
randomization.

In the context of flight control, authors in \cite{molchanov2019sim} investigate domain
randomization for a RL-based stabilization flight controller. Particularity their focus is in
developing a policy that can be transferred to multiple different quadcopter
configurations. In this work they randomize the mass, the motor distance, motor
response, torque and thrust characteristics.  
Training was conducted in their own simulation using mathematical models for
the quadcopter dynamics. A Tensorflow based learning framework was used
for training and the trained policy was transfer to hardware by extracting
the trained \nn parameters from the Tensorflow model to a custom \nn C library.
Policy evaluation was performed on three different quadcopters.
Their results show the policy trained for a specific aircraft, without
randomization performed best. 
Similar observations to \cite{andrychowicz2018learning} were reported in which 
domain randomization provided moderate improvements. 
Full randomization generalized better but other policies provided 
better performance for each particular aircraft.

To further reduce the reality gap and easy the transfer to hardware it is essential to increase the accuracy of
 the aircraft model (\ie digital twin) used in simulation during training.

\section{Digital Twinning}
\label{sec:bg:twin}
The concept of digital twinning was first introduced in Michael Grieves's course
on Product Lifecycle Management (PLM) in 2003~\cite{grieves2014digital}. 
He defines the digital twin concept to consist of three main parts, the physical asset in
the real space, the virtual asset in virtual space, and a data connection link
between these two spaces. 
With the rise of CPS, there is a plethora of sensor data
available fueling  new applications for digital twins. 

In work provided by \cite{gabor2016simulation}  a 
generic software architecture for the integration of digital twins is proposed.  There has 
been a paradigm shift from classical simulation architectures as the cognitive 
system (\ie the system consisting of the logic to perform some desired 
functionality) now as the ability to communicate with both the physical world 
(\ie the hardware) and a simulator (\ie digital twin). From the CPS's software
perspective it should be indistinguishable  whether 
it is interacting with hardware or its digital twin.  Thus 
it is required the hardware and digital twin must implement  identical interfaces.  The authors introduce 
an observer design pattern to allow subcomponents in the software architecture 
to communicate. 

Although the digital twinning concept was initially described in
the context of manufacturing, in regards to aviation it has been adopted  by NASA for vehicle health
management~\cite{glaessgen2012digital} and GE Aviation for jet engine analytics
and modelling.  

Digital twinning has been proposed as a method to optimize practices regarding 
certification, fleet management and sustainment of future NASA and U.S. Air Force
vehicles~\cite{glaessgen2012digital}. Current approaches are inefficient. Based
on insufficient data of the aircraft, assumptions about system health are made
based on statistics and heuristics from past observations and experiences.
This  can lead to unnecessary
inspections, or worse, result in damage for an
aircraft that has a unique, previously unseen experience. 
As next generation aircraft become more sophisticated, greater introspection of
the individual aircraft will be required.
A digital twin can address these issues by providing near real-time analytics
and state of an individual aircraft. More specifically the authors describe the
use of digital twins to provide a method to continuously predict the health of
the aircraft. This has remarkable benefits such as the ability to predict
future failures and address them early on before they become severe. 

A digital twin is just one of the technologies used as part of larger vision of NASA's
to create self-aware vehicles~\cite{tuegel2011reengineering}.
The authors define a self-aware vehicle as ``an aircraft, spacecraft or system
is one that is aware of its internal state, has situational awareness of its
environment, can assess its capabilities currently and project them into the
future, understands its mission objectives, and can make decisions under
uncertainty regarding its ability to achieve its mission objectives."

Digital twinning provides the self-aware vehicle with the ability to monitor
system health in real-time and forecast failures before they occur.
This results in unparalleled degree of safety. 
Depending on the current aircraft state, a flight envelope can be uniquely
establish to
ensure predictable performance while operating in that range.
Furthermore sensor data is 
relayed back to a ground stations to utilize the collective computational
power of server farms to further assess the state of the aircraft.

In this dissertation we incorporate digital twinning concepts as a method to 
synthesize optimal flight controller policies that are unique to each individual
aircraft.

\cleardoublepage
\newcommand\ppoaa{0}
\newcommand\ppoab{0}
\newcommand\ppoac{0}
\newcommand\ppoad{0}
\newcommand\ppoae{0}
\newcommand\ppoaf{0}
\newcommand\ppoag{0}
\newcommand\ppoah{0}
\newcommand\ppoai{0}
\newcommand\ppoaj{0}
\newcommand\ppoak{0}
\newcommand\ppoal{0}
\newcommand\ppoam{0}
\newcommand\ppoan{0}
\newcommand\ppoao{0}
\newcommand\ppoap{0}
\newcommand\ppoaq{0}
\newcommand\ppoar{0}
\newcommand\ppoas{0}
\newcommand\ppoat{0}
\newcommand\ppoau{0}
\newcommand\ppoav{0}
\newcommand\ppoaw{0}
\newcommand\ppoax{0}
\newcommand\ppoba{0}
\newcommand\ppobb{0}
\newcommand\ppobc{0}
\newcommand\ppobd{0}
\newcommand\ppobe{0}
\newcommand\ppobf{0}
\newcommand\ppobg{0}
\newcommand\ppobh{0}
\newcommand\ppobi{0}
\newcommand\ppobj{0}
\newcommand\ppobk{0}
\newcommand\ppobl{0}
\newcommand\ppobm{0}
\newcommand\ppobn{0}
\newcommand\ppobo{0}
\newcommand\ppobp{0}
\newcommand\ppobq{0}
\newcommand\ppobr{0}
\newcommand\ppobs{0}
\newcommand\ppobt{0}
\newcommand\ppobu{0}
\newcommand\ppobv{0}
\newcommand\ppobw{0}
\newcommand\ppobx{0}
\newcommand\ppoca{0}
\newcommand\ppocb{0}
\newcommand\ppocc{0}
\newcommand\ppocd{0}
\newcommand\ppoce{0}
\newcommand\ppocf{0}
\newcommand\ppocg{0}
\newcommand\ppoch{0}
\newcommand\ppoci{0}
\newcommand\ppocj{0}
\newcommand\ppock{0}
\newcommand\ppocl{0}
\newcommand\ppocm{0}
\newcommand\ppocn{0}
\newcommand\ppoco{0}
\newcommand\ppocp{0}
\newcommand\ppocq{0}
\newcommand\ppocr{0}
\newcommand\ppocs{0}
\newcommand\ppoct{0}
\newcommand\ppocu{0}
\newcommand\ppocv{0}
\newcommand\ppocw{0}
\newcommand\ppocx{0}
\newcommand\trpoaa{0}
\newcommand\trpoab{0}
\newcommand\trpoac{0}
\newcommand\trpoad{0}
\newcommand\trpoae{0}
\newcommand\trpoaf{0}
\newcommand\trpoag{0}
\newcommand\trpoah{0}
\newcommand\trpoai{0}
\newcommand\trpoaj{0}
\newcommand\trpoak{0}
\newcommand\trpoal{0}
\newcommand\trpoam{0}
\newcommand\trpoan{0}
\newcommand\trpoao{0}
\newcommand\trpoap{0}
\newcommand\trpoaq{0}
\newcommand\trpoar{0}
\newcommand\trpoas{0}
\newcommand\trpoat{0}
\newcommand\trpoau{0}
\newcommand\trpoav{0}
\newcommand\trpoaw{0}
\newcommand\trpoax{0}
\newcommand\trpoba{0}
\newcommand\trpobb{0}
\newcommand\trpobc{0}
\newcommand\trpobd{0}
\newcommand\trpobe{0}
\newcommand\trpobf{0}
\newcommand\trpobg{0}
\newcommand\trpobh{0}
\newcommand\trpobi{0}
\newcommand\trpobj{0}
\newcommand\trpobk{0}
\newcommand\trpobl{0}
\newcommand\trpobm{0}
\newcommand\trpobn{0}
\newcommand\trpobo{0}
\newcommand\trpobp{0}
\newcommand\trpobq{0}
\newcommand\trpobr{0}
\newcommand\trpobs{0}
\newcommand\trpobt{0}
\newcommand\trpobu{0}
\newcommand\trpobv{0}
\newcommand\trpobw{0}
\newcommand\trpobx{0}
\newcommand\trpoca{0}
\newcommand\trpocb{0}
\newcommand\trpocc{0}
\newcommand\trpocd{0}
\newcommand\trpoce{0}
\newcommand\trpocf{0}
\newcommand\trpocg{0}
\newcommand\trpoch{0}
\newcommand\trpoci{0}
\newcommand\trpocj{0}
\newcommand\trpock{0}
\newcommand\trpocl{0}
\newcommand\trpocm{0}
\newcommand\trpocn{0}
\newcommand\trpoco{0}
\newcommand\trpocp{0}
\newcommand\trpocq{0}
\newcommand\trpocr{0}
\newcommand\trpocs{0}
\newcommand\trpoct{0}
\newcommand\trpocu{0}
\newcommand\trpocv{0}
\newcommand\trpocw{0}
\newcommand\trpocx{0}
\newcommand\ddpgaa{0}
\newcommand\ddpgab{0}
\newcommand\ddpgac{0}
\newcommand\ddpgad{0}
\newcommand\ddpgae{0}
\newcommand\ddpgaf{0}
\newcommand\ddpgag{0}
\newcommand\ddpgah{0}
\newcommand\ddpgai{0}
\newcommand\ddpgaj{0}
\newcommand\ddpgak{0}
\newcommand\ddpgal{0}
\newcommand\ddpgam{0}
\newcommand\ddpgan{0}
\newcommand\ddpgao{0}
\newcommand\ddpgap{0}
\newcommand\ddpgaq{0}
\newcommand\ddpgar{0}
\newcommand\ddpgas{0}
\newcommand\ddpgat{0}
\newcommand\ddpgau{0}
\newcommand\ddpgav{0}
\newcommand\ddpgaw{0}
\newcommand\ddpgax{0}
\newcommand\ddpgba{0}
\newcommand\ddpgbb{0}
\newcommand\ddpgbc{0}
\newcommand\ddpgbd{0}
\newcommand\ddpgbe{0}
\newcommand\ddpgbf{0}
\newcommand\ddpgbg{0}
\newcommand\ddpgbh{0}
\newcommand\ddpgbi{0}
\newcommand\ddpgbj{0}
\newcommand\ddpgbk{0}
\newcommand\ddpgbl{0}
\newcommand\ddpgbm{0}
\newcommand\ddpgbn{0}
\newcommand\ddpgbo{0}
\newcommand\ddpgbp{0}
\newcommand\ddpgbq{0}
\newcommand\ddpgbr{0}
\newcommand\ddpgbs{0}
\newcommand\ddpgbt{0}
\newcommand\ddpgbu{0}
\newcommand\ddpgbv{0}
\newcommand\ddpgbw{0}
\newcommand\ddpgbx{0}
\newcommand\ddpgca{0}
\newcommand\ddpgcb{0}
\newcommand\ddpgcc{0}
\newcommand\ddpgcd{0}
\newcommand\ddpgce{0}
\newcommand\ddpgcf{0}
\newcommand\ddpgcg{0}
\newcommand\ddpgch{0}
\newcommand\ddpgci{0}
\newcommand\ddpgcj{0}
\newcommand\ddpgck{0}
\newcommand\ddpgcl{0}
\newcommand\ddpgcm{0}
\newcommand\ddpgcn{0}
\newcommand\ddpgco{0}
\newcommand\ddpgcp{0}
\newcommand\ddpgcq{0}
\newcommand\ddpgcr{0}
\newcommand\ddpgcs{0}
\newcommand\ddpgct{0}
\newcommand\ddpgcu{0}
\newcommand\ddpgcv{0}
\newcommand\ddpgcw{0}
\newcommand\ddpgcx{0}
\newcommand\pida{0}
\newcommand\pidb{0}
\newcommand\pidc{0}
\newcommand\pidd{0}
\newcommand\pide{0}
\newcommand\pidf{0}
\newcommand\pidg{0}
\newcommand\pidh{0}
\newcommand\pidi{0}
\newcommand\pidj{0}
\newcommand\pidk{0}
\newcommand\pidl{0}
\newcommand\pidm{0}
\newcommand\pidn{0}
\newcommand\pido{0}
\newcommand\pidp{0}
\newcommand\pidq{0}
\newcommand\pidr{0}

\newcommand\trials{3\xspace}
\newcommand\totalcommands{3,000\xspace}
\newcommand\ci{95\%\xspace}
\newcommand\thresholdband{10\%\xspace}
\newcommand\thresholdriselow{10\%\xspace}
\newcommand\thresholdrisehigh{90\%\xspace}
\newcommand\rtddpg{33 hours\xspace}
\newcommand\rtppo{9 hours\xspace}
\newcommand\rttrpo{13 hours\xspace}

\newcommand{\CHone}{\textbf{C1}}
\newcommand\CHtwo{\textbf{C2}}
\newcommand\CHthree{\textbf{C3}}

\newcommand\extra{(Section~\ref{sec:continuous})\xspace}

\chapter{Reinforcement Learning for UAV Attitude Control}
\label{chapter:gymfc}
\thispagestyle{myheadings}

\graphicspath{ {2_GymFC/figures/} }

Over the last decade there has been an uptrend in the popularity of UAVs. In particular, quadcopters have
received significant attention in the research community where a
significant number of seminal results and applications have been proposed and experimented.
This recent growth is primarily attributed to the drop in cost of
onboard sensors, actuators and small-scale embedded computing
platforms. Despite the significant progress, flight control is still
considered an open research topic. On the one hand, flight control
inherently implies the ability to perform highly time-sensitive
sensory data acquisition, processing and computation of forces to
apply to the aircraft actuators. On the other hand, it is desirable
that UAV flight controllers are able to tolerate faults; adapt to
changes in the payload and/or the environment; and to optimize flight
trajectory, to name a few.

Autopilot systems for UAVs are typically composed of an ``inner loop" 
responsible for aircraft stabilization and control, and an ``outer loop" to 
provide mission level objectives (\eg way-point navigation).  Flight control 
systems for UAVs are predominately implemented using the Proportional, Integral 
Derivative~(PID) control systems. PIDs have demonstrated exceptional 
performance in many circumstances, including in the context of drone racing, 
where precision and agility are key. In stable 
environments a PID controller exhibits close-to-ideal performance. 
When exposed to unknown dynamics (\eg wind, variable payloads, voltage 
sag, etc), however, a PID controller can be far from optimal~\cite{maleki2016reliable}.
For next generation flight control systems to be intelligent, a way needs to be devised to
incorporate adaptability to mutable dynamics and environment.

The development of intelligent flight control systems is an active area of 
research~\cite{santoso2017state}, specifically through the use of \nns which are an attractive option given they are universal 
approximators and resistant to noise~\cite{miglino1995evolving}. 

Online learning methods~(\eg \cite{dierks2010output}) have the advantage of 
learning the aircraft dynamics in real-time. The main limitation with online learning 
is that the flight control system is 
only knowledgeable of its past experiences. It follows that its performances are limited when 
exposed to a new event.  Training models offline using supervised learning is 
problematic as data is expensive to obtain and derived from inaccurate 
representations of the underlying aircraft dynamics (\eg flight data from a 
similar aircraft using PID control) which can lead to suboptimal control policies~\cite{bobtsov2016hybrid,shepherd2010robust,williams2005flight}.
To construct high-performance intelligent flight control systems it is necessary 
to use a hybrid approach. First, accurate offline models are used to construct a
baseline controller, while online learning provides fine tuning and real-time adaptation. 

An alternative to supervised learning for creating offline models is RL.
Using RL it is possible to develop optimal control policies for a 
UAV without making any assumptions about the aircraft dynamics. 
Recent work has shown RL to be effective for UAV autopilots, providing adequate 
path tracking~\cite{hwangbo2017control}. Nonetheless, previous  work on intelligent 
flight control systems has primarily focused on guidance and navigation.

\new{
\textbf{Open Challenges in RL for Attitude Control} RL is currently being 
applied to a wide range of applications.
each with its own set of challenges.  Attitude control for UAVs is a 
particularly interesting RL problem for a number of reasons.  We've highlighted 
three areas we find important below:}
 \begin{itemize}
	 \item[\textbf{C1}] \new{\textbf{Precision and Accuracy:}
Many RL tasks can be solved in a variety of ways. For example, to win a game 
there may be a number of sequential moves that will lead to the same outcome.  
In the case of optimal attitude control there is little tolerance and 
flexibility as to the sequence of control signals that will achieve the desired 
attitude (\emph{e.g.} angular rate) of the aircraft.  Even the slightest 
deviations can lead to instabilities.
It remains unclear what level of control accuracy can be achieved when using 
intelligent control trained with RL for time-sensitive attitude control --- 
\emph{i.e.} the ``inner loop".  
Therefore determining the achievable level of accuracy is critical in 
establishing if RL is suitable for attitude flight control.
} 

	\item[\textbf{C2}]\new {\textbf{Robustness and Adaptation:} In the context 
			 of control, robustness refers to the controllers performance in the 
			 presence of uncertainty when control parameters are fixed while 
			 adaptiveness refers to the controllers performance to adapt to the 
			 uncertainties by adjusting the control 
			 parameters~\cite{wang2001fundamental}. 
			 It is assumed the \nn trained with RL will face 
			 uncertainties  when transfer to physical hardware due to the
             reality gap. 
			 However it remains unknown what range of uncertainty the controller 
			 can operate safely before adaptation is necessary.  
			 Characterizing the controllers robustness will provide valuable 
			 insight into the design of the intelligent flight control system 
			 architecture.  For instance what will be the necessary adaptation 
			 rate and what sensor data can be collected from the real world to 
			 update the RL environment.
			 %to update the RL environment to reduce this gap and if 
			 %modifications can be make to the RL environment to increase 
			 %robustness.  
			% 
			 %Given training is time consuming, investigating whether a baseline 
			 %neural network can be constructed with knowledge of basic flight 
			 %dynamics and then later fine tuned for specific aircraft 
			 %configuration would support development and adoption of these 
			 %types of controllers~(this is referred to as domain adaptation in 
			 %the machine learning communities CITE). 
%
%
%
%
%
%
%
%
%
			%A neural network controller unique to each individual aircraft has 
			%the potential not possible     	
		}
	
\item[\textbf{C3}]\new{\textbf{Reward Engineering:} 
		%Reward engineering also affects the speed in which the agent will 
		%learn, more expressive rewards can lead to the agent reaching the goal 
		%quicker.  
	%	
		In the context of attitude control, the reward must encapsulate the 
		agent's performance achieving the desired attitude goals.  As goals 
		become more complex and demanding~(\emph{e.g.} minimizing energy 
		consumption, or stability in presence of damage ) identifying which 
		performance metrics are most expressive  will be necessary to push the 
		performance of intelligent control systems trained with RL.
%
%
%
		%Making inaccurate assumptions can cause  adverse and unpredictable 
		%affects in flight performance.	Although multiple reward functions may 
		%jbe adequate, as performance goals and aircraft models become more 
		%complex identifying and developing expressive reward functions  
%
		%Attitude control is an interesting task for RL because it actually 
		%composed into two sequential goals. The first goal requires the 
		%controller to reach the target angular velocity and then the second to 
		%maintain stability at the desired target.    
	}

  \end{itemize}

\textbf{Our Contributions} In this chapter we study \new{in-depth \CHone}, the 
accuracy and precision of attitude control provided by intelligent flight 
controllers trained using RL.  While we specifically focus on the creation of 
controllers for the Iris quadcopter~\cite{iris}, the methods developed hereby 
apply to a wide range of multi-rotor UAVs, and can also be extended to 
fixed-wing aircraft.  We develop a novel training environment called
\gymfcGeneric with 
the use of a high fidelity physics simulator for the agent to learn attitude 
control. This being the initial release, it will be referred to as \gym for the
remainder of the chapter. \gym is an OpenAI Environment~\cite{brockman2016openai} providing a 
common interface for researchers to develop intelligent flight control systems.  
The simulated environment consists of an Iris quadcopter digital twin~\cite{gabor2016simulation}. The intention is to
eventually be able to transfer the trained policy to physical hardware.
Controllers are trained using state-of-the-art RL algorithms: Deep Deterministic 
Policy Gradient~(DDPG), Trust Region
Policy Optimization~(TRPO), and Proximal Policy Optimization~(PPO). We then 
compare the performance of our synthesized controllers with that of a PID 
controller.  Our evaluation finds that controllers trained using PPO outperform 
PID control and are capable of exceptional performance. 
To summarize, this chapter makes the following contributions: 

\begin{itemize}
\item \gym, an open source~\cite{gymfc} environment for developing intelligent 
	attitude flight controllers while providing the research community a tool to
	progress performance.

\item A learning architecture for attitude control utilizing digital twinning 
	concepts for minimal effort when transferring trained controllers into 
	hardware.

\item An evaluation for state-of-the-art RL algorithms, such as Deep 
	  Deterministic Policy Gradient~(DDPG), Trust Region Policy 
	  Optimization~(TRPO),
	  and Proximal Policy Optimization~(PPO), learning policies for
      aircraft attitude control. As a first work in this direction,
      our evaluation also establishes a baseline for future work.

\item An analysis of intelligent flight control performance developed with RL 
      compared to traditional PID control.

\end{itemize}

The remainder of this chapter is organized as follows. In 
Section~\ref{sec:gymfc:related} we  review simulation environments and architectures currently
used for training RL policies. In Section~\ref{sec:gymfc:env} we present our training environment and 
use this environment to evaluate RL performance for flight control in 
Section~\ref{sec:gymfc:eval}. Finally Section~\ref{sec:gymfc:conclusion}
concludes the chapter
and provides a number of future research directions.

\section{Background and Related Work}
\label{sec:gymfc:related}
The release of OpenAI Gym~\cite{brockman2016openai}
made a huge splash in the RL community providing a common API for
RL environments and a repository of various environments implementing this API.
This common API has had a large impact on RL algorithm evaluations and has become
the staple for  benchmarking new algorithms.
Since its release a number of popular RL algorithm libraries have added
supported for OpenAI Gym including OpenAI Baselines~\cite{baselines}, 
Stable Baselines~\cite{stable-baselines},
 Tensorforce~\cite{schaarschmidt2017tensorforce},
Keras-RL~\cite{plappert2016kerasrl}, and
TF-Agents~\cite{TFAgents}.

Creating an instance of the environment is as easy as calling
\texttt{gym.make(env\_id)} in which \texttt{env\_id} is a string representing the unique ID of
the environment. The simplistic environment creation is beneficial for
benchmarking purposes as it  
provides a consistent environment.  Nonetheless, this is an issue 
for more complex environments that have the intention of using the trained
policy in the real world. One could argue for a specific application there is
no need for a common API. However one of the advantages of the Gym API as we
previously mentioned is the vast adoption of the API by RL algorithm libraries.
This allows
one to stand up a training environment with only a few lines of code and easily
allow users to switch from one RL algorithm to another. 

Within the collection of environments, a number of continuous control
environments exist such as controlling a lunar lander, race car, and
a bipedal walker. Additionally there exist robotic tasks such
as hand manipulation using the MuJoCo physics engine~\cite{todorov2012mujoco}.  
Using OpenAI Gym's API, researchers and developers have begun to
create their own environments. 

Gazebo~\cite{koenig2004design} is a mature open source high fidelity simulator
and has been used as a simulator backend for training environments.  
 It is also a popular simulator choice
for SITL and HITL testing of flight control firmware projects, for example
Betaflight~\cite{betaflight}, PX4~\cite{meier2015px4} and
Ardupilot~\cite{ardupilot}.
Gazebo supports the open source physics engines ODE~\cite{ode}, Bullet~\cite{bullet},
Simbody~\cite{sherman2011simbody} and
DART~\cite{lee2018dart} giving
the user the flexibility to choose the best one for their application.
Gazebo also provides a C++ API for developing custom models and dynamics as well as
a Google Protobuf API for externally interacting with the simulation
environment. Simulation worlds and models are constructed via the SDF file format~\cite{sdf} which is an XML file with a schema specific for describing robots
and their environments.

In \cite{zamora2016extending} the authors
present a gym learning framework for the robotic operating system~(ROS) and Gazebo.
This project contains an environment for the Erle-Copter~\cite{erle} to learn obstacle
avoidance. The user must provide a autopilot backend such as PX4 to interface
with the quadcopter.  
However since the release of this whitepaper, the project has been depreciated and the
authors placed a focus on environments for robotics arms rather than flight
control. 

Airsim~\cite{shah2018airsim}, a flight simulator developed by Microsoft, yields realistic
visualizations which can reduce the reality gap for flight control systems
using visual navigation.
This is achieved using the Unreal Engine, due to the difficulties involved in
trying to build
large scale realistic environments using Gazebo.
The architecture is designed in such a way to be interchangeable with various
vehicles and protocols.  Furthermore the simulator is capable of running at high
frequencies to support HITL simulations. 
However Airsim on its own does not provide training environments.

To support RL training tasks, AirLearning~\cite{krishnan2019air} introduces
a benchmarking platform for synthesizing high-level navigation flight controllers.
The authors address challenges with generating random environments and provide
a configurable way to change  the difficulty of the generated environment. 
The architecture is developed with HITL simulation in mind with a unique
approach of decoupling the policy with the hardware to allow evaluations to be
conducted for a variety of hardware configurations. 
This work also evaluates trained policies with quality of flight metrics such
as flight time, energy consumed and distance traveled.

\section{Reinforcement Learning Architecture}
\label{sec:gymfc:rl}
\begin{figure*}
    \centering
	{\includegraphics[trim=3 400 335 0, clip, 
	width=0.9\textwidth]{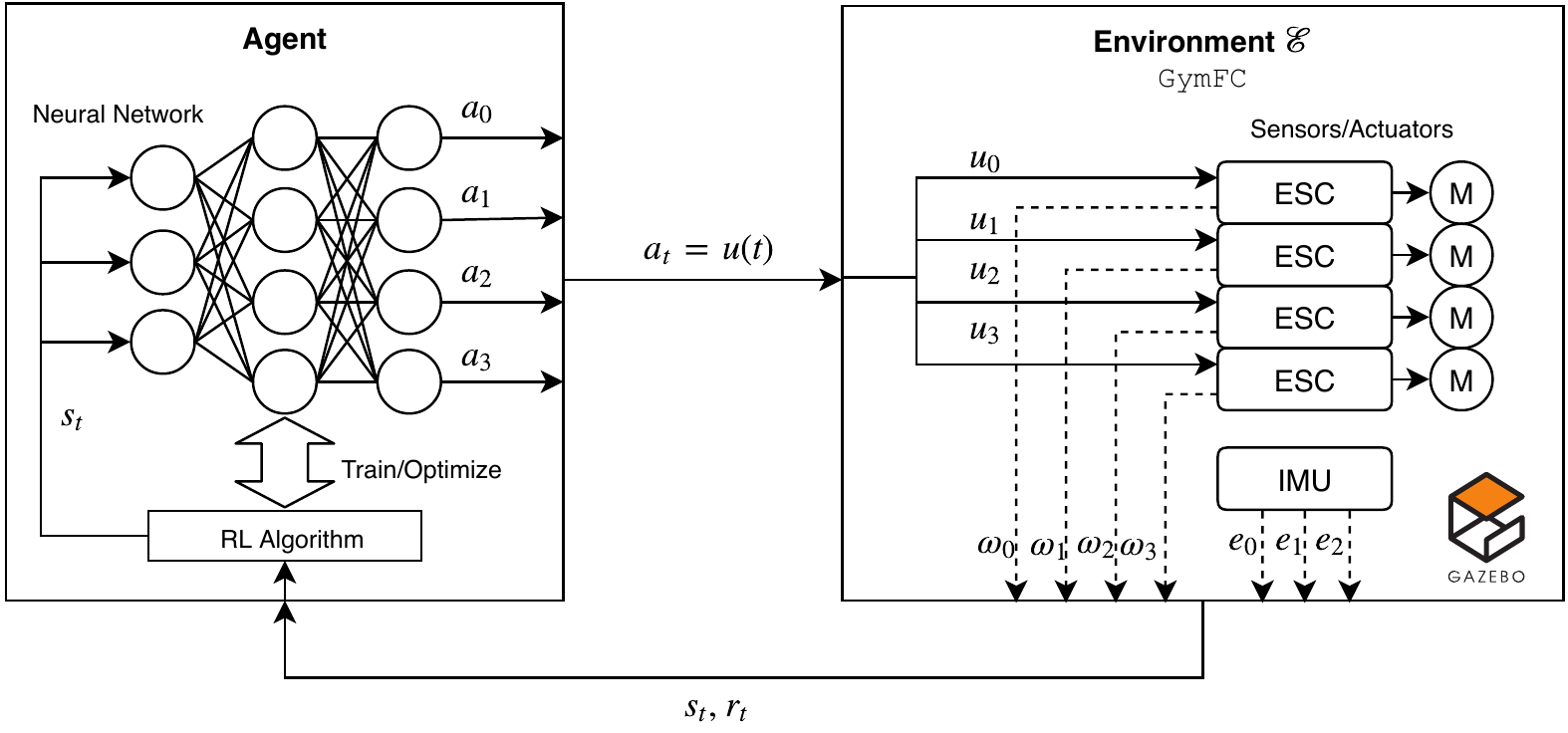}}
	\caption{\new{RL architecture using the \textsc{GymFC} environment 
	for training intelligent attitude flight controllers.}}
	\label{fig:rlenv}
\end{figure*}

In this work we consider \new{an RL architecture depicted in 
Figure~\ref{fig:rlenv} consisting of} a \nn-based flight controller as an 
agent  interacting with an Iris quadcopter~\cite{iris} in a high fidelity 
physics simulated  environment $\mathcal{E}$, more specifically using the
Gazebo simulator~\cite{koenig2004design}.   
Given our goal is developing low level attitude controllers, we do not need a
simulator with realistic visualizations. 
In this 
work we use the Gazebo simulator in light of its maturity, 
flexibility, extensive documentation, and active community.  

At each discrete time-step $t$, the agent receives an observation $x_t$ from the 
environments  consisting of the angular velocity error of each axis $e = 
\Omega^* - \Omega$ and the angular velocity of each rotor $\omega_i$ which are 
obtained from the quadcopter's emulated  \new{inertial measurement unit (IMU)} and 
electronic speed controller (ESC) sensors respectively. These observations are 
in the continuous observation spaces $x_t \in \mathbb{R}^{(M+D)}$ where $D=3$
degrees 
of rotational freedom. 
Once the observation  is received, the agent executes an action $a_t$ within 
$\mathcal{E}$.  In return the agent receives a single numerical reward $r_t$ 
indicating the performance of this action.  The action is also in a
continuous action space $a_t \in  \mathbb{R}^M$ and corresponds to the four 
control signals \new{$u(t)$ sent to each ESC driving the attached motor {\small 
		\circled{M}}}.  
Because the agent is only receiving this sensor data it is unaware of the 
physical environment and the aircraft dynamics and therefore $\mathcal{E}$ is 
only partially observed by the agent. 
Motivated by \cite{mnih2013playing} we consider the state to be a sequence of 
the past observations and actions $s_t = x_i, a_i, \dots, a_{t-1}, x_t$.

\section{\gym}
\label{sec:gymfc:env}
In this section we describe our learning environment \gym for developing 
intelligent flight control systems using RL.  The goal of the proposed environment is to
allow the agent to learn attitude control of an aircraft with only the 
knowledge of the number of actuators.  \gym includes both an {\bf episodic task} and a 
{\bf continuous task}. In an episodic task, the agent is required to learn a policy for 
responding to individual angular velocity commands. This allows the agents to 
learn the step response from rest for a given command, allowing its performance 
to be accurately measured. Episodic tasks however are not reflective of realistic
flight conditions. For this reason, in a continuous task, pulses with random widths 
and amplitudes are continuously generated, and correspond to angular velocity set-points. 
The agent must respond accordingly and track the desired target over time.
In Section~\ref{sec:gymfc:eval} we evaluate our synthesized controllers via episodic tasks, 
but we have strong experimental evidence that training via episodic tasks produces
controllers that behave correctly in continuous tasks as well~\extra.

\begin{figure}
\centering
{\includegraphics[trim=0 120 230 0, clip, width=0.85\columnwidth]{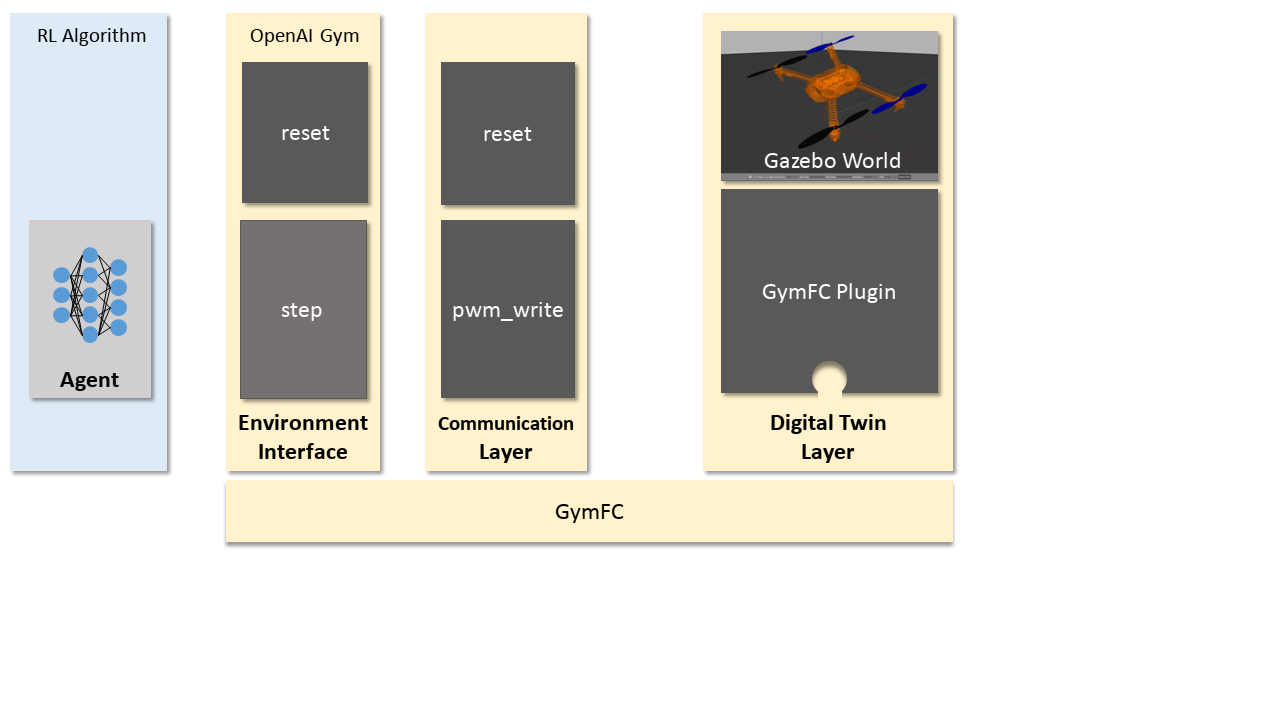}}
\caption{Overview of \gym environment architecture. }
\label{fig:env}
\end{figure}

\gym has a multi-layer hierarchical architecture composed of three layers: (i) a 
digital twin layer, (ii) a communication layer, and (iii) an agent-environment interface layer.  
This design decision was made to clearly establish roles and allow layer 
implementations to change (\eg to use a different simulator) without affecting 
other layers as long as the layer-to-layer interfaces remain intact.  A high 
level overview of the environment architecture is illustrated in 
Figure~\ref{fig:env}.  We will now discuss in greater detail each layer with a 
bottom-up approach.

\subsection{Digital Twin Layer}
\label{sec:gymfc:twin}
At the heart of the learning environment is a high fidelity physics simulator 
which provides functionality and realism that is hard to achieve with an
abstract mathematical model of the aircraft and environment. One of the primary
design goals of \gym is to minimize the effort required to transfer a controller
from the learning environment into the final platform.
For this reason, the simulated environment exposes identical interfaces to  
actuators and sensors as they would exist in the physical world.  In the ideal 
case the agent should not be able to distinguish between interaction with the 
simulated world (\ie its digital twin) and its hardware counter part.  

In a nutshell, the {\bf digital twin layer} is defined by (i) the
simulated world, and (ii) its interfaces to the above communication
layer (see Figure~\ref{fig:env}).

\begin{figure}
	\centering
	{\includegraphics[trim=45 0 65 30, clip, width=0.48\textwidth]{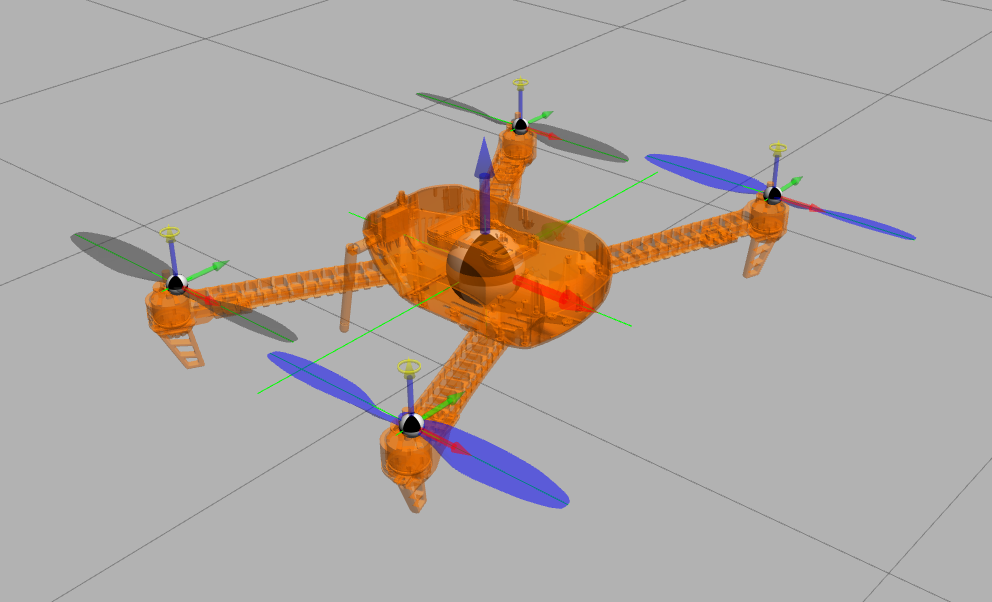}}
	\caption{The Iris quadcopter in Gazebo one meter above the ground.  The body 
	is transparent to show where the center of mass is linked as a ball joint to 
	the world.  Arrows represent the various joints used in the model. }
	\label{fig:iris}
\end{figure}

\textbf{Simulated World} The simulated world is constructed specifically for UAV 
attitude control in mind.  The technique we developed allows attitude control to 
be accomplished independently of guidance and/or navigation control.  This is 
achieved by fixing the center of mass of the aircraft to a ball joint in the 
world, allowing it to rotate freely in any direction, which would be impractical 
if not impossible to achieved in the real world due to gimbal lock and friction 
of such an apparatus. In this work the aircraft to be controlled in the 
environment is modeled off of the Iris quadcopter~\cite{iris}  with a weight of 
1.5~Kg, and 550~mm motor-to-motor distance. An illustration of the quadcopter in 
the environment is displayed in Figure~\ref{fig:iris}. Note during training 
Gazebo runs in headless mode without this user interface to increase simulation 
speed.  This architecture however can be used with any multicopter as long as a 
digital twin can be constructed.  Helicopters and multicopters represent 
excellent candidates for our setup because they can achieve a full range of 
rotations along all the three axes.  This is typically not the case with 
fixed-wing aircraft.  Our design can however be expanded to support fixed-wing 
by simulating airflow over the control surfaces for attitude control.  Gazebo 
already integrates a set of tools for modelling lift and drag.

\textbf{Interface} The digital twin layer provides two command interfaces to the 
communication layer:  simulation reset and motor update. Simulation reset 
commands are supported by Gazebo's API and are not part of our implementation.  
Motor updates are provided by a UDP server. We hereby discuss our approach to 
developing this interface.

In order to keep synchronicity between the simulated world and the controller
of the digital twin, the pace at which simulation should progress is directly 
enforced. This is possible by controlling the simulator step-by-step. In our 
initial approach, Gazebo's Google Protobuf~\cite{protobuf} API was used, with a 
specific message to progress by a single simulation step.  By subscribing to 
status messages (which include the current simulation step) it is possible to 
determine when a step has completed and to ensure synchronization.  However as 
we attempted to increase the rate of advertising step messages, we discovered 
that the rate of status messages is capped at 5~Hz.
Such a limitation introduces a consistent bottleneck in the simulation/learning pipeline. 
Furthermore it was found that Gazebo silently drops messages it cannot process.

A set of important modifications were made to increase experiment
throughput.  The key idea was to allow motor update commands to
directly drive the simulation clock.  By default Gazebo comes
pre-installed with an ArduPilot Arducopter~\cite{ardupilot} plugin to
receive motor updates through a UDP server. These motor updates are in
the form of  pulse width modulation (PWM) signals. At the same time,
sensor readings from the inertial measurement unit (IMU) on board the
aircraft is sent over a second UDP channel.  Arducopter is an open
source multicopter firmware and its plugin was developed to support
SITL testing.

We derived our \gymfcGeneric aircraft plugin from the Arducopter plugin with the
following modifications (as well as those discussed in
Section~\ref{sec:comm}). Upon receiving a motor command, the motor
forces are updated as normal but then a simulation step is executed.
Sensor data is read and then sent back as a response to the client
over the same UDP channel.  In addition to the IMU sensor data we also
simulate sensor data obtained from the electronic speed
controller~(ESC). The ESC provides the angular velocities of each
rotor, which are relayed to the client too.   Implementing our GymFC 
Plugin with this approach successfully allowed us to work around
the limitations of the Google Protobuf API and increased step throughput
by over 200 times.

\subsection{Communication Layer}
\label{sec:comm}
The communication layer is positioned in between the digital twin and the 
agent-environment interface.  This layer manages the low-level communication channel to 
the aircraft and simulation control. The primary function of this layer is to 
export a high-level synchronized API to the higher layers for interacting with 
the digital twin which uses asynchronous communication protocols. This layer 
provides the commands \texttt{pwm\_write} and \texttt{reset} to the 
agent-environment interface layer.  

The function call \texttt{pwm\_write} takes as input a vector  of PWM values for 
each actuator, corresponding to the control input $u(t)$. These PWM values 
correspond to the same values that would be sent to an ESC on a physical UAV.    
The PWM values are translated to a normalized format expected by the GymFC 
Plugin, and then packed into a UDP packet for transmission to the \gymfcGeneric Plugin UDP 
server.  The communication layer blocks until a response is received from the 
\gymfcGeneric Plugin, forcing synchronized writes for the above layers.  The UDP 
reply is unpacked and returned in response.  

During the learning process the simulated environment must be reset at the 
beginning of each learning episode. Ideally one could use the \texttt{gz} 
command line utility included with the Gazebo installation which is lightweight 
and does not require additional dependencies. Unfortunately there is a known 
socket handle leak~\cite{gzbug} that causes Gazebo to crash if the command is 
issued more than the maximum number of open files allowed by the operating 
system.  Given we are running thousands episodes during training this was not an 
option for us.  Instead we opted to use the Google Protobuf interface so we 
did not have to deploy a patched version of the utility on our test servers. 
Because resets only occur at the beginning of a training session and are not in 
the critical processing loop, using the Google Protobuf API here is acceptable.

Upon start of the communication layer, a connection is established with the 
Google Protobuf API server and we subscribe to world statistics messages which 
includes the current simulation iteration. To reset the simulator, a world 
control message is advertised instructing the simulator to reset the simulation 
time. The communication layer blocks until it receives a world statistics 
message indicating the simulator has been reset and then returns back control to 
the agent-environment interface layer.  Note the world control message is only 
resetting the simulation time, not the entire simulator (\ie models and 
sensors). This is because we found that in some cases when a world control 
message was issued to perform a full reset the sensor data took a few additional 
iterations for reset. To ensure proper reset to the above layers this time reset 
message acts as a signalling mechanism to the \gymfcGeneric Plugin.  When the plugin 
detects a time reset has occurred it resets the whole simulator and most 
importantly steps the simulator until the sensor values have also reset ensuring 
above layers that when a new training session starts, reading sensor values 
accurately reflect the current state and not the previous state from stale 
values.

\subsection{Environment Interface Layer}

The topmost layer interfacing with the agent is the environment interface layer 
which  implements the OpenAI Gym~\cite{brockman2016openai} environment API.  
Each OpenAI Gym environment defines an observation space and an action space.  
These are used to inform the agent of the bounds to expect for environment 
observations and what are legal bounds for the action input, respectively.
As previously mentioned in Section~\ref{sec:gymfc:rl} \gym is in both the continuous 
observation space and action space domain. The state is of size $m\times(M+D)$ 
where $m$ is the memory size indicating the number of past observations; $M = 4$ 
as we consider a four-motor configuration; and $D = 3$ since each measurement 
is taken in the 3 dimensions.
Each observation value is in $[-\infty : \infty]$. The action space 
is of size $M$ equivalent to the number of control actuators of the aircraft  (\ie 
four for a quadcopter), where each value is normalized between $[-1:1]$ to be 
compatible with most agents who squash their output using the hyperbolic
tangent function.  

\gym implements two primary OpenAI functions, namely \texttt{reset} and 
\texttt{step}.  The \texttt{reset} function is called at the start of an episode 
to reset the environment and returns the initial environment state. This is 
also when the desired target angular velocity $\Omega^*$ or setpoint is 
computed.
The setpoint is randomly sampled from a uniform distribution between 
$[\Omega_{min}, \Omega_{max}]$. For the continuous task this is also set at a 
random interval of time. Selection of these bounds may refer to the desired 
operating region of the aircraft.  Although it is highly unlikely during normal 
operation that a quadcopter will be expected to reach the majority of these target 
angular velocities, the intention of these tasks are to push and stress the 
performance of the aircraft. 

The \texttt{step} function executes a single simulation step with the specified 
actions and returns to the agent the new state vector, together with a reward 
indicating how well the given action was performed.
Reward engineering can be challenging. If careful design is not performed, the derived 
policy may not reflect what was originally intended. Recall from 
Section~\ref{sec:gymfc:rl} that the reward is ultimately what shapes the policy. For 
this work, with the goal of establishing a baseline of accuracy, we develop a 
reward to reflect the current angular velocity error (\ie $e = \Omega^* - 
\Omega$).  In the future \gym will be expanded to include additional 
environments aiding in the development of more complex policies particularity to 
showcase the advantages of using RL to adapt and learn.
We translate the current error $e_t$ at time $t$ into into a derived reward  
$r_t$ normalized between $[-1, 0]$ as follows,
\begin{equation}
r_t = -clip\left( sum(|\Omega^*_t - \Omega_t| )/ 3 \Omega_{max} \right)
\label{eq:sumabserr}
\end{equation}
where the $sum$  function sums the absolute value of the error of each axis, and 
the $clip$  function clips the result between the $[0,1]$ in cases where there 
is an overflow in the error. Since the reward is negative, it signifies a 
penalty, the agent maximizes the rewards (and thus minimizing error) overtime in 
order to track the target as accurately as possible. Rewards are normalized to 
provide standardization and stabilization during training \cite{normalize}. 

Additionally we also experimented with a variety of other rewards. We found 
sparse binary rewards\footnote{A reward structured so that $r_t=0$ if 
$sum(|e_t|) < threshold$, otherwise $r_t = -1$.} 
to give poor performance. We believe this to be due to complexity of 
quadcopter control. In the early stages of learning the agent explores its 
environment. However the event of randomly reaching the target angular velocity 
within some threshold was rare and thus did not provide the agent with enough 
information to converge. Conversely, we found that signalling at each timestep 
was best.

\section{Evaluation}
\label{sec:gymfc:eval}
In this section we present our evaluation on the accuracy of studied
\nn-based attitude flight controllers trained with RL. 
To our knowledge, this is the first RL baseline
conducted for quadcopter attitude control.

\subsection{Setup}

We evaluate the RL algorithms DDPG, TRPO, and PPO using the implementations in 
the
OpenAI Baselines project~\cite{baselines}. The goal of the OpenAI Baselines project 
is to establish a reference implementation of RL algorithms, providing baselines for 
researchers to compare approaches and build upon. Every algorithm is run with 
defaults except for the number of simulations steps which we increased to 10~million. 
For reference the hyperparameters can be found in Table~\ref{tab:gymfc:ppo},
Table~\ref{tab:gymfc:trpo}, and Table~\ref{tab:gymfc:ddpg} for PPO, TRPO and
DDPG respectively. The PPO, TRPO \nn architectures have two hidden layers with
32 nodes each using hyperbolic tangent functions. The DDPG actor network  
has two hidden layers of 64 nodes using  rectified linear units, while the
output layer  uses hyperbolic tangent functions. The DDPG critic layer also has
the same internal structure however the output layer is unbounded.
\begin{table}
    \centering
\begin{tabular}{l|c}
Hyperparameter            & Value          \\ \hline
Horizon (T)               & 2048  \\
Adam stepsize             & $3 \times 10^{-4}\times \rho$  \\
Num. epochs               & 10    \\
Minibatch size            & 64  \\
Discount ($\gamma$)       & 0.99  \\
GAE parameter ($\lambda$) & 0.95   
\end{tabular}
\caption{\new{PPO hyperparameters where $\rho$ is linearly annealed over the
course of training from 1 to 0.}}
\label{tab:gymfc:ppo}
\end{table}

\begin{table}
    \centering
\begin{tabular}{l|c}
Hyperparameter            & Value          \\ \hline
Horizon & 1024 \\
Max KL-divergence & 0.01 \\
Value function learning rate             & $1 \times 10^{-3}$\\
Num. epochs & 5 \\
Discount ($\gamma$)       & 0.99  \\
GAE parameter ($\lambda$) & 0.98   
\end{tabular}
\caption{TRPO hyperparameters.}
\label{tab:gymfc:trpo}
\end{table}

\begin{table}
    \centering
\begin{tabular}{l|c}
Hyperparameter            & Value          \\ \hline
Num. epochs & 5000 \\
Num. epochs per cycle & 20\\
Num. rollout steps & 100 \\
Batch size & 64 \\
Noise type & adaptive-param\_0.2 \\
Actor learning rate & $1 \times 10^{-4}$ \\
Critic learning rate & $1 \times 10^{-3}$ \\
Discount ($\gamma$)       & 0.99  \\
\end{tabular}
\caption{DDPG hyperparameters.}
\label{tab:gymfc:ddpg}
\end{table}

The episodic task parameters were configured to run each episode for a maximum 
of 1~second of simulated time allowing enough time for the controller to 
respond to the command as well as additional time
to identify if a steady state has been reached.  The bounds the target angular 
 velocity is sampled from is set to $\Omega_{min}=-5.24$ rad/s, 
 $\Omega_{max}=5.24$ rad/s ($\pm$ 300 deg/s). These limits were constructed by 
 examining PID's performance to make sure we expressed physically feasible 
 constraints.  The max step size of the Gazebo simulator, which specifies the 
 duration of each physics update step was set to 1~ms to develop highly accurate 
 simulations. In other words, our physical world ``evolved'' at 1~kHz.
Training and evaluations were run on Ubuntu~16.04 with an eight-core 
i7-7700 CPU and an NVIDIA GeForce GT~730 graphics card.

For our PID controller, we ported the mixing and SITL implementation from 
Betaflight~\cite{betaflight} to Python to be compatible with \gym. 
The PID controller was first tuned using the classical Ziegler-Nichols 
method~\cite{ziegler1942optimum} and then manually adjusted to improve 
performance of the step response sampled around the midpoint $\pm\Omega_{max}/2$.
We obtained the following gains for each axis of rotation: 
$K_\phi = [2, 10, 0.005], K_\theta = [10, 10, 0.005], K_\psi = [4, 50, 0.0]$, where each vector 
contains to the $[K_P,K_I,K_D]$ (proportional, integrative, derivative) gains, respectively.
Next we measured the distances between the arms of the quadcopter to calculate 
the mixer values for each motor $m_i, i \in \{0, \ldots, 3\}$.  Each vector $m_i$ 
is of the form $m_i = [ m_{(i,\phi)}, m_{(i,\theta)}, m_{(i,\psi)}]$, \ie   
roll, pitch, and yaw (see Section~\ref{sec:bg:pid}). The final values were: 
$m_0 = [ -1.0,  0.598, -1.0 ]$, $m_1 = [ -0.927, -0.598,  1.0 ]$,  $m_2 = [ 1.0,  
0.598,  1.0 ]$ and lastly $m_3 = [ 0.927, -0.598, -1.0 ]$. The mix values and 
PID sums are then used to compute each motor signal $u_i$ according to 
Equation~\ref{eq:mix}, where $\mathbf{T}=0$ for no additional throttle.  

To evaluate and compare the accuracy of the different algorithms we used a 
set of metrics. First, we define ``initial error'' as the distance between the rest velocities
and the current setpoint. A notion of progress toward the setpoint from rest can then be expressed as
the percentage of the initial error that has been ``corrected''. Correcting 0\% of the initial
error means that no progress has been made; while 100\% indicates that the setpoint
has been reached. Each metric value is independently computed for each axis. We hereby list our metrics.
\textbf{Success} captures the
number of experiments (in percentage) in which the controller eventually settles in 
an band within 90\% an 110\% of the initial error, \ie $\pm10\%$ from the setpoint. 
\textbf{Failure} captures the average percent error relative to the initial error 
after $t=500~ms$, for those experiments that do not make it in the $\pm10\%$ error band. 
The latter metric quantifies the magnitude of unacceptable controller performance.
The delay in the measurement ($t>500~ms$) is to exclude the rise regime. The underlying
assumption is that a steady state is reached before $500~ms$. 
\textbf{Rise} is the average time 
in milliseconds it takes the controller to go from \thresholdriselow to 
\thresholdrisehigh of the initial error. \textbf{Peak} is the max achieved angular 
velocity represented as a percentage relative to the initial error.  
Values greater than 100\% indicate overshoot, while values less than 100\% represent 
undershoot.
\textbf{Error} is the mean sum of the absolute value error of each episode in 
radians per second. This provides a generic metric for performance. Our last 
metric is \textbf{Stability}, which captures how stable the response is halfway 
through the simulation, \ie at $t>500ms$. Stability is calculated by taking the 
linear regression of the angular velocities and reporting the slope of the 
calculated line. Systems that are unstable have a non-zero slope.

\begin{figure*}[htp]
	\centering
	\begin{subfigure}[b]{0.3\textwidth}
		{\includegraphics[width=1.75in]{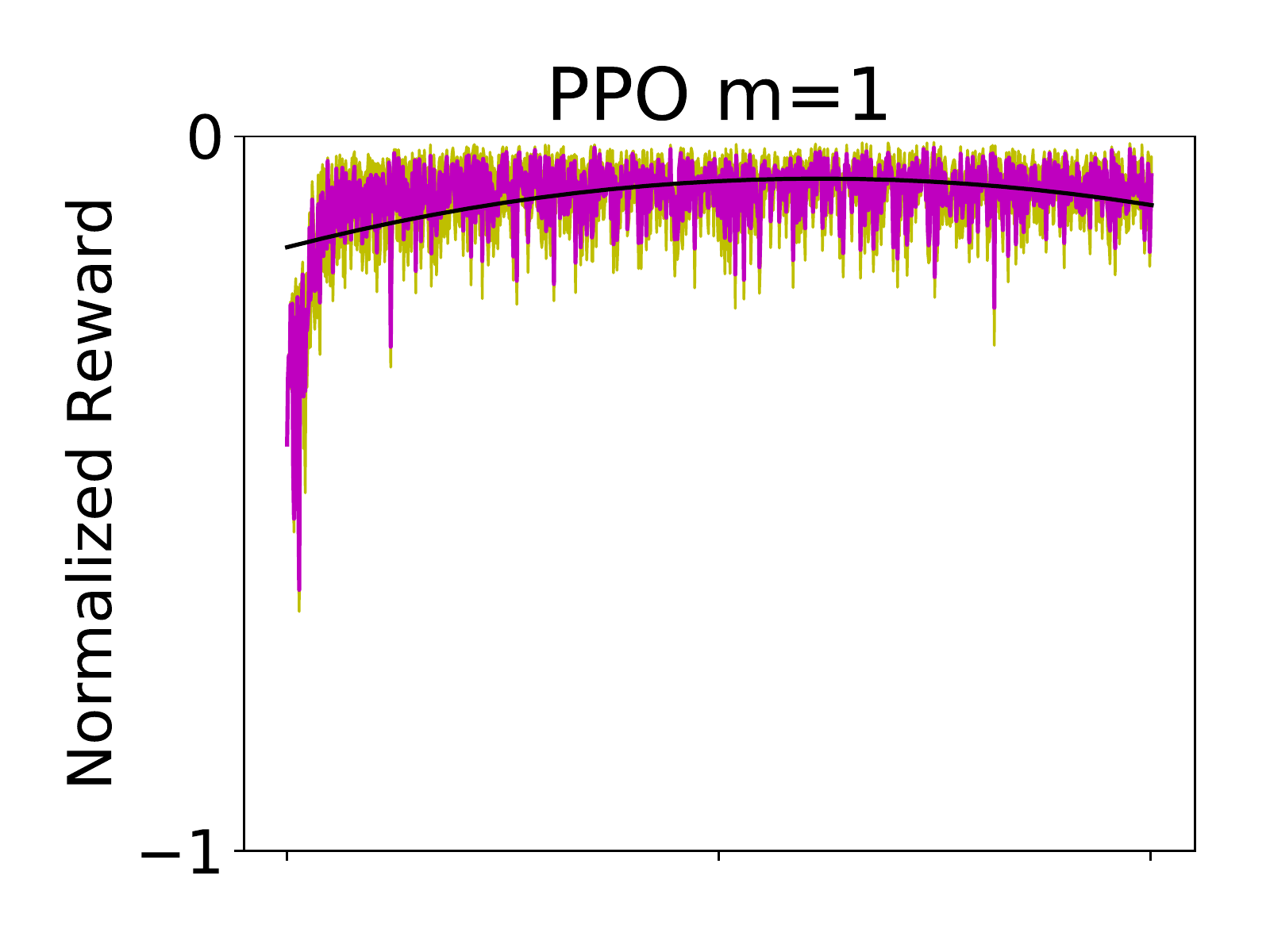}}
		%	\caption{PPO}
	\end{subfigure}
	\hfill
	\begin{subfigure}[b]{0.3\textwidth}
		{\includegraphics[width=1.75in]{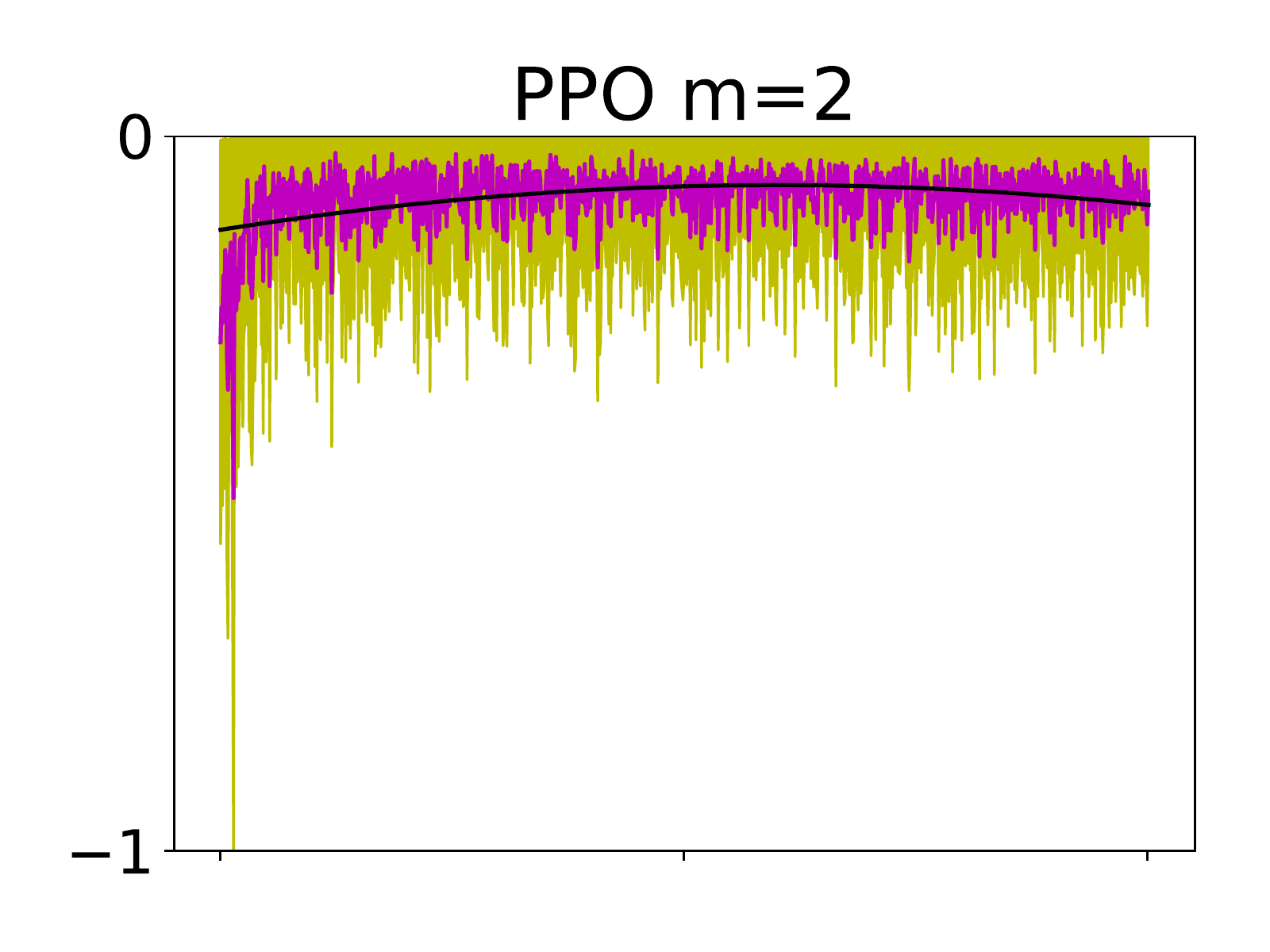}}
		%\caption{PPO}
	\end{subfigure}
	\hfill
	\begin{subfigure}[b]{0.3\textwidth}
		{\includegraphics[width=1.75in]{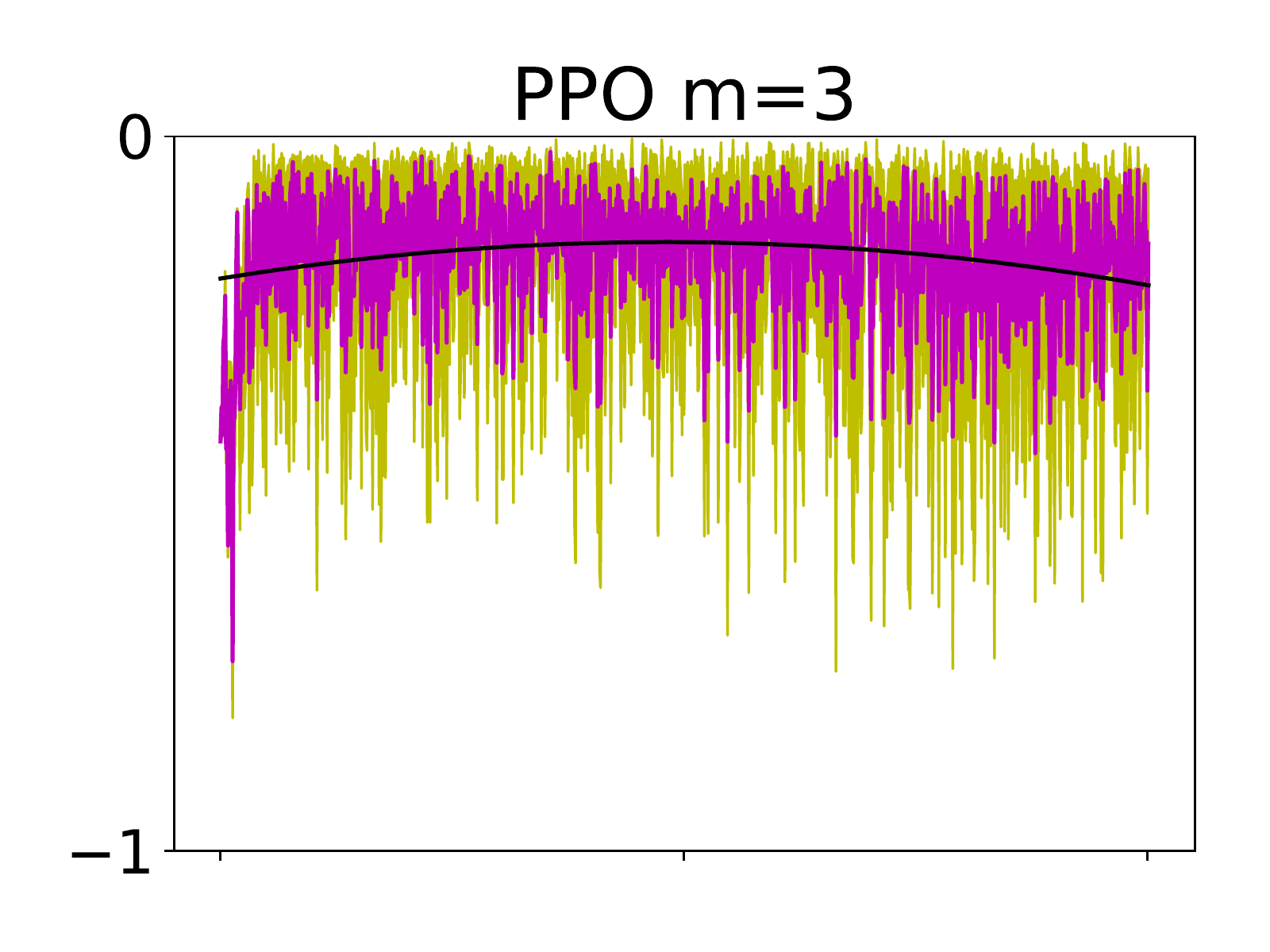}}
		%\caption{PPO}
	\end{subfigure}
	\\
	\begin{subfigure}[b]{0.3\textwidth}
		{\includegraphics[width=1.75in]{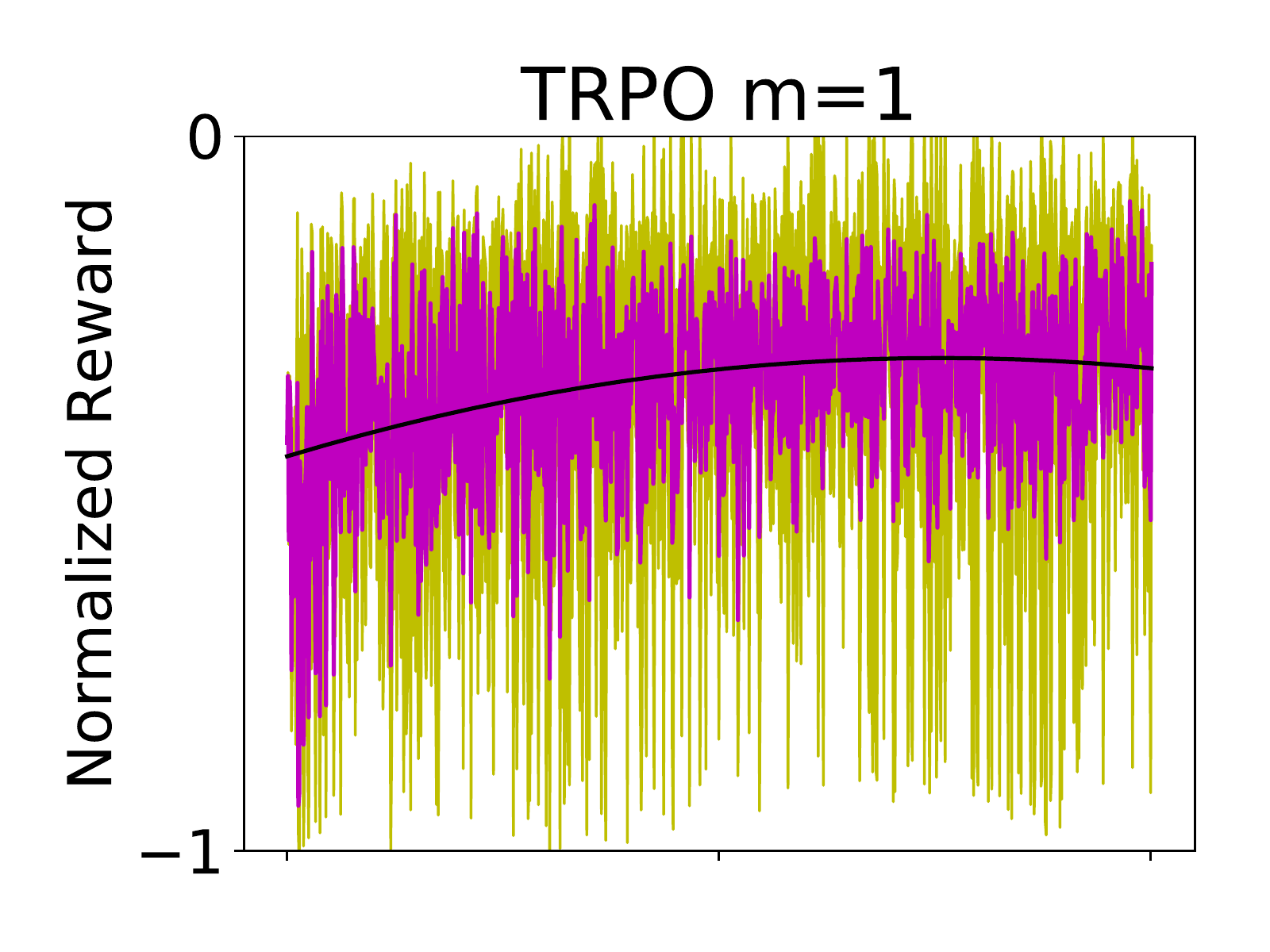}}
		%\caption{TRPO}
	\end{subfigure}
	\hfill
	\begin{subfigure}[b]{0.3\textwidth}
		{\includegraphics[width=1.75in]{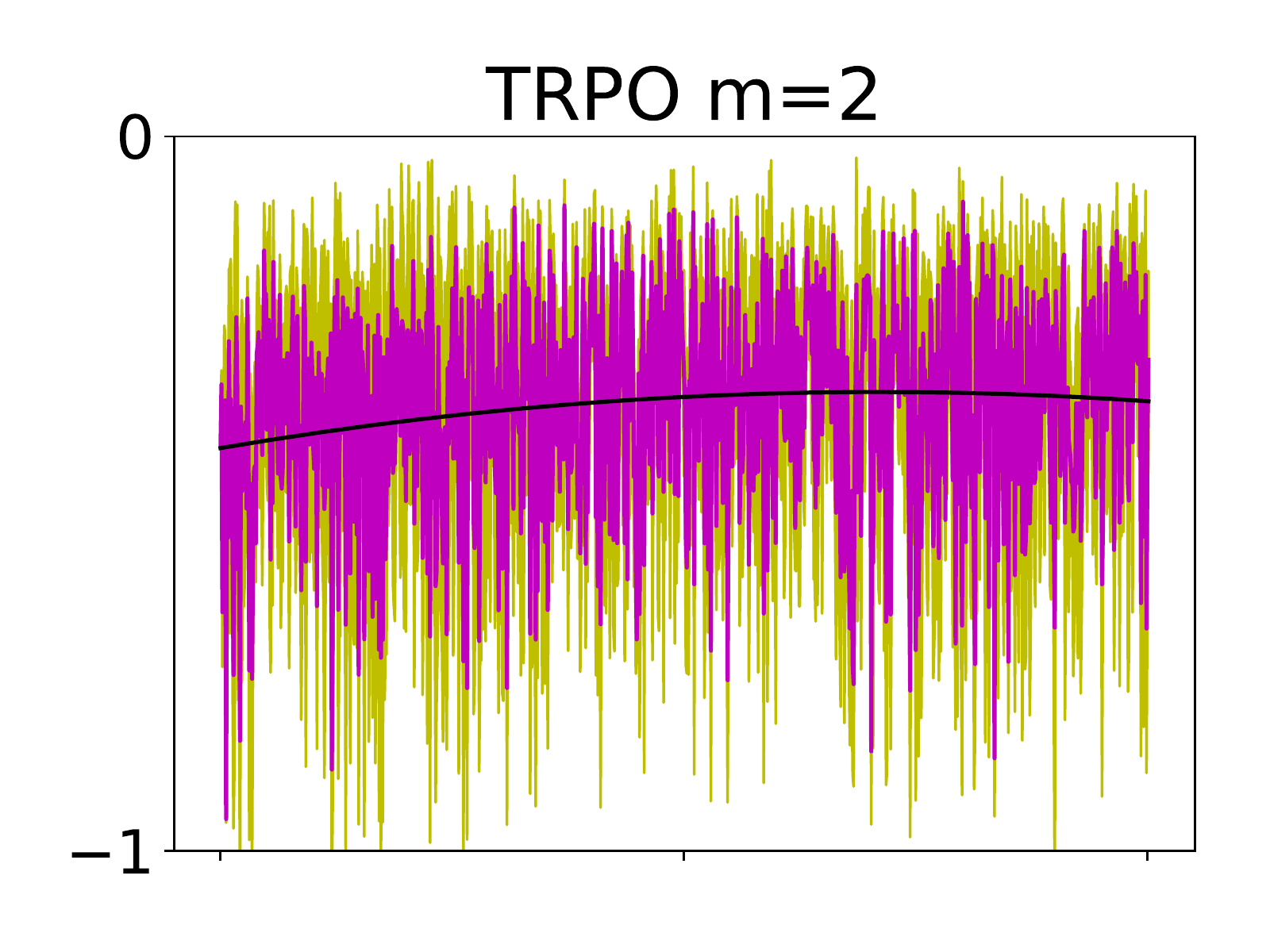}}
		%\caption{TRPO}
	\end{subfigure}
	\hfill
	\begin{subfigure}[b]{0.3\textwidth}
		{\includegraphics[width=1.75in]{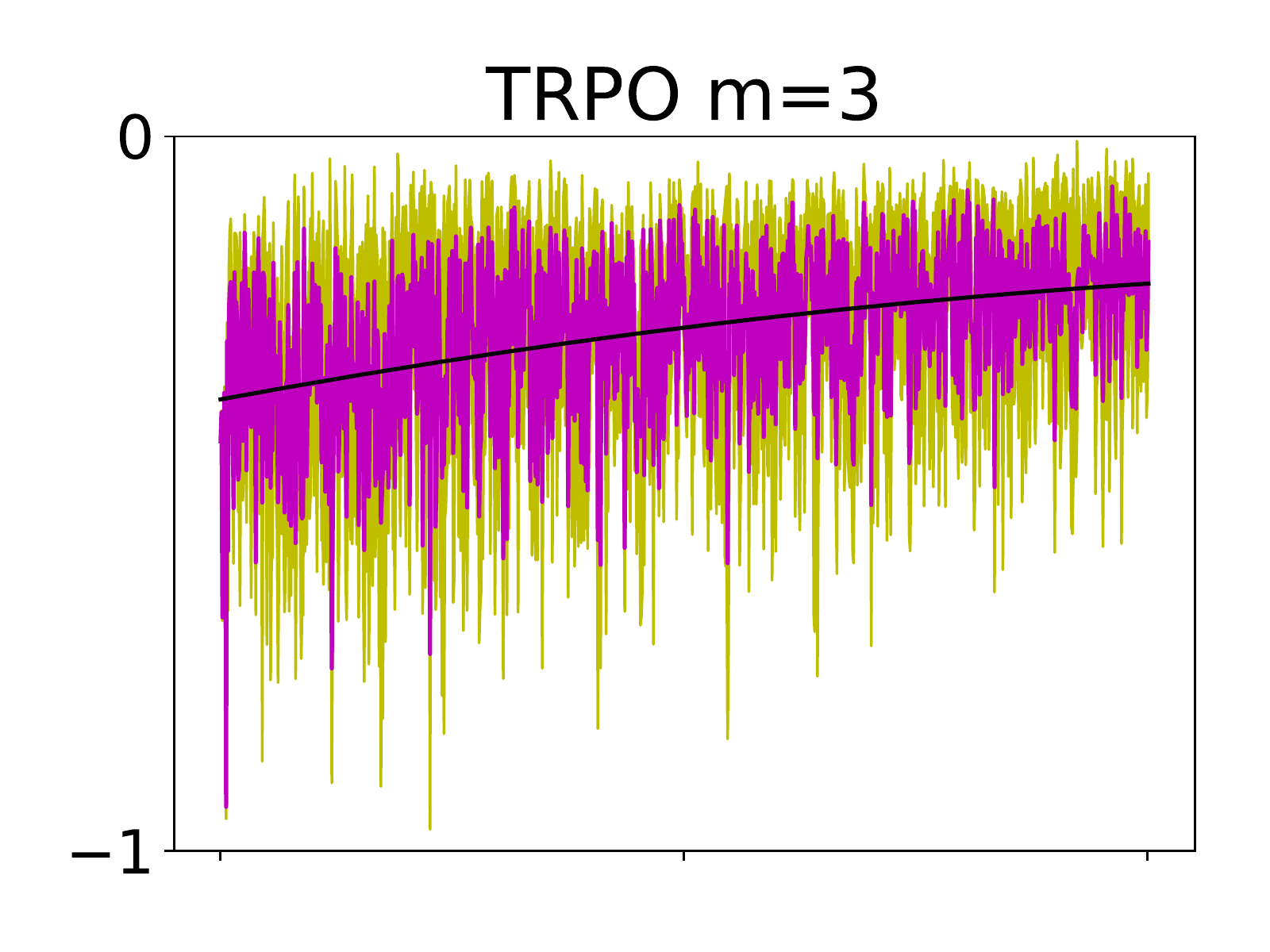}}
		%\caption{TRPO}
	\end{subfigure}
	\\
	\begin{subfigure}[b]{0.3\textwidth}
		{\includegraphics[width=1.75in]{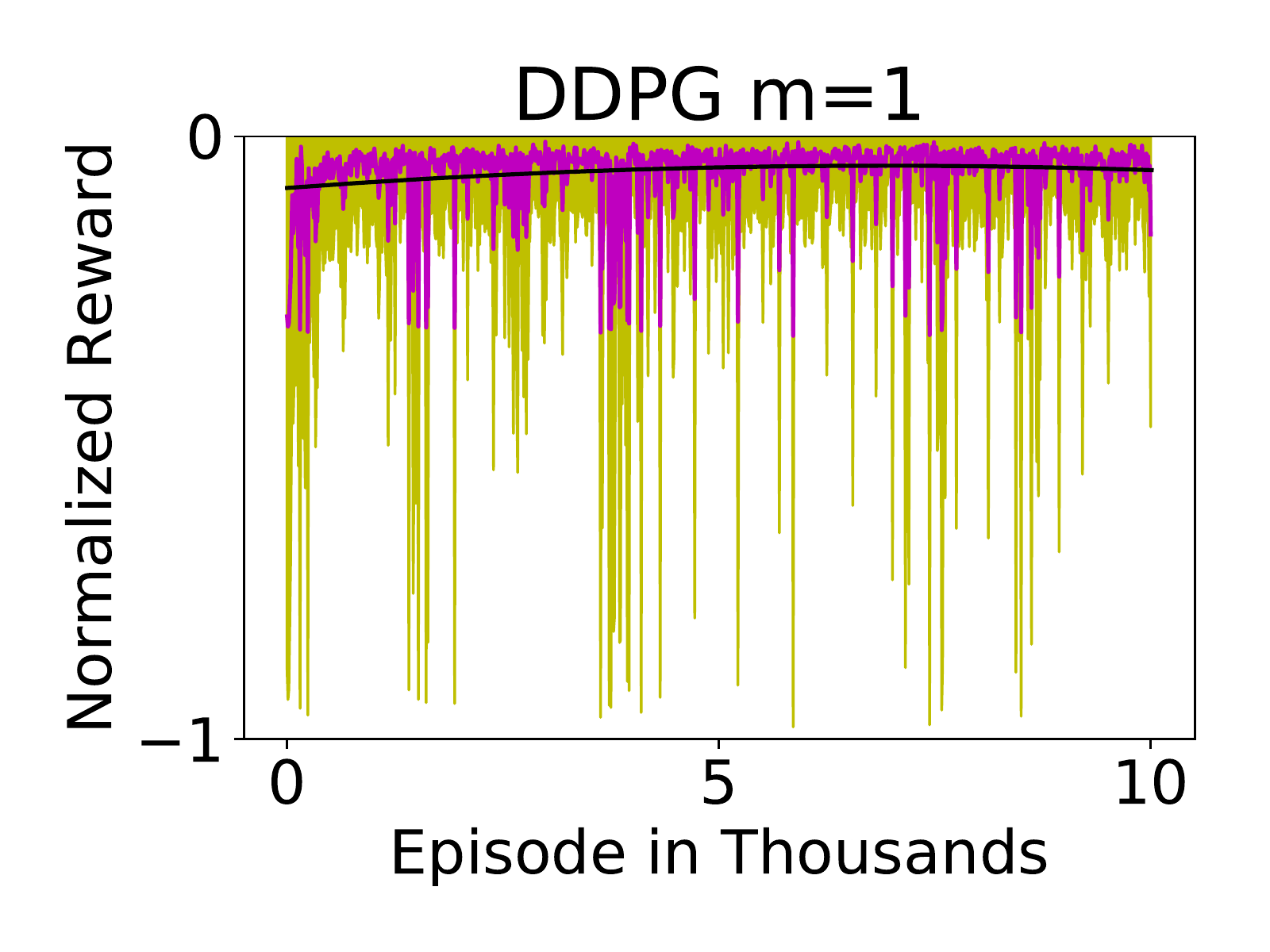}}
		%\caption{DDPG}
	\end{subfigure}
	\hfill
	\begin{subfigure}[b]{0.3\textwidth}
		{\includegraphics[width=1.75in]{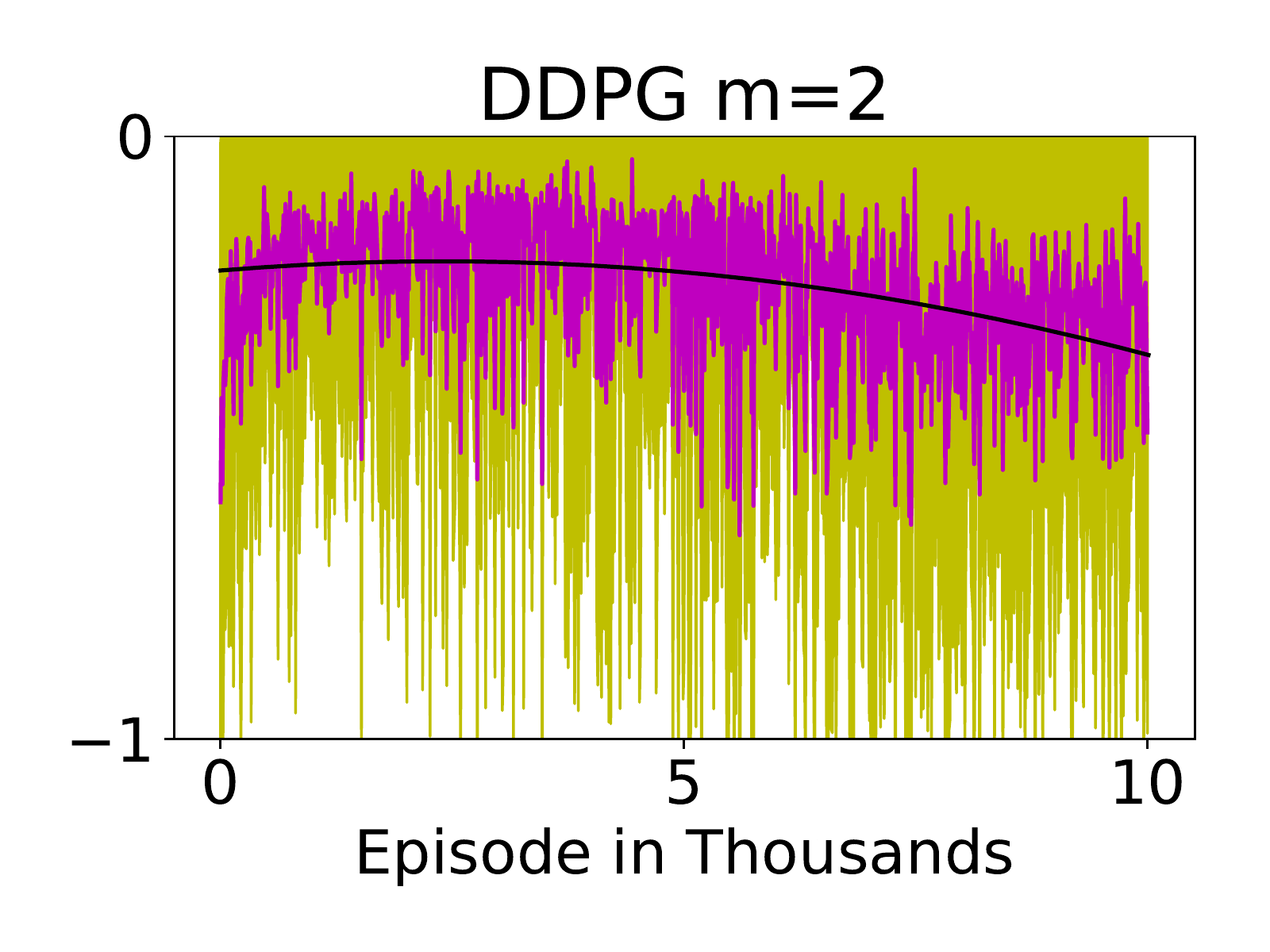}}
		%\caption{DDPG}
	\end{subfigure}
	\hfill
	\begin{subfigure}[b]{0.3\textwidth}
		%\missingfigure{DDPG m3}
		{\includegraphics[width=1.75in]{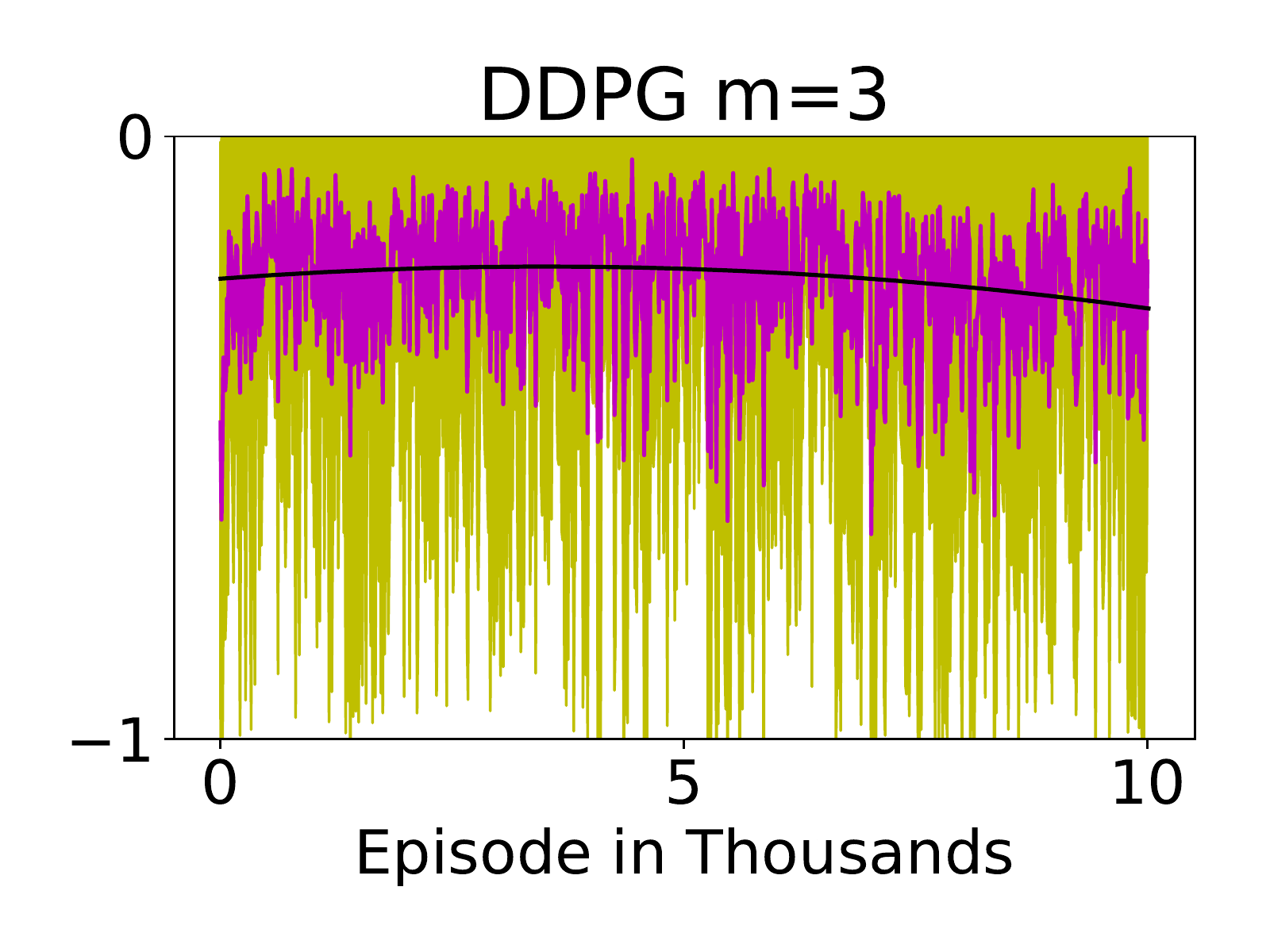}}
		%\caption{DDPG}
	\end{subfigure}
	\caption{Average normalized rewards \new{shown in magenta} received during 
	training of 10,000 episodes (10 million steps) for each RL algorithm and 
	memory $m$ sizes 1, 2 and 3. Plots share common $y$ and $x$ axis.  
	\new{Additionally, yellow} represents the \ci confidence interval \new{and 
		the black line is a two degree polynomial added to illustrate the trend 
		of the rewards over time.}  }
	\label{fig:training}
\end{figure*}

\renewcommand\ppoaa{99.9$\pm$0.1}
\renewcommand\ppoab{84.8$\pm$1.6}
\renewcommand\ppoac{100.0$\pm$0.0}
\renewcommand\ppoad{0.0$\pm$0.0}
\renewcommand\ppoae{3.2$\pm$0.4}
\renewcommand\ppoaf{0.0$\pm$0.0}
\renewcommand\ppoaa{65.9$\pm$2.4}
\renewcommand\ppoab{94.1$\pm$4.3}
\renewcommand\ppoac{73.4$\pm$2.7}
\renewcommand\ppoad{113.8$\pm$2.2}
\renewcommand\ppoae{107.7$\pm$2.2}
\renewcommand\ppoaf{128.1$\pm$4.3}
\renewcommand\ppoag{309.9$\pm$7.9}
\renewcommand\ppoah{440.6$\pm$13.4}
\renewcommand\ppoai{215.7$\pm$6.7}
\renewcommand\ppoaj{0.0$\pm$0.0}
\renewcommand\ppoak{0.0$\pm$0.0}
\renewcommand\ppoal{0.0$\pm$0.0}

\renewcommand\ppoba{100.0$\pm$0.1}
\renewcommand\ppobb{53.3$\pm$2.2}
\renewcommand\ppobc{99.2$\pm$0.4}
\renewcommand\ppobd{0.0$\pm$0.1}
\renewcommand\ppobe{11.1$\pm$0.7}
\renewcommand\ppobf{0.2$\pm$0.1}
\renewcommand\ppoba{58.6$\pm$2.5}
\renewcommand\ppobb{125.4$\pm$6.0}
\renewcommand\ppobc{105.0$\pm$5.0}
\renewcommand\ppobd{116.9$\pm$2.5}
\renewcommand\ppobe{103.0$\pm$2.7}
\renewcommand\ppobf{126.8$\pm$3.7}
\renewcommand\ppobg{305.2$\pm$7.9}
\renewcommand\ppobh{674.5$\pm$19.1}
\renewcommand\ppobi{261.3$\pm$7.6}
\renewcommand\ppobj{0.0$\pm$0.0}
\renewcommand\ppobk{0.0$\pm$0.0}
\renewcommand\ppobl{0.0$\pm$0.0}

\renewcommand\ppoca{88.6$\pm$1.4}
\renewcommand\ppocb{47.5$\pm$2.2}
\renewcommand\ppocc{99.3$\pm$0.4}
\renewcommand\ppocd{2.9$\pm$0.5}
\renewcommand\ppoce{45.6$\pm$3.8}
\renewcommand\ppocf{0.3$\pm$0.2}
\renewcommand\ppoca{101.5$\pm$5.0}
\renewcommand\ppocb{128.8$\pm$5.8}
\renewcommand\ppocc{79.2$\pm$3.3}
\renewcommand\ppocd{108.9$\pm$2.2}
\renewcommand\ppoce{94.2$\pm$5.3}
\renewcommand\ppocf{119.8$\pm$2.7}
\renewcommand\ppocg{405.9$\pm$10.9}
\renewcommand\ppoch{1403.8$\pm$58.4}
\renewcommand\ppoci{274.4$\pm$5.3}
\renewcommand\ppocj{0.0$\pm$0.0}
\renewcommand\ppock{0.0$\pm$0.0}
\renewcommand\ppocl{0.0$\pm$0.0}

\renewcommand\trpoaa{42.2$\pm$2.2}
\renewcommand\trpoab{54.1$\pm$2.2}
\renewcommand\trpoac{66.5$\pm$2.1}
\renewcommand\trpoad{48.5$\pm$5.5}
\renewcommand\trpoae{21.8$\pm$2.5}
\renewcommand\trpoaf{29.5$\pm$2.3}
\renewcommand\trpoaa{103.9$\pm$6.2}
\renewcommand\trpoab{150.2$\pm$6.7}
\renewcommand\trpoac{109.7$\pm$8.0}
\renewcommand\trpoad{125.1$\pm$9.3}
\renewcommand\trpoae{110.4$\pm$3.9}
\renewcommand\trpoaf{139.6$\pm$6.8}
\renewcommand\trpoag{1644.5$\pm$52.1}
\renewcommand\trpoah{929.0$\pm$25.6}
\renewcommand\trpoai{1374.3$\pm$51.5}
\renewcommand\trpoaj{-0.4$\pm$0.1}
\renewcommand\trpoak{-0.2$\pm$0.0}
\renewcommand\trpoal{-0.1$\pm$0.0}

\renewcommand\trpoba{43.4$\pm$2.2}
\renewcommand\trpobb{46.4$\pm$2.2}
\renewcommand\trpobc{39.8$\pm$2.1}
\renewcommand\trpobd{44.2$\pm$3.0}
\renewcommand\trpobe{109.3$\pm$12.6}
\renewcommand\trpobf{55.9$\pm$4.4}
\renewcommand\trpoba{161.3$\pm$6.9}
\renewcommand\trpobb{162.7$\pm$7.0}
\renewcommand\trpobc{108.4$\pm$9.6}
\renewcommand\trpobd{100.1$\pm$5.1}
\renewcommand\trpobe{144.2$\pm$13.8}
\renewcommand\trpobf{101.7$\pm$5.4}
\renewcommand\trpobg{1432.9$\pm$47.5}
\renewcommand\trpobh{2375.6$\pm$84.0}
\renewcommand\trpobi{1475.6$\pm$46.4}
\renewcommand\trpobj{0.1$\pm$0.0}
\renewcommand\trpobk{0.4$\pm$0.0}
\renewcommand\trpobl{-0.1$\pm$0.0}

\renewcommand\trpoca{68.0$\pm$2.0}
\renewcommand\trpocb{59.4$\pm$2.2}
\renewcommand\trpocc{82.6$\pm$1.7}
\renewcommand\trpocd{19.1$\pm$1.5}
\renewcommand\trpoce{30.4$\pm$5.0}
\renewcommand\trpocf{12.8$\pm$1.8}
\renewcommand\trpoca{130.4$\pm$7.1}
\renewcommand\trpocb{150.8$\pm$7.8}
\renewcommand\trpocc{129.1$\pm$8.9}
\renewcommand\trpocd{141.3$\pm$7.2}
\renewcommand\trpoce{141.2$\pm$8.1}
\renewcommand\trpocf{147.1$\pm$6.8}
\renewcommand\trpocg{1120.1$\pm$36.4}
\renewcommand\trpoch{1200.7$\pm$34.3}
\renewcommand\trpoci{824.0$\pm$30.1}
\renewcommand\trpocj{0.1$\pm$0.0}
\renewcommand\trpock{-0.1$\pm$0.1}
\renewcommand\trpocl{-0.1$\pm$0.0}

\renewcommand\ddpgaa{59.0$\pm$2.2}
\renewcommand\ddpgab{50.5$\pm$2.2}
\renewcommand\ddpgac{73.9$\pm$1.9}
\renewcommand\ddpgad{29.1$\pm$2.2}
\renewcommand\ddpgae{41.7$\pm$3.6}
\renewcommand\ddpgaf{22.4$\pm$2.4}
\renewcommand\ddpgaa{68.2$\pm$3.7}
\renewcommand\ddpgab{100.0$\pm$5.4}
\renewcommand\ddpgac{79.0$\pm$5.4}
\renewcommand\ddpgad{133.1$\pm$7.8}
\renewcommand\ddpgae{116.6$\pm$7.9}
\renewcommand\ddpgaf{146.4$\pm$7.5}
\renewcommand\ddpgag{1201.4$\pm$42.4}
\renewcommand\ddpgah{1397.0$\pm$62.4}
\renewcommand\ddpgai{992.9$\pm$45.1}
\renewcommand\ddpgaj{0.0$\pm$0.0}
\renewcommand\ddpgak{-0.1$\pm$0.0}
\renewcommand\ddpgal{0.1$\pm$0.0}

\renewcommand\ddpgba{49.2$\pm$1.5}
\renewcommand\ddpgbb{99.1$\pm$4.9}
\renewcommand\ddpgbc{40.7$\pm$1.8}
\renewcommand\ddpgbd{42.0$\pm$5.5}
\renewcommand\ddpgbe{46.7$\pm$8.0}
\renewcommand\ddpgbf{71.4$\pm$7.0}
\renewcommand\ddpgbg{2388.0$\pm$63.9}
\renewcommand\ddpgbh{2607.5$\pm$72.2}
\renewcommand\ddpgbi{1953.4$\pm$58.3}
\renewcommand\ddpgbj{-0.1$\pm$0.0}
\renewcommand\ddpgbk{-0.1$\pm$0.0}
\renewcommand\ddpgbl{-0.0$\pm$0.0}

\renewcommand\ddpgca{85.3$\pm$5.9}
\renewcommand\ddpgcb{124.3$\pm$7.2}
\renewcommand\ddpgcc{105.1$\pm$8.6}
\renewcommand\ddpgcd{101.0$\pm$8.2}
\renewcommand\ddpgce{158.6$\pm$21.0}
\renewcommand\ddpgcf{120.5$\pm$7.0}
\renewcommand\ddpgcg{1984.3$\pm$59.3}
\renewcommand\ddpgch{3280.8$\pm$98.7}
\renewcommand\ddpgci{1364.2$\pm$54.9}
\renewcommand\ddpgcj{0.0$\pm$0.1}
\renewcommand\ddpgck{0.2$\pm$0.1}
\renewcommand\ddpgcl{0.0$\pm$0.0}

\begin{table}[]
\centering
\caption{Rise time averages from \totalcommands command inputs 
	per configuration with \ci confidence.  }
\label{table:averise}
{\setlength{\tabcolsep}{0.2em}
\def\arraystretch{1.15}%
\begin{tabular}{lc|c|c|c|}
\cline{3-5}
\multicolumn{1}{c}{}                        &     & \multicolumn{3}{c|}{Rise (ms)} \\ \cline{2-5} 
\multicolumn{1}{c|}{}                       & $m$ & $\phi$   & $\theta$  & $\psi$  \\ \hline
\multicolumn{1}{|l|}{\multirow{3}{*}{PPO}}  & 1   & \ppoaa   & \ppoab    & \ppoac  \\ \cline{2-5} 
\multicolumn{1}{|l|}{}                      & 2   & \ppoba   & \ppobb    & \ppobc  \\ \cline{2-5} 
\multicolumn{1}{|l|}{}                      & 3   & \ppoca   & \ppocb    & \ppocc  \\ \hline
\multicolumn{1}{|l|}{\multirow{3}{*}{TRPO}} & 1   & \trpoaa  & \trpoab   & \trpoac \\ \cline{2-5} 
\multicolumn{1}{|l|}{}                      & 2   & \trpoba  & \trpobb   & \trpobc \\ \cline{2-5} 
\multicolumn{1}{|l|}{}                      & 3   & \trpoca  & \trpocb   & \trpocc \\ \hline
\multicolumn{1}{|l|}{\multirow{3}{*}{DDPG}} & 1   & \ddpgaa  & \ddpgab   & \ddpgac \\ \cline{2-5} 
\multicolumn{1}{|l|}{}                      & 2   & \ddpgba  & \ddpgbb   & \ddpgbc \\ \cline{2-5} 
\multicolumn{1}{|l|}{}                      & 3   & \ddpgca  & \ddpgcb   & \ddpgcc \\ \hline
\end{tabular}
}
\end{table}

\begin{table}[]
\centering
\caption{Peak averages from \totalcommands command inputs 
	per configuration with \ci confidence.  }
\label{table:avepeak}
{\setlength{\tabcolsep}{0.2em}
\def\arraystretch{1.15}%
\begin{tabular}{lc|c|c|c|}
\cline{3-5}
\multicolumn{1}{c}{}                        &     & \multicolumn{3}{c|}{Peak (\%)} \\ \cline{2-5} 
\multicolumn{1}{c|}{}                       & $m$ & $\phi$   & $\theta$  & $\psi$  \\ \hline
\multicolumn{1}{|l|}{\multirow{3}{*}{PPO}}  & 1   & \ppoad   & \ppoae    & \ppoaf  \\ \cline{2-5} 
\multicolumn{1}{|l|}{}                      & 2   & \ppobd   & \ppobe    & \ppobf  \\ \cline{2-5} 
\multicolumn{1}{|l|}{}                      & 3   & \ppocd   & \ppoce    & \ppocf  \\ \hline
\multicolumn{1}{|l|}{\multirow{3}{*}{TRPO}} & 1   & \trpoad  & \trpoae   & \trpoaf \\ \cline{2-5} 
\multicolumn{1}{|l|}{}                      & 2   & \trpobd  & \trpobe   & \trpobf \\ \cline{2-5} 
\multicolumn{1}{|l|}{}                      & 3   & \trpocd  & \trpoce   & \trpocf \\ \hline
\multicolumn{1}{|l|}{\multirow{3}{*}{DDPG}} & 1   & \ddpgad  & \ddpgae   & \ddpgaf \\ \cline{2-5} 
\multicolumn{1}{|l|}{}                      & 2   & \ddpgbd  & \ddpgbe   & \ddpgbf \\ \cline{2-5} 
\multicolumn{1}{|l|}{}                      & 3   & \ddpgcd  & \ddpgce   & \ddpgcf \\ \hline
\end{tabular}
}
\end{table}

\begin{table}[]
\centering
\caption{Error averages from \totalcommands command inputs 
	per configuration with \ci confidence.  }
\label{table:aveerror}
{\setlength{\tabcolsep}{0.2em}
\def\arraystretch{1.15}%
\begin{tabular}{lc|c|c|c|}
\cline{3-5}
\multicolumn{1}{c}{}                     &     & \multicolumn{3}{c|}{Error (rad/s)} \\ \cline{2-5} 
\multicolumn{1}{c|}{}                    & $m$ & $\phi$    & $\theta$   & $\psi$    \\ \hline
\multicolumn{1}{|l|}{\multirow{3}{*}{PPO}} & 1   & \ppoag    & \ppoah     & \ppoai    \\ \cline{2-5} 
\multicolumn{1}{|l|}{}                   & 2   & \ppobg    & \ppobh     & \ppobi    \\ \cline{2-5} 
\multicolumn{1}{|l|}{}                   & 3   & \ppocg    & \ppoch     & \ppoci    \\ \hline
\multicolumn{1}{|l|}{\multirow{3}{*}{TRPO}} & 1   & \trpoag   & \trpoah    & \trpoai   \\ \cline{2-5} 
\multicolumn{1}{|l|}{}                   & 2   & \trpobg   & \trpobh    & \trpobi   \\ \cline{2-5} 
\multicolumn{1}{|l|}{}                   & 3   & \trpocg   & \trpoch    & \trpoci   \\ \hline
\multicolumn{1}{|l|}{\multirow{3}{*}{DDPG}} & 1   & \ddpgag   & \ddpgah    & \ddpgai   \\ \cline{2-5} 
\multicolumn{1}{|l|}{}                   & 2   & \ddpgbg   & \ddpgbh    & \ddpgbi   \\ \cline{2-5} 
\multicolumn{1}{|l|}{}                   & 3   & \ddpgcg   & \ddpgch    & \ddpgci   \\ \hline
\end{tabular}
}
\end{table}

\begin{table}[]
\centering
\caption{Stability averages from \totalcommands command inputs 
	per configuration with \ci confidence.  }
\label{table:avestable}
{\setlength{\tabcolsep}{0.2em}
\def\arraystretch{1.15}%
\begin{tabular}{lc|c|c|c|}
\cline{3-5}
\multicolumn{1}{c}{}                        &     & \multicolumn{3}{c|}{Stability} \\ \cline{2-5} 
\multicolumn{1}{c|}{}                       & $m$ & $\phi$   & $\theta$  & $\psi$  \\ \hline
\multicolumn{1}{|l|}{\multirow{3}{*}{PPO}}  & 1   & \ppoaj   & \ppoak    & \ppoal  \\ \cline{2-5} 
\multicolumn{1}{|l|}{}                      & 2   & \ppobj   & \ppobk    & \ppobl  \\ \cline{2-5} 
\multicolumn{1}{|l|}{}                      & 3   & \ppocj   & \ppock    & \ppocl  \\ \hline
\multicolumn{1}{|l|}{\multirow{3}{*}{TRPO}} & 1   & \trpoaj  & \trpoak   & \trpoal \\ \cline{2-5} 
\multicolumn{1}{|l|}{}                      & 2   & \trpobj  & \trpobk   & \trpobl \\ \cline{2-5} 
\multicolumn{1}{|l|}{}                      & 3   & \trpocj  & \trpock   & \trpocl \\ \hline
\multicolumn{1}{|l|}{\multirow{3}{*}{DDPG}} & 1   & \ddpgaj  & \ddpgak   & \ddpgal \\ \cline{2-5} 
\multicolumn{1}{|l|}{}                      & 2   & \ddpgbj  & \ddpgbk   & \ddpgbl \\ \cline{2-5} 
\multicolumn{1}{|l|}{}                      & 3   & \ddpgcj  & \ddpgck   & \ddpgcl \\ \hline
\end{tabular}
}
\end{table}

\renewcommand\ppoaa{99.8$\pm$0.3}
\renewcommand\ppoab{100.0$\pm$0.0}
\renewcommand\ppoac{100.0$\pm$0.0}
\renewcommand\ppoad{0.1$\pm$0.1}
\renewcommand\ppoae{0.0$\pm$0.0}
\renewcommand\ppoaf{0.0$\pm$0.0}

\renewcommand\ppoba{100.0$\pm$0.0}
\renewcommand\ppobb{53.3$\pm$3.1}
\renewcommand\ppobc{99.8$\pm$0.3}
\renewcommand\ppobd{0.0$\pm$0.0}
\renewcommand\ppobe{20.0$\pm$2.4}
\renewcommand\ppobf{0.0$\pm$0.0}

\renewcommand\ppoca{98.7$\pm$0.7}
\renewcommand\ppocb{74.7$\pm$2.7}
\renewcommand\ppocc{99.3$\pm$0.5}
\renewcommand\ppocd{0.4$\pm$0.2}
\renewcommand\ppoce{5.4$\pm$0.7}
\renewcommand\ppocf{0.2$\pm$0.2}

\renewcommand\trpoaa{32.8$\pm$2.9}
\renewcommand\trpoab{59.0$\pm$3.0}
\renewcommand\trpoac{87.4$\pm$2.1}
\renewcommand\trpoad{72.5$\pm$10.6}
\renewcommand\trpoae{17.4$\pm$3.7}
\renewcommand\trpoaf{9.4$\pm$2.6}

\renewcommand\trpoba{19.7$\pm$2.5}
\renewcommand\trpobb{48.2$\pm$3.1}
\renewcommand\trpobc{56.9$\pm$3.1}
\renewcommand\trpobd{76.6$\pm$5.0}
\renewcommand\trpobe{43.0$\pm$6.5}
\renewcommand\trpobf{38.6$\pm$7.0}

\renewcommand\trpoca{96.8$\pm$1.1}
\renewcommand\trpocb{60.8$\pm$3.0}
\renewcommand\trpocc{73.2$\pm$2.7}
\renewcommand\trpocd{1.5$\pm$0.8}
\renewcommand\trpoce{20.6$\pm$4.1}
\renewcommand\trpocf{20.6$\pm$3.4}

\renewcommand\ddpgaa{84.1$\pm$2.3}
\renewcommand\ddpgab{52.5$\pm$3.1}
\renewcommand\ddpgac{90.4$\pm$1.8}
\renewcommand\ddpgad{11.1$\pm$2.2}
\renewcommand\ddpgae{41.1$\pm$5.5}
\renewcommand\ddpgaf{4.6$\pm$1.0}

\renewcommand\ddpgba{26.6$\pm$2.7}
\renewcommand\ddpgbb{26.1$\pm$2.7}
\renewcommand\ddpgbc{50.2$\pm$3.1}
\renewcommand\ddpgbd{82.7$\pm$8.5}
\renewcommand\ddpgbe{112.2$\pm$12.9}
\renewcommand\ddpgbf{59.7$\pm$7.5}

\renewcommand\ddpgca{39.2$\pm$3.0}
\renewcommand\ddpgcb{44.8$\pm$3.1}
\renewcommand\ddpgcc{60.7$\pm$3.0}
\renewcommand\ddpgcd{52.0$\pm$6.4}
\renewcommand\ddpgce{101.8$\pm$13.0}
\renewcommand\ddpgcf{33.9$\pm$3.4}

\renewcommand\pida{100.0$\pm$0.0}
\renewcommand\pidb{100.0$\pm$0.0}
\renewcommand\pidc{100.0$\pm$0.0}
\renewcommand\pidd{0.0$\pm$0.0}
\renewcommand\pide{0.0$\pm$0.0}
\renewcommand\pidf{0.0$\pm$0.0}

\begin{table}[]
	\footnotesize
\centering
\caption{Success and Failure results for considered algorithms. The row 
	highlighted in blue refers to our best-performing learning agent PPO, while the rows 
         highlighted in yellow correspond to the best agents for the other two algorithms. }
\label{table:fails}
{\setlength{\tabcolsep}{0.2em}

\def\arraystretch{1.15}%
\begin{tabular}{lc|c|c|c|c|c|c|}
\cline{3-8}
\multicolumn{1}{c}{}                        &     & \multicolumn{3}{c|}{Success 
(\%)} & \multicolumn{3}{c|}{Failure (\%)} \\ \cline{2-8} \multicolumn{1}{c|}{}                       
& $m$ & $\phi$    & $\theta$   & $\psi$   & $\phi$     & $\theta$    & $\psi$    
\\ \hline \rowcolor{blue!30}
\multicolumn{1}{|l|}{\cellcolor{white}\multirow{3}{*}{PPO}}  & 1   & \ppoaa    & 
\ppoab     & \ppoac   & \ppoad     & \ppoae      & \ppoaf    \\ \cline{2-8} 
\multicolumn{1}{|l|}{}                      & 2   & \ppoba    & \ppobb     & 
\ppobc   & \ppobd     & \ppobe      & \ppobf    \\ \cline{2-8} 
\multicolumn{1}{|l|}{}                      & 3   & \ppoca    & \ppocb     & 
\ppocc   & \ppocd     & \ppoce      & \ppocf    \\ \hline
\multicolumn{1}{|l|}{\cellcolor{white}\multirow{3}{*}{TRPO}} & 1   & \trpoaa   & 
\trpoab    & \trpoac  & \trpoad    & \trpoae     & \trpoaf   \\ \cline{2-8} 
\multicolumn{1}{|l|}{}                      & 2   & \trpoba   & \trpobb    & 
\trpobc  & \trpobd    & \trpobe     & \trpobf   \\ \cline{2-8} 
\rowcolor{yellow!30}
\multicolumn{1}{|l|}{\cellcolor{white}}                      & 3   & \trpoca   & 
\trpocb    & \trpocc  & \trpocd    & \trpoce     & \trpocf   \\ \hline
\rowcolor{yellow!30}
\multicolumn{1}{|l|}{\cellcolor{white}\multirow{3}{*}{DDPG}} & 1   & \ddpgaa   & 
\ddpgab    & \ddpgac  & \ddpgad    & \ddpgae     & \ddpgaf   \\ \cline{2-8} 
\multicolumn{1}{|l|}{}                      & 2   & \ddpgba   & \ddpgbb    & 
\ddpgbc  & \ddpgbd    & \ddpgbe     & \ddpgbf   \\ \cline{2-8} 
\multicolumn{1}{|l|}{}                      & 3   & \ddpgca   & \ddpgcb    & 
\ddpgcc  & \ddpgcd    & \ddpgce     & \ddpgcf   \\ \hline
\multicolumn{2}{|l|}{PID}                         & \pida     & \pidb      & \pidc    & \pidd      & \pide       & \pidf     \\ \hline
\end{tabular}}
\end{table}

\renewcommand\ppoaa{66.6$\pm$3.2}
\renewcommand\ppoab{70.8$\pm$3.6}
\renewcommand\ppoac{72.9$\pm$3.7}
\renewcommand\ppoad{112.6$\pm$3.0}
\renewcommand\ppoae{109.4$\pm$2.4}
\renewcommand\ppoaf{127.0$\pm$6.2}
\renewcommand\ppoag{317.0$\pm$11.0}
\renewcommand\ppoah{326.3$\pm$13.2}
\renewcommand\ppoai{217.5$\pm$9.1}
\renewcommand\ppoaj{0.0$\pm$0.0}
\renewcommand\ppoak{0.0$\pm$0.0}
\renewcommand\ppoal{0.0$\pm$0.0}

\renewcommand\ppoba{64.4$\pm$3.6}
\renewcommand\ppobb{102.8$\pm$6.7}
\renewcommand\ppobc{148.2$\pm$7.9}
\renewcommand\ppobd{118.4$\pm$4.3}
\renewcommand\ppobe{104.2$\pm$4.7}
\renewcommand\ppobf{124.2$\pm$3.4}
\renewcommand\ppobg{329.4$\pm$12.3}
\renewcommand\ppobh{815.3$\pm$31.4}
\renewcommand\ppobi{320.6$\pm$11.5}
\renewcommand\ppobj{0.0$\pm$0.0}
\renewcommand\ppobk{0.0$\pm$0.0}
\renewcommand\ppobl{0.0$\pm$0.0}

\renewcommand\ppoca{97.9$\pm$5.5}
\renewcommand\ppocb{121.9$\pm$7.2}
\renewcommand\ppocc{79.5$\pm$3.7}
\renewcommand\ppocd{111.4$\pm$3.4}
\renewcommand\ppoce{111.1$\pm$4.2}
\renewcommand\ppocf{120.8$\pm$4.2}
\renewcommand\ppocg{396.7$\pm$14.7}
\renewcommand\ppoch{540.6$\pm$22.6}
\renewcommand\ppoci{237.1$\pm$8.0}
\renewcommand\ppocj{0.0$\pm$0.0}
\renewcommand\ppock{0.0$\pm$0.0}
\renewcommand\ppocl{0.0$\pm$0.0}

\renewcommand\trpoaa{119.9$\pm$8.8}
\renewcommand\trpoab{149.0$\pm$10.6}
\renewcommand\trpoac{103.9$\pm$9.8}
\renewcommand\trpoad{103.0$\pm$11.0}
\renewcommand\trpoae{117.4$\pm$5.8}
\renewcommand\trpoaf{142.8$\pm$6.5}
\renewcommand\trpoag{1965.2$\pm$90.5}
\renewcommand\trpoah{930.5$\pm$38.4}
\renewcommand\trpoai{713.7$\pm$34.4}
\renewcommand\trpoaj{0.7$\pm$0.1}
\renewcommand\trpoak{0.3$\pm$0.0}
\renewcommand\trpoal{0.0$\pm$0.0}

\renewcommand\trpoba{108.0$\pm$8.3}
\renewcommand\trpobb{157.1$\pm$9.9}
\renewcommand\trpobc{47.3$\pm$6.5}
\renewcommand\trpobd{69.4$\pm$7.4}
\renewcommand\trpobe{117.7$\pm$9.2}
\renewcommand\trpobf{126.5$\pm$7.2}
\renewcommand\trpobg{2020.2$\pm$71.9}
\renewcommand\trpobh{1316.2$\pm$49.0}
\renewcommand\trpobi{964.0$\pm$31.2}
\renewcommand\trpobj{0.1$\pm$0.1}
\renewcommand\trpobk{0.5$\pm$0.1}
\renewcommand\trpobl{0.0$\pm$0.0}

\renewcommand\trpoca{115.2$\pm$9.5}
\renewcommand\trpocb{156.6$\pm$12.7}
\renewcommand\trpocc{176.1$\pm$15.5}
\renewcommand\trpocd{153.5$\pm$8.1}
\renewcommand\trpoce{123.3$\pm$6.9}
\renewcommand\trpocf{148.8$\pm$11.2}
\renewcommand\trpocg{643.5$\pm$20.5}
\renewcommand\trpoch{895.0$\pm$42.8}
\renewcommand\trpoci{1108.9$\pm$44.5}
\renewcommand\trpocj{0.1$\pm$0.0}
\renewcommand\trpock{0.0$\pm$0.0}
\renewcommand\trpocl{0.0$\pm$0.0}

\renewcommand\ddpgaa{64.7$\pm$5.2}
\renewcommand\ddpgab{118.9$\pm$8.5}
\renewcommand\ddpgac{51.0$\pm$4.8}
\renewcommand\ddpgad{165.6$\pm$11.6}
\renewcommand\ddpgae{135.4$\pm$12.8}
\renewcommand\ddpgaf{150.8$\pm$6.2}
\renewcommand\ddpgag{929.1$\pm$39.9}
\renewcommand\ddpgah{1490.3$\pm$83.0}
\renewcommand\ddpgai{485.3$\pm$25.4}
\renewcommand\ddpgaj{0.1$\pm$0.1}
\renewcommand\ddpgak{-0.2$\pm$0.1}
\renewcommand\ddpgal{0.1$\pm$0.0}

\renewcommand\ddpgba{nan$\pm$0.0}
\renewcommand\ddpgbb{nan$\pm$0.0}
\renewcommand\ddpgbc{nan$\pm$0.0}
\renewcommand\ddpgbd{0.0$\pm$0.0}
\renewcommand\ddpgbe{0.0$\pm$0.0}
\renewcommand\ddpgbf{0.0$\pm$0.0}
\renewcommand\ddpgbg{2701.9$\pm$90.1}
\renewcommand\ddpgbh{2716.2$\pm$93.2}
\renewcommand\ddpgbi{2569.9$\pm$90.4}
\renewcommand\ddpgbj{-0.0$\pm$0.0}
\renewcommand\ddpgbk{0.0$\pm$0.0}
\renewcommand\ddpgbl{-0.0$\pm$0.0}

\renewcommand\ddpgba{49.2$\pm$2.1}
\renewcommand\ddpgbb{99.1$\pm$6.9}
\renewcommand\ddpgbc{40.7$\pm$2.5}
\renewcommand\ddpgbd{84.0$\pm$10.4}
\renewcommand\ddpgbe{93.5$\pm$15.4}
\renewcommand\ddpgbf{142.7$\pm$12.5}
\renewcommand\ddpgbg{2074.1$\pm$86.4}
\renewcommand\ddpgbh{2498.8$\pm$109.8}
\renewcommand\ddpgbi{1336.9$\pm$50.1}
\renewcommand\ddpgbj{-0.1$\pm$0.0}
\renewcommand\ddpgbk{-0.2$\pm$0.1}
\renewcommand\ddpgbl{-0.0$\pm$0.0}

\renewcommand\ddpgca{73.7$\pm$8.4}
\renewcommand\ddpgcb{172.9$\pm$12.0}
\renewcommand\ddpgcc{141.5$\pm$14.5}
\renewcommand\ddpgcd{103.7$\pm$11.5}
\renewcommand\ddpgce{126.5$\pm$17.8}
\renewcommand\ddpgcf{119.6$\pm$8.2}
\renewcommand\ddpgcg{1585.4$\pm$81.4}
\renewcommand\ddpgch{2401.3$\pm$109.8}
\renewcommand\ddpgci{1199.0$\pm$74.0}
\renewcommand\ddpgcj{-0.1$\pm$0.1}
\renewcommand\ddpgck{-0.2$\pm$0.1}
\renewcommand\ddpgcl{0.1$\pm$0.0}

\renewcommand\pida{79.0$\pm$3.5}
\renewcommand\pidb{99.8$\pm$5.0}
\renewcommand\pidc{67.7$\pm$2.3}
\renewcommand\pidd{136.9$\pm$4.8}
\renewcommand\pide{112.7$\pm$1.6}
\renewcommand\pidf{135.1$\pm$3.3}
\renewcommand\pidg{416.1$\pm$20.4}
\renewcommand\pidh{269.6$\pm$11.9}
\renewcommand\pidi{245.1$\pm$11.5}
\renewcommand\pidj{0.0$\pm$0.0}
\renewcommand\pidk{0.0$\pm$0.0}
\renewcommand\pidl{0.0$\pm$0.0}

\begin{table}[]
\centering
\caption{RL rise time evaluation compared to PID of best-performing agent.  
	Values reported are the average of 1,000 command inputs with \ci confidence.  PPO $m=1$ highlighted in blue outperforms 
	all other agents, including PID control.  Metrics highlighted in red for PID 
	control are outpreformed by the PPO agent.}
\label{table:besttime}
{\setlength{\tabcolsep}{0.2em}
\def\arraystretch{1.15}%
\begin{tabular}{lc|c|c|c|}
\cline{3-5}
\multicolumn{1}{c}{}                        &     & \multicolumn{3}{c|}{Rise (ms)} \\ \cline{2-5} 
\multicolumn{1}{c|}{}                       & $m$ & $\phi$   & $\theta$  & $\psi$  \\ \hline
\rowcolor{blue!30}
\multicolumn{1}{|l|}{\multirow{3}{*}{\cellcolor{white}PPO}}  & 1   & \ppoaa   & \ppoab    & \ppoac  \\ \cline{2-5} 
\multicolumn{1}{|l|}{}                      & 2   & \ppoba   & \ppobb    & \ppobc  \\ \cline{2-5} 
\multicolumn{1}{|l|}{}                      & 3   & \ppoca   & \ppocb    & \ppocc  \\ \hline
\multicolumn{1}{|l|}{\multirow{3}{*}{TRPO}} & 1   & \trpoaa  & \trpoab   & \trpoac \\ \cline{2-5} 
\multicolumn{1}{|l|}{}                      & 2   & \trpoba  & \trpobb   & \trpobc \\ \cline{2-5} 
\multicolumn{1}{|l|}{}                      & 3   & \trpoca  & \trpocb   & \trpocc \\ \hline
\multicolumn{1}{|l|}{\multirow{3}{*}{DDPG}} & 1   & \ddpgaa  & \ddpgab   & \ddpgac \\ \cline{2-5} 
\multicolumn{1}{|l|}{}                      & 2   & \ddpgba  & \ddpgbb   & \ddpgbc \\ \cline{2-5} 
\multicolumn{1}{|l|}{}                      & 3   & \ddpgca  & \ddpgcb   & \ddpgcc \\ \hline
\multicolumn{2}{|l|}{PID}                         & \cellcolor{red!30}\pida & \cellcolor{red!30} \pidb     & \pidc   \\ \hline
\end{tabular}
}
\end{table}

\begin{table}[]
\centering
\caption{RL peak angular velocity percentage evaluation compared to PID of best-performing agent.  
	Values reported are the average of 1,000 command inputs with \ci confidence.  PPO $m=1$ highlighted in blue outperforms 
	all other agents, including PID control.  Metrics highlighted in red for PID 
	control are outpreformed by the PPO agent.}
\label{table:bestpeak}
{\setlength{\tabcolsep}{0.2em}
\def\arraystretch{1.15}%
\begin{tabular}{lc|c|c|c|}
\cline{3-5}
\multicolumn{1}{c}{}                        &     & \multicolumn{3}{c|}{Peak (\%)} \\ \cline{2-5} 
\multicolumn{1}{c|}{}                       & $m$ & $\phi$   & $\theta$  & $\psi$  \\ \hline
\rowcolor{blue!30}
\multicolumn{1}{|l|}{\multirow{3}{*}{\cellcolor{white}PPO}}  & 1   & \ppoad   & \ppoae    & \ppoaf  \\ \cline{2-5} 
\multicolumn{1}{|l|}{}                      & 2   & \ppobd   & \ppobe    & \ppobf  \\ \cline{2-5} 
\multicolumn{1}{|l|}{}                      & 3   & \ppocd   & \ppoce    & \ppocf  \\ \hline
\multicolumn{1}{|l|}{\multirow{3}{*}{TRPO}} & 1   & \trpoad  & \trpoae   & \trpoaf \\ \cline{2-5} 
\multicolumn{1}{|l|}{}                      & 2   & \trpobd  & \trpobe   & \trpobf \\ \cline{2-5} 
\multicolumn{1}{|l|}{}                      & 3   & \trpocd  & \trpoce   & \trpocf \\ \hline
\multicolumn{1}{|l|}{\multirow{3}{*}{DDPG}} & 1   & \ddpgad  & \ddpgae   & \ddpgaf \\ \cline{2-5} 
\multicolumn{1}{|l|}{}                      & 2   & \ddpgbd  & \ddpgbe   & \ddpgbf \\ \cline{2-5} 
\multicolumn{1}{|l|}{}                      & 3   & \ddpgcd  & \ddpgce   & \ddpgcf \\ \hline
\multicolumn{2}{|l|}{PID}                         & \cellcolor{red!30}\pidd    
&\cellcolor{red!30} \pide     &\cellcolor{red!30} \pidf    \\ \hline
\end{tabular}
}
\end{table}

\begin{table}[]
\centering
\caption{RL error evaluation compared to PID of best-performing agent.  
	Values reported are the average of 1,000 command inputs with \ci confidence.  PPO $m=1$ highlighted in blue outperforms 
	all other agents, including PID control.  Metrics highlighted in red for PID 
	control are outpreformed by the PPO agent.}
\label{table:besterror}
{\setlength{\tabcolsep}{0.2em}
\def\arraystretch{1.15}%
\begin{tabular}{lc|c|c|c|}
\cline{3-5}
\multicolumn{1}{c}{}                     &     & \multicolumn{3}{c|}{Error (rad/s)} \\ \cline{2-5} 
\multicolumn{1}{c|}{}                    & $m$ & $\phi$    & $\theta$   & $\psi$    \\ \hline
\rowcolor{blue!30}
\multicolumn{1}{|l|}{\multirow{3}{*}{\cellcolor{white}PPO}} & 1   & \ppoag    & \ppoah     & \ppoai    \\ \cline{2-5} 
\multicolumn{1}{|l|}{}                   & 2   & \ppobg    & \ppobh     & \ppobi    \\ \cline{2-5} 
\multicolumn{1}{|l|}{}                   & 3   & \ppocg    & \ppoch     & \ppoci    \\ \hline
\multicolumn{1}{|l|}{\multirow{3}{*}{TRPO}} & 1   & \trpoag   & \trpoah    & \trpoai   \\ \cline{2-5} 
\multicolumn{1}{|l|}{}                   & 2   & \trpobg   & \trpobh    & \trpobi   \\ \cline{2-5} 
\multicolumn{1}{|l|}{}                   & 3   & \trpocg   & \trpoch    & \trpoci   \\ \hline
\multicolumn{1}{|l|}{\multirow{3}{*}{DDPG}} & 1   & \ddpgag   & \ddpgah    & \ddpgai   \\ \cline{2-5} 
\multicolumn{1}{|l|}{}                   & 2   & \ddpgbg   & \ddpgbh    & \ddpgbi   \\ \cline{2-5} 
\multicolumn{1}{|l|}{}                   & 3   & \ddpgcg   & \ddpgch    & \ddpgci   \\ \hline
\multicolumn{2}{|l|}{PID}                &\cellcolor{red!30} \pidg     & \pidh      &\cellcolor{red!30} \pidi  \\ \hline
\end{tabular}
}
\end{table}

\begin{table}[]
\centering
\caption{RL stability evaluation compared to PID of best-performing agent.  
	Values reported are the average of 1,000 command inputs with \ci confidence.  PPO $m=1$ highlighted in blue outperforms 
	all other agents, including PID control.  Metrics highlighted in red for PID 
	control are outpreformed by the PPO agent.}
\label{table:beststable}
{\setlength{\tabcolsep}{0.2em}
\def\arraystretch{1.15}%
\begin{tabular}{lc|c|c|c|}
\cline{3-5}
\multicolumn{1}{c}{}                        &     & \multicolumn{3}{c|}{Stability} \\ \cline{2-5} 
\multicolumn{1}{c|}{}                       & $m$ & $\phi$   & $\theta$  & $\psi$  \\ \hline
\rowcolor{blue!30}
\multicolumn{1}{|l|}{\multirow{3}{*}{\cellcolor{white}PPO}}  & 1   & \ppoaj   & \ppoak    & \ppoal  \\ \cline{2-5} 
\multicolumn{1}{|l|}{}                      & 2   & \ppobj   & \ppobk    & \ppobl  \\ \cline{2-5} 
\multicolumn{1}{|l|}{}                      & 3   & \ppocj   & \ppock    & \ppocl  \\ \hline
\multicolumn{1}{|l|}{\multirow{3}{*}{TRPO}} & 1   & \trpoaj  & \trpoak   & \trpoal \\ \cline{2-5} 
\multicolumn{1}{|l|}{}                      & 2   & \trpobj  & \trpobk   & \trpobl \\ \cline{2-5} 
\multicolumn{1}{|l|}{}                      & 3   & \trpocj  & \trpock   & \trpocl \\ \hline
\multicolumn{1}{|l|}{\multirow{3}{*}{DDPG}} & 1   & \ddpgaj  & \ddpgak   & \ddpgal \\ \cline{2-5} 
\multicolumn{1}{|l|}{}                      & 2   & \ddpgbj  & \ddpgbk   & \ddpgbl \\ \cline{2-5} 
\multicolumn{1}{|l|}{}                      & 3   & \ddpgcj  & \ddpgck   & \ddpgcl \\ \hline
\multicolumn{2}{|l|}{PID}                         & \pidj    & \pidk     & \pidl   \\ \hline
\end{tabular}
}
\end{table}

\subsection{Results}
Each learning agent was trained with an RL algorithm for a total of 10 million 
simulation steps, equivalent to 10,000 episodes or about 2.7 simulation hours.  
The agents configuration is defined as the RL algorithm used for training and 
its memory size $m$.
Training for DDPG took approximately~\rtddpg, while PPO and TRPO took 
approximately~\rtppo and \rttrpo respectively. The average sum of rewards for 
each episode is  normalized between $[-1,0]$ and displayed in 
Figure~\ref{fig:training}. This computed average \new{in magenta} is from 
\trials independently trained agents with the same configuration, while the 95\% 
confidence is shown in \new{yellow. Additionally we have added a two degree 
	polynomial in black fit to the data to illustrate the reward trend over 
	time.}  Training results show clearly that PPO converges consistently 
compared to TRPO and DDPG, and overall PPO accumulates higher rewards.  What is 
also interesting and counter-intuitive is that the larger memory size actually 
\emph{decreases} convergence and stability among all trained algorithms.  
Recall from Section~\ref{sec:bg:rl} that RL algorithms learn a policy to map 
states to action.  A reason for the decrease in convergence could be attributed 
to the state space increasing causing the RL algorithm to take longer to learn 
the mapping to the optimal action.  As part of our future work, we plan to 
investigate using separate memory sizes for the error and rotor velocity to 
decrease the state space. Additionally increasing the size of the \nn could
compensate for the increase in state space.
Reward gains during training of TRPO and DDPG are 
quite inconsistent with large confidence intervals. \new{Although performance 
	for DDPG $m=1$ looks promising, upon further investigation into the large 
	confidence interval we found this was due to the algorithm completely 
	failing to respond to certain command inputs thus questioning whether the 
	algorithm has learned the underlying flight dynamics (this is emphasized 
	later in Table~\ref{table:fails})}.

\new{In the future we plan to investigate methods to decrease training times by 
	addressing challenges \textbf{C2} and \textbf{C3}. Specific 
	to \textbf{C2} to support a large range of aircraft, we will explore whether 
	we can construct a generic \nn taught general flight 
	dynamics~(Section~\ref{sec:bg:dynamics}) which will provide a baseline to 
	extend training to create intelligent controllers unique to an aircraft   
	(otherwise known as domain adaptation~\cite{blitzer2008learning}).  
	Additionally considering \textbf{C3} we will experiment with developing more 
	expressive reward functions to decrease training times.  }

Each trained agent was then evaluated on 1,000 never before seen command inputs 
in an episodic task. Since there are \trials agents per configuration, each 
configuration was evaluated over a total of \totalcommands episodes. The average 
performance metrics are reported in Table~\ref{table:averise} for Rise, Table~\ref{table:avepeak} for Peak, Table~\ref{table:aveerror} for Error and Table~\ref{table:avestable} for Stability.
Results show that the agent trained with PPO outperforms TRPO and DDPG in every 
measurement.  In fact, PPO is the only one that is able to achieve stability 
(for every $m$), while all other agents have at least one axis where the Stability 
metric is non-zero. 

Next the best performing agent for each algorithm and memory size is compared  
to the PID controller. The best agent was selected based on the lowest sum of 
errors of all three axis reported by the Error metric. The Success and Failure 
metrics are compared in Table~\ref{table:fails}.
Results show that agents trained with PPO would be the only ones good enough for 
flight, with a success rate close to perfect, and where the roll failure of 
$0.2\%$ is only off by about $0.1\%$ from the setpoint. However the best 
trained agents for TRPO and DDPG are often significantly far away from the 
desired angular velocity. For example TRPO's best agent, $39.2\%$~(60.8\% 
success, see Table~\ref{table:fails}) of the time does not reach the desired 
pitch target with upwards of a $20\%$ error from the setpoint.

Next we provide our thorough analysis comparing the best agents  
in Table~\ref{table:besttime} for Rise, Table~\ref{table:bestpeak} for Peak, 
Table~\ref{table:besterror} for Error and Table~\ref{table:beststable} for Stability. 
We have found that RL agents trained with PPO using 
$m=1$ provide performance and accuracy exceeding that of our PID controller in 
regards to rise time, peak velocities achieved, and total error. What is 
interesting is that usually a fast rise time could cause overshoot however the 
PPO agent has on average a faster rise time and less overshoot. 
This is most likely explained by the faster switching and oscillations causes
in the PWM control signal output of the PPO controller, allowing it to
compensate quicker than PID control. However if transferred to the real world,
the addition of these oscillations could be problematic. 
Both PPO and PID 
reach a stable state measured halfway through the simulation.

\begin{figure*}
	\includegraphics[width=\textwidth]{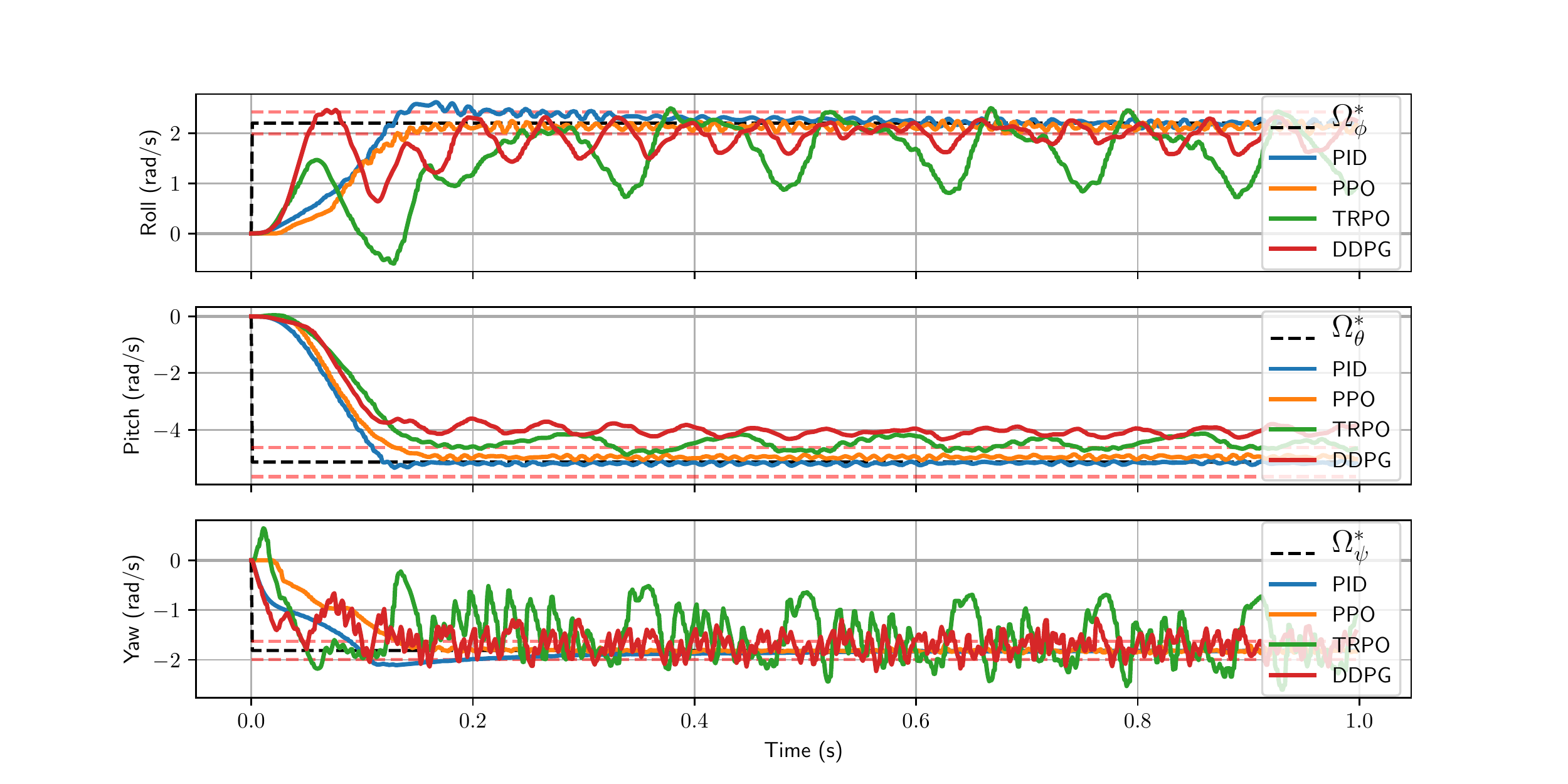}
	\caption{Step response of best trained RL agents compared to
          PID. Target angular velocity is $\Omega^*=[2.20, -5.14,
            -1.81]$ rad/s shown by dashed black line.  Error bars
          $\pm$\thresholdband of initial error from $\Omega^*$ are
          shown in dashed red.}
	\label{fig:stepcompare}
\end{figure*}

To illustrate the performance of each of the best agents a random simulation is 
sampled and the step response for each attitude command is displayed in 
Figure~\ref{fig:stepcompare} along with the target angular velocity to achieve 
$\Omega^*$. All algorithms reach some steady state however only PPO and PID do 
so within the error band indicated by the dashed red lines.  TRPO and DDPG have 
extreme oscillations in both the roll and yaw axis, which would cause 
instability during flight. \new{In this particular example we can observe PID to 
	perform better with a 19\% decrease in error compared to PPO, most visibly 
	in yaw control.	However globally speaking, in terms of error, PPO has shown 
	to be a more accurate attitude controller.}

\begin{figure*}
	\includegraphics[width=\textwidth]{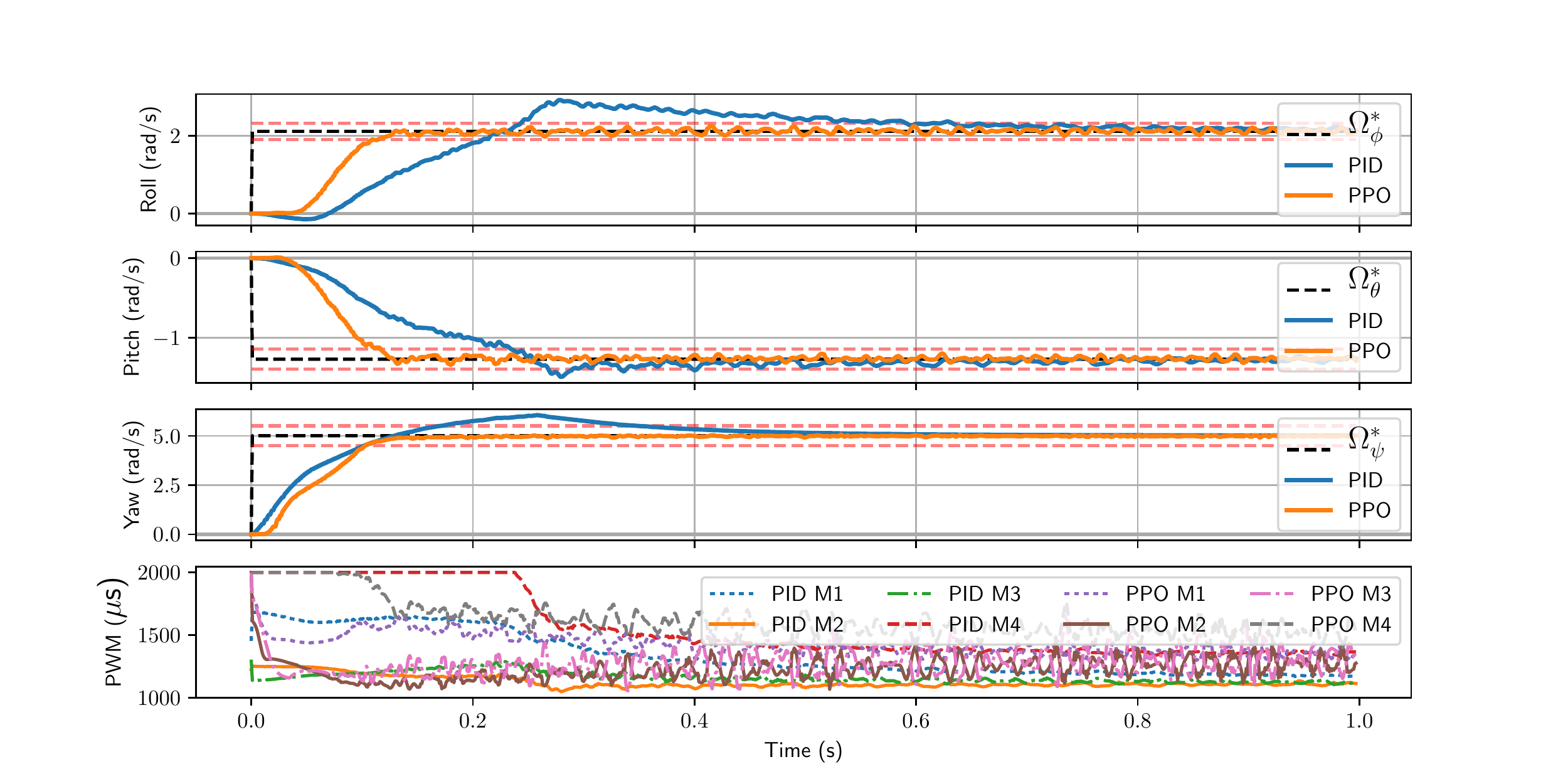}
	\caption{Step response and PWM motor signals \new{in microseconds ($\mu$s)} 
		of the best trained PPO agent compared to PID.  Target angular velocity 
		is  $\Omega^*=[2.11, -1.26, 5.00]$ rad/s shown by dashed black line.  
		Error bars $\pm$\thresholdband of initial error from $\Omega^*$ are 
		shown in dashed red.}
	\label{fig:bestcompare}
\end{figure*}

To highlight the performance and accuracy of the PPO agent we sample another 
simulation and show the step response and also the PWM control signals generated 
by each controller in Figure~\ref{fig:bestcompare}. In this figure we can see 
the PPO agent has exceptional tracking capabilities of the desired attitude.  
\new{Compared to PID, the PPO controller has a 44\% decrease in error.} The PPO 
agent has a  2.25 times faster rise time on the roll axis, 2.5 times faster on 
the pitch axis and 1.15 time faster on the yaw axis.
Furthermore the PID controller experiences slight overshoot in both the roll and 
yaw axis while the PPO agent does not. In regards to the control output, the PID 
controller exerts more power to motor three but then motor values eventually 
level off while the PPO control signal oscillates comparably more.

\subsection{Continuous Task Evaluation}
\label{sec:continuous}

\begin{figure*}
	\centering
	{\includegraphics[trim=55 0 65 40, clip, width=\textwidth]{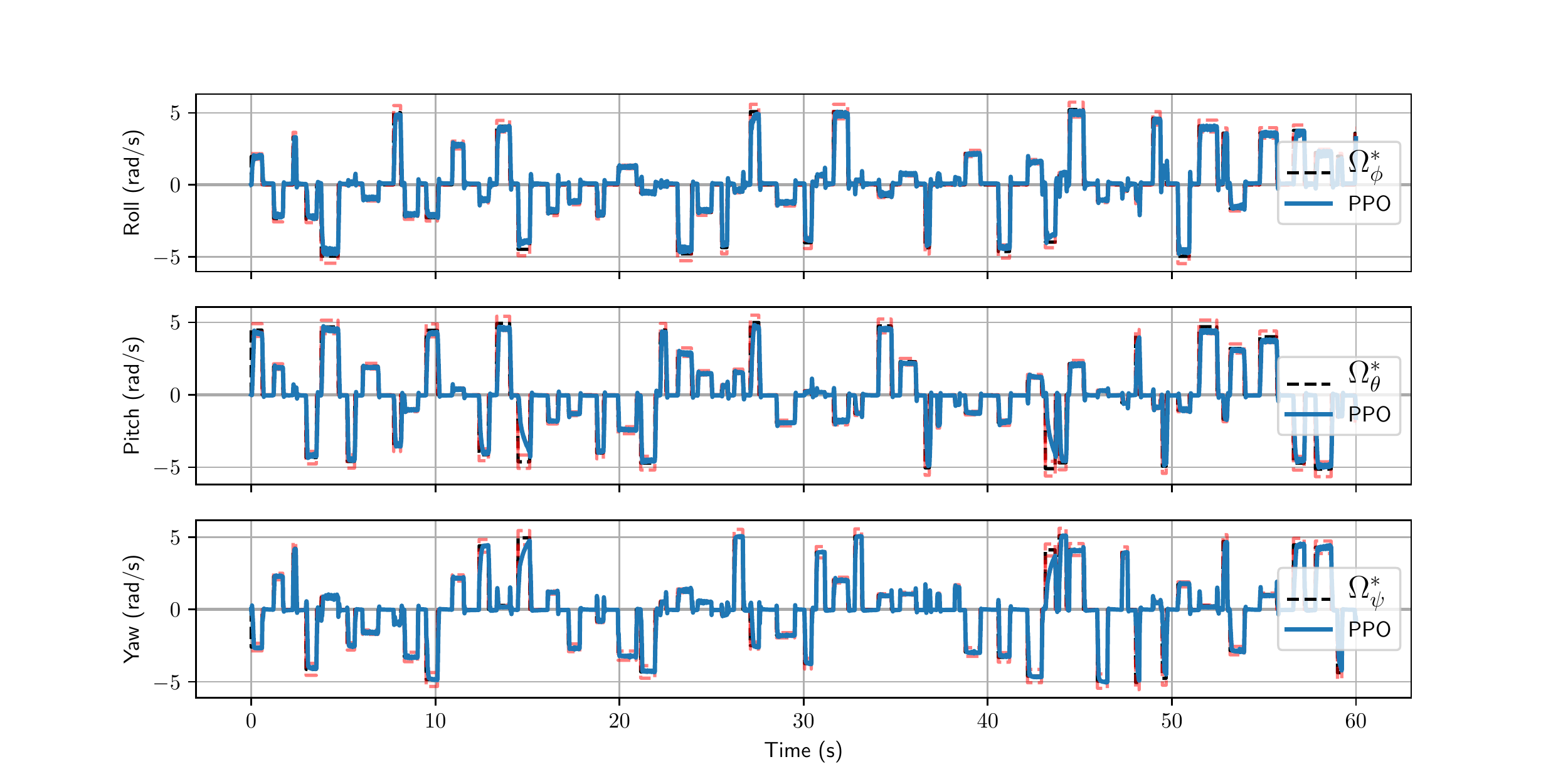}}
	\caption{Performance of PPO agent trained with episodic tasks but evaluated 
	using a continuous task for a duration of 60 seconds. The time in seconds at which a 
	new command is issued is randomly sampled from the interval $[0.1,1]$ and each issued command
	is maintained for a random duration also sampled from $[0.1, 1]$.  Desired 
	angular velocity is specified by the black line while the red line is the 
	attitude tracked by the agent.}
	\label{fig:ppo1}
\end{figure*}

\begin{figure*}
	\centering
	{\includegraphics[trim=45 0 65 30, clip, width=\textwidth]{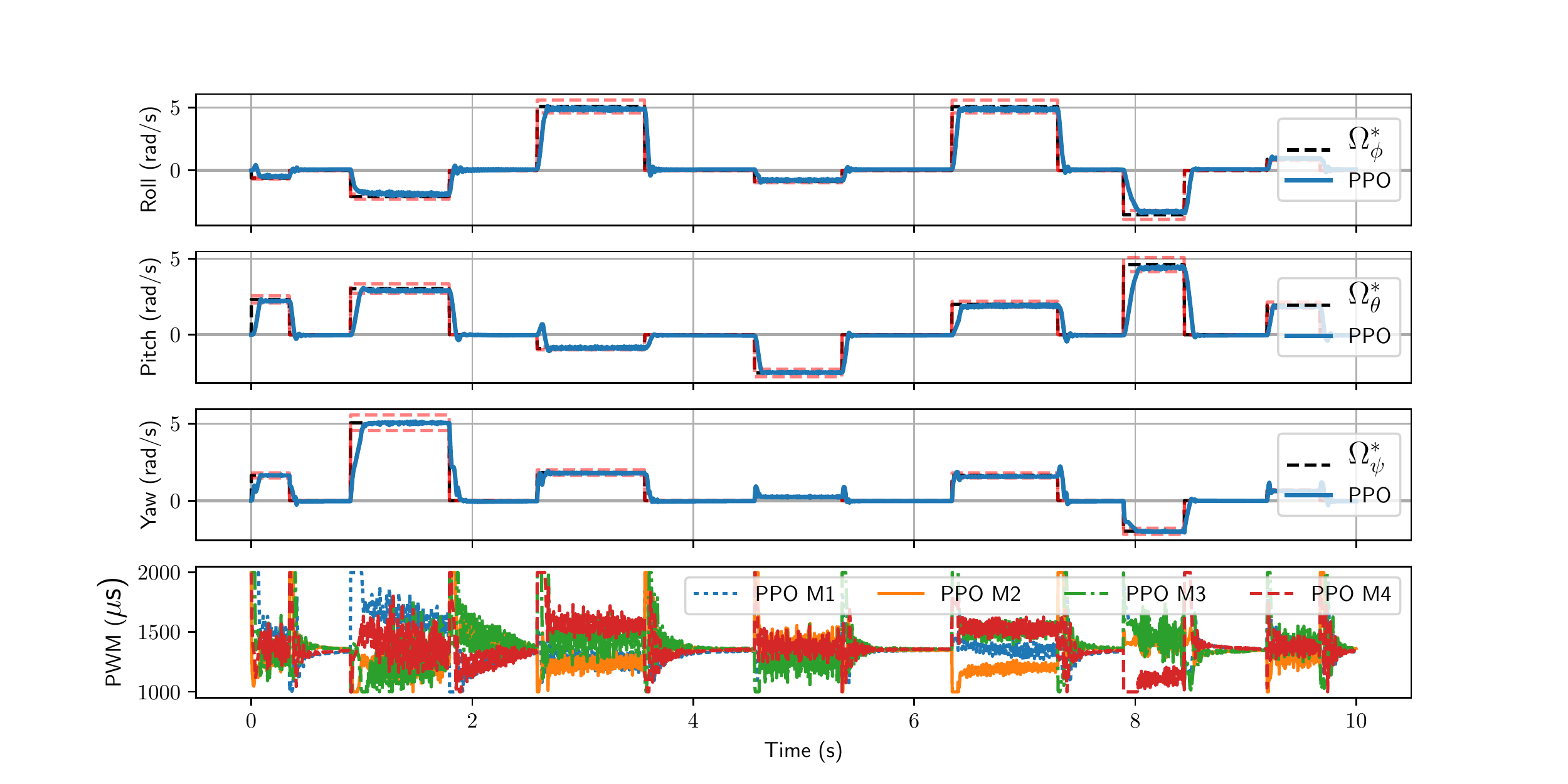}}
	\caption{Close up of continuous task results for PPO agent with PWM values.}
	\label{fig:ppo2}
\end{figure*}

\begin{figure*}
	\centering
	{\includegraphics[trim=45 0 65 30, clip, width=\textwidth]{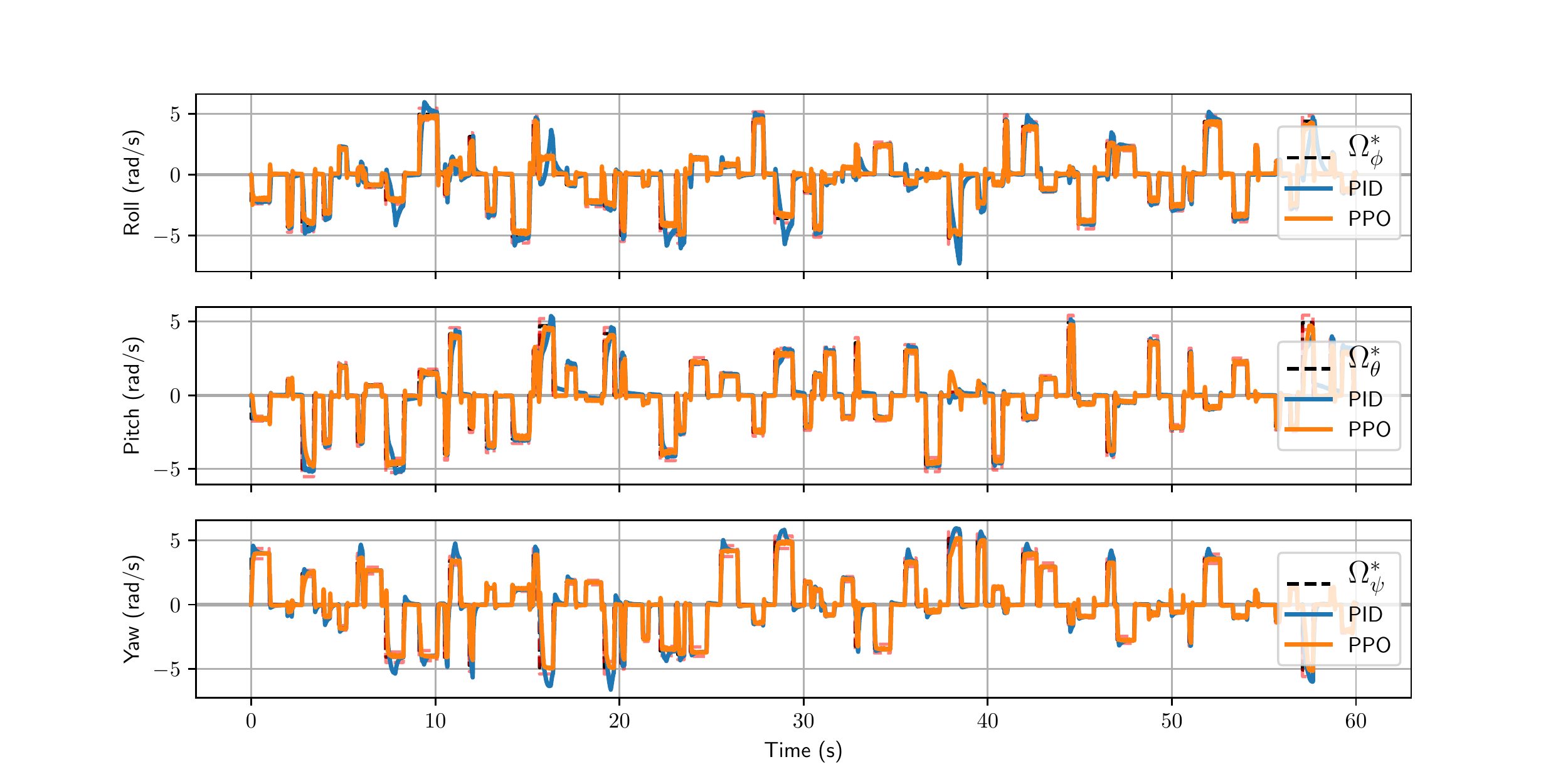}}
	\caption{Response comparison of a PID and PPO agent evaluated in continuous 
		task environment.  The PPO agent, however, is only trained using 
		episodic tasks.}
	\label{fig:ppopid}
\end{figure*}

In this section we briefly expand on our findings that show that even  if agents 
are trained through episodic tasks their performance transfers to continuous 
tasks without the need for additional training.  
Figure~\ref{fig:ppo1}  shows that an agent trained with PPO using episodic tasks has exceptional performance when 
evaluated in a continuous task. Figure~\ref{fig:ppo2} is a close up of another 
continuous task sample showing the details of the tracking and corresponding 
motor output. These results are quite remarkable as they suggest that training 
with episodic tasks is sufficient for developing intelligent attitude flight 
controller systems capable of operating in a continuous environment. In 
Figure~\ref{fig:ppopid} another continuous task is sampled and the PPO agent is 
compared to a PID agent. The performance evaluation shows the PPO agent to have 
22\% decrease in overall error in comparison to the PID agent.

\section{Future Work and Conclusion}
\label{sec:gymfc:conclusion}
In this chapter we presented our RL training environment \gym for developing 
intelligent attitude controllers for UAVs \new{and addressed in-depth 
\textbf{C1: Precision and Accuracy}, which identifies if \nns trained 
with RL can produce accurate attitude controllers.} We placed an emphasis on 
digital twinning concepts to allow transferability to real hardware.  We used 
\gym to evaluate the performance of state-of-the-art RL algorithms PPO, TRPO and 
DDPG to identify if they are appropriate to synthesize high-precision attitude 
flight controllers.  Our results highlight that: (i) RL can train accurate 
attitude controllers; and (ii) that those trained with PPO outperformed a fully 
tuned PID controller on almost every metric.  
It is important to note that although our analysis found our TRPO and DDPG policies to be
insufficient in providing stable flight we did
not perform any hyperparameter tuning in this work. Thus in future work further
benchmarking will be required to discover the true capabilities and potential of these other algorithms.

Although we base our evaluation on 
results obtained in episodic tasks, we found that trained agents were able to 
perform exceptionally well also in continuous tasks without retraining~\extra.  
This suggests that training using episodic tasks is sufficient for developing 
intelligent attitude controllers.  The results presented in this work can be 
considered as a first
milestone and a good motivation to further inspect the boundaries of RL 
for flight control. 

With this premise, we plan to develop
our future work along three main avenues. On the one hand, we plan to  
investigate \new{\textbf{C2: Robustness and Adaptation} and \textbf{C3: Reward 
Engineering}} to harness the true power of RL's 
ability to adapt and learn in environments
with dynamic properties (\eg wind, variable payload, system damage and failure).  
On the other hand we intend
to transfer our trained agents onto a real aircraft to evaluate their live
performance \new{including timing and memory analysis of the \nn.  
This will allow us to define the minimum hardware specifications required to use 
\nn attitude control.}
Furthermore, we plan to expand \gym to support other aircraft such as fixed 
wing, while continuing to increase the realism of the simulated
environment by improving the accuracy of our digital twins.

\cleardoublepage

\newcommand\depsize{24.86 KB\xspace}

\newcommand\ChNoiseYDeltaDecrease{87.95\%\xspace}
\newcommand\ChNoiseYError{XXX\%\xspace}

\newcommand\ChYDecrease{90.56\%\xspace}

\newcommand\rOmegaDeltaError{45.07\%\xspace}
\newcommand\rOmegaDeltaPower{3.41\%\xspace}

\newcommand\rPpoHorizon{500\xspace}
\newcommand\rPpoStepsize{$1 \times 10^{-4} \times \rho$\xspace}
\newcommand\rPpoEpochs{5\xspace}
\newcommand\rPpoMinibatch{32\xspace}
\newcommand\rPpoDiscount{0.99\xspace}
\newcommand\rPpoGae{0.95\xspace}

\newcommand\rNNLooptime{XXX\xspace}
\newcommand\rFCLooptime{XXX\xspace}

\newcommand\nnBenchmarkOpt{57.67\xspace}
\newcommand\nnBenchmarkNormal{58.77\xspace}

\newcommand\rCheckpointTotalSize{XXX\xspace}
\newcommand\rGraphSize{12KB\xspace}
\newcommand\rOptGraphDecrease{16\%\xspace}
\newcommand\rHexSize{913KB\xspace}
\newcommand\rHexSizeBetaflight{895KB\xspace}
\newcommand\rTimeNfAlgDisarmedWCET{204\xspace}
\newcommand\rTimeNfAlgDisarmedBCET{194\xspace}
\newcommand\rTimeNfAlgDisarmedP{4.9\xspace}

\newcommand\rTimeNfAlgArmedWCET{210\xspace}
\newcommand\rTimeNfAlgArmedBCET{195\xspace}
\newcommand\rTimeNfAlgArmedP{7.1\xspace}

\newcommand\rTimeNfLoopDisarmedWCET{244\xspace}
\newcommand\rTimeNfLoopDisarmedBCET{229\xspace}
\newcommand\rTimeNfLoopDisarmedP{6.1\xspace}

\newcommand\rTimeNfLoopArmedWCET{423\xspace}
\newcommand\rTimeNfLoopArmedBCET{263\xspace}
\newcommand\rTimeNfLoopArmedP{37.8\xspace}

\newcommand\rTimeBfAlgDisarmedWCET{14\xspace}
\newcommand\rTimeBfAlgDisarmedBCET{9\xspace}
\newcommand\rTimeBfAlgDisarmedP{35.7\xspace}

\newcommand\rTimeBfAlgArmedWCET{15 \xspace}
\newcommand\rTimeBfAlgArmedBCET{9 \xspace}
\newcommand\rTimeBfAlgArmedP{40.0 \xspace}

\newcommand\rTimeBfLoopDisarmedWCET{58\xspace}
\newcommand\rTimeBfLoopDisarmedBCET{45\xspace}
\newcommand\rTimeBfLoopDisarmedP{22.4\xspace}

\newcommand\rTimeBfLoopArmedWCET{238\xspace}
\newcommand\rTimeBfLoopArmedBCET{78\xspace}
\newcommand\rTimeBfLoopArmedP{67.2\xspace}

\newcommand\rMaxNNTaskFreq{2.4kHz\xspace}
\newcommand\rNFTaskFreq{2.4kHz\xspace}
\newcommand\rGyroRate{4kHz\xspace}
\newcommand\rLoopDenom{2\xspace}

\newcommand\rMaxLoopHertz{2kHz\xspace}
\newcommand\rMaxGyroHertz{4kHz\xspace}

\newcommand\rFasterThanPX{8\xspace}
\newcommand\rFastThanPWM{40\xspace}
\newcommand\rSlowerThanBF{1.78\xspace}

\newcommand\rRealR{15.17\xspace}
\newcommand\rRealP{21.05\xspace}
\newcommand\rRealY{11.26\xspace}

\newcommand\rSimR{2.88\xspace}
\newcommand\rSimP{1.52\xspace}
\newcommand\rSimY{4.07\xspace}

\newcommand\rSimPidR{3.90\xspace}
\newcommand\rSimPidP{5.25\xspace}
\newcommand\rSimPidY{3.42\xspace}

\newcommand\powerVoltage{16.78\xspace}
\newcommand\ampsNFdisarmed{0.37\xspace}
\newcommand\ampsNFarmed{0.67\xspace}

\newcommand\powerNFdisarmed{6.21\xspace}
\newcommand\powerNFarmed{11.24\xspace}

\newcommand\ampsBFdisarmed{0.37\xspace}
\newcommand\ampsBFarmed{0.6\xspace}

\newcommand\powerBFdisarmed{6.21\xspace}
\newcommand\powerBFarmed{10.07\xspace}

\newcommand\tunableWeights{1,344\xspace}

\newcommand\rSimError{$[1.47, 2.59, 1.14]$\xspace}

\newcommand\videourl{\cite{neuroflightproject}\xspace}
\newcommand\sourcecode{\cite{neuroflight}\xspace}

\newcommand\numFlights{5\xspace}

\graphicspath{ {3_Neuroflight/figures/} }

\chapter{Neuroflight: Next Generation Flight Control Firmware}
\label{chapter:nf}
\thispagestyle{myheadings}

Recently there has been explosive growth in user-level applications
developed for UAVs. However little
innovation has been made to the UAV's low-level attitude flight
controller which still predominantly uses classic PID
control. Although PID control has proven to be sufficient for a
variety of applications, it falls short in dynamic flight conditions
and environments (\eg in the presence of wind, payload changes and
voltage sags). In these cases, more sophisticated control strategies
are necessary, that are able to adapt and learn.
The use of \nns for flight control~(\ie neuro-flight control) has been
actively researched for decades to overcome limitations in other control
algorithms such as PID control. However the vast majority of research has
focused on developing autonomous neuro-flight controller autopilots capable of tracking
trajectories~\cite{shepherd2010robust,nicol2008robust,dierks2010output,bagnell2001autonomous,kim2004autonomous,abbeel2007application,hwangbo2017control,dos2012experimental}.

In Chapter~\ref{chapter:gymfc} we introduced our  OpenAI gym
environment  \gymfc. 
Via \gym it is possible to train \nns attitude control of a quadcopter
in simulation using RL.  Neuro-flight
controllers trained with PPO~\cite{schulman2017proximal} were shown to exceed
the performance of a PID controller. Nonetheless the attitude
neuro-flight controllers were not validated in the {real world},
thus it remained as an open question if the \nns trained in \gym are
capable of flight.  As such, this chapter makes the following
contributions:
\begin{itemize}
\item We introduce Neuroflight, the first open source neuro-flight
controller firmware for multi-rotors and fixed wing aircraft.  The \nn
embedded in Neuroflight replaces attitude control and motor mixing
commonly found in traditional flight control
firmwares~(Section~\ref{sec:nf:overview}).
\item To train neuro-flight controllers capable of stable flight
in the {real world we introduce \newgym}{, a modified
environment addressing several challenges} in making the transition
from simulation to reality~(Section~\ref{sec:nf:challenge}).
\item We propose a toolchain for compiling a trained \nn to run on
embedded hardware. To our knowledge this is the first work that
consolidates a neuro-flight attitude controller on a microcontroller,
rather than a multi-purpose onboard computer, thus allowing deployment
on lightweight micro-UAVs~(Section~\ref{sec:nf:sys}).
\item Lastly, we provide an evaluation showing the
\nn can execute at over \rMaxLoopHertz on an Arm Cortex-M7 processor and
flight tests demonstrate that a quadcopter running \fc can achieve
stable flight and execute aerobatic maneuvers such as rolls, flips,
and the Split-S~(Section~\ref{sec:nf:eval}).  Source code for the project
can be found at
\sourcecode 
and videos of our test flights can be viewed at \videourl.
\end{itemize}

The goal of this work is to provide the community with a stable
platform to innovate and advance development of neuro-flight control
design for UAVs, and to take a step towards making neuro-flight
controllers mainstream. In the future we hope to establish \nn powered
attitude control as a convenient alternative to classic PID control
for UAVs operating in harsh environments or that require particularly
competitive set point tracking performance~(\eg drone racing).

\section{Background and Related Work}
\label{sec:nf:related}

Over time there has been a number of successes transferring controllers trained
with RL to multicopters. % onboard computer to autonomously track trajectories in the real world.
This includes helicopters~\cite{bagnell2001autonomous,kim2004autonomous,abbeel2007application}
and quadcopters~\cite{hwangbo2017control,dos2012experimental}.
Unfortunately none of these works have published any code thereby making it
difficult to reproduce results and to build on top of their research. Furthermore
evaluations are only in respect to the accuracy of position therefore it is
still unknown how well attitude is controlled. 
Of the open source flight control firmwares currently available, every single one uses PID control~\cite{ebeid2018survey}.

In regards to methods and techniques  
for transferring trained policies to hardware, these are neglected in
the 
helicopter
control literature~
\cite{bagnell2001autonomous,kim2004autonomous,abbeel2007application}.
Given the resource constrained hardware onboard a quadcopter, hardware details are
more commonly discussed, however strategies for policy transfer are still
lagging behind. 
A common strategy for executing high-level navigation tasks is to use 
a separate companion (compute) board which computes the desired attitude commands and
sends them over a serial connection to an off-the-shelf flight controller. 
For example the default
configuration of the Intel Aero~\cite{intelaero} uses an Intel compute board
which communicates with a microcontroller running PX4. 
Previous research has used companion boards for onboard
computation of RL controllers. In \cite{hwangbo2017control} an Intel computer stick is
used for the RL controller which outputs the desired motor
thrust values. These are then provided as input to a software library for interfacing
over serial to the separate flight control board. 
In~\cite{palossi201964mw} the authors present an impressive vision based
navigation system using an RL controller for the Crazyflie quadcopter. A
companion board  executes the \nn and interfaces with the Crazyflie flight
controller over the serial peripheral interface~(SPI).
Additionally this work
provides an extensive evaluation of the architecture required to successfully
perform vision navigation in
such a resource constrained hardware environment.

To reduce weight and increase communication throughput a single control board
should be used. 
Work by \cite{molchanov2019sim}
executes their policy directly on the flight controller for a Crazyflie
quadcopter. Nonetheless aircraft state estimation is offloaded to a ground station.
Using a postprocessing stage after policy training, the network parameters from the
trained model are extracted and compiled into a C function to be linked into
the Crazyflie's flight control firmware.

Developing a generic all-in-one flight control board capable of complex
navigation tasks is challenging due to timing guarantees of time
sensitive tasks.
Advances made by \cite{cheng2018end} have ported the flight control firmware
Cleanflight~\cite{cleanflight}
to run within a real-time operating system. Their analysis on the Intel Aero
compute board shows their approach is able to bound end-to-end latencies from
sensor input to motor output.

\section{\fc Overview}
\label{sec:nf:overview}
\fc is a fork of Betaflight version 3.3.3~\cite{betaflight}, a high 
performance flight controller firmware used extensively in
first-person-view~(FPV) multicopter racing.  
Internally Betaflight uses a two-degree-of-freedom PID controller~(not to be
confused with rotational degrees-of-freedom) for attitude 
control and includes other enhancements such 
as gain scheduling for increased stability when battery voltage
is low and  throttle is high.  Betaflight runs on a wide variety of flight 
controller hardware  based on the Arm  Cortex-M family of 
microcontrollers.  Flight control tasks are scheduled using a non-preemptive
cooperative scheduler. 
The main PID controller task consists of multiple
subtasks, including: (1) reading the remote control (RC) command for the desired
angular velocity, (2) reading
and filtering the angular velocity from the onboard
gyroscope sensor, (3) evaluating the PID controller, (4) applying motor mixing to the
PID output to account for asymmetries in the motor locations~(see
Section~\ref{sec:bg:pid} for
further details on mixing), and (5) writing 
the motor control signals to the ESCs. 

\begin{figure*}
 %<left> <lower> <right> <upper>
 \centering
 {\includegraphics[trim=0 280 0
    0,clip,width=0.78\paperwidth]{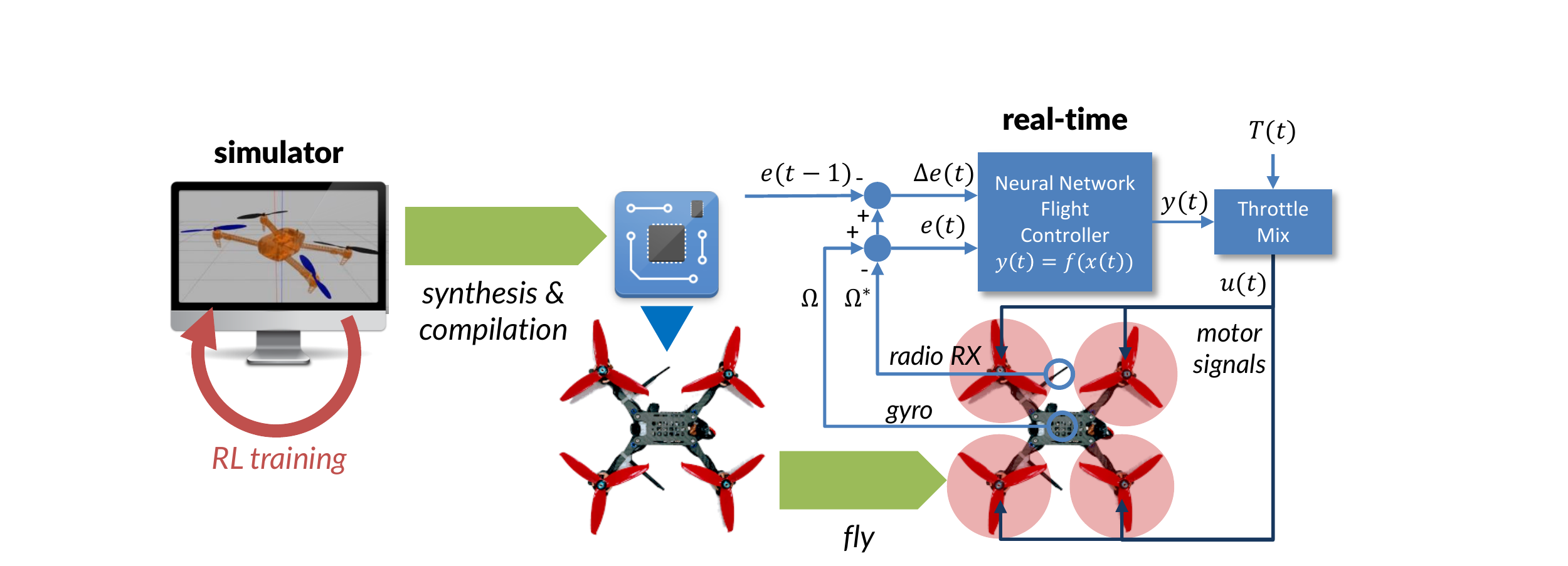}}
 \caption{Overview of the \fc architecture.}
 \label{fig:overview}
\end{figure*}

\fc replaces Betaflight's PID controller task with a neuro-flight controller task.
This task uses a single \nn for attitude control and motor mixing. The
architecture of \fc decouples the \nn from the rest of the firmware
allowing the \nn to be trained and compiled independently.  An
overview of the architecture is illustrated in
Fig.~\ref{fig:overview}.
The compiled \nn is then later linked into \fc to produce a firmware image for the
target flight controller hardware. 

To \fc, the \nn appears to be a generic function $y(t) = f(x(t))$. The 
input is $x(t)=[e(t), \Delta e(t)]$ where  $\Delta e(t)=e(t) - e(t-1)$. 
The output $y(t)=[y_0, \dots, y_{M-1}]$ where $M$ is the number of
aircraft actuators to be controlled and $y_i \in [0,1]$ is the control
signal representing the percent power to be applied to the $i^{th}$
actuator. This output representation is protocol agnostic and is not
compatible with \nns trained with \gym from Chapter~\ref{chapter:gymfc} whose output is the PWM to be
applied to the actuator. PWM is seldomly used in high performance
flight control firmware and has been replaced by digital protocols
such as DShot for improved accuracy and speed~\cite{betaflight}.

At time $t$, the \nn inputs are resolved; $\Omega^*(t)$ is read from the
RX serial port which is either connected to a radio receiver in the case of manual
flight or an onboard companion computer operating as an autopilot in the case of autonomous flight, and $\Omega(t)$ is read from the gyroscope sensor.
 The \nn is then evaluated to obtain the control signal outputs $y(t)$. However the \nn has no concept of thrust
($\mathbf{T}$),
therefore to achieve translational movement the thrust command must be mixed into
the \nn output to produce the final control signal output to the ESC, $u(t)$. 
The logic of throttle mixing is to uniformly apply additional power across all  
motors proportional to
the available range in the \nn output,
while giving priority to achieving $\Omega^*(t)$. 
This approach does make the assumption the performance of each motor is
identical, which may not always be the case.   
If any output value is over 
saturated (\ie $\exists y_i(t)  : y_i(t) \ge 1$)  no additional throttle will be
added. 
The input throttle value is scaled
depending on the available output range to obtain the actual throttle value:
\begin{equation}
\widehat{\mathbf{T}}(t)= \mathbf{T}(t)\left(1 - \mathrm{max}_i\{y_i(t)\}\right)
\end{equation}
where the function $\mathrm{max}$ returns the max value from the \nn output.
The readjusted throttle value is then proportionally added to each \nn
output to form the final control signal output:
\begin{equation}
    u_i(t)  =
\widehat{\mathbf{T}}(t) + y_i(t).
\end{equation}

\section{\newgym}
\label{sec:nf:challenge}
In this section we discuss the enhancements made to \gym to create \newgym.
These changes primarily consist of a new
state representation and reward system.

\subsection{State Representation}
\label{sec:challenge:motor_dep}
\gym returns the state $x(t)=(e(t), \omega(t))$ to the agent at each time step.  
However not all UAVs have the sensors to measure motor velocity
$\omega(t)$ as this typically involves digital ESC protocols.  Even in
an aircraft with compatible hardware, including the  motor
velocity as an input to the \nn introduces additional challenges. This
is because a \nn trained on RPMs does not easily transfer
from simulation to the real world, unless an accurate propulsion
subsystem model is available for the digital twin.  A mismatch between
the physical propulsion system~(\ie motor/propeller combination) and
the digital twin will result in the inability to achieve stable
flight.
Developing an accurate motor models is time-consuming and
expensive. Specialized equipment is required to capture the relations
between voltage, power consumption, temperature, rotor velocity, torque, and thrust.

To address these issues we investigated training using alternative
environment states that do not rely on any specific characteristic of
the motor(s). We posited that reducing the entire state to just
angular velocity errors would carry enough information for the \nn to
achieve stable flight. At the same time, we expected that the
obtained \nn would transfer well to the real aircraft. Thus, our \nn
is trained by replacing $\omega(t)$ with the error differences $\Delta
e(t)$.
To identify the performance impact of this design choice, we trained
two \nns. A first \nn was trained with $\omega(t)$ in input. Its
behavior was compared to a second \nn trained in an environment that
provides $\Delta e(t)$ instead.
Both \nns were trained with PPO using
hyperparameters from~\cite{gymfc} for 10 million steps.
After training, each \nn was validated against 10 never before seen
random target angular velocities.  Results show the \nn trained in an
environment with,

\begin{equation}
x(t)=(e(t),\Delta e(t))
\label{eq:nf:input}
\end{equation}

experienced on
average \rOmegaDeltaError \textit{less} error with only an increase
of \rOmegaDeltaPower in its control signal outputs.

In RL the interaction between the agent and the environment can be
formally modeled as a Markov Decision Process~(MDP) in which the
probability that the agent transitions to the next state depends on
its current state and action to be taken. The behavior of the agent is
defined by its policy which is essentially a mapping of states to
actions.
There may be multiple different state representations that are able to
map to actions resulting in similar performance.
For instance, it emerged from our experiments that using a history of
errors as input to the \nn also led to satisfactory performance. This
approach has the disadvantage of requiring a \new{state
history table to be maintained}, which ultimately made the approach
less desirable.

The intuition why a state representation comprised of only angular
velocity errors works can be summarized as follows. First, note that a
PD controller (a PID controller with the integrative gain set to zero)
is also a function computed over the angular velocity error. Because
an \nn is essentially a universal approximator, the expectation is
that the \nn would also be able to find a suitable control strategy
based on these same inputs.

However, modifying the environment state alone is not
enough to achieve stable flight. The RL task also needs to be
adjusted.
Training using episodic tasks, in which the aircraft is at rest and
must reach an angular velocity never exposes the agent to scenarios in
which the quadcopter must return to still from some random angular
velocity.
With the new state input consisting of the previous state, this is a
significant difference from \gym which only uses the current state.
For this purpose, a continuous task is constructed to mimic real
flight, continually issuing commands.\footnote{Technically this is
still considered an episodic task since the simulation time is finite.
However in the {real world} flight time is typically finite as
well.} This task randomly samples a command and sets the target
angular velocity to this command for a random amount of time. This
command is then followed by an idle~(\ie $\Omega^*=[0, 0, 0]$) command
to return the aircraft to still for a random amount of time. This is
repeated until a max simulation time is reached.

\subsection{Reward System}
\label{sec:reward}
\new{  Reward engineering is a
        particularly difficult problem. As reward systems increase in
        complexity, they may present unintended side affects resulting in the
        agent behaving in an unexpected manner.}

\newgym reinforces stable flight behavior through our reward
system defined as:
\begin{equation}
r = r_e + r_y + r_\Delta.
\end{equation} 
The agent is penalized
for its angular velocity error, similar to \gym, along each axis with:
\begin{equation}
r_e = -(e_{\phi}^2 + e_{\theta}^2 + e_{\psi}^2).
\end{equation}

However we have
identified the remaining two variables in the reward system as
critical for transferability to the {real world} and achieving
stable flight. Both rewards are a function of the agents control
output. First $r_y$ rewards the agent for minimizing the control
output, and next, $r_\Delta$ rewards the agent for minimizing
oscillations.

The rewards as a function of the control signal are able to aid in the
transferability by compensating for limitations in the training
environment and unmodelled dynamics in the motor model.

\textbf{Minimizing Output Oscillations.}
\label{sec:challenge:noise}
In the real world high frequency oscillations in the control output
can damage motors. Rapid switching of the control output causes the
ESC to rapidly change the angular velocity of the motor drawing
excessive current into the motor windings. The increase in current
causes high temperatures which can lead to the insulation of the motor
wires to fail. Once the motor wires are exposed they will produce a
short and ``burn out" the motor.

The reward system used by \gym is strictly a function of the angular
velocity error.  This is inadequate in developing neuro-flight
controllers that can be used in the real world. Essentially this
produces controllers that closely resemble the behavior of an
over-tuned PID controller. The controller is stuck in a state in which
it is always correcting itself, leading to output oscillation.

In order to construct networks that produce smooth control signal
outputs, the control signal output must be introduced into the reward
system. This turned out to be quite challenging.
Ultimately we were able to construct \nns outputting stable control outputs
with the inclusion of the following reward:
\begin{equation}
r_\Delta =
    \beta \sum_{i=0}^{N-1} \text{max}\{ 0, \Delta y_\text{max} -  \left(\Delta
            y_i\right)^2\}.
\end{equation}
This reward is only applied if the absolute angular velocity error for
every axis is less than some threshold~(\ie the error band). This
allows the agent to be signaled by $r_e$ to reach the target without
the influence from this reward.
Maximizing $r_\Delta$ will drive the agent's change in output to zero
when in the error band. To derive $r_\Delta$, the change in the
control output $y_i$ from the previous simulation step is squared to
magnify the effect. This is then subtracted from a constant $\Delta
y_\text{max}$ defining an upper bound for the change in the control
output. The \texttt{max} function then forces a positive reward,
therefore if $(\Delta y_i)^2$ exceeds the limit no reward will be
given. The rewards for each control output $N-1$ are summed and then
scaled by a constant $\beta$, where $\beta > 0$.  Using the same
training and validation procedure previously discussed, we found a \nn
trained in \newgym compared to \gym resulted in
a \ChNoiseYDeltaDecrease decrease in $\Delta y$.

\textbf{Minimizing Control Signal Output Values.}
\label{sec:challenge:power}
Recall from Section~\ref{sec:gymfc:twin}, that the \gym environment fixes 
the aircraft to the simulation world about its center of mass, allowing it to
only perform rotational movements. \new{Due to this constraint} the agent can achieve
 $\Omega^*$ with a number of different control signal outputs (\eg when
$\Omega^*=[0,0,0]$ this can be achieved as long as $y_0 \equiv y_1 \equiv y_2
\equiv y_3$).
However this poses a significant problem when transferred to the real world as an aircraft
is not fixed about its center of mass. Any additional power to the motors will result in an
unexpected change in translational movement. This is immediately evident when
arming the quadcopter which should remain idle until RC commands are
received. At idle, the power output (typically ~4\% of the throttle value) must not result in
any translational movement.
Another byproduct of inefficient control signals is a decreased throttle
range~(Section~\ref{sec:nf:overview}). Therefore it
is desirable to have the \nn control signals minimized while still
maintaining the desired angular velocity. 
In order to teach the agent to minimize control outputs we
introduce the reward function:
\begin{equation}
    r_y =\alpha \left( 1-\bar{y} \right)
\end{equation}
providing the agent a positive reward as the output decreases. Since $y_i \leq 1$ 
we first compute the average output $\bar{y}$. Next $1-\bar{y}$ is calculated
as a positive reward for low output usage which is scaled by a constant
$\alpha$, where $\alpha > 0$. \nns trained using this reward experience on average a \ChYDecrease
decrease in their control signal output.

    \new{\textbf{Challenges and Lessons Learned.} The fundamental challenge we faced was managing high amplitude oscillations in
the control signal.  
In stochastic continuous control problems it is standard for the network to
output the mean from a Gaussian distribution~\cite{schulman2017proximal,chou2017improving}.
However this poses problems for control tasks with bounded outputs such as
flight control. The typical strategy is to clip the output to the target bounds
yet we have observed this to significantly contribute to oscillations in the control output. 

Through our experience  
we learned that due to the output being stochastic~(which aids in exploration),
the rewards must encapsulate the general trend of the
performance and not necessarily at a specific time~(\eg} \new{the stochastic output
naturally oscillates). 
Additionally we found the reward system must include performance metrics 
other than (but possibly in addition to)
traditional time domain step response characteristics (\eg} \new{overshoot, rise time,
settling time, etc.). Given the agent initially knows nothing, there 
is no step
response to analyze. In future work we will explore the use of goal based
learning in an attempt to develop a hybrid solution in which the agent learns
enough to track a step response, then use traditional metrics for fine tuning.} 

\new{Although our reward system was
sufficient in achieving flight, we believe this is still an open area of
research worth exploring. In addition to aforementioned rewards, we
experimented with several other rewards 
including penalties for over saturation of the
control output (\ie} \new{if the network output exceeded the
clipped region), control output jerk (\ie} \new{change in
acceleration), and the number of oscillations in the output.    
When combining multiple rewards, balancing these rewards can be an
exercise of its own. 
For example if penalizing for number of oscillations or jerk this can lead to
an output that
resembles a low frequency square wave if penalizing the amplitude is not
considered.
    }

\section{Toolchain}
\label{sec:nf:sys}
\begin{figure}
\centering
{\includegraphics[trim=0 70 240 0,clip, width=0.8\textwidth]{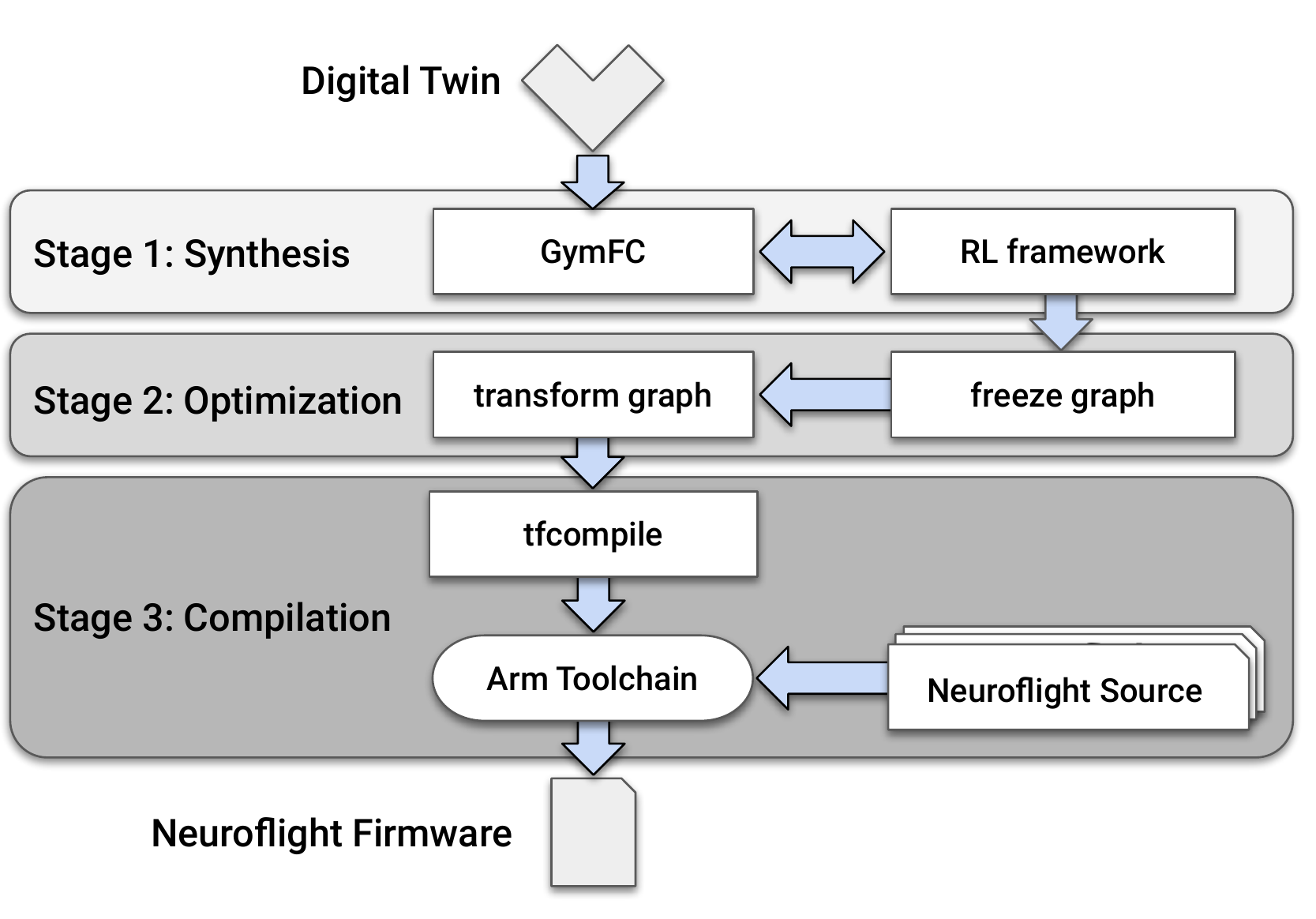}}
\caption{Overview of the \fc toolchain.}
\label{fig:stages}
\end{figure}

In this section we introduce our toolchain for building the \fc firmware.
\fc is based on the philosophy that each flight control firmware should be
 customized for the target aircraft to achieve maximum performance.
To train a \nn optimal 
attitude control of an aircraft, a
digital twin of the
aircraft must be constructed to be used in simulation.  
This work begins to address how digital twin fidelity 
affects flight performance, however it is still an open question that we will
address in future work.
The toolchain displayed in {Fig.}~\ref{fig:stages} consists of three stages and takes as input a digital
twin and outputs a \fc firmware unique to the digital twin. 
In the remainder of this section we
will discuss each stage in detail.

\subsection{Synthesis} 

The synthesis stage takes as input a digital twin of an aircraft and
synthesizes a
\nn  attitude flight controller  capable of achieving stable flight in the
real world.
Our toolchain can support any RL library that interfaces
	with  OpenAI environment APIs and allows for the \nn state to be saved  as a
    Tensorflow graph. 
Currently our toolchain uses RL algorithms provided by
OpenAI baselines~\cite{baselines} which has been modified to save the \nn state. 
In Tensorflow, the saved state of a \nn is known as a checkpoint and consists
of three files describing the structure and values in the graph.
Once training has completed, the checkpoint is provides as input to Stage 2: Optimization.

\subsection{Optimization} The optimization stage is an intermediate stage between training and compilation that prepares the \nn graph to be run on hardware. The optimization stage (and compilation stage) require a number of Tensorflow tools which can all be found in the Tensorflow repository~\cite{tensorflow}.  The first step in the optimization stage is to \textit{freeze} the
graph. Freezing the graph accomplishes two tasks: (1) condenses the three checkpoint files into a single Protobuf file by replacing
variables with their equivalent constant values~(\eg numerical weight values)
and (2) extracts the subgraph containing the trained \nn by trimming unused nodes and operations that were only used during training. Freezing is done with Tensorflow's
\texttt{freeze\_graph.py} tool  which takes as input the checkpoint and the
output node of the graph so the tool can identify and extract the subgraph.

Unfortunately the Tensorflow input and output nodes are not documented by RL
libraries (OpenAI baselines~\cite{baselines}, Stable
baselines~\cite{stable-baselines},
TensorForce~\cite{schaarschmidt2017tensorforce})  and in most cases it is not
trivial to identify them. 
We reverse engineered the graph produced by {OpenAI} Baselines (specifically the PPO1 implementation) using a
combination of tools and cross referencing with the source code. A Tensorflow graph can
be visually inspected using Tensorflow's Tensorboard tool.
OpenAI Baselines does not support Tensorboard thus we created a script
to convert a checkpoint to a Probobuf file and then used Tensorflow's
\texttt{import\_pb\_to\_tensorboard.py}  tool to  view the graph in Tensorboard.
Additionally we used Tensorflow's \texttt{summarize\_graph} tool to 
summarize the inputs and outputs of the graph. 
Ultimately we identified the input node to be ``pi/ob'',  and the output to be 
``pi/pol/final/BiasAdd''.

Once the graph is frozen, it is optimized to run on hardware by running the Tensorflow \texttt{transform\_graph} 
tool. Optimization provided by this tool allows graphs to execute faster and
reduce its overall footprint by further removing unnecessary nodes. 
The optimized frozen ProtoBuf file is provided as input to Stage 3: Compilation.

\subsection{Compilation}
\label{sec:toolchain:compile}
A significant challenge was developing a method to integrate a trained \nn into
\fc to be able to run on the limited resources provided by a 
microcontroller.  The most powerful of the microcontrollers supported
by Betaflight \new{and Neuroflight} consists of 1MB of flash memory\new{, 320KB
of SRAM} and an ARM Cortex-M7
processor with a clock speed of 216MHz~\cite{STM32F745VG}. Recently
there has been an increase in interest for running \nns on embedded
devices but few solutions have been proposed \new{and no standard solution
    exists}.
We found Tensorflow's tool \texttt{tfcompile} to work
best for our toolchain. \texttt{tfcompile} provides 
ahead-of-time (AOT) compilation of Tensorflow graphs into executable code 
primarily motivated as a method to execute graphs on mobile devices.  Normally executing 
graphs requires the Tensorflow runtime  which is far too heavy  
for a microcontroller.  Compiling graphs using \texttt{tfcompile} does 
not use the Tensoflow runtime which results in a self contained executable and
a reduced footprint. 

Tensorflow uses the
Bazel~\cite{bazel} build system and  expects you
will be using the \texttt{tfcompile} Bazel macro in your project. \fc on
the other hand is using \texttt{make} with the GNU Arm Embedded Toolchain. Thus
it was necessary for us to integrate \texttt{tfcompile} into the toolchain by calling the \texttt{tfcompile} binary directly.
When invoked, an object file representing the compiled
graph  and an accompanying header file is produced. Examining the header file we identified three additional Tensorflow dependencies that must be included
in \fc~(typically this is automatically included if using the Bazel build system): the AOT runtime 
(\texttt{runtime.o}), an interface to run the compiled functions
(\texttt{xla\_compiled\_cpu\_function.o}), and running options
(\texttt{executable\_run\_options.o}) for a total of \depsize. 
In 
Section~\ref{sec:nf:eval} we will analyze the size of the generated object file
for the specific neuro-flight controller.

To perform fast floating point calculations \fc must be compiled with
ARM's hard-float application binary interface~(ABI). Betaflight core,
inherited by
\fc already defines the proper compilation flags in the Makefile however it is
required that the entire firmware must be compiled with the same ABI
meaning the Tensorflow graph must also be compiled with the same ABI.
Yet \texttt{tfcompile} does not currently allow for setting arbitrary
compilation flags which required us to modify the code.  Under the
hood, \texttt{tfcompile} uses the LLVM backend for code generation. We
were able to enable hard floating points through the ABIType attribute
in the \texttt{llvm::TargetOptions} class.

\section{Evaluation}
\label{sec:nf:eval}

\begin{figure}
\centering
\begin{subfigure}[b]{0.4\textwidth}
    {\includegraphics[trim=20 0 53 30,clip,
        width=\textwidth]{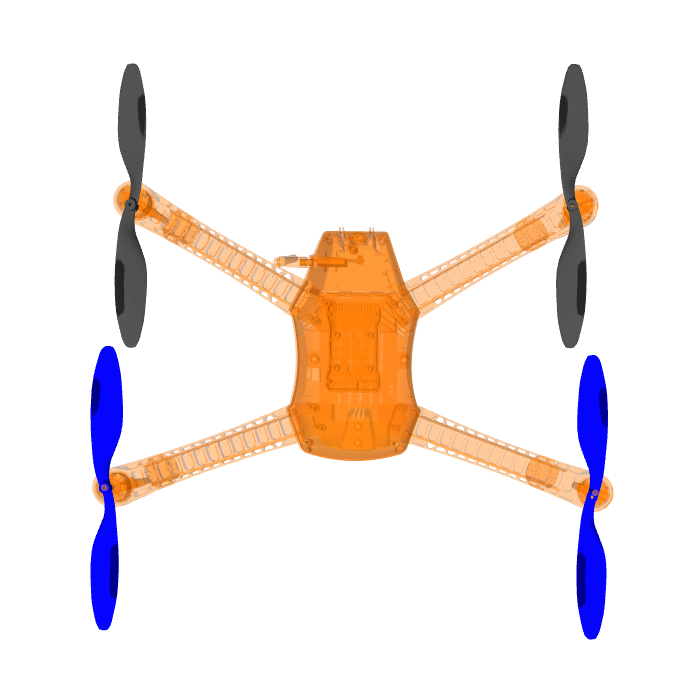}}
    \caption{Iris }
    \label{fig:nf:iris}
\end{subfigure}
~
\begin{subfigure}[b]{0.4\textwidth}
    {\includegraphics[trim=20 0 53 30,clip, width=\textwidth]{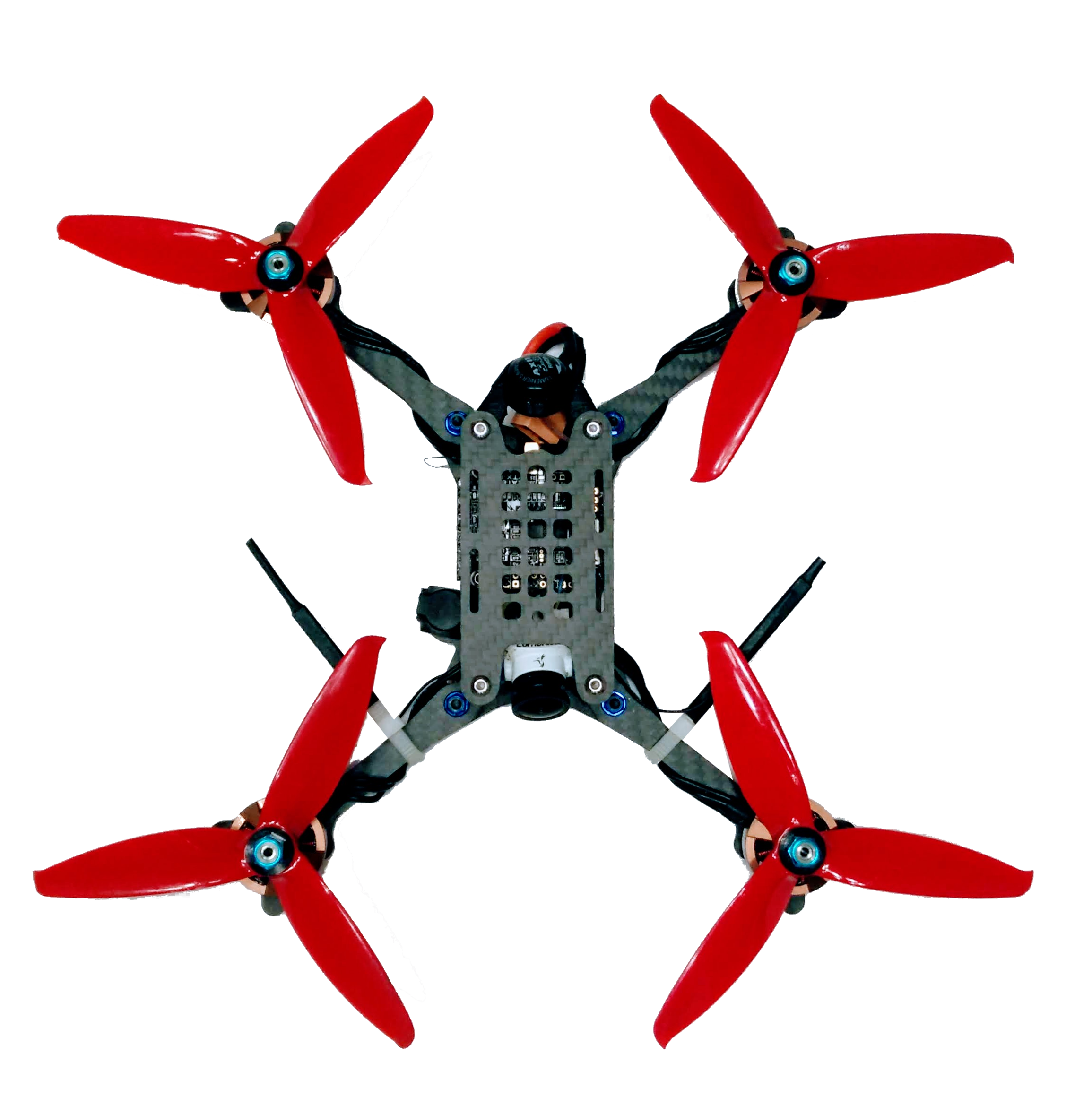}}
    \caption{\aircraft}
    \label{fig:quad}
\end{subfigure}
\caption{{Iris simulated quadcopter compared to the \aircraft}{ real
        quadcopter.}} 
\label{fig:quadcompare}
\end{figure}

In this section we evaluate \fc controlling a high performance \new{custom} FPV
racing quadcopter \new{named \aircraft}, pictured in Fig.~\ref{fig:quad}. First and foremost, we show that it is capable of
maintaining stable flight. Additionally, we demonstrate that the
synthesized \nn controller is also able to stabilize the aircraft even
when executing advanced aerobatic maneuvers.  Additional images of \new{\aircraft}
and its entire build log have been published to
RotorBuilds~\cite{rotorbuild}.

\subsection{Firmware Construction} 
\label{sec:eval:firmware}
We used the Iris quadcopter model included with the Gazebo
simulator \new{(which is also used by \gym)} with modifications to the motor model \new{to more accurately reflect \aircraft} for our digital twin.
The digital twin motor model used by Gazebo is quite simple. Each
control signal is multiplied by a maximum rotor
velocity constant to derive the target rotor velocity while each rotor is
associated with a PID
controller to achieve this target rotor velocity. 
We obtained an estimated maximum 33,422 RPMs for our propulsion system
from Miniquad Test Bench~\cite{miniquad} to update the maximum rotor
velocity constant. We also modified the rotor PID controller (P=0.01,
I=1.0) to achieve a similar throttle ramp.

{\aircraft}{ is in stark contrast with the Iris quadcopter
  model used by \gym}{ which is advertised for autonomous flight
  and imaging~\cite{iris}. We have provided a visual comparison in
  Fig.~\ref{fig:quadcompare} and a comparison between the aircraft
  specifications in Table~\ref{tbl:specs}.  In this table, weight
  includes the battery, while the wheelbase is the motor to motor
  diagonal distance. Propeller specifications are in the format
  ``LL:PPxB" where LL is the propeller length in inches, PP is the
  pitch in inches and B is the number of blades. Brushless motor sizes
  are in the format ``WWxHH" where WW and HH is the stator width and
  height respectively. The motors $K_v$ value is the motor velocity
  constant and is defined as the inverse of the motors back-EMF
  constant which roughly indicates the RPMs per volt on an unloaded
  motor~\cite{learnrc}. Flight controllers are classified by the
  version of the embedded ARM Cortex-M processor prefixed by the
  letter `F' (\eg}{ F4 flight controller uses an ARM Cortex-M4).}

\begin{table}[]
    \centering
    \begin{tabular}{l|ll}
                          & Iris                    & \aircraft
        \\ \hline
		Weight			  & 1282g       & 432g
		\\
		Wheelbase		  & 550mm  					& 212mm
		\\ \hline
        Propeller         & 10:47x2                  & 51:52x3
        \\
        Motor             & 28x30 850$K_v$              & 22x04 2522$K_v$
        \\
        Battery           & 3-cell 3.5Ah LiPo & 4-cell 1.5Ah LiPo
        \\
        Flight Controller & F4                      & F7                     
    \end{tabular}
\caption{{Comparison between Iris and \aircraft}{ specifications.} }
\label{tbl:specs}
\end{table}

\begin{figure*}
  \begin{subfigure}[b]{0.5\textwidth}
    \includegraphics[width=\textwidth]{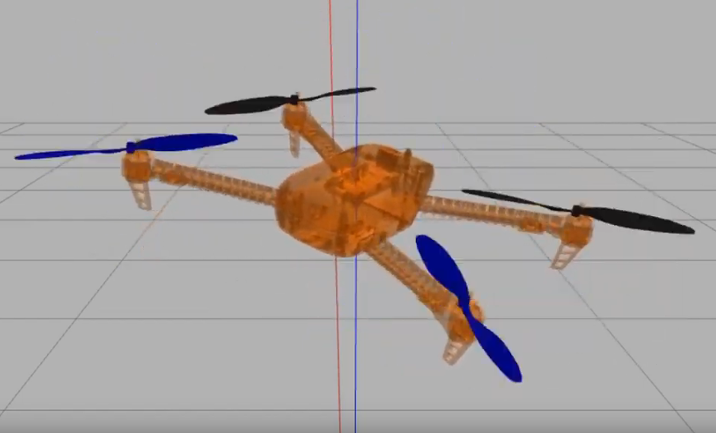}
    \caption{Screenshot of the Iris quadcopter flying in simulation.}
    \label{fig:iris_sim}
  \end{subfigure}
  ~
  \begin{subfigure}[b]{0.5\textwidth}
    \includegraphics[trim=0 0 0
      0,clip,width=\textwidth]{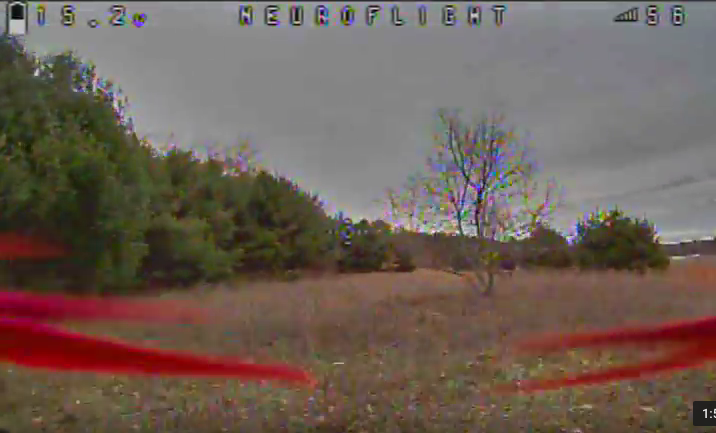}
    \caption{Still frame of the FPV video footage acquired during a test
        flight.}
    \label{fig:nf1_fpv}
  \end{subfigure}
  \caption{Flight in simulation (left) and in the real world (right).}
\end{figure*}

Our \nn architecture consists of \new{6 inputs, 4 outputs,} 2 hidden layers with 32 nodes each
using hyperbolic tangent activation functions \new{ resulting in a total of
    \tunableWeights} \new{ tunable weights}. \new{The network outputs the mean of a Gaussian distribution with a variable standard deviation as defined by PPO for continuous domains~\cite{schulman2017proximal}.}
Training was performed with the OpenAI Baseline version 0.1.4 implementation of PPO1
due to its previous success in Chapter~\ref{chapter:gymfc} \new{which showed
    PPO to out perform DDPG~\cite{lillicrap2015continuous}, and TRPO~\cite{schulman2015trust} in regards to attitude control
in simulation}. A picture of the quadcopter during  
trained in \newgym can be seen in Fig.~\ref{fig:iris_sim}.

The reward system hyperparameters used were $\alpha=300$, $\beta=0.5$, and
$\Delta y_{max}=100^2$ \new{and the PPO hyperparameters used are reported in
    Table~\ref{tab:ppo}. The reward hyperparameter $\Delta y_{max}$ is defined
    as the maximum delta in the output we are willing to accept, while $\alpha$
    and $\beta$ were found through experimentation to find the desired balance between
    minimizing the output and minimizing the output oscillations.  The discount
    and Generalized Advantage Estimate~(GAE) parameters were taken from
\cite{schulman2017proximal} while the remaining parameters were found using
random search. The agent was particularity sensitive to the selection
of the horizon and minibatch size. To account for sensor noise in the real
world we added noise to the angular velocity measurements which was sampled from a
Gaussian distribution with $\mu=0$ and $\sigma=5$. The standard deviation was
obtained by
incrementing $\sigma$ until it began to impact the controllers ability to track
the setpoint in simulation. We observed 
this to reduce motor oscillations in the real world.}

A challenge we faced transferring the trained policy to hardware was that 
we were unable to get the quadcopter to idle.
The control signals generated at
 at idle, $\Omega^* = [0,0,0]$, was producing a net force greater than the 
 downward force of our aircraft. As a result, the quadcopter would not stay on
 the ground. A possible explanation to this behavior could be due to the differences
 between the simulated quadcopter and the real quadcopter.  
 %If the thrust to weight ratio is lower in simulation, when transferred to the
 %real quadcopter, the
 %same generated control signals will provide more thrust than expected.   
%
As a work around to make it easier for the agent to generate small control signals, 
we disabled gravity in the training environment. By doing so the agent does not have to
fight the additional force of gravity while still being able to learn the
relationship between the angular velocity and control outputs. In the real
world, as long as a minimum throttle value is mixed in to the output of the
\nn during flight~(\eg either manually by the
pilot or by configuring the firmware) such that
the net force is greater or equal to zero,
it will provide
stable flight.
Of course neglecting this force results in a less accurate representation of
the real world. However our immediate goal is to show transferability. In future work we plan to investigate 
alternative environments to teach the quadcopter to idle without sacrificing
real world dynamics.  
One possibility is to
include a quaternion $q$ defining the quadcopters orientation, and the current
throttle value as part of the aircraft state. Therefore the agent can be taught when $\textbf{T}=0$ and
$q=[x=0, y=0, z=0, w=1]$ (\ie no thrust and no rotation), to minimize the
output small enough to idle.

\begin{table}[]
    \centering
\begin{tabular}{l|c}
Hyperparameter            & Value          \\ \hline
Horizon (T)               & \rPpoHorizon  \\
Adam stepsize             & \rPpoStepsize  \\
Num. epochs               & \rPpoEpochs    \\
Minibatch size            & \rPpoMinibatch \\
Discount ($\gamma$)       & \rPpoDiscount  \\
GAE parameter ($\lambda$) & \rPpoGae      
\end{tabular}
\caption{\new{PPO hyperparameters where $\rho$ is linearly annealed over the
course of training from 1 to 0.}}
\label{tab:ppo}
\end{table}

\begin{figure}
\centering
{\includegraphics[trim=0 0 0
    0,clip,width=0.8\textwidth]{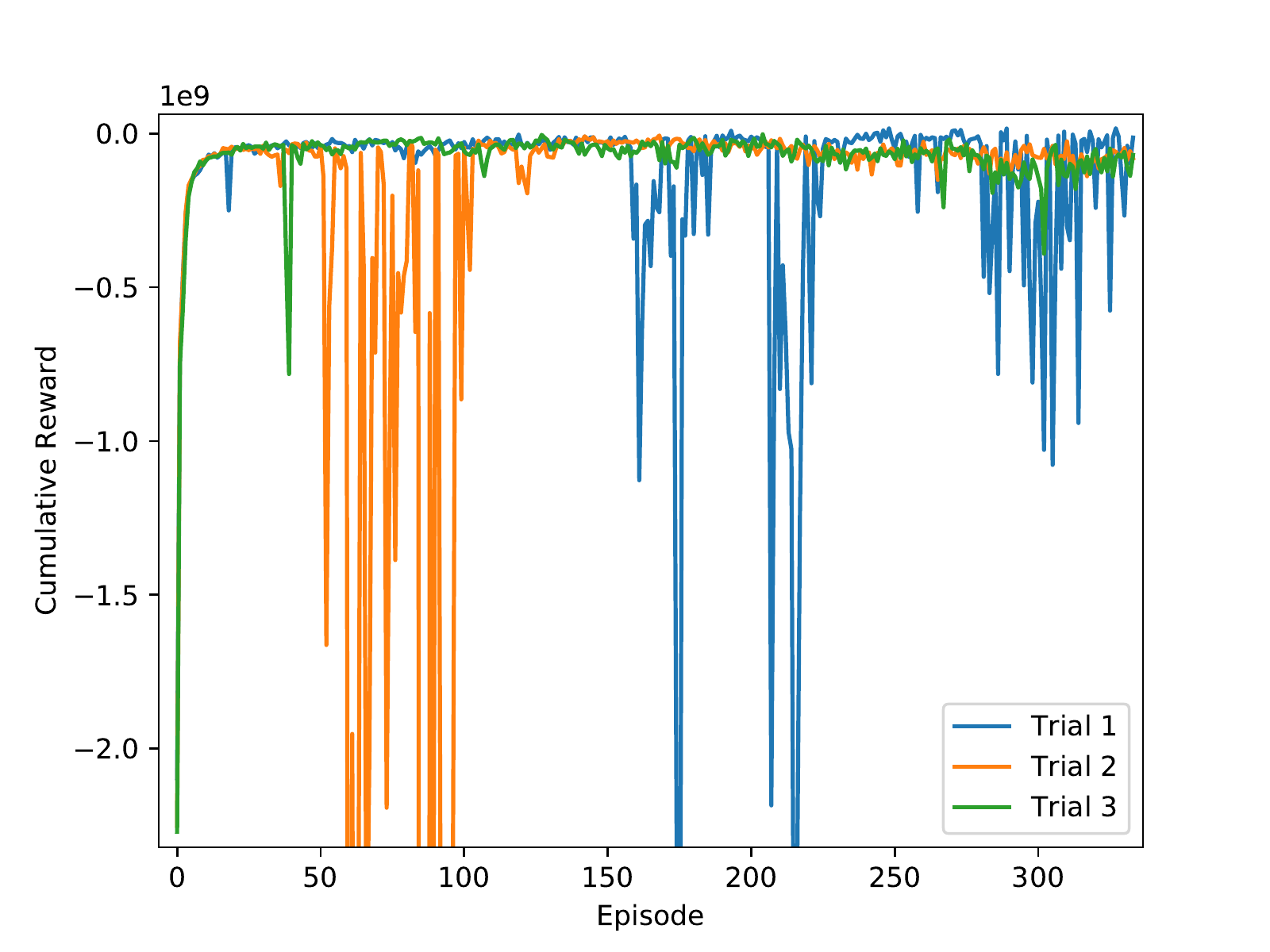}}
\caption{Cumulative rewards for each training episode.}
\label{fig:rewards}
\end{figure}

Each training task/episode ran for 30
seconds in simulation. The simulator is configured to take simulation steps
every 1ms which results in a total of 30,000 simulation steps per episode.
\new{Training ran for a total of 10 million time steps (333 episodes) on a desktop computer
    running Ubuntu 16.04 with an eight-core i7-7700 CPU and an NVIDIA GeForce GT
    730 graphics card which took approximately 11 hours. However training
    converged much earlier at around 1 million time steps (33 episodes) in just
    over an hour~(Fig.~\ref{fig:rewards}).}
We trained a total of three \nns which each used a different random seed for
the RL training algorithm and selected the \nn that received the highest
cumulative reward to use in \fc. Fig.~\ref{fig:rewards} shows a plot of the cumulative
rewards of each training episode for each of the \nns. 
The
plot illustrates how drastic training episodes can  vary simply due to the use
of a different seed.

The optimization stage reduced the frozen Tensorflow graph of the best
performing \nn by
\rOptGraphDecrease to a size of \rGraphSize.
The graph was compiled with Tensorflow version 1.8.0-rc1 and the firmware was compiled for the MATEKF722 target corresponding to the
manufacturer and model of our flight controller MATEKSYS Flight Controller
F722-STD. \new{Our flight controller uses the STM32F722RET6 microcontroller
with 512KB flash memory, and 256KB of SRAM.}

\new{
We inspected the \texttt{.text}, \texttt{.data} and \texttt{.bss} section
    headers of the firmware's ELF file to derive a lower bound of the memory
    utilization. These sections totalled 380KB, resulting in at least 74\% utilization
    of the flash memory. Graph optimization accounted for a reduction
    of 280B, all of which  was reduced from the \texttt{.text} section.
    Although in terms of memory utilization the  optimization stage was not
    necessary, this however will be more important for
    larger networks in the future.
    Comparing this to the parent project, Betaflight's sections totalled 375KB.

    Using Tensorflow's benchmarking tool we performed one million evaluations of the
    graph with and without optimization and found the optimization processes
    to reduce execution time on average by $1.1 \mu s$.

}

\subsection{Simulation Evaluation}
\begin{figure}
\centering
{\includegraphics[trim=25 0 70
    0,clip,width=\textwidth]{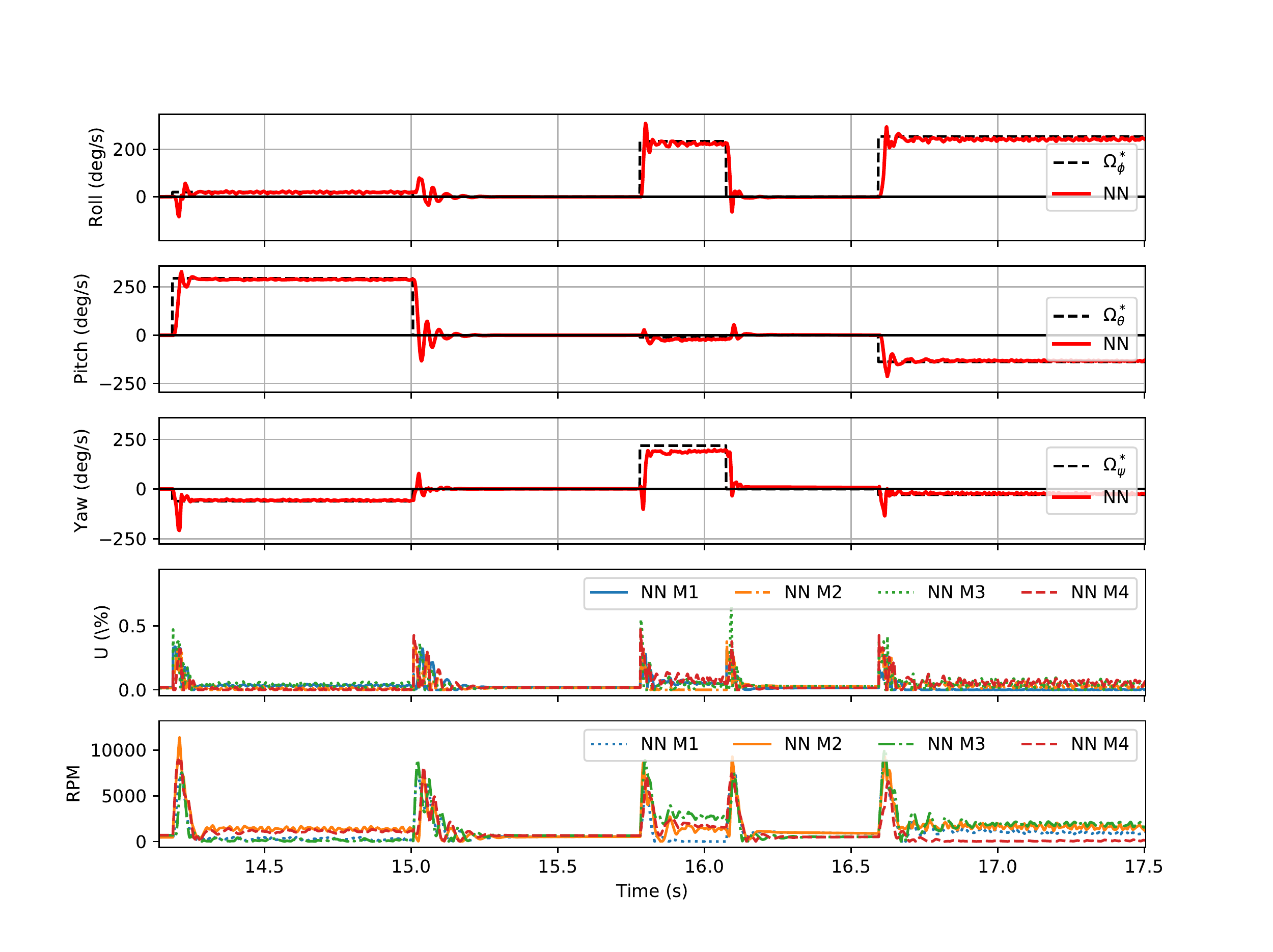}}
\caption{Simulation validation of trained \nn in \newgym training environment. Actual aircraft angular velocity is represented by the red line, while the desired angular velocity is the dashed black line. Control signal and motor velocity is also shown. }
\label{fig:sim_eval}
\end{figure}

In this section we validate the best performing \nn in simulation using the
\newgym environment. 
We execute the trained \nn for five episodes in the environment for a total of 2.5 minutes simulation time. 
A zoomed in portion of one of the episodes is illustrated in Figure~\ref{fig:sim_eval}.
This figure also displays the control signals generated by the \nn which is the value sent to the ESC.
Note this is a different representation than that used in Chapter~\ref{chapter:gymfc} 
which used PWM control signals. This is because the output must match that of the target flight control firmware for seamless transferability. If we compare the control output to that of the trained agent in 
Chapter~\ref{chapter:gymfc}, for example in Figure~\ref{fig:ppo2},  we can observe the impact this reward system has on reducing control signal values and oscillations.  

From these validation episodes we computed the average performance metrics in  
Table~\ref{tab:training_validation}. The controller does a decent job tracking the trajectory however it does suffer from overshooting the target particularly for the yaw axis which results in an increased error.

\begin{table}[]
\centering
\begin{tabular}{l|ccc|c|}
\cline{2-5}
                             & \multicolumn{4}{c|}{NN Controller (PPO)}                                                                 \\ \hline
\multicolumn{1}{|l|}{Metric} & Roll ($\phi$) & Pitch($\theta$) & Yaw ($\psi$)  & \cellcolor[HTML]{DAE8FC}Average       \\ \hline
\multicolumn{1}{|l|}{MAE}    & 12         & 10           & 21         & \cellcolor[HTML]{DAE8FC}14         \\
\multicolumn{1}{|l|}{MSE}    & 989        & 902          & 3,033      & \cellcolor[HTML]{DAE8FC}1,641      \\
\multicolumn{1}{|l|}{IAE}    & 12,557     & 10,491       & 21,711     & \cellcolor[HTML]{DAE8FC}14,919     \\
\multicolumn{1}{|l|}{ISE}    & 989,863    & 902,243      & 3,033,486  & \cellcolor[HTML]{DAE8FC}1,641,864  \\
\multicolumn{1}{|l|}{ITAE}   & 180,688    & 152,279      & 324,266    & \cellcolor[HTML]{DAE8FC}219,078    \\
\multicolumn{1}{|l|}{ITSE}   & 12,944,056 & 12,507,006   & 40,928,038 & \cellcolor[HTML]{DAE8FC}22,126,367 \\ \hline
\end{tabular}
\caption{Performance metric for NN training validation. Metric is reported for each individual axis, along with the average. Lower values are better.}
\label{tab:training_validation}
\end{table}

\subsection{Timing Analysis}
\label{sec:timing}

Running a flight control task with a fast control rate allows for the use of a
high speed ESC protocol, reducing write latency to the motors and thus resulting in higher precision flight. 
Therefore it is critical to analyze the
execution time of the neuro-flight control task so the optimal
control rate of the task can be determined. Once this is identified it can be
used to select which 
ESC protocol will provide the best performance. 
We collect timing data for \fc and compare
this to its parent project Betaflight. Times are taken for when the quadcopter is disarmed and also armed under
load for the control algorithm  (\ie evaluation of the \nn and PID
equation) and also the entire flight control task which in addition to the
control algorithm includes reading the gryo, reading the RC commands and
writing to the motors.
\begin{table}[]
    \centering
{\setlength{\tabcolsep}{0.2em}
\def\arraystretch{1.15}%
\begin{tabular}{ll|c|c|c|}
\cline{3-5}
                                                &             & WCET~($\mu s$)
& BCET~($\mu s$)                     & Var. Window (\%)         \\ \hline
\multicolumn{1}{|l|}{\multirow{2}{*}{Disarmed}} & \fc & \rTimeNfAlgDisarmedWCET & \rTimeNfAlgDisarmedBCET & \rTimeNfAlgDisarmedP \\ \cline{2-5} 
\multicolumn{1}{|l|}{}                          & Betaflight  & \rTimeBfAlgDisarmedWCET  & \rTimeBfAlgDisarmedBCET & \rTimeBfAlgDisarmedP \\ \hline
\multicolumn{1}{|l|}{\multirow{2}{*}{Armed}}    & \fc & \rTimeNfAlgArmedWCET    & \rTimeNfAlgArmedBCET    & \rTimeNfAlgArmedP    \\ \cline{2-5} 
\multicolumn{1}{|l|}{}                          & Betaflight  & \rTimeBfAlgArmedWCET    & \rTimeBfAlgArmedBCET    & \rTimeBfAlgArmedP    \\ \hline
\end{tabular}
}
\caption{Control algorithm timing analysis.}
\label{table:timing_alg}
\end{table}

We instrumented the firmware to calculate the timing measurement and
wrote the results to an unused serial port on the flight control
board.  Connecting to the serial port on the flight control board via
an FTDI adapter we are able to log the data on an external PC running
\texttt{minicom}. We recorded 5,000 measurements and report the
worst-case execution time~(WCET), best-case execution time~(BCET) and
the variability window in Table~\ref{table:timing_alg} for the control
algorithm and Table~\ref{table:timing_loop} for the control task.  The
variability window is calculated as the difference between the WCET
and BCET, normalized by the WCET, \ie
$(\text{WCET}-\text{BCET})/\text{WCET}$. This provides indication of
how predicable is the execution of the flight control logic, as it
embeds information about the relative fluctuation of execution
times. Two remarks are important with respect to the results in
Table~\ref{table:timing_alg}. First, the \nn  compared to PID
is about 14x slower (armed case), although the predictability of the
controller increases. It is important to remember that, while
executing the PID is much simpler than evaluating an \nn, our approach
allows removing additional logic that is required by a PID, such as motor mixing. Thus, a more meaningful comparison
needs to be performed by looking at the overall WCET and
predictability of the whole flight control task, which we carry out in
Table~\ref{table:timing_loop}. Second, because the \nn evaluation always
involve the same exact steps, an improvement in terms of
predictability can be observed under Neuroflight.

\begin{table}[]
    \centering
{\setlength{\tabcolsep}{0.2em}
\def\arraystretch{1.15}%
\begin{tabular}{ll|c|c|c|}
\cline{3-5}
                                                &             & WCET~$(\mu s)$
& BCET~$(\mu s)$           & Var. Window (\%)           \\ \hline
\multicolumn{1}{|l|}{\multirow{2}{*}{Disarmed}} & \fc & \rTimeNfLoopDisarmedWCET & \rTimeNfLoopDisarmedBCET & \rTimeNfLoopDisarmedP \\ \cline{2-5} 
\multicolumn{1}{|l|}{}                          & Betaflight  & \rTimeBfLoopDisarmedWCET & \rTimeBfLoopDisarmedBCET & \rTimeBfLoopDisarmedP \\ \hline
\multicolumn{1}{|l|}{\multirow{2}{*}{Armed}}    & \fc & \rTimeNfLoopArmedWCET    & \rTimeNfLoopArmedBCET    & \rTimeNfLoopArmedP    \\ \cline{2-5} 
\multicolumn{1}{|l|}{}                          & Betaflight  & \rTimeBfLoopArmedWCET    & \rTimeBfLoopArmedBCET    & \rTimeBfLoopArmedP    \\ \hline
\end{tabular}
}
\caption{Flight control task timing analysis.}
\label{table:timing_loop}
\end{table}

The timing analysis reported in Table~\ref{table:timing_loop} reveals
that the neuro-flight control task has a WCET of \rTimeNfLoopArmedWCET
$\mu s$ which would allow for a max execution rate of \rMaxNNTaskFreq.
However in \fc (and in Betaflight), the flight control task frequency
must be a division of the gyro update frequency, thus with
\rMaxGyroHertz gyro update and a denominator of \rLoopDenom, the
neuro-flight control task can be configured to execute at
\rMaxLoopHertz. To put this into perspective this is \rFasterThanPX
times faster\footnote{According to the default loop rate of 250Hz.}
than the popular PX4 firmware~\cite{meier2015px4}.

Furthermore this control rate is \rFastThanPWM times faster than the traditional PWM ESC
protocol used by commercial quadcopters (50Hz \cite{abdulrahim2019defining}) 
thereby allowing us to configure \fc  to use the ESC protocol DShot600 which has
a max frequency of 37.5kHz ~\cite{looptime}.   

\begin{figure*}
\centering
{\includegraphics[trim=50 0 70
    0,clip,width=\textwidth]{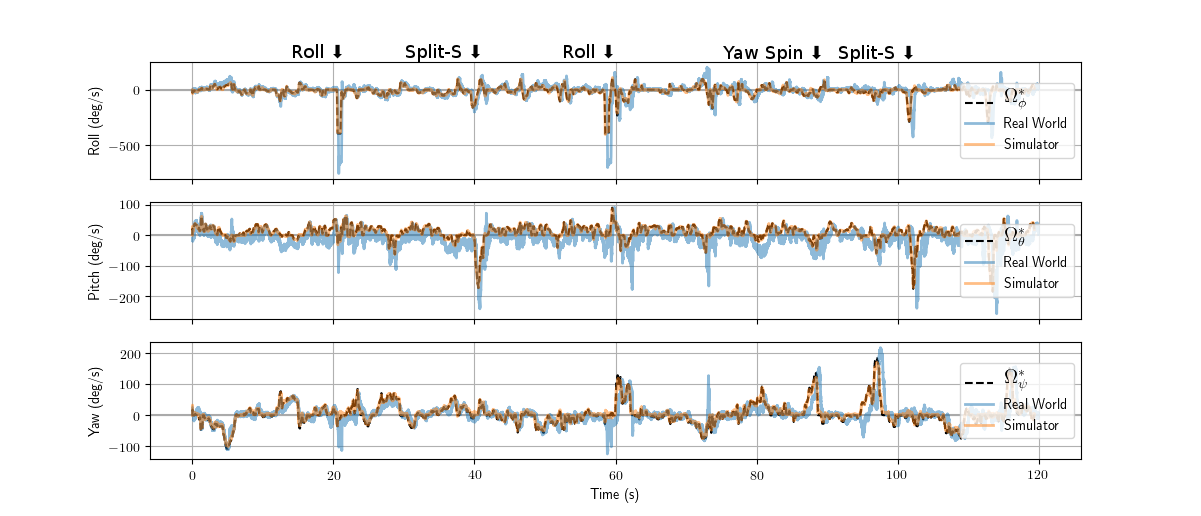}}
\caption{Flight test log demonstrating \fc tracking a desired angular velocity
    in the {real world} compared to
    in simulation. Maneuvers during this flight are annotated. }
\label{fig:flight}
\end{figure*}

Given the simplicity of the PID algorithm it 
came as no surprise that the Betaflight flight control task is faster, yet this is only by a
factor of \rSlowerThanBF when armed. As we can see comparing  
Table~\ref{table:timing_alg} to Table~\ref{table:timing_loop} the additional
subprocesses tasks  are the bottleneck of the
Betaflight flight control ask. 
However referring to the variability window, the \fc control algorithm
and control task are far more stable than Betaflight. The Betaflight
flight control task exhibits little predictability when armed.

Recent research has shown there is no measurable improvements for control task
loop rates that are faster than 4kHz~\cite{abdulrahim2019defining}.
Our timing analysis has shown that \fc is close of this goal.  To
reach this goal there are three approaches we can take: (1) Support
future microcontrollers with faster processor speeds, (2) experiment
with different \nn architectures to reduce the number of arithmetic
operations and thus reduce the computational time to execute the \nn,
and (3) optimize the flight control sub tasks to reduce the flight
control task's WCET and variability window. In future work we
immediately plan to explore (2) and (3), results obtained in these
directions would not depend on the specific hardware used in the final
assembly.

\begin{figure}
\centering
\begin{subfigure}[b]{0.8\columnwidth}
    {\includegraphics[trim=0 0 20 20,clip, width=\textwidth]{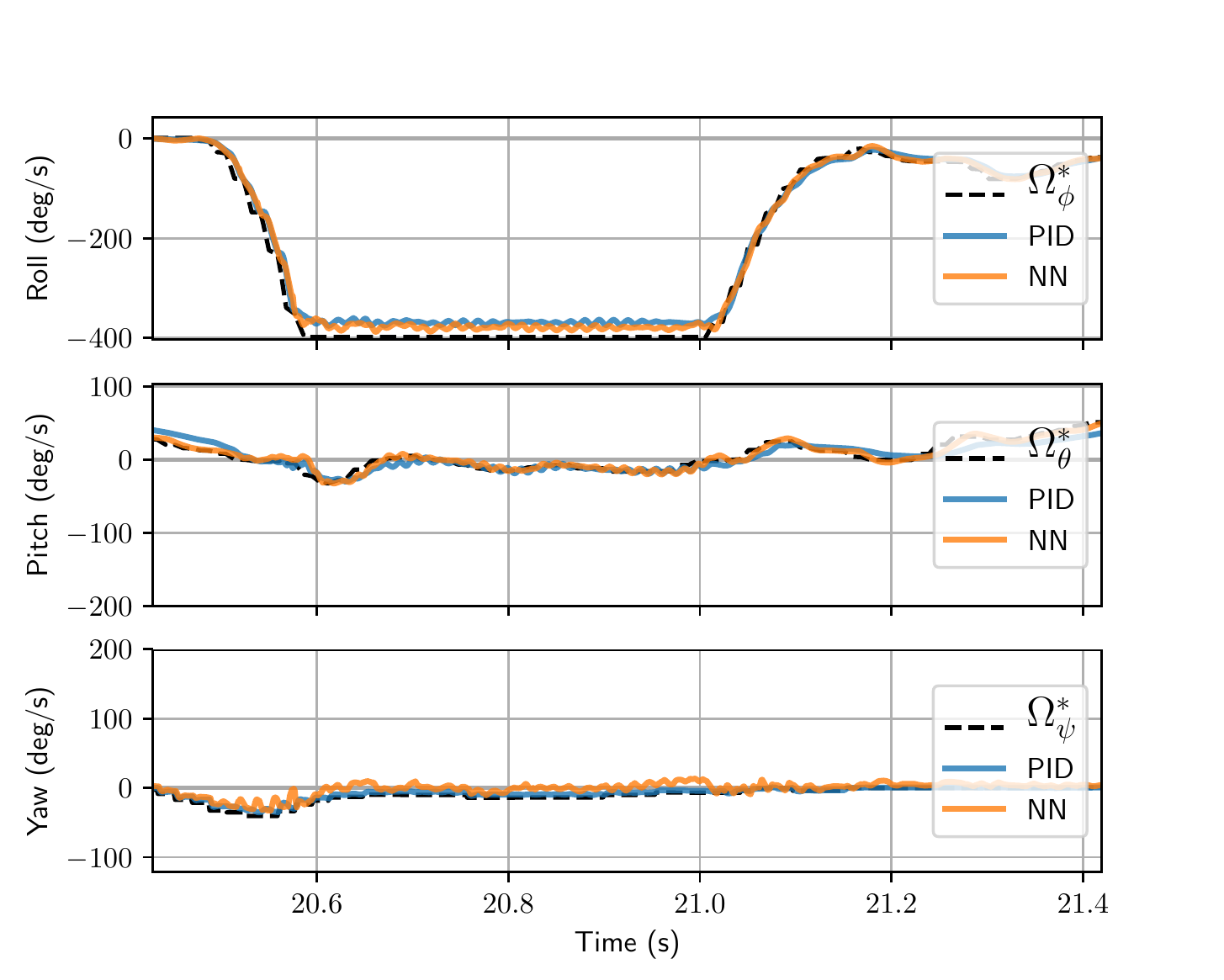}}
    \caption{\new{Roll}}
    \label{fig:roll}
\end{subfigure}
\\
\begin{subfigure}[b]{0.8\columnwidth}
    {\includegraphics[trim=0 0 20 20,clip,
        width=\textwidth]{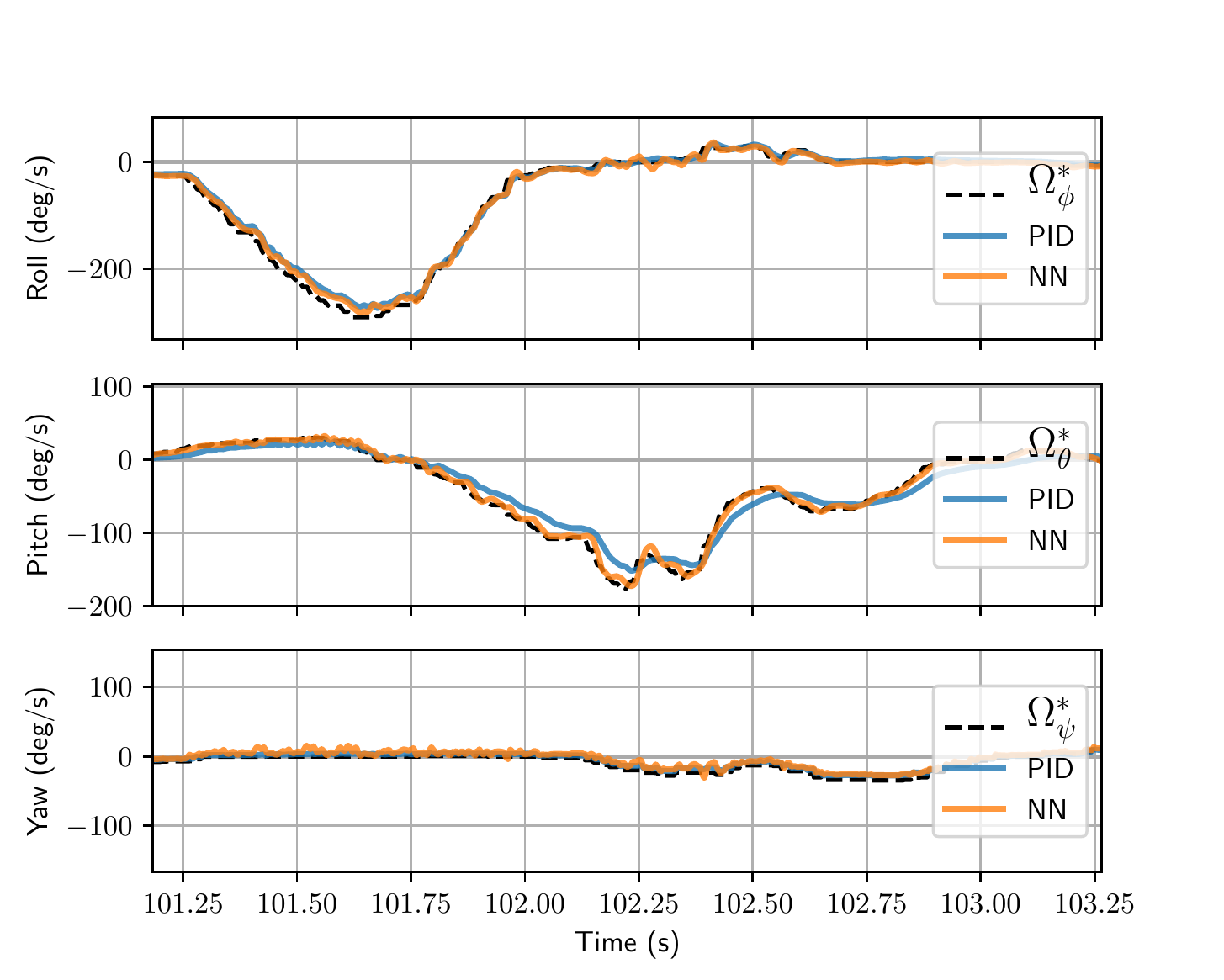}}
    \caption{\new{Split-S} }
    \label{fig:splits}
\end{subfigure}
\caption{\new{Performance comparison of the \nn} \new{ controller versus a PID controller tracking a desired angular velocity
    in  simulation to execute the Split-S and roll aerobatic maneuvers.}} 
\label{fig:zoom}
\end{figure}

\subsection{Power Analysis}
\label{sec:power}
\new{The flight controller affects power consumption directly and indirectly.
    The direct power draw is a result of the execution of the control
    algorithm/task, while the indirect power draw is due to the generated control
    signals which determines the amount of power the ESC will draw.} 
    
\new{As a first attempt to understand and compare the power consumption of a \nn} \new{based controller  to a standard PID controller, we performed a
    static power analysis. For \aircraft} \new{running Neuroflight, we connected a multimeter
    inline with the battery power supply to measure the current
    draw and report the measurements for both when the quadcopter is
    disarmed~(direct power consumption) 
    and armed idling~(indirect power consumption), similarly done to our timing
    analysis. We then take the same measurements
    for the \aircraft} \new{running Betaflight~(PID control). Results reported in
    Table~\ref{table:power} show there is no change using the \nn} \new{based
    controller in regards to direct power draw of the control algorithm. This result
    was expected as the flight control firmware does not execute sleep
    instructions. However for the indirect power draw, there is a measurable
    70mA~(approximately 11\%) increase in
    current draw for the \nn} \new{controller. 
It is important to remember this particular \nn} \new{controller has been trained to
optimize its ability to track a desired angular velocity. Thus the increase in
current draw does not come as a surprise as the control signals will be
required to switch
quickly to maintain the set point which results in increased current draw.}

\new{An advantage a \nn} \new{
controller has over a traditional PID controller is that it has the ability to
optimize its performance based on a number of conditions and characteristics, such as power
consumption.  In the future we will investigate alternative optimization goals
for the controller and instrument \aircraft} \new{ with sensors to record power
consumption in flight to perform a thorough power analysis.
}

\begin{table}[]
    \centering
{\setlength{\tabcolsep}{0.2em}
\def\arraystretch{1.15}%
\begin{tabular}{ll|c|c|c|}
\cline{3-5}
                                                &    & Voltage (V)   & Current (A)     & Power (W)        \\ \hline
\multicolumn{1}{|l|}{\multirow{2}{*}{Disarmed}} & \fc & \powerVoltage & \ampsNFdisarmed & \powerNFdisarmed \\ \cline{2-5} 
\multicolumn{1}{|l|}{}                          & Betaflight & \powerVoltage & \ampsBFdisarmed & \powerBFdisarmed \\ \hline
\multicolumn{1}{|l|}{\multirow{2}{*}{Armed}}    & \fc & \powerVoltage & \ampsNFarmed    & \powerNFarmed    \\ \cline{2-5} 
\multicolumn{1}{|l|}{}                          & Betaflight & \powerVoltage & \ampsBFarmed    & \powerBFarmed    \\ \hline
\end{tabular}
}
\caption{\new{Power analysis of  Neuroflight compared to Betaflight.}}
\label{table:power}
\end{table}

\subsection{Flight Evaluation}
\label{sec:nf:flighteval}

To test the performance of \fc we had an experienced drone racing
pilot conduct five test flights for us. The FPV videos of the test flights can be viewed
at \videourl. A still image extracted from the
FPV video feed shows the view point of the pilot of one of the test
flights can be seen in Fig.~\ref{fig:nf1_fpv}. In FPV flying the
aircraft has a camera which transmits the analog video feed back to
the pilot who is wearing goggles with a monitor connected to a video
receiver. This allows the pilot to control the aircraft from the
perspective of the aircraft.

\fc supports real-time logging
during flight allowing us to collect gyro and RC command data to analyze how
well the neuro-flight controller is able to track the desired angular velocity. 
We asked the pilot to fly a mix of basic maneuvers such as loops and figure
eights and advanced maneuvers such as rolls, flips, dives and the Split-S. To execute
a Split-S the pilot inverts the quadcopter and descends in a half loop dive, exiting the loop so they are flying in the opposite horizontal direction. 
Once we collected the flight logs we played the desired angular rates back to
the \nn in the \newgym environment to evaluate the performance in simulation.
This allows the performance gap between the two environments to be measured and
identify the reality gap.
Comparison between the simulated and {real world} performance for one of the
test flights is illustrated in
{Fig.}~\ref{fig:flight} while specific maneuvers that occur during this test
flight are annotated. 
\new{
We computed various error metrics for the flights  including the
Mean Absolute Error~(MAE), 
and 
Mean Squared Error~(MSE),
as well as the discrete form of the
Integral Absolute Error~(IAE), 
Integral Squared Error~(ISE), 
Integral Time-weighted Absolute Error~(ITAE),
and
Integral Time-weighted Squared Error~(ITSE).
These values are reported 
 in Table~\ref{tab:real_flight} are an average for the real flights
and in Table~\ref{tab:sim_flight_metrics_nn} for the simulated flight .
}

\newcommand\realUMu{$0.27 \pm 1e-4$\xspace}
\newcommand\realUDiffMu{$0.01 \pm 5.5e-5$\xspace}

\def\meta{15\xspace}
\def\metb{21\xspace}
\def\metc{13\xspace}
\def\metd{16\xspace}
\def\mete{1,720\xspace}
\def\metf{1,860\xspace}
\def\metg{686\xspace}
\def\meth{1,422\xspace}
\def\meti{15,176\xspace}
\def\metj{21,160\xspace}
\def\metk{13,478\xspace}
\def\metl{16,605\xspace}
\def\metm{1,711,764\xspace}
\def\metn{1,851,450\xspace}
\def\meto{682,914\xspace}
\def\metp{1,415,376\xspace}
\def\metq{705,614\xspace}
\def\metr{1,001,476\xspace}
\def\mets{638,513\xspace}
\def\mett{781,868\xspace}
\def\metu{98,725,074\xspace}
\def\metv{90,438,678\xspace}
\def\metw{37,397,559\xspace}
\def\metx{75,520,437\xspace}

\begin{table}[]
    \centering
{\setlength{\tabcolsep}{0.2em}
\def\arraystretch{1.15}%
\begin{tabular}{l|ccc|c|}
\cline{2-5}
                             & \multicolumn{4}{c|}{NN Controller (PPO)}                                                \\ \hline
\multicolumn{1}{|l|}{Metric} & Roll ($\phi$) & Pitch($\theta$) & Yaw ($\psi$)  & \cellcolor[HTML]{DAE8FC}Average       \\ \hline
\multicolumn{1}{|l|}{MAE}    & \meta         & \metb           & \metc         & \cellcolor[HTML]{DAE8FC}\metd         \\
\multicolumn{1}{|l|}{MSE}    & \mete      & \metf          & \metg        &
\cellcolor[HTML]{DAE8FC}\meth        \\
\multicolumn{1}{|l|}{IAE}    & \meti     & \metj       & \metk     &
\cellcolor[HTML]{DAE8FC}\metl     \\
\multicolumn{1}{|l|}{ISE}    & \metm  & \metn      & \meto    &
\cellcolor[HTML]{DAE8FC}\metp    \\
\multicolumn{1}{|l|}{ITAE}   
& \metq    & \metr    & \mets    & 
\cellcolor[HTML]{DAE8FC}\mett    \\
\multicolumn{1}{|l|}{ITSE}   & \metu & \metv   & \metw &
\cellcolor[HTML]{DAE8FC}\metx \\ \hline
\end{tabular}
}
\caption{\new{Error metrics of the NN controller from \numFlights flight in the real
world. Metrics are reported for each individual axis, along with the average.
Lower values are better.}}
\label{tab:real_flight}
\end{table}

\newcommand\simUMu{$0.037 \pm 1e-4$\xspace}
\newcommand\simUDiffMu{$0.007 \pm 1e-4$\xspace}

\begin{table}[]
\centering
{\setlength{\tabcolsep}{0.2em}
\def\arraystretch{1.15}%
\begin{tabular}{l|ccc|c|}
\cline{2-5}
                             & \multicolumn{4}{c|}{NN Controller (PPO)}                                              \\ \hline
\multicolumn{1}{|l|}{Metric} & Roll ($\phi$) & Pitch($\theta$) & Yaw ($\psi$) & \cellcolor[HTML]{DAE8FC}Average      \\ \hline
\multicolumn{1}{|l|}{MAE}    & 3          & 2            & 4         &
\cellcolor[HTML]{DAE8FC}3         \\
\multicolumn{1}{|l|}{MSE}    & 23         & 6            & 27         &
\cellcolor[HTML]{DAE8FC}19        \\
\multicolumn{1}{|l|}{IAE}    & 2,888      & 1,523        & 4,072     & \cellcolor[HTML]{DAE8FC}2,827     \\
\multicolumn{1}{|l|}{ISE}    & 23,227     & 5,589        & 27,203    &
\cellcolor[HTML]{DAE8FC}18,673    \\
\multicolumn{1}{|l|}{ITAE}   & 179,945    & 93,339       & 261,947   & \cellcolor[HTML]{DAE8FC}178,410   \\
\multicolumn{1}{|l|}{ITSE}   & 1,499,076  & 369,577      & 1,893,954 & \cellcolor[HTML]{DAE8FC}1,254,202 \\ \hline
\end{tabular}
}
\caption{Error metrics for simulation playback using \nn controller. Metric is reported for each individual axis, along with the average. Lower values are better.}
\label{tab:sim_flight_metrics_nn}
\end{table}

\begin{table}[]
\centering
{\setlength{\tabcolsep}{0.2em}
\def\arraystretch{1.15}%

\begin{tabular}{l|ccc|c|}
\cline{2-5}
                             & \multicolumn{4}{c|}{PID}                                                              \\ \hline
\multicolumn{1}{|l|}{Metric} & Roll ($\phi$) & Pitch($\theta$) & Yaw ($\psi$) & \cellcolor[HTML]{DAE8FC}Average      \\ \hline
\multicolumn{1}{|l|}{MAE}    & 4          & 5            & 3         & \cellcolor[HTML]{DAE8FC}4         \\
\multicolumn{1}{|l|}{MSE}    & 35         & 46           & 21        &
\cellcolor[HTML]{DAE8FC}34        \\
\multicolumn{1}{|l|}{IAE}    & 3,905     & 5,258        & 3,423     & \cellcolor[HTML]{DAE8FC}4,195     \\
\multicolumn{1}{|l|}{ISE}    & 34,811     & 45,590       & 20,549    & \cellcolor[HTML]{DAE8FC}33,650    \\
\multicolumn{1}{|l|}{ITAE}   & 236,408    & 320,205      & 217,343   & \cellcolor[HTML]{DAE8FC}257,985   \\
\multicolumn{1}{|l|}{ITSE}   & 2,100,576  & 2,927,031    & 1,419,391 & \cellcolor[HTML]{DAE8FC}2,148,999 \\ \hline
\end{tabular}
}
\caption{Error metrics for simulation playback using PID controller. Metric is reported for each individual axis, along with the average. Lower values are better.}
\label{tab:sim_flight_metrics_pid}
\end{table}

As we can see there is a considerable increase in error (16 degrees per second
on average) transferring from
simulation from reality, however this was expected because the digital twin does not perfectly
model the real system.
\new{There is a large increase in error for the integral
    measurements.
    A partial explanation for this is if we refer to  {Fig.}~\ref{fig:flight} (particularly the pitch axis) we can
    see the controller is consistently off by about 10
    degrees which will continually add error to these measurements. }
Additionally the difference in squared errors is quite significant which will
emphasize larger errors that occur.
The increased error on the pitch axis appears to be due to the
differences in frame shape between the digital twin and real
quadcopter, which are both asymmetrical but in relation to a different
axis.
This discrepancy may have resulted in pitch control
lagging in the real world as more torque and power is required to pitch in our
real quadcopter.   

We also compared the average absolute difference in the control
signals~($\overline{|\Delta u|}$) between
the two worlds.  In simulation we found this to be \simUDiffMu, while in the
real world there was a minor increase to \realUDiffMu but we found this did not result
in any
harm to the motors such as a noticeable increase in output oscillations or
heat being generated.

A more accurate digital twin model can boost accuracy. 
Furthermore, during this particular flight wind gusts exceeded 30mph, while in the
simulation world there are no external disturbances acting upon the aircraft.
In the future we plan to deploy an array of sensors to measure wind speed so we
can correlate wind gusts with excessive error. 
Nonetheless, as shown in the video, stable flight can be maintained
demonstrating the transferability of a \nn trained with our approach.

\textbf{PID vs \nn Control.} Next we performed an experiment to
compare the performance of the \nn controller used in \fc to a PID
controller in simulation using the \newgym environment.  
\new{Although other control algorithms may exist in literature that out perform
    PID, of the open source flight controllers available for
    benchmarking, every single one uses PID~\cite{ebeid2018survey}.
    A major contribution of this work is providing the research community an
    additional flight control algorithm for benchmarking.
}

The PID
controller was tuned in simulation using the classical
Ziegler-Nichols method~\cite{ziegler1942optimum} and then manually
adjusted to \new{reduce overshoot} to obtained the following gains for
each axis of rotation: $K_\phi = [0.032029, 0,0.000396 ]$, $K_\theta =
[0.032029, 0, 0.000396 ]$, $K_\psi = [0.032029, 0, 0]$, where
$K_\text{axis} = [K_p,K_i,K_d]$ for each proportional, integrative,
and derivative gains, respectively.
\new{It took  approximately a half hour to manually tune the 9 gains with the
    bottleneck being the time to execute the simulator in order to obtain the
    parameters to calculate Ziegler-Nichols. In comparison to training a \nn}
\new{via PPO, there is not a considerable overhead difference given this is an
    offline task. In fact the tuning rate by PPO is significantly
    faster by a factor of 75.}

The RC commands from the real test flight where then replayed back to the
simulator similar to the previous experiment, however this time using the tuned PID controller. A 
\new{zoomed in comparison of
the \nn} \new{ and PID controller tracking the desired angular velocity for
two aerobatic maneuvers is
shown in Fig.~\ref{fig:zoom}. Although the performance is quite close,
we can most visibly the \nn} \new{controller tracking  the pitch axis during a
Split-S maneuver more accurately. }

\new{We also computed the same control measurements for the PID controller and
reported them in Table~\ref{tab:sim_flight_metrics_pid}. Results show, on average, the
\nn} \new{controller to outperform the PID controller for every one of our metrics. }

It is important to note PID tuning is a challenging task and the PID
controller's accuracy and ability to control the quadcopter is only as
good as the tune. The \nn controller on the other hand did not require
any manually tuning, instead through RL and interacting with the
aircraft over time it is able to teach itself attitude control.  As
we continue to the reduce the gap between simulation and the real
world, the performance of the \nn controller will continue to improve
in the real world.

\section{Future Work and Conclusion}
\label{sec:nf:con}
In this chapter we introduced \fc, the first open-source neuro-flight control
firmware for  multicopters and fixed wing aircraft and its
accompanying toolchain. %
There are \new{four} main directions we plan to
pursue in future work.
\begin{enumerate}

\item \textbf{Digital twin development.} In this work we synthesized our \nn
using an existing quadcopter model that did not match NF1. Although stable flight was achieved demonstrating the \nns
robustness, comparison between the simulated flight verse the actual flight
is evidence inaccuracies in the digital twin has a negative affect in flight control accuracy.  
In future work we will develop an accurate digital twin of NF1
and investigate how the fidelity
of a digital twin affects flight performance in an effort to reduce costs during
development. 
\item \textbf{Adaptive and predictive control.} With a stable platform in place
we can now begin to harness the \nn's true potential.  We will enhance the
training environment to teach adaptive control to account for excessive sensor noise, voltage
sag, change in flight dynamics due to high throttle input, payload changes, external disturbances such as
wind, and propulsion system failure. 
\item \textbf{Continuous learning.} Our current approach trains \nns exclusively using offline learning. However, in order to 
reduce the performance gap between the simulated and {real world},
we expect that a hybrid architecture involving online incremental
learning will be necessary. % to provide continuouslearning.
Online learning will allow the aircraft to adapt, in real-time, and
to compensate for any modelling errors that existed during synthesis of
the \nn during offline (initial) training.
Given the payload restrictions of micro-UAVs and weight associated
with hardware necessary for online learning we will investigate
methods to off-load the computational burden of incremental learning
to the cloud.
\item \textbf{\nn architecture development.}
     Several performance benefits can be realized from an optimal
         network architecture for flight control including improved
         accuracy~(Section~\ref{sec:nf:flighteval}) and
         faster execution~(Section~\ref{sec:timing}). In future work we plan to
         explore recurrent architectures utilizing long short-term memory~(LSTM)
         to improve accuracy. Additionally we will investigate alternative
         distributions such as the beta function which is naturally bounded~\cite{chou2017improving}.
         Furthermore we will explore the use of the rectified linear unit~(ReLU) activation functions to increase execution
         time which is more computationally efficient than the hyperbolic tangent 
         function. 
\end{enumerate}
The economic costs associated with developing neuro-flight control
will foreshadow its future, determining whether its use will remain
confined to special purpose applications, or if it will be adopted in
mainstream flight control architectures.
Nonetheless, we strongly believe that \fc is a major milestone in
neuro-flight control and will provide the required foundations for
next generation flight control firmwares.

\cleardoublepage

\definecolor{dkgreen}{rgb}{0,0.6,0}
\definecolor{gray}{rgb}{0.5,0.5,0.5}
\definecolor{mauve}{rgb}{0.58,0,0.82}
\definecolor{gray}{rgb}{0.4,0.4,0.4}
\definecolor{darkblue}{rgb}{0.0,0.0,0.6}
\definecolor{lightblue}{rgb}{0.0,0.0,0.9}
\definecolor{cyan}{rgb}{0.0,0.6,0.6}
\definecolor{darkred}{rgb}{0.6,0.0,0.0}

\lstset{
  basicstyle=\ttfamily\footnotesize,
  columns=fullflexible,
  showstringspaces=false,
  %numbers=left,                   % where to put the line-numbers
  %numberstyle=\tiny\color{gray},  % the style that is used for the line-numbers
  %stepnumber=1,
  %numbersep=5pt,                  % how far the line-numbers are from the code
  backgroundcolor=\color{white},      % choose the background color. You must add \usepackage{color}
  showspaces=false,               % show spaces adding particular underscores
  showstringspaces=false,         % underline spaces within strings
  showtabs=false,                 % show tabs within strings adding particular underscores
  frame=none,                   % adds a frame around the code
  rulecolor=\color{black},        % if not set, the frame-color may be changed on line-breaks within not-black text (e.g. commens (green here))
  tabsize=2,                      % sets default tabsize to 2 spaces
  captionpos=b,                   % sets the caption-position to bottom
  breaklines=true,                % sets automatic line breaking
  breakatwhitespace=false,        % sets if automatic breaks should only happen at whitespace
  title=\lstname,                   % show the filename of files included with \lstinputlisting;
                                  % also try caption instead of title
  commentstyle=\color{gray}\upshape
}

\lstdefinelanguage{XML}
{
  morestring=[s][\color{mauve}]{"}{"},
  morestring=[s][\color{black}]{>}{<},
  morecomment=[s]{<?}{?>},
  morecomment=[s][\color{dkgreen}]{<!--}{-->},
  stringstyle=\color{black},
  identifierstyle=\color{lightblue},
  keywordstyle=\color{red},
  morekeywords={xmlns,xsi,noNamespaceSchemaLocation,type,id,x,y,source,target,version,tool,transRef,roleRef,objective,eventually}% list your attributes here
}

\newcommand\pluginsource{\cite{gymfcplugins}\xspace}
\newcommand\plugin{GymFC plugin\xspace}
\newcommand\simcontrol{simulation controller\xspace}

\newcommand\maxrpm{25042 RPM\xspace}
\newcommand\maxQ{0.0565 $\pm$ 0.0008 N $\cdot$ m\xspace}
\newcommand\maxT{6.59 $\pm$ 0.09 N\xspace}
\newcommand\ct{$2.87\times 10^{-2}$\xspace}
\newcommand\cq{$1.38\times 10^{-3}$\xspace}
\newcommand\kt{$9.37 \times 10^{-7}$\xspace}
\newcommand\kq{$8.64 \times 10^{-3}$\xspace}
\newcommand\kp{0.0001\xspace}
\newcommand\ki{0\xspace}
\newcommand\kd{0\xspace}
\newcommand\fmin{-0.1\xspace}
\newcommand\fmax{0.05\xspace}

\newcommand\fminCW{-0.05\xspace}
\newcommand\fminCCW{-0.1\xspace}
\newcommand\fmaxCW{0.1\xspace}
\newcommand\fmaxCCW{0.05\xspace}

\newcommand\torqueMAE{0.003 N $\cdot$ m\xspace} % or 5% of average
\newcommand\thrustMAE{0.588 N\xspace} % or 9% of average
\newcommand\velocityMAE{XXX\xspace}

\newcommand\rpmTwoFiveMAEPercent{4.11\%\xspace}
\newcommand\rpmFiveZeroMAEPercent{3.51\%\xspace} 
\newcommand\rpmSevenFiveMAEPercent{3.31\%\xspace}
\newcommand\rpmFullMAEPercent{3.90\%\xspace}

\newcommand\odeUnstableRate{XXX\xspace}
\newcommand\odeMaxDrift{XXX\xspace}

\newcommand\instableODEtwo{$\Omega=[-87, 85, 147]$\xspace}
\newcommand\instableODEone{$\Omega=[-263, 269, 364]$\xspace}
\newcommand\instableODEfive{$\Omega=[-617, 428, 693]$\xspace}

\newcommand\instableODETwoMax{95mm\xspace}
\newcommand\instableODEoneMax{39mm\xspace}
\newcommand\instableODEFiveMax{10mm\xspace}

\newcommand\gyroBins{11\xspace}
\newcommand\gyroTime{37 seconds\xspace}
\newcommand\gyroN{26,777\xspace}

\newcommand\gyroRollMu{-0.2546\xspace}
\newcommand\gyroRollSigma{1.3373\xspace}

\newcommand\gyroPitchMu{0.2419\xspace}
\newcommand\gyroPitchSigma{0.9990\xspace}

\newcommand\gyroYawMu{0.079\xspace}
\newcommand\gyroYawSigma{1.4516\xspace}

\newcommand\gyroCI{0.01\xspace}

\newcommand\cmdN{786,022\xspace}
\newcommand\cmdBins{100\xspace}
\newcommand\cmdAve{-2.3 deg/s\xspace}
\newcommand\cmdStd{12.4 deg/s\xspace}

\newcommand\ppoUMu{$0.12 \pm 0.01$\xspace}
\newcommand\ppoUDiffMu{$0.02 \pm 0.01$\xspace}

\newcommand\pidUMu{$0.03 \pm 0.019$\xspace}
\newcommand\pidUDiffMu{$0.04 \pm 0.02$\xspace}
\newcommand\realFlightCount{7\xspace}
\newcommand\realFlightTime{8.5 minutes\xspace}
\def\realUMu{$0.30 \pm 2e-4$\xspace}
\newcommand\realTwoUDiffMu{$0.15 \pm 4e-4$\xspace}

\newcommand\playbackTwoUMu{$0.30 \pm 2e-4$\xspace}
\newcommand\playbackTwoUDiffMu{$0.08 \pm 2e-4$\xspace}

\graphicspath{ {4_Twin/figures/} }
\chapter{Aircraft Modelling for \textit{In Silico} Neuro-flight Controller Synthesis}
\label{chapter:twin}
\thispagestyle{myheadings}
Tuning controllers \textit{in silico} (\ie in simulation) has numerous advantages over tuning in
the real world. It is cost effective, runs faster than real time, allows for
rapid prototyping and testing, and it is easily automated. Additionally, the controller can be exposed to environments and conditions that would otherwise be difficult and expensive to do in the real world~(\eg part failure, extreme weather, etc). 
Unfortunately it can be very challenging to obtain the same level of performance from the controller when  transferred to hardware operating in the real world.
This is primarily due to the simulator failing to capture all of the dynamics
in the controller's  real world operating environment.
To provide seamless transferability to hardware, the ultimate goal would be to
eliminate the reality gap. But the world is a highly complex place with many unknowns. Modelling  the known dynamics can require an extraordinary level of computation. 

Several methods have been proposed to aid in transferring the \nn trained in
simulation to the real
world such as sampling data from the real world environment, and integrating it
into the simulation environment~\cite{jakobi1995noise}.  Additionally, injecting
noise and domain randomizing have also been shown to improve performance in the
real world~\cite{tobin2017domain,andrychowicz2018learning,molchanov2019sim}. The idea
behind these techniques is to train the \nn on copious variations of the
environment such that the actual real world just appears as another variation to
the \nn. This essentially blurs the reality gap for the controller.

To further improve performance,  an ideal control system would, in addition,
provide online tuning to account for unknown dynamics found in the real world. 
These tuning strategies form building blocks for hierarchical tuning (learning) frameworks. 
However before online tuning can be utilized, the controller  must first be
tuned \textit{in silico}  well enough to operate  in the real world.

In Chapter~\ref{chapter:nf} our quadcopter achieved stable flight in the real world with a
\nn-based controller trained by \newgym via RL. However there was a significant, but expected, gap between the performance observed in simulation compared to the real world due to the inaccuracies in the aircraft model used during training. In this chapter 
we propose our methodology for creating a
digital twin for a multicopter and use this methodology to create a digital twin of
our aircraft, \aircraft.
In summary this chapter makes the following contributions:

\begin{itemize}
\item \gymfcTwo, a universal flight control tuning framework. As a prerequisite
to creating the digital twin, it was first necessary to revise \newgym to
easily support any aircraft. This update provides a framework for tuning any
control algorithm, not just \nn-based flight controllers. In addition to using
the framework for training neuro-flight controller policies, we also demonstrate its
modular design implementing a dynamometer simulation for validating motor
performance, and a PID tuning platform.

\item A methodology for creating multicopter digital twins.
We outline, from the ground up, how to create a digital twin of a multicopter.
This consists of creating the rigid bodies  and modelling motor dynamics.
To measure performance characteristics of our propulsion system we develop a
dynamometer for collecting rotor velocity, thrust and torque measurements.  

\item Propulsion system modelling enhancements. Building upon the PX4 SITL motor
models~\cite{px4sitl}, this work 
introduces enhancements to modelling motor response and throttle curves.
These models have been ported to the \gymfcTwo framework and have been made open source available from \cite{gymfcplugins}.

\item A simulation stability analysis. Multicopters are extremely agile, due to
having full rotational range of motion, independent of translational motion.
Aggressive angular velocity maneuvers are subjected to high centripetal
forces, in simulation, and also in the real world. However in simulation,
significant forces can introduce simulation instabilities. 
In this work we introduce a tool for measuring model stability in simulation and
compare these results using two different physics engines used by the
Gazebo simulator. 

\item Implementation of \gymfcTwo for synthesizing neuro-flight
    controllers trained via RL. We propose our user modules consisting of a new 
    environment and reward function to further reduce errors and aid in
    transferring the trained policy to the real world.  

\item Evaluation of a neuro-flight controller synthesized with its digital twin.
    We first evaluate the neuro-flight controller
    in simulation and find it to exceed the performance of a PID controller, in
    regards to minimizing error, and also having a larger flight envelope.
    Next,
    we transfer the trained policy to hardware and perform a number of flight
    tests. Although our flight logs show control signals oscillations are high, 
    they do not have any impact on the stability of the aircraft. In fact, in
    regards to tracking error, our analysis finds 
    training on the digital twin greatly reduces error, resulting in a smoother
    more accurate controller than previously obtained.
\end{itemize}

The remainder of this chapter is organized as follows. 
In  Section~\ref{sec:twin:gymfc} we introduce \gymfcTwo as a means to
standardize flight control tuning in silico. Next, in Section~\ref{sec:twin}, we
propose our methodology for developing multicopter digital twins and walk through
the processes of creating a
digital twin for our aircraft \aircraft. 
In Section~\ref{sec:stable} we verify the stability of our digital twin in
simulation before it is used for training. Next, we describe the changes we
made to the
training environment in Section~\ref{sec:twin:impl} and then we evaluate the
performance of the  synthesized neuro-controller  
in Section~\ref{sec:twin:eval}.  In Section~\ref{sec:twin:related} we review other flight simulators,
aircraft models and data sources. Finally, in Section~\ref{sec:twin:con}, we conclude with our final remarks and future work.

\section{G\lowercase{ym}FC\lowercase{v}2}
\label{sec:twin:gymfc}

\begin{figure*}
\centering
{\includegraphics[trim=0 100 0
    0,clip,width=\textwidth]{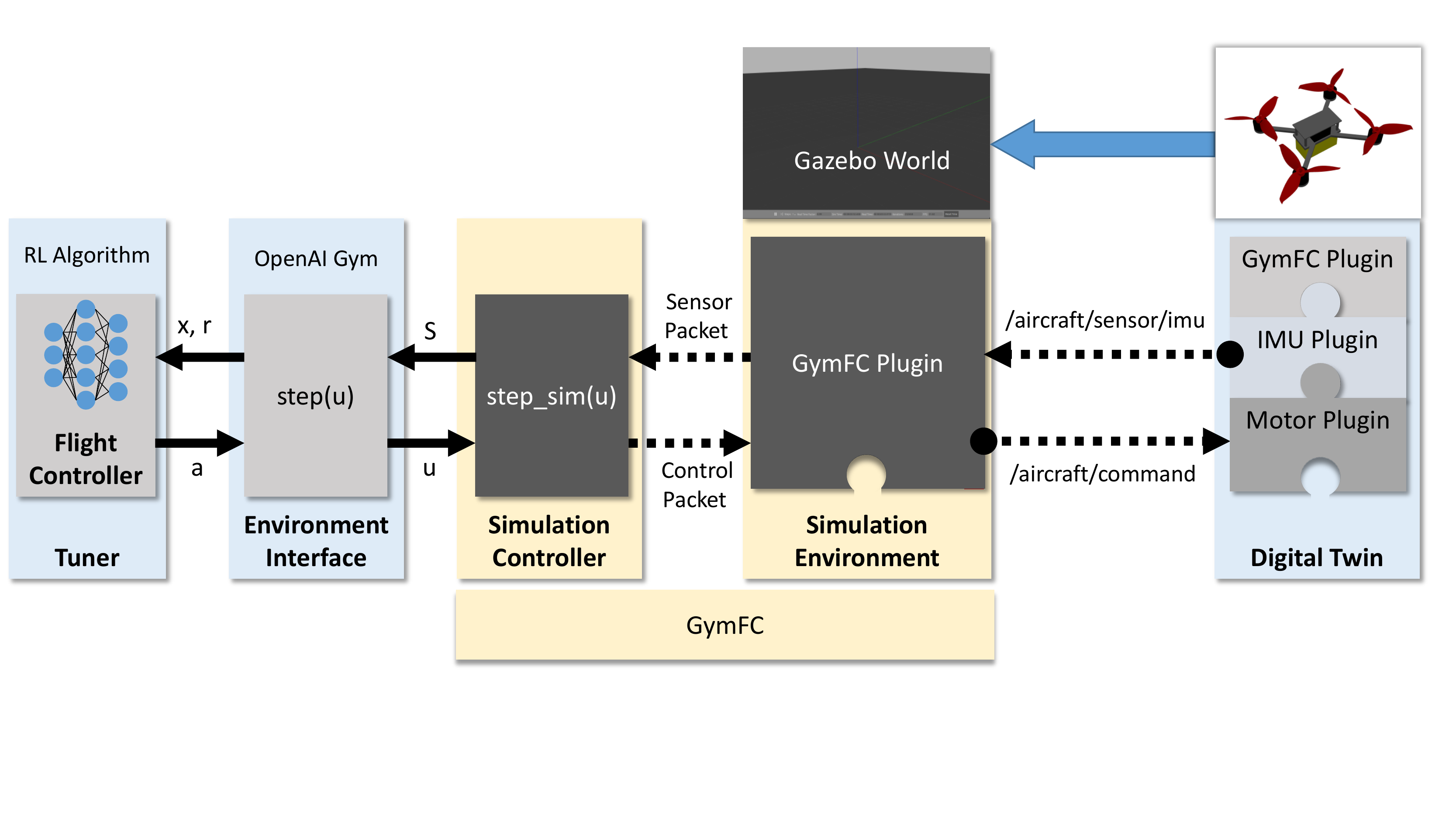}}
\caption{Instance of \gymfcTwo architecture for synthesizing RL-based flight
    controller.}
\label{fig:gymfc2}
\end{figure*}

In this section we introduce \gymfcTwo, a powerful tool for flight control
development. The new version has a redesigned architecture to address
limitations in the previous versions. 
An illustration of its typically usage is depicted in Fig.~\ref{fig:gymfc2}.

The main drawbacks with the previous versions of \gymfcGeneric is that it is tightly coupled to the aircraft model and
was specifically developed as an RL environment. The new architecture of
\gymfcTwo is aircraft agnostic~( meaning it does not care what type of aircraft
is being controlling)
and is a generic tool for 
flight controller development~(that is, it is not strictly for \nn-based flight
controllers). 
To synthesize optimal flight controllers, each controller must be trained for  
its unique aircraft
digital twin. Thus the primary motivation for the  new architecture 
was to provide an easy way to use any aircraft model.

To support a more generic framework, \gymfc was reduced to only the core implementation
for providing the training environment and interfaces with the simulation
environment and the aircraft. The remaining functionality has been moved
to user provided modules. 
This increases 
flexibility allowing the client to provide their own controller environments and
aircraft models.  
For example, this allows  a user to test and develop any type of flight controller, not only for neuro-based controllers
but also more traditional controllers such as PID.  
Additionally, for those developing neuro-based controllers, this allows the user
to develop, maintain and version control their training interfaces independent
of \gymfcGeneric.
Furthermore, reward engineering for RL-based training is a challenging  
problem dependent on many factors such as aircraft type and performance
optimization goals. During development it will be common for users to be
experimenting with different implementation which is easier to do within the
new architecture. 

In the remainder of this section we will first discuss the details of the \gymfcTwo
architecture and then the user provided modules.

\subsection{Architecture}

\gymfcTwo consists of two modules, a simulation controller providing a client interface for interacting with the simulator and the simulator environment which provides the  tuning environment and an aircraft interface.

\subsubsection{Simulation Controller}
\label{sec:controller}
The simulation controller is the client facing module in the form of a Python library.
Its purpose is to provide an interface for the user to configure and control the tuning environment.

\textbf{Configuration.} 
\gymfcTwo is initialized with an aircraft configuration file.
The aircraft configuration file is
in the SDF file format~\cite{sdf} which is an XML file with a schema specific for describing robots
and their environments made popular by Gazebo. The configuration file describes the aircraft model for use by the Gazebo simulator such as the locations to the 3D mesh files, geometric properties, and also the definitions of the plugins to be loaded for modelling dynamics.
In an SDF file, the plugin element contains a filename attribute that points to the name  of a shared library to be loaded at run time.

To simply user configuration, without requiring multiple configuration files, information specifically needed by the \gymfcTwo simulation environment is also embedded in the aircraft configuration.
However due to constraints in the SDF schema, arbitrary XML elements are not allowed in the file. Fortunately, 
the SDF plugin element does allow for arbitrary
elements to be defined. Thus as a workaround, the user must define 
our dummy plugin  
\texttt{libAircraftConfigPlugin.so} that contains the information needed by the
\gymfcTwo simulation environment plugin. 
This plugin does not provide any dynamic capabilities, it is merely a method to
provide \gymfcTwo configuration information.

This plugin defines the number of actuators the aircraft
uses for control as well as the sensors that are supported by the aircraft. 
Knowledge of the supported sensors is strictly for optimization purposes which
will be discussed later in this section.

For attitude controllers, the configuration must also specify the aircraft's  center of thrust which the simulator environment will use to fix the aircraft to in the simulation world.
An example of this plugin for our quadcopter, \aircraft, is displayed in Listing~\ref{lst:sdf}. 
Although our aircraft supports additional sensors, for training and tuning purposes we only require angular velocity values. 

{
\lstset{language=XML}
\centering
\begin{lstlisting}[caption={\aircraft configuration for
		\gymfcTwo},captionpos=b, label=lst:sdf]
<plugin name="cfg" filename="libAircraftConfigPlugin.so">
  <motorCount>4</motorCount>
  <centerOfThrust> 
    <link>battery</link>
    <offset>0 0 0.058</offset>
  </centerOfThrust>
  <sensors>
    <sensor type="imu">
		<enable_angular_velocity>true</enable_angular_velocity>
    </sensor>
  </sensors>
</plugin>
\end{lstlisting}
}

\textbf{Simulation Control.} The client can control the simulator in two ways (1) stepping the simulator through the 
\texttt{step\_sim} function  and
(2) resetting the simulator and aircraft state by the \texttt{reset} function. The \texttt{step\_sim} function  takes as input an array of control signals
$u$ for each aircraft actuator and performs a single simulation step,
returning a flattened array of the aircraft sensor values in order as defined in the aircraft configuration file. The controller also exposes class attributes for the sensor values to be accessed directly.  

The simulation
controller communicates with the simulation environment  through a UDP
network channel which encodes the control signal and sensor
messages in Google Protobuf messages. 

\subsubsection{Simulation Environment}
\label{sec:twin:simenv}
The simulation environment (specifically the Gazebo \gymfcTwo plugin) provides the majority of the heavy lifting and 
is constructed specifically for the task of tuning flight
controllers. The environment supports attitude control tuning as in the initial
version, as well as environments for motor modelling
and navigation tasks. The new architecture also allows users to provide their
own simulation worlds for more complex training such as obstacle avoidance. 
The simulation environment can be thought of as a Gazebo simulation wrapper with
custom APIs for interacting with an aircraft in simulation.

Upon launch, the environment reads the location of aircraft configuration file
from an environment variable set by the simulation controller. The environment then dynamically loads the aircraft
model into the simulator and is ready to start accepting motor control 
messages from the controller. These motor messages also doubles as the simulation clock, every call to \texttt{step\_sim} sends a motor message triggering a simulation step. 

A challenge encountered dynamically loading the aircraft model was developing a
communication channel to 
 send and receive messages from the aircraft while still remaining decoupled from \gymfcTwo.
 We solved this problem by developing 
a topic based publish-subscribe messaging API which is summarized in Table~\ref{tab:sensors}.
This API provides messages for sending the motor control signals, as well as
reading sensors. Additionally, values such as motor torque and force exist which
can be beneficial for motor model validation and reward engineering.
In the future we plan to support additional sensors to
aid in navigation tasks 
such video, sonar, and
LIDAR.  

During initialization, the \gymfcTwo simulation plugin 
initializes a publisher for the \texttt{/aircraft/command}
topic, and will also subscribe to every sensor topic of the sensors enabled in the  aircraft configuration file.
The enabled senors are required by the aircraft configuration to allow the
\gymfcTwo plugin to know it has received all of the sensor messages before returning the state back to the controller.
At a high level, the following events complete  a single simulation step,
\begin{enumerate}
    \item Upon receiving a motor control message  from the simulation controller, publish topic \texttt{/aircraft/command} with
        an array of the control signals, where the array index corresponds to
        the motor/actuator identifier.
    \item Increment the simulation one time step. This
        triggers any digital twin plugin to execute.
    \item Wait to receive sensor messages from the  enabled \texttt{/aircraft/sensor}
        topics.
    \item Pack received sensor values and simulation state into single message and
        send back  to the simulation controller.
\end{enumerate}

This decoupled communication channel provides the aircraft model designer the freedom to implement a variety of
different aircraft architectures, without requiring \gymfcTwo to know these details. For
example, a designer may choose to model a single virtual ESC as one plugin which will subscribe to
the \texttt{/aircraft/command} topic (\ie one to one) while another option
would be have a separate ESC/motor plugin instances for each motor who will each subscribe to
the command topic and extract their value at the corresponding array index (\ie one to many).

Although the publish-subscribe API provides a modular, flexible channel, it
does increase complexity due to its asynchronous behavior. Messages are 
received out of order thus the \gymfcTwo plugin
uses a rendezvous point which blocks the state from being sent to the simulation
controller until all sensor value are received.
This enforces the required sequential time steps between the simulation controller and its
environment.  
\begin{table}[]
	\centering
\begin{tabular}{|l|c|c|}
\hline
\multicolumn{1}{|c|}{\textbf{Topic}}       & \textbf{Direction}            & \textbf{Values}                      \\ \hline
/aircraft/command                          & $\rightarrow$                 & Control Signals \\ \hline
\multirow{3}{*}{/aircraft/sensor/imu}      & \multirow{3}{*}{$\leftarrow$} & Angular Velocity                     \\ \cline{3-3} 
                                           &                               & Linear Acceleration                  \\ \cline{3-3} 
                                           &                               & Orientation                          \\ \hline
\multirow{6}{*}{/aircraft/sensor/esc/$<$id$>$} & \multirow{6}{*}{$\leftarrow$} & Angular Velocity                     \\ \cline{3-3} 
                                           &                               & Temperature                          \\ \cline{3-3} 
                                           &                               & Voltage                              \\ \cline{3-3} 
                                           &                               & Current                              \\ \cline{3-3}
                                           &                               & Force                              \\ \cline{3-3}
                                           &                               & Torque                              \\ \hline
/aircraft/sensor/current                   & $\leftarrow$                  & Current                              \\ \hline
/aircraft/sensor/voltage                   & $\leftarrow$                  & Voltage                              \\ \hline
/aircraft/sensor/gps                   & $\leftarrow$                  &
Longitude and Latitude                              \\ \hline
\end{tabular}
\caption{Digital twin API. This table summarizes the topics and their corresponding message values. Direction specifies who is the publisher where $\rightarrow$ is a message published by the flight controller plugin and $\leftarrow$ is a message published by a sensor.}
\label{tab:sensors}
\end{table}

\subsection{User Provided Modules}
A typical instance of \gymfcTwo is composed of four additional user provided 
modules: a flight control algorithm, a flight control algorithm tuner, an
environment interface, and a digital twin.
The modules provide researchers and developers an easy way to share. A number of off-the-self solutions exist for the first two modules,
however custom implementations are typically required for developing the
environment interface and the aircraft model.
This section will describe each in detail.

\textbf{Flight controller algorithm.} The
flight control algorithm performs some evaluation to derive the motor control
signals.
The algorithm can generically be represented as 
the function $u(t)=f(S(t),w)$ which takes as
input the current state representation of the aircraft $S(t)$ and a set of tunable
parameters $w$ and outputs an array of control signals $u(t)$ for each aircraft
actuator. 
For example, this can be a \nn-based controller with adjustable network
weights $w=W$, or a PID controller with tunable gains
$w=\{K_P, K_I, K_D\}$ for each roll, pitch and yaw axis. 
Our goal is to find $w$.

\textbf{Flight controller algorithm tuner.} The tuner interacts with the flight
control algorithm and the environment interface to find an optimized $w$
depending on some performance goals~(\eg minimizing error, increasing flight
time, etc). 
For \nns trained using RL, a number of off-the-self solutions exist such as
OpenAI Baselines~\cite{baselines}, 
Tensorforce~\cite{schaarschmidt2017tensorforce}, and others. These RL
frameworks also provide
the \nn implementation.% which will be trained to be the flight controller. 

\textbf{Environment interface.} The environment interface is intended to be  a light
weight shim     
that either inherits or creates an instance of the \gymfcTwo
simulation controller and performs any additional implementation required for
interfacing with the control algorithm, and to support tuning. 

It is common for the input and output of the control algorithm to differ from the
aircraft state, and the actuator control signal respectively. 
For example, a \nn controller with an output layer consisting
of hyperbolic tangent activation
functions (\ie in the range $[-1,1])$ may be synthesized for a flight control firmware
requiring each control signals to be in the range $u \in [0,1]$.
Furthermore, for PID control (and also our \nn), the input is a function of the
error. The error must be computed from the angular velocity of the aircraft
state. 
This module must provide a transformation function to provide these required
mappings. 
The transformation function should 
implement the same API found in the  
target flight control firmware.

When executing, the flight control algorithm should not be able to distinguish
between the environment interface module, and the firmware. 
The goal of this framework is to provide seamless transfer from the simulation
environment
to hardware. Once the flight control algorithm is tuned, it can be ``dropped''
into the firmware without any modification.

This module shall also provide any additional information required by the tuner. 
For RL-based tuners, one of the most important functions of this module is to provide the
reward function.
Additionally if the user wishes be compatible
with OpenAI Gym environments, this module would also need to  
inherit \texttt{gym.Env}. Note, this is a change from \gymfcOne which was an OpenAI environment by default.

\textbf{Digital Twin.} The digital twin is a digital replica of the real aircraft 
the flight control algorithm will ultimately control.
It consists of the aircraft configuration, 3D meshes, and the plugins for
modelling the sensors and actuators. 
Unlike the previous user modules that have more freedom defining the
interfaces between them, the digital twin interacts only with \gymfcTwo and has a
strict API that must be followed as previously
outlined in Table~\ref{tab:sensors}. 
At a minimum to achieve flight, the digital twin must implement an IMU plugin which publishes
angular velocities to the
\texttt{/aircraft/sensor/imu} topic, and a motor plugin which subscribes to
the \texttt{/aircraft/command} topic.
\gymfcTwo does not have knowledge of the unit of measure for the data provided
by the sensors, it
is up to the user to ensure consistency between the values published by the
digital twin and the other user provided modules.

In the following section we will discuss our method for
creating a digital twin of our aircraft.

\section{Digital Twin Modelling}
\label{sec:twin}
In this section we will discuss our method for developing an aircraft
model~(\ie \emph{digital twin}) for our real quadcopter, \aircraft, for which the neuro-flight controller will be uniquely synthesized for.
At a high level this involves
 defining the rigid bodies~(Section~\ref{sec:twin:rigid}) of each aircraft part
 (known in Gazebo as a link), developing models for the motor dynamics and
 modelling sensor noise (\eg from the gyro).

\subsection{Rigid Body}
\label{sec:twin:rigid}

One of the challenges of developing a rigid body for the aircraft is 
computing the moments of inertia. 
One approach is to experimentally measure the moments of inertia using
techniques such as a torsional pendulum~\cite{ringegni2001experimental},
however this does not scale well.
The second approach is to compute the moments of inertia using a computer model of
the object. 
Most software tools for computing the inertia of an object  assume a uniform mass
distribution~\cite{cignoni2008meshlab}. 
However for a quadcopter, the majority of the mass is located at the center (battery) and end of the arms (motors).
To account for the non-homogeneous mass distribution, the aircraft can be
decomposed into its
individual parts and a rigid body can be created for each one assuming the mass
density is more uniform in the individual part.
There is a trade-off associated with the number of parts to model. In
one hand we will gain a higher fidelity model, yet on the other hand this will require more
computation power for simulation.

\begin{figure}
\centering
\begin{subfigure}{0.5\textwidth}
	{\includegraphics[trim=0 0 0
		0,clip,width=\columnwidth]{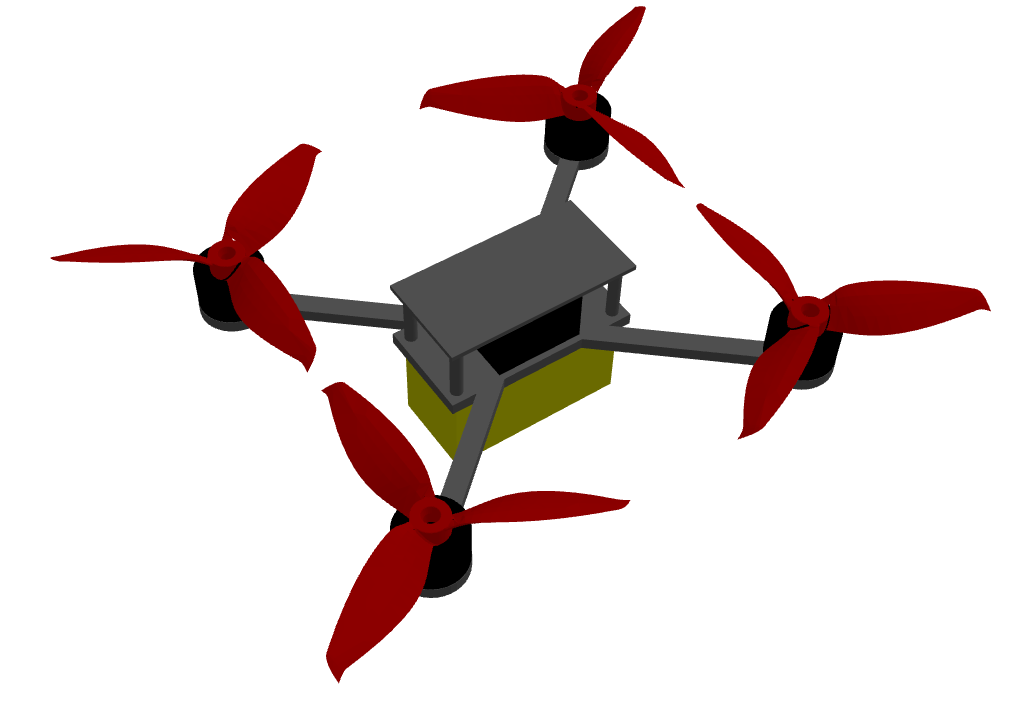}}
	\caption{Digital twin of  \aircraft}
	\label{fig:model}
\end{subfigure}
\begin{subfigure}{0.45\textwidth}
	{\includegraphics[trim=0 0 0
		0,clip,width=\columnwidth]{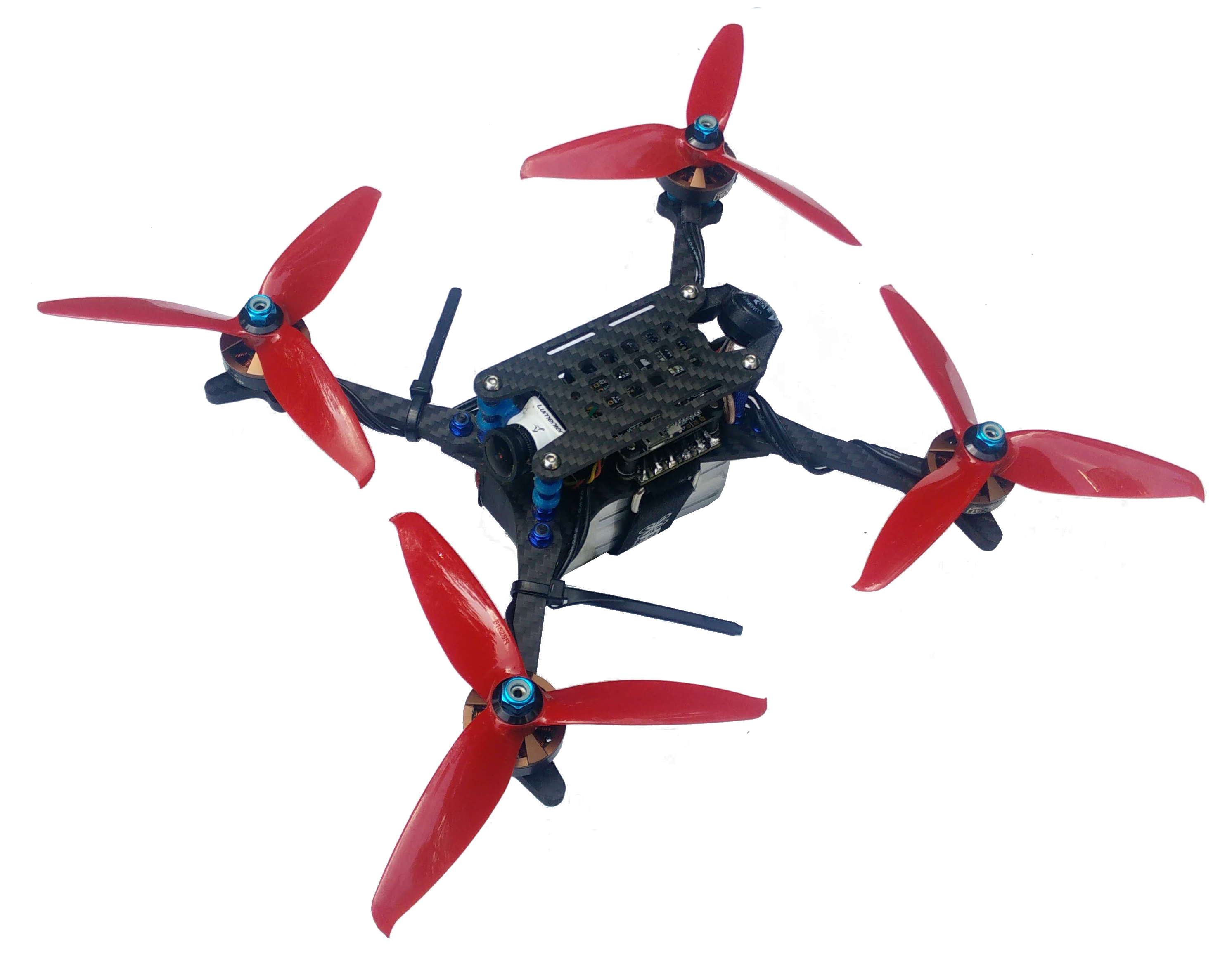}}
	\caption{\aircraft}
	\label{fig:twin:nf1}
\end{subfigure}

\caption{Digital twin of \aircraft compared to real quadcopter.}
\label{fig:twin:twincompare}
\end{figure}

Given we require the aircraft computer models for training in simulation, we
use the second approach for computing the mass properties via software.
We used FreeCAD~\cite{freecad} to develop models for the frame, motor, battery, flight control stack.
For simplicity, the flight control stack was modelled as a single component
however in reality the flight control stack is composed of the ESC,
flight controller, and video transmitter~(VTX). Additionally we omitted models for the VTX
antenna, and camera. We obtained the propellers from GRABCAD~\cite{gemfan}. 
A picture of the digital twin, compared to the real aircraft  is displayed in
Fig.~\ref{fig:twin:twincompare}.

The frame geometry is particularity important as it
affects the aircraft's flight performance. In modern UAV flight
controllers, asymmetries in the placement of actuators is accounted for through
\textit{mixing} which is essentially a lookup table that scales the
control signal depending on the distance the motor is from each axis of rotation. For RL synthesized controllers, the agent will learn the
geometry of the frame and encode this into the \nn. 

The mass of each individual part of the quadcopter
was measured. We then used Numpy-STL~\cite{numpystl} to compute the volume and mass
properties for each part, including the center of mass~(CoM) and the moments of
inertia, $I'$. As documented by Gazebo~\cite{inertia}, the computed moments of inertia must be scaled
by the length units (unit\_scale), and the density to derive the actual inertia
tensor $I$,
\begin{equation}
I = I'  \text{unit\_scale}^2  m/V 
\end{equation}

The individual aircraft parts, and their corresponding mesh and mass properties are 
added to a single SDF file. The position of each aircraft part is then adjusted
by modifying the \texttt{pose} XML element to correctly assemble the aircraft. 
Loading the model in Gazebo, we were able to validate the position of all the
parts.
When assembling the aircraft, it is essential to make sure the aircraft aligns
with the correct axis of rotation, otherwise the IMU will not report the
expected 
values (discussed in detail the following section). In Gazebo the axis lines, (R)ed, (G)reen,
and
(B)lue, map to the axis (R)oll, (P)itch and (Y)aw respectively. 
This SDF file also includes the \gymfcTwo plugin definition provided  in Listing~\ref{lst:sdf}.

We obtain the center of thrust value by measuring offset from
the bottom of the model to the base of the rotor. In the follow sections we
discuss our method for configuring the motor model and IMU plugins which will
also be added to the SDF.

\subsection{IMU Model}
\label{sec:imu}
To model the IMU we  ported over the IMU plugin provided by PX4~\cite{px4sitl},
and implemented the digital twin API. 
For angular velocity measurements, essentially all this plugin does is query
the Gazebo API for the angular velocity for a particular link in the world.
Thus the IMU plugin must be configured with a link that will emulate the flight
control stack on the real quadcopter. 
We assigned it to our FC stack link, however one must pay special attention to
validate the orientation of the part. If assembled according to the procedure
in the previous section, there should be no problems. However this can be confirmed  using the
test scripts included with \gymfcTwo to step the simulator with specific
control signals to rotate the aircraft while monitoring the IMU values
provided by the plugin. For example, set $u=[0,0,1,1]$ to 
roll right, the IMU values should match this movement. 

In order to increase the fidelity of our digital twin, we introduce gyro
noise. In past literature~\cite{jakobi1995noise,andrychowicz2018learning,molchanov2019sim}
noise has been sampled from a Gaussian distribution. 
To introduce noise into the model we must identify the gyro noise mean and variance for
each axis, 
$\eta_{(\phi, \mu)}$, $\eta_{(\phi, \sigma)}$,
$\eta_{(\theta, \mu)}$, 
$\eta_{(\theta, \sigma)}$,
$\eta_{(\psi, \mu)}$, and  
$\eta_{(\psi, \sigma)}$. 

We would like to point out that we introduce gyro noise during training from within our environment
interface user module, not from within the IMU plugin. This provided us with additional
flexibility such as easily evaluating the performance of controllers with
different noise parameters than having to modify the SDF file to make these
changes. In the future we will explore ways to make noise configuration easier
for the plugin.

\subsection{Motor Model}
\label{sec:model}
In this section we will discuss our method for developing the motor model
for \aircraft. 
In Gazebo, model dynamics are implemented by C++ plugins. 
Each plugin definition is associated with a set of configurable options that
are defined in the models SDF file.
Our motor models are based on the PX4 Gazebo SITL motor model plugins~\cite{px4sitl} that have been
ported to \gymfcTwo.  We have made our motor
plugins open source at the following link \pluginsource 
allowing the community to utilize them in their own research and improve upon
them.
In this section
we discuss the values that must be configured in the plugin, and the
methodology for deriving the values in order to use the motor model
plugins.  
Given the modular architecture of \gymfcTwo,
researchers can also easily use their own motor models.

The PX4 motor models derive force and torque approximations for a propeller
propulsion system using blade
element theory~\cite{mccormick1995aerodynamics}.
The propeller performance can be defined by two dimensionless coefficients $C_T$
and $C_Q$ for the thrust and torque coefficient respectively. The thrust
coefficient is given as, 
\begin{equation}
	C_T = \frac{T}{\rho n^2 D^4}
\end{equation}
where $T$ is the thrust, $\rho$ is the air mass density, $n$ is the propeller rotational speed in
revolutions per second, and $D$
is the propeller diameter.
The torque coefficient is given as,
\begin{equation}
	C_Q = \frac{Q}{\rho n^2 D^5}
\end{equation}
where  $Q$ is the torque.
The values for $T$, $Q$, $\rho$, and $D$ must have
consistent units. 

The thrust and torque coefficients are a function of the dimensionless advanced ratio $J$
which quantifies the effects of the propeller in forward motion in relation to
its angular velocity
given by,
\begin{equation}
    J=\frac{V_\infty}{nD}
\end{equation}
where $V_\infty$ is the freestream fluid velocity.
When $J=0$, this is the static case in which $V_\infty=0$.

To develop a  model for a propeller driven propulsion system to be used in simulation, an approximation of the thrust and torque
for a given propeller rotational speed must be derived. 
The PX4 Gazebo SITL plugin 
computes the motor thrust
in Netwons (N), for each motor  by,
\begin{equation}
    T(\omega) = \omega^2 K_T  
    \label{eq:t}
\end{equation}
which is a function of the rotor's current
angular velocity, $\omega$, in radians per second for a configurable thrust
constant $K_T$.
Given $C_T$,  one can derive the constant $K_T$ to be, 
\begin{equation}
    K_T = \frac{C_T \rho D^4}{(2\pi)^2} 
    \label{eq:kt}
\end{equation}
where  $\rho$ is in $\text{kg/}\text{m}^3$ and 
the propeller diameter $D$ is in meters.
The PX4 Gazebo SITL plugin 
computes the torque in Newton meters ($N \cdot m$) as a function of the thrust,
\begin{equation}
    Q(T) = TK_Q  
\end{equation}
where $K_Q$ is a configurable torque constant. Given $C_T$ and $C_Q$, $K_Q$ is
defined as follows, 
\begin{equation}
K_Q = \frac{C_Q D}{C_T}
    \label{eq:kq}
\end{equation}
The PX4 SITL motor model requires us to  find $C_T$ and $C_Q$ experimentally
for $J=0$ in order to calculate the constants $K_T$ and $K_Q$~\footnote{In the source these constants are referred to as the
    motor and moment constants respectively, they have been altered to stay
    consistent with the previous notation. The PX4 SITL plugin attempts to model other dynamics
    such as rotor drag
    that we will not go into detail. The reader is invited to read the source code
    if they are interested in these details.}.

\textbf{Motor Response.} In addition to modelling the thrust and torque of the propulsion system,  we also need to model the motor
response for a given control input. 
Most research related to quadcopter control do not model the motor response and assume the motor response to be
instant, which can lead to inaccuracies~\cite{molchanov2019sim}.

For a known maximum rotational velocity, which is found experimentally, a
 PID controller can be used to model the motor response. We found this to
 provide a more realistic response than other methods, such as a discrete first
 order filter used by the PX4 SITL motor model. The PID controller computes the additive force $F(t)'$ at time $t$ to apply to the
 rotor  as follows,
\begin{equation}
    F(t)' = K_p e(t) + K_i \int_0^t e(\tau) d \tau + K_d \frac{d e(t)}{dt}
\end{equation}
where the error is defined as,
\begin{equation}
e(t) = \omega(t) - \omega(t)^* 
\label{eq:pide}
\end{equation}
which is the difference between the current rotor angular velocity $\omega(t)$,
and desired rotor velocity $\omega(t)^* = H(u)$.
Here, $H$ is the rotor velocity transfer function which is necessary to create
the mapping $u \rightarrow \omega^*$  as there may not be a linear relationship
between the control signal and the motor 
 angular velocity.

To control the acceleration and deceleration of the motor response 
the output of the PID controller is clamped to a minimum value $F_\text{min}$
and maximum value $F_\text{max}$. Essentially these values control the 
slope of the response. 
\begin{equation}
    F(t) = \text{clamp}(F(t)', F_\text{min}, F_\text{max})
\end{equation}
The clamped force $F(t)$ is then added to the propeller joint in the Gazebo
simulator.
The three PID gains, $K_P$, $K_I$, and $K_D$ along with $F_\text{min}$,
$F_\text{max}$ must be tuned  to achieve the desired step response.

In summary, to configure the motor model plugin  we must
derive the constants $K_T$, $K_Q$, $H$,  $K_P$, $K_I$, $K_D$, $F_\text{min}$, and
$F_\text{max}$ through experimental measurements. In the following section we
will discuss our methodology for  obtaining these values.

\subsection{Experimental Methodology}
In this section we introduce our experimental methodology for deriving the
motor model configuration constants which requires specially designed equipment
and procedures to obtain. 
For the IMU model, we did not require any special equipment or procedures to
derive the values for our model.
\begin{figure*}
\centering
{\includegraphics[trim=40 200 50
    110,clip,width=\textwidth]{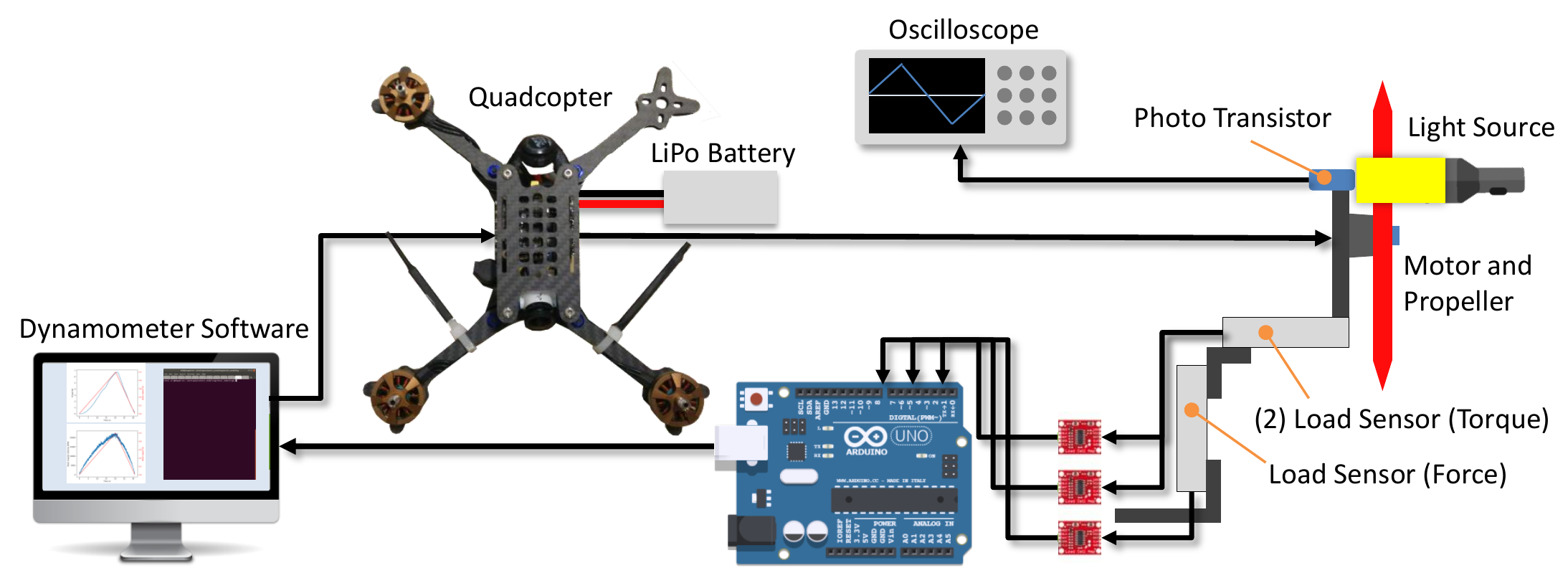}}
\caption{Dynamometer diagram.}
\label{fig:dyno}
\end{figure*}

\subsubsection{Equipment}
To derive the motor constants defined in Section~\ref{sec:model} we constructed
a dynamometer~(dyno) to measure thrust, torque and rotor angular velocity.
A diagram of our system is illustrated in Fig.~\ref{fig:dyno}. 
Our custom dyno software consists of two modules that run in parallel. The first
module controls the motor and the second
module captures and records sensor data. 
Our motor control module uses a
unique approach in which the electronics from the aircraft
are repurposed for controlling the speed of the motor. 
A complete build log, including the electronics of our aircraft, can be found at~\cite{rotorbuild}.
This solution is cost
effective and reduces any errors that may be introduced if using a dyno that
uses hardware that differs from that found in the aircraft~(\eg latency caused
by ESC protocols, power delivery of the ESC, etc.).   
The flight controller ran the Neuroflight firmware and our ESC uses the
firmware BLHeli\_32.
The motor control module sends motor commands to the aircraft's flight
controller via the MultiWiiSerial~(MSP) protocol over USB. %that is supported by \nf. 
The flight controller interprets the MSP command
and writes the motor command to the ESC which applies the necessary power to
achieve the desired output to the motor.

The sensor data capture module interfaces with an Arduino which is responsible
for aggregating the sensor data obtained from the motor.
The motor is mounted to a static testing apparatus from
RCBenchmark~\cite{rcbenchmark}
that is outfitted with the sensors to collect thrust, torque and rotor velocity
measurements. The motor mount is attached perpendicular to two 1Kg load sensors
that are separated  from one another by 80mm for
measuring torque. The torque  is calculated from the average of
the two load sensors
$\text{LS}_{\tau_1}$ and $\text{LS}_{\tau_2}$ using the
following equation, 
\begin{equation}
\tau = \frac{|k_{\tau_1} \text{LS}_{\tau_1}| + |k_{\tau_2}
\text{LS}_{\tau_2}| }{2}
\end{equation} 
where $k_{\tau_1}$ and $k_{\tau_2}$ are constants found during calibration. 
The absolute value of each is taken as one load sensor will experience a pull
(outputting a negative value),
while the other will experience a push (outputting a positive value).

The load sensors for measuring torque is attached perpendicular to a 2Kg load
sensor $\text{LS}_{T}$ for measuring thrust. The resulting force is
calculated by, 
\begin{equation}
T = |k_{T} \text{LS}_{T}|
\end{equation} 
where $k_{T}$ is a constant found during calibration. The absolute value is
taken to support both  push and pull propellers. Each of the load sensors
is connected to an amplifier to boost the signal to be read by the Arduino. 

To measure rotor angular velocity, a photo transistor and a light source is
used which triggers a pulse every time a propeller blade passes between the
transistor and light source. 
Our first approach attempted to connect the output of the photo transistor to
an interrupt pin on the Arduino  which would cause a interrupt handler to be
invoked every time a blade passed the photo transistor and light source. Based
on the number of interrupts that occur within a predefined time window, the
RPM could then be calculated.  This approach was ideal as it would allow the
entire system to be automated. 
However during validation using 
a Tektronix MDO3034 oscilloscope we found the readings from the Arduino were
limited to about 75\% throttle. Upon further inspection we discovered as the
angular velocity increased, the voltage emitted from the sensor
would decrease. This drop in voltage was enough to be below the $0.6\text{Vcc}$
threshold for what is considered a logic high on the Arduino. 
Due to this limitation, we decided to manually collect the velocity data using
the oscilloscope which also has the added benefit of having a higher sampling rate. 
Using the oscilloscope, the voltage values were recorded during each measurement. 
Post processing of the data is performed to derive the RPM values.
This is accomplished by parsing every 
$b=3$ voltage pulse  as a single rotation. The RPMs were then calculated by the intermediate times between
each complete propeller rotation. 

\begin{figure*}
\centering
{\includegraphics[trim=0 100 150
    0,clip,width=0.8\textwidth]{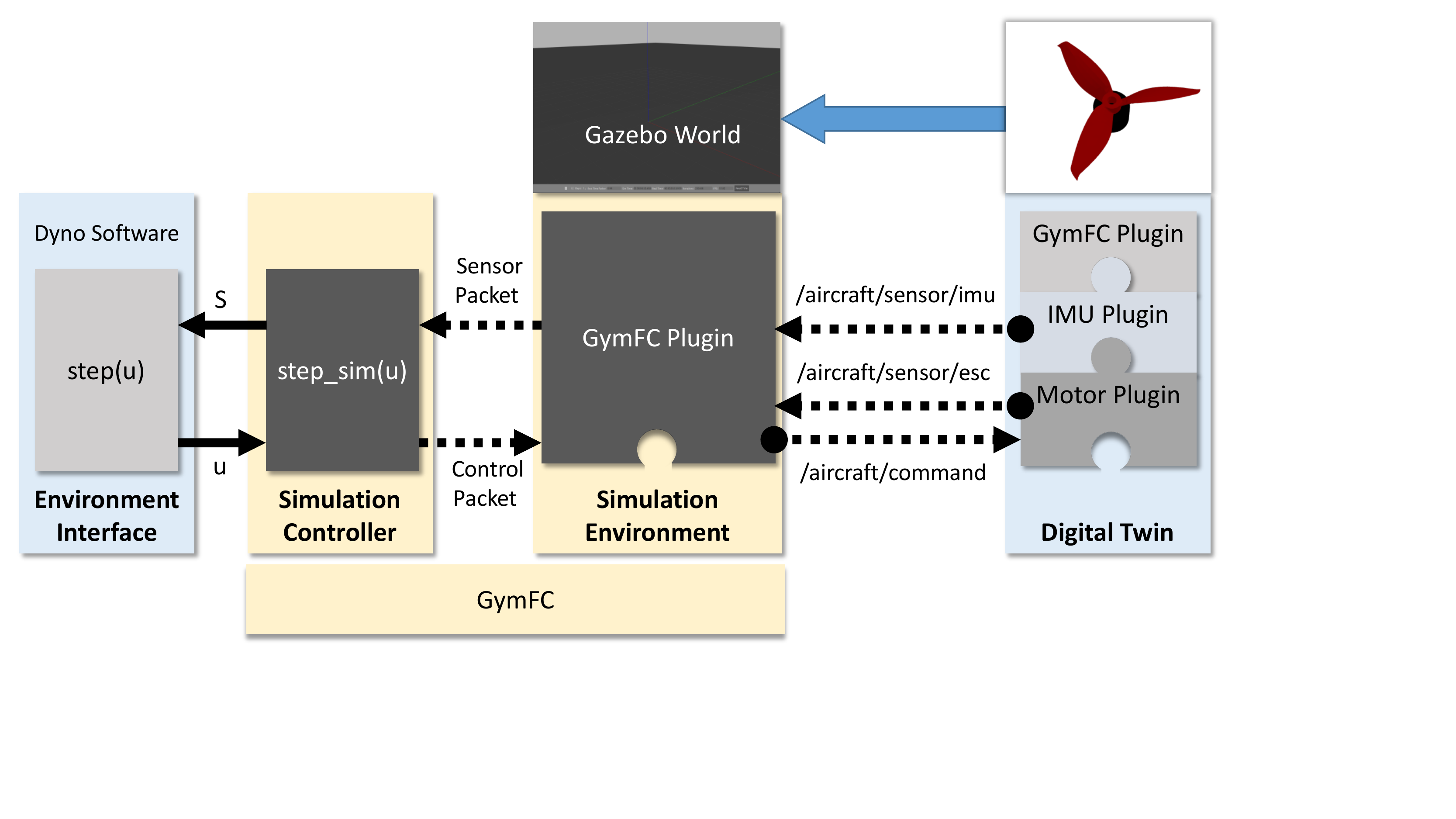}}
\caption{Instance of \gymfcTwo architecture for dyno validation.}
\label{fig:dynosim}
\end{figure*}

\textbf{Dynamometer simulator.} To validate  and develop our motor model, we
used \gymfcTwo to implement a dyno
simulator to measure the
motors thrust, torque, and RPMs in simulation. The dyno architecture is depicted
in Fig.~\ref{fig:dynosim}. A motor model was created extracting the motor and propeller
links used in the \aircraft model. The aircraft configuration enabled the ESC sensor
to obtain the  thrust, torque, and RPMs measurements. A dyno 
software module interfaces with \gymfcTwo to replicate 
the
control inputs provided by the real dyno. At every simulation step  the dyno module
records the measurements and at the end of the simulation saves the data to a
file for later processing. This dyno software is open source and is available
from the  \gymfcGeneric code repository~\cite{gymfccode}. 

\subsubsection{Calibration} Calibration of the dyno was required to obtain accurate thrust and torque
measurements from the load cells. 
For torque calibration, a lever 130mm in length was mounted to the center of
the motor mounting plate, extending to the left, to allow torque to be applied
to the motor mounting plate. Payloads starting at 200 grams   were
hung from the lever in increments of 200 grams until the max rating of the load sensor  was
reached. Both calculated torque
load sensor readings were recorded. 
Once a measurement was recorded for a
given payload, the payload was removed before the next incremented payload was
measured to check for hysteresis. This process was then repeated with the lever
extending to the right.  A linear fit was then applied to each of the load cell data
to derive a transfer function for each load cell torque measurements.

For thrust calibration, the dyno was rotated
90 degrees counter clockwise such that the motor mounting plate faced upward.
Payloads were  then added on  top of the motor mount thus applying a positive force on the force load
sensor.
The sensor recording procedure was conducted in the same manner as the torque
calibration and a
linear fit was also then applied to the sensor data
to derive a transfer function for thrust measurements.

\subsubsection{Procedure}

Using the dyno we have designed two experiments to measure and collect the
necessary data to derive the motor model constants.  The first experiment
performs  a step response and the second experiment performs a throttle ramp. 

The step response experiment is conducted to identify the motor response
parameters~(\ie $H$, $K_P$, $K_I$, $K_D$, $F_\text{min}$, and $F_\text{max}$).
To perform these measurements a fixed throttle value is applied for one
second to capture the acceleration, followed by a throttle value of zero for another
additional second to capture deceleration. Four target throttle values are 
selected: 25\%,
50\%, 75\%, and 100\%. 

Using the captured step response data, a throttle curve is generated to identify
the relationship between the control signal~(\ie throttle value) and the
corresponding achieved rotor velocity. 
This data is fitted to a polynomial function to derive the rotor velocity
transfer function $H$. 

Once the control signal to rotor velocity mapping is modelled to derive $H$, the
dyno simulator can then be used to manually tune the motor model PID controller
to fit the measured step response. 
We can do this independently of having a complete motor model because 
we are only interested in the rotor velocity and its response, not of its
thrust and torque output. 

The motor model plugin configuration is first updated with
$H$.
The $K_P$ term is then incremented until the desired
target velocity was reached, while $F_\text{min}$ and $F_\text{max}$ are tuned to
match the slope during acceleration and deceleration. If we recall from
Section~\ref{sec:model} the reason $F_\text{min}$ and
$F_\text{max}$ cannot be computed directly from the
experimentally measured slope is due to the fact that the output of the PID
controller sets the accumulated force on the rotor, not the absolute RPM velocity
of the rotor. % which could result in additional jitter.
In this work we set $K_I$ and $K_D$ to zero.

The throttle ramp  experiment is used to measure the torque and thrust.
The throttle ramp
increments the throttle from 0 to 100\%
over the course of 20 seconds and then decrements the throttle from 100\% to 0 for an
additional 20 seconds.
Using the maximum rotor velocity obtained from the step response experiment,
the maximum thrust and torque values are used to calculate $K_T$, and $K_Q$.

With all of the constants identified and updated in the motor model plugin
configuration, the dyno
simulator is used to validate the motor model plugin against the real
world measured data.

\subsection{Experimental Results}
In this section we report our empirical experimental results. Our gyro noise
parameters are summarized in Table~\ref{tab:twin:noise}. The parameters obtained
from the motor experimental measurements are summarized  in Table~\ref{tab:params}
while the derived motor constants are summarized in Table~\ref{tab:constants}.
\begin{table}[]
\centering
\begin{tabular}{l|c|c}
Axis (ax)  & Mean ($\mu$)   & Variance ($\sigma$) \\ \hline
Roll ($\phi$)  & \gyroRollMu  & \gyroRollSigma    \\
Pitch ($\theta$) & \gyroPitchMu & \gyroPitchSigma   \\
Yaw  ($\psi$) & \gyroYawMu   & \gyroYawSigma    
\end{tabular}
\caption{Normal PDF parameters for gyro noise mean~($\eta_{(\text{ax}, \mu)}$)
and variance  ($\eta_{(\text{ax}, \sigma)}$) in degrees per second.}
\label{tab:twin:noise}
\end{table}

\begin{table}[]
	\centering
\begin{tabular}{l|l}
Parameter & Value \\ \hline
Max $T$   & \maxT      \\
Max $Q$   &  \maxQ     \\
Max RPM   &  \maxrpm \\
$C_{T0}$      &  \ct \\
$C_{Q0}$      &   \cq \\    
\end{tabular}
\caption{Propeller propulsion system parameters.}
\label{tab:params}
\end{table}

\begin{table}[]
	\centering
\begin{tabular}{l|l}
Parameter & Value \\ \hline
$K_T$   & \kt      \\
$K_Q$   & \kq      \\
$K_p$   & \kp      \\
$K_i$   & \ki      \\
$K_d$   & \kd      \\
$F_\text{min}$   & \fmin     \\
$F_\text{max}$   & \fmax     \\
$H$           &  Eq.~\ref{eq:fu} \\
\end{tabular}
\caption{Propeller propulsion system model constants.}
\label{tab:constants}
\end{table}

\subsubsection{Gyro Noise}
\begin{figure}
\centering
\begin{subfigure}{0.5\textwidth}
    {\includegraphics[trim=0 0 35
		35,clip,width=\textwidth]{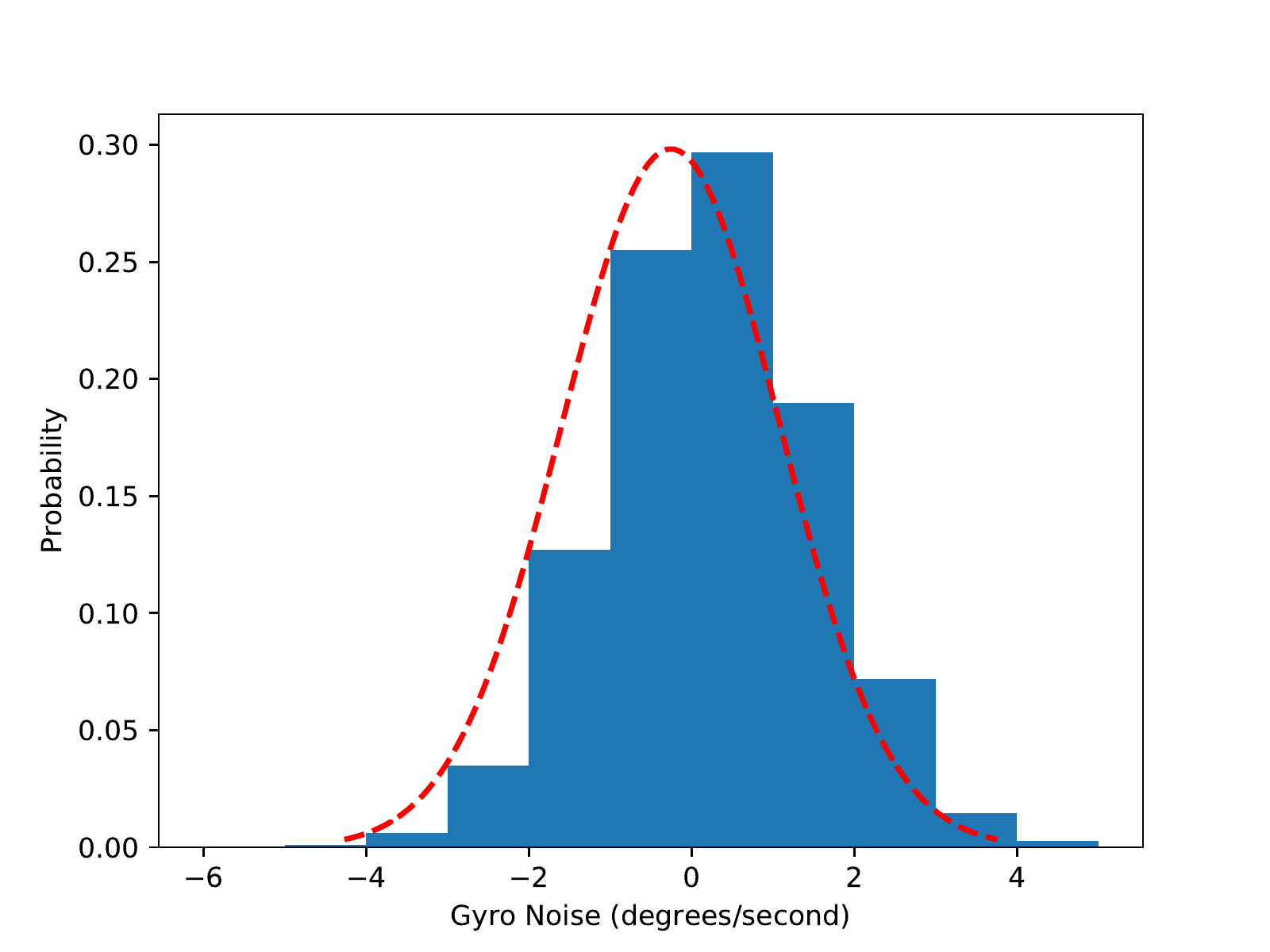}}
	\caption{Roll}
\end{subfigure}
\\
\begin{subfigure}{0.5\textwidth}
    {\includegraphics[trim=0 0 35
		35,clip,width=\textwidth]{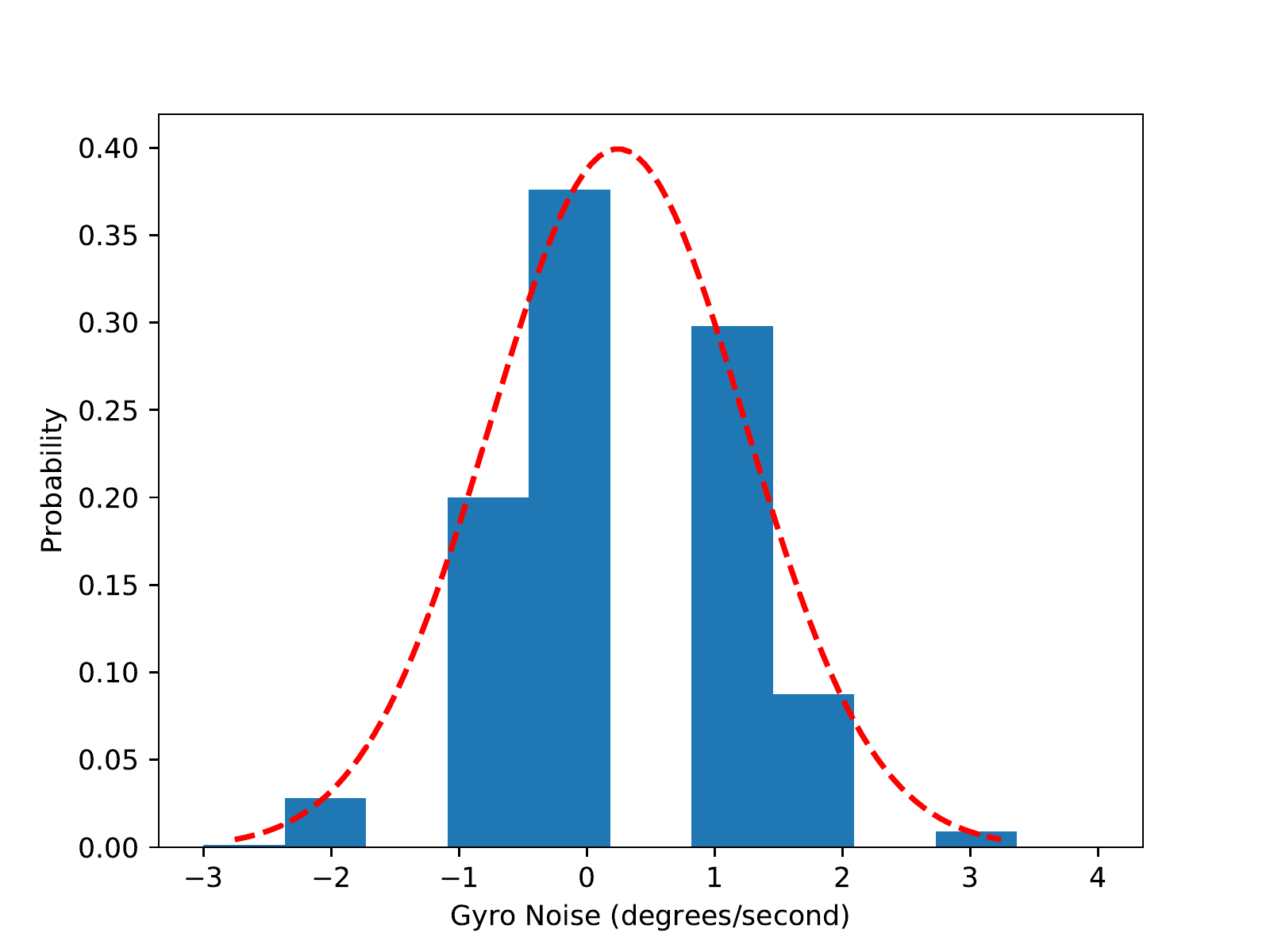}}
	\caption{Pitch}
\end{subfigure}
\\
\begin{subfigure}{0.5\textwidth}
    {\includegraphics[trim=0 0 35
		35,clip,width=\textwidth]{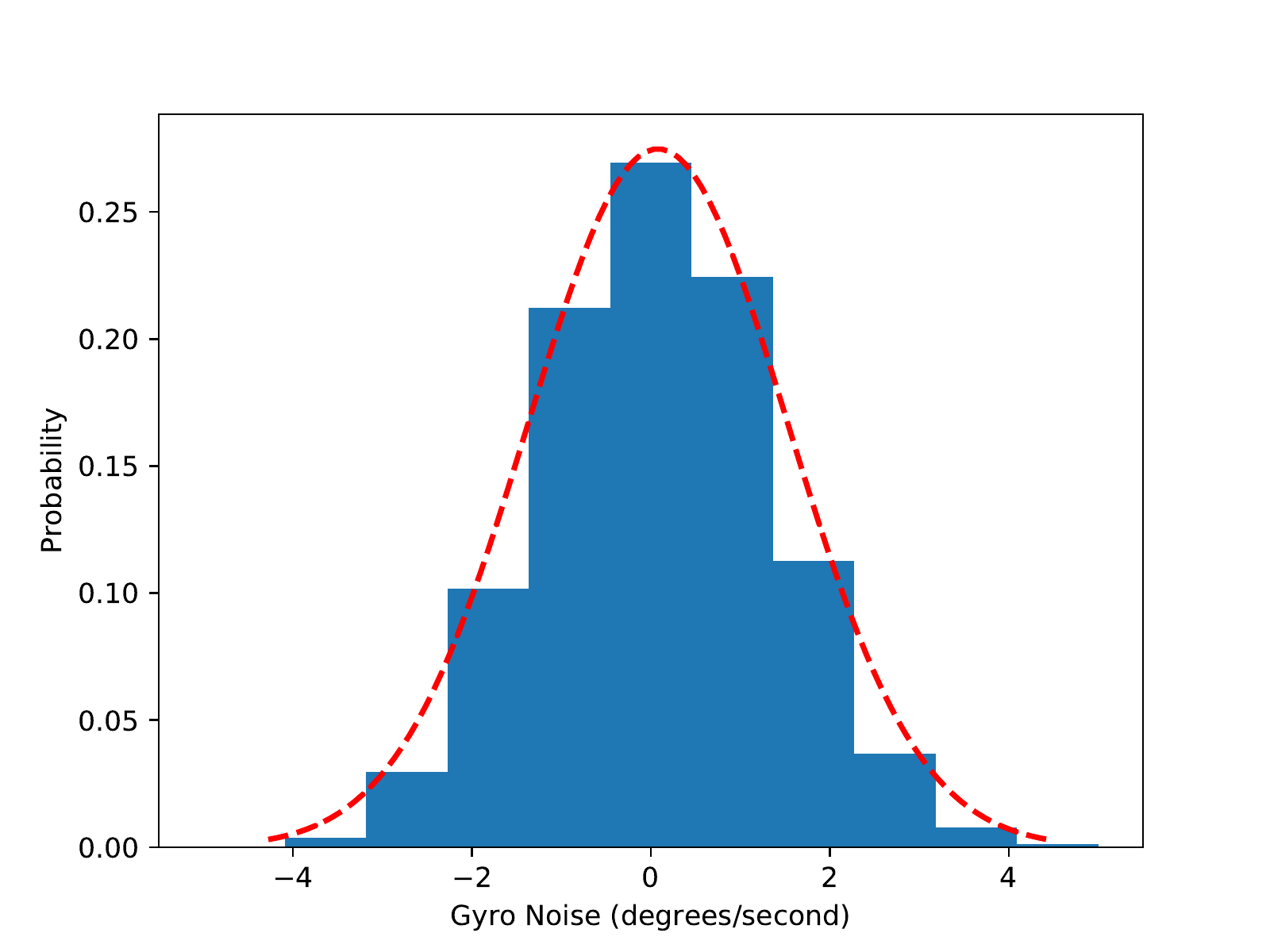}}
	\caption{Yaw}
\end{subfigure}
\caption{Gyro Noise}
\label{fig:gyro}
\end{figure}

To obtain the parameters for they IMU noise model
we recorded the gyroscope values from
our real aircraft, \aircraft, when armed, for over 30 seconds to obtain  \gyroN
samples. We then plotted a histogram of the data for each axis. These plots are
displayed in Fig.~\ref{fig:gyro}.
As we can see from the figure, we verify the data fits well to a normal distribution. 
Next we fit the data to the normal distribution probability
density function~(PDF) to obtain the mean and variance values for each axis as reported in
Table~\ref{tab:twin:noise}. 

\subsubsection{Step Response}
\begin{figure}
\centering
{\includegraphics[trim=0 0 0
    0,clip,width=0.8\textwidth]{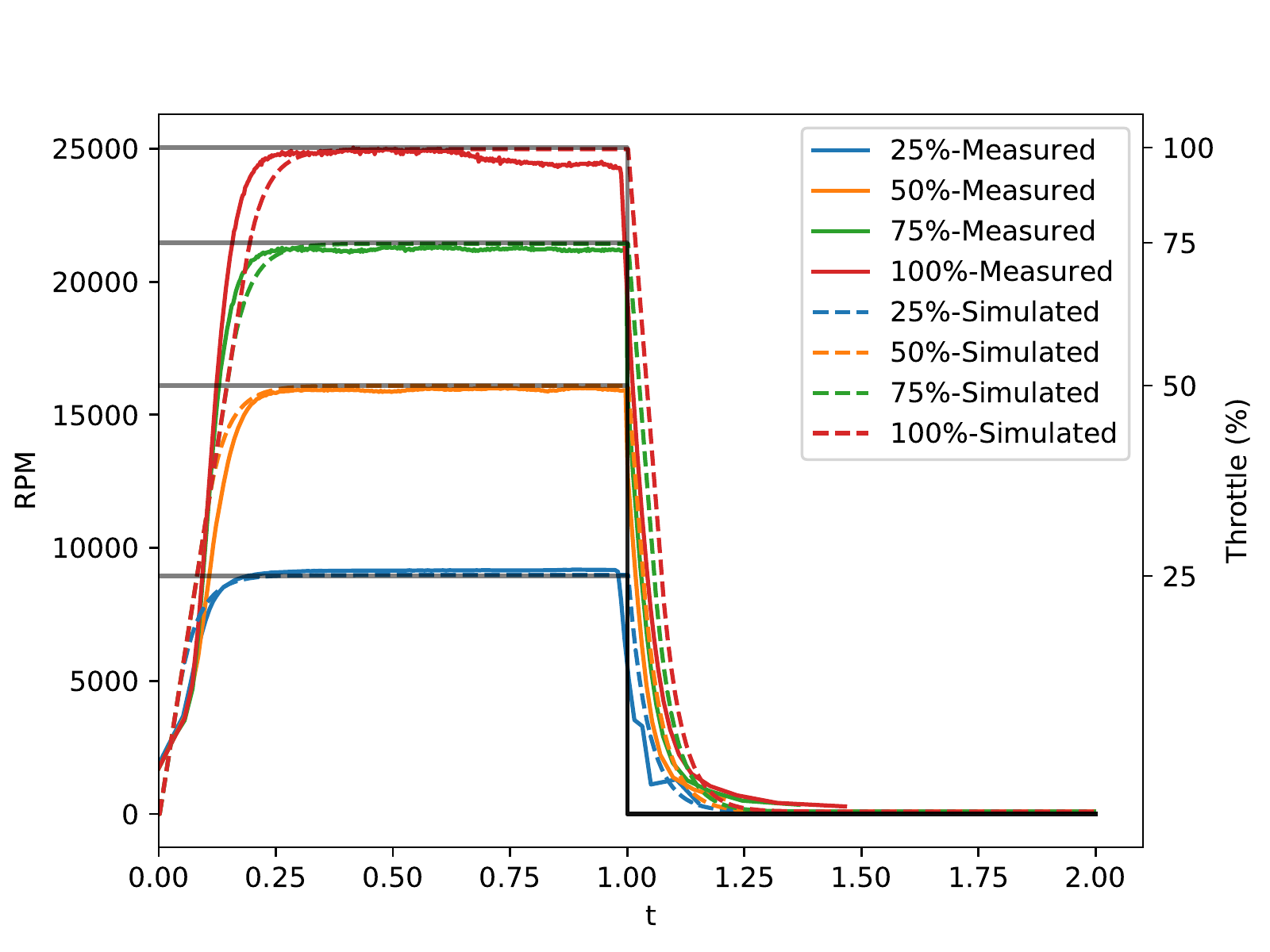}}
\caption{Step response of motor model compared to real motor.}
\label{fig:rpmfitted}
\end{figure}

\begin{figure}
\centering
{\includegraphics[trim=0 0 0
    0,clip,width=0.8\textwidth]{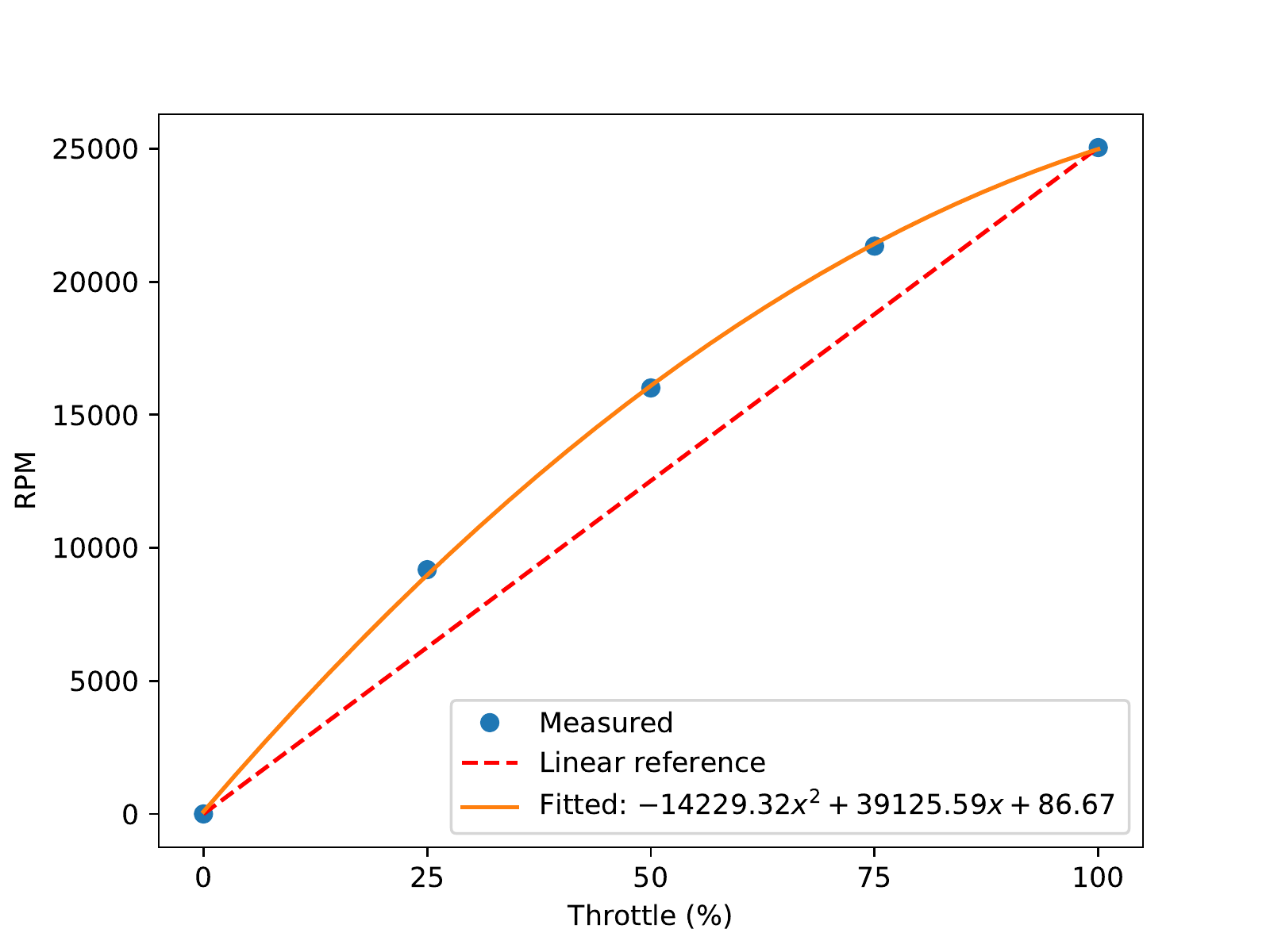}}
\caption{Throttle curve.}
\label{fig:nonlinearthrottle}
\end{figure}

Results from the step response experiment are displayed in
Fig.~\ref{fig:rpmfitted}
 while the throttle curve is displayed in
Fig.~\ref{fig:nonlinearthrottle} which is  fitted to a two degree polynomial
function to obtain the transfer function defined in Eq.~\ref{eq:fu}. 

\begin{equation}
H(u) = -14,229.32u^2 + 39,125.59u + 86.67
\label{eq:fu}
\end{equation}

These results signify the 
importance of using the ESC (and ESC firmware)  used during flight for deriving the motor
measurements in order to generate an
accurate model. Each propulsion
system will result in a unique motor response due to the current drawn for a
given propeller and the capabilities of the ESC to delivery this power to the
motor.
Most ESC firmware for UAVs use an
open-loop controller, that is, there is no feedback to reach its target.
Unlike our simulated propulsion system, the real ESC is unaware of the maximum achievable
rotor velocity as this will vary depending on the motor and propeller
combination. The ESC will map the control signal to a duty cycle (\ie switching
frequency) to reach a particular angular velocity. It is up to the higher level
attitude controller to compute the
 control signals to send to the ESC in order achieve the desired aircraft angular velocity.

After updating the motor model plugin configuration with the identified transfer function, the motor
PID controller was tuned to obtain the desired motor response.  
Fig.~\ref{fig:rpmfitted} also shows a comparison of the measured step response 
with the motor model plugin validated in simulation. Our analysis finds each  simulation
step response to have an angular velocity percent error (\ie the MAE divided by the max
RPMs) of  \rpmTwoFiveMAEPercent, \rpmFiveZeroMAEPercent, \rpmSevenFiveMAEPercent
and \rpmFullMAEPercent for the 25\%, 50\%, 75\%, and 100\% throttle values
respectively. These results show the motor response of the digital twin is
accurate to less than 5\% error of the real motor response
across all throttle values tested.

\subsubsection{Throttle Ramp}
\begin{figure*}
\centering
    \begin{subfigure}{0.8\textwidth}
        \includegraphics[width=\textwidth]{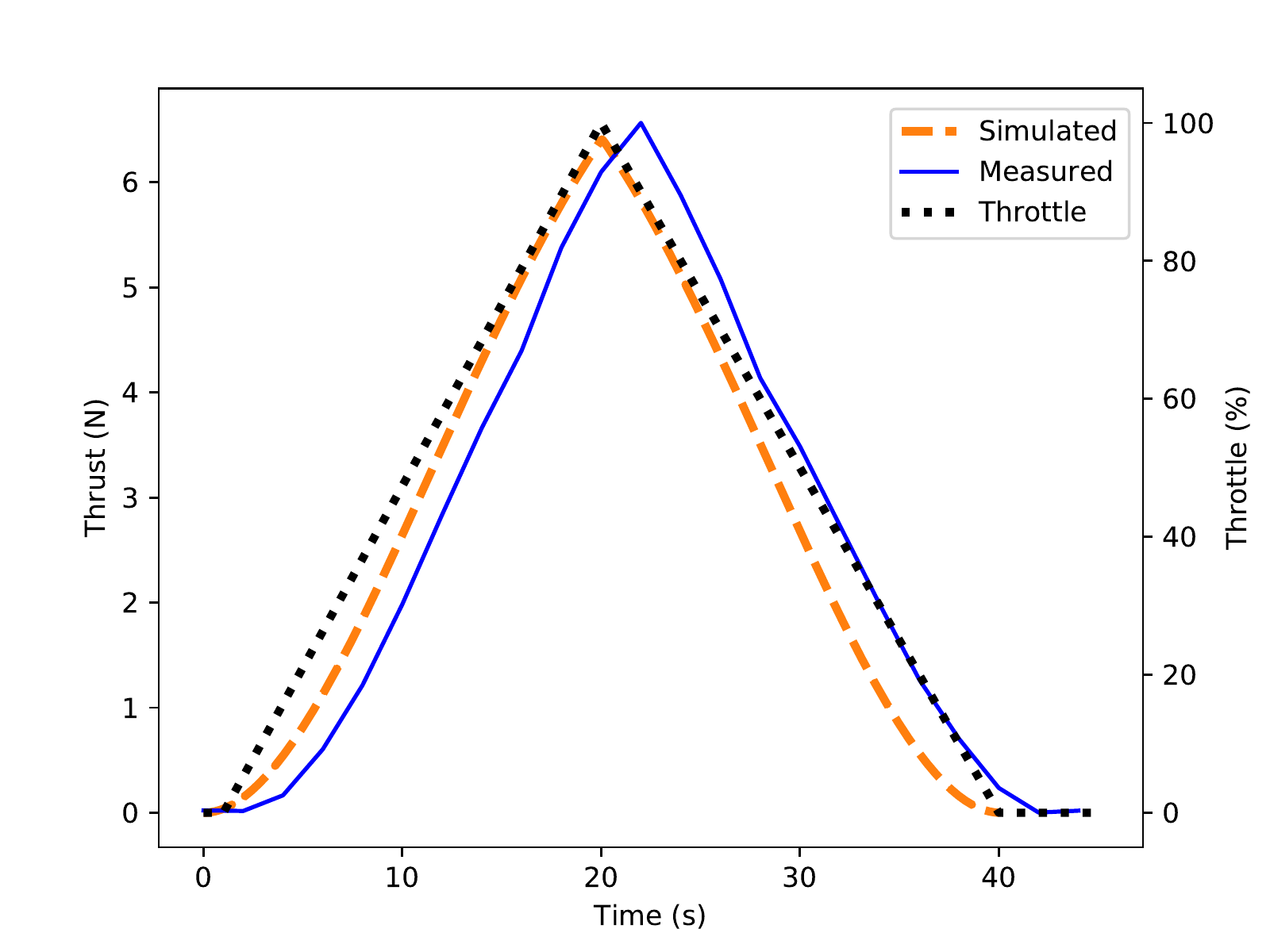}
        \caption{Thrust}
        \label{fig:dyno:thrust}
    \end{subfigure}\\
    \begin{subfigure}{0.8\textwidth}
        \includegraphics[width=\textwidth]{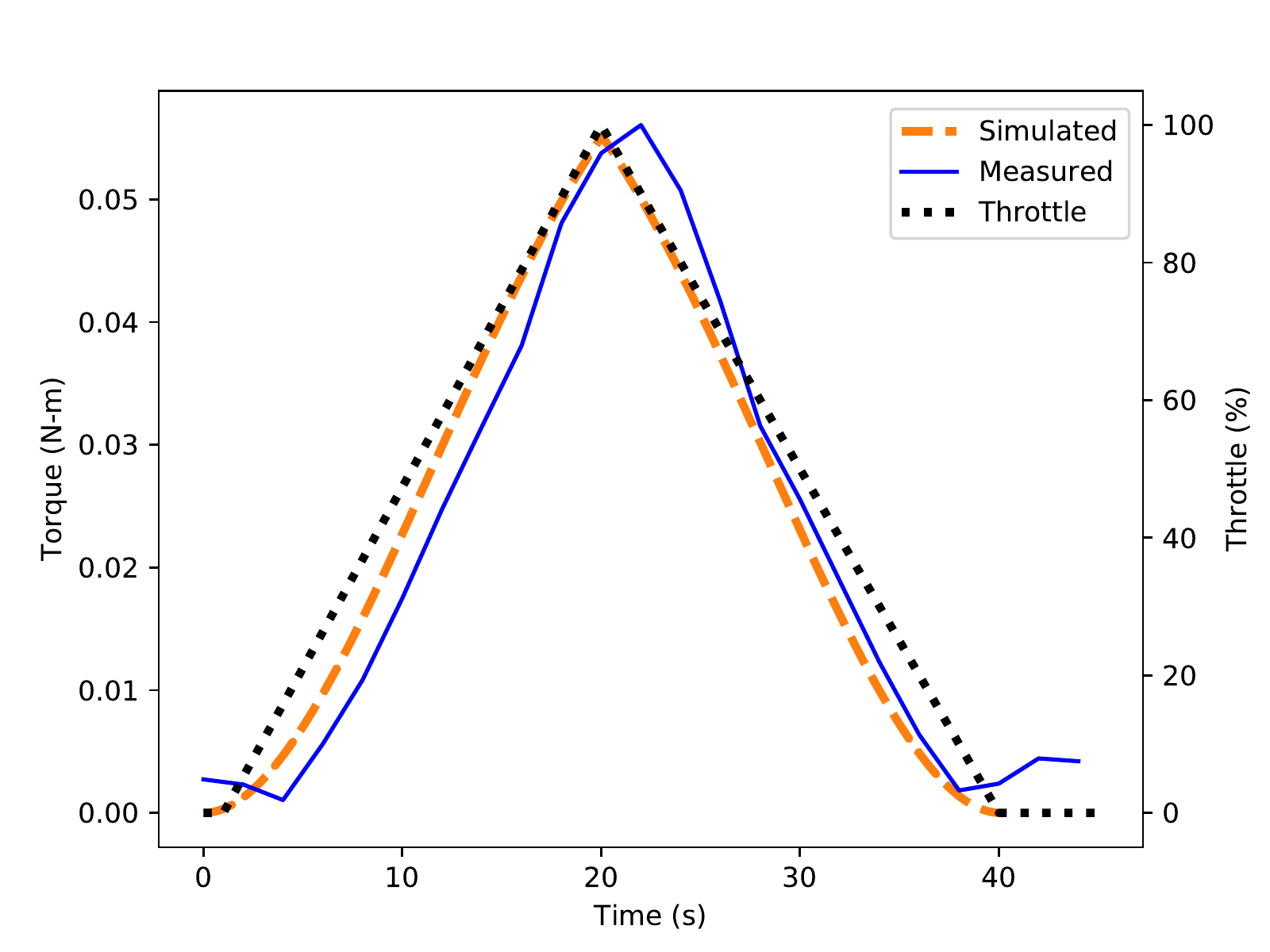}
        \caption{Torque}
        \label{fig:dyno:torque}
    \end{subfigure}
    \caption{Throttle ramp measurements. }
    \label{fig:motor_plots}
\end{figure*}

We performed $N=20$  independent measurements and report the maximum thrust and
torque values along with the 95\% confidence interval in Table~\ref{tab:params}. 
Additionally, the data was averaged together to generate the thrust response displayed in Fig.~\ref{fig:dyno:thrust} and the torque
response is displayed in Fig.~\ref{fig:dyno:torque}.
In these figures, the dashed black line is the percent throttle value applied.

\subsubsection{Motor Constants and Validation}
\begin{figure}
\centering
    \begin{subfigure}{0.8\textwidth}
        \includegraphics[width=\textwidth]{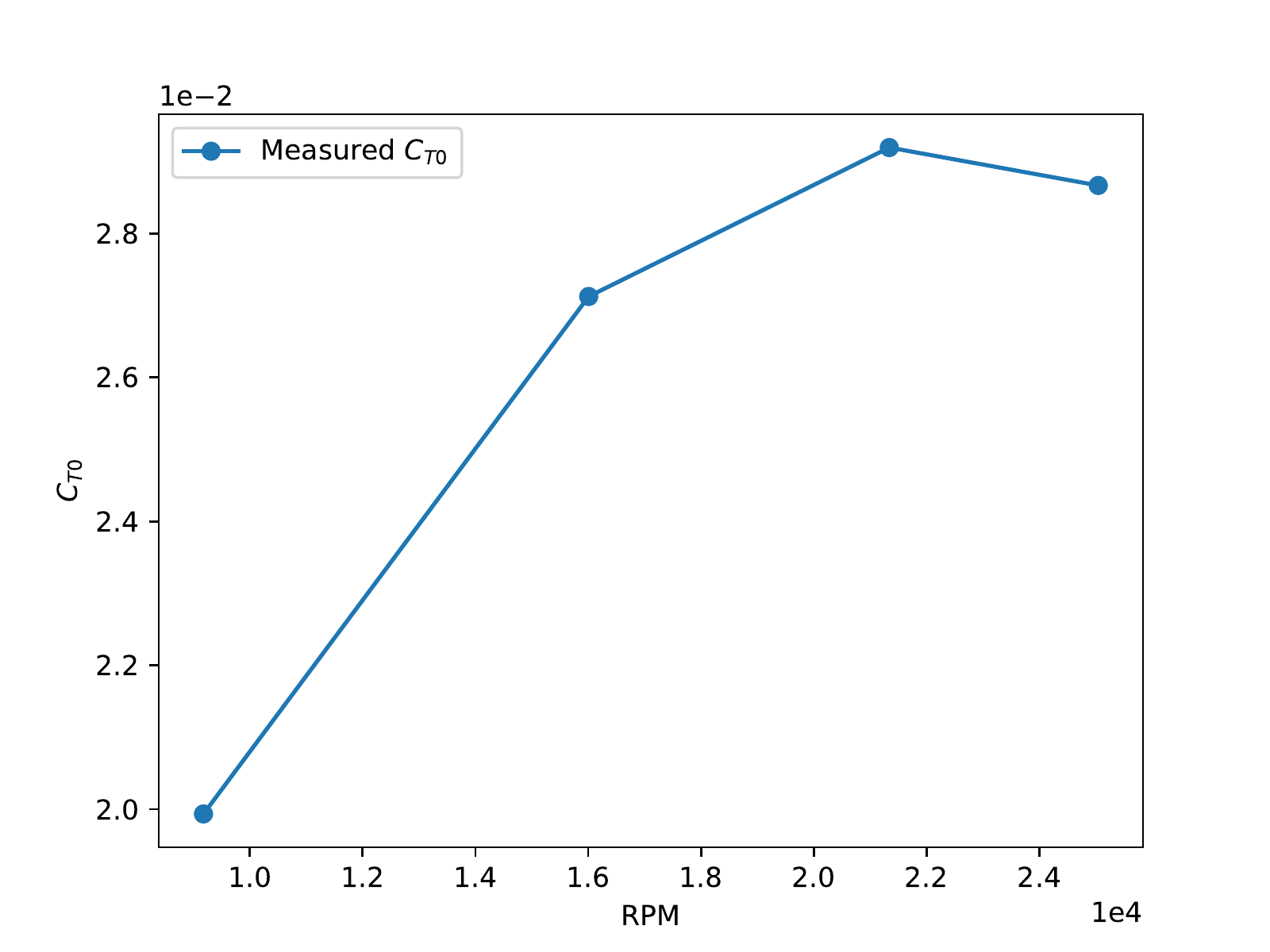}
        \caption{Thrust coefficient}
        %\label{fig:thrust}
    \end{subfigure}\\
    \begin{subfigure}{0.8\textwidth}
        \includegraphics[width=\textwidth]{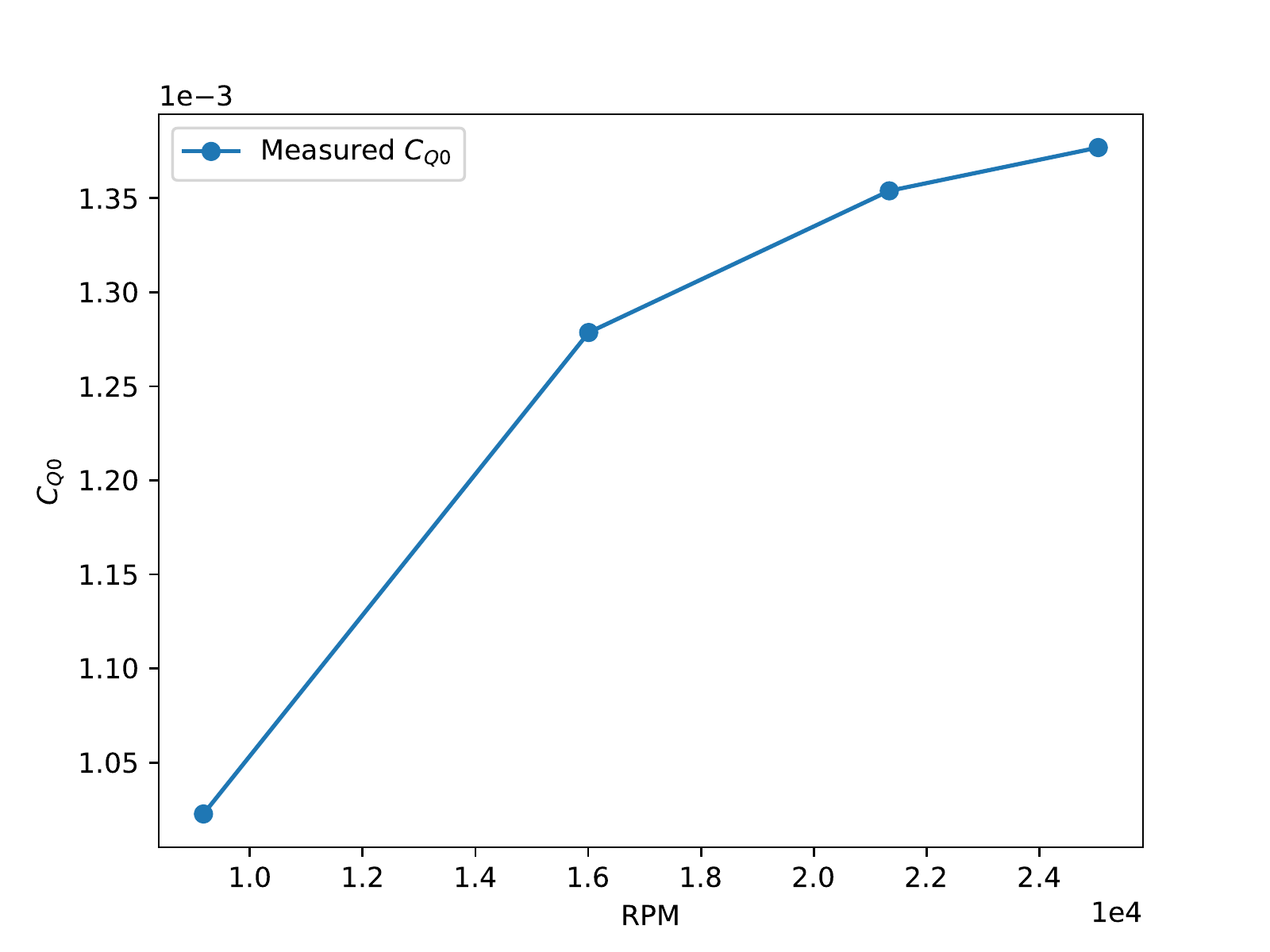}
        \caption{Torque coefficient}
        %\label{fig:torque}
    \end{subfigure}
\caption{Propeller coefficients}
\label{fig:coeff}
\end{figure}

\begin{figure}
\centering
    \begin{subfigure}{0.8\textwidth}
        \includegraphics[width=\textwidth]{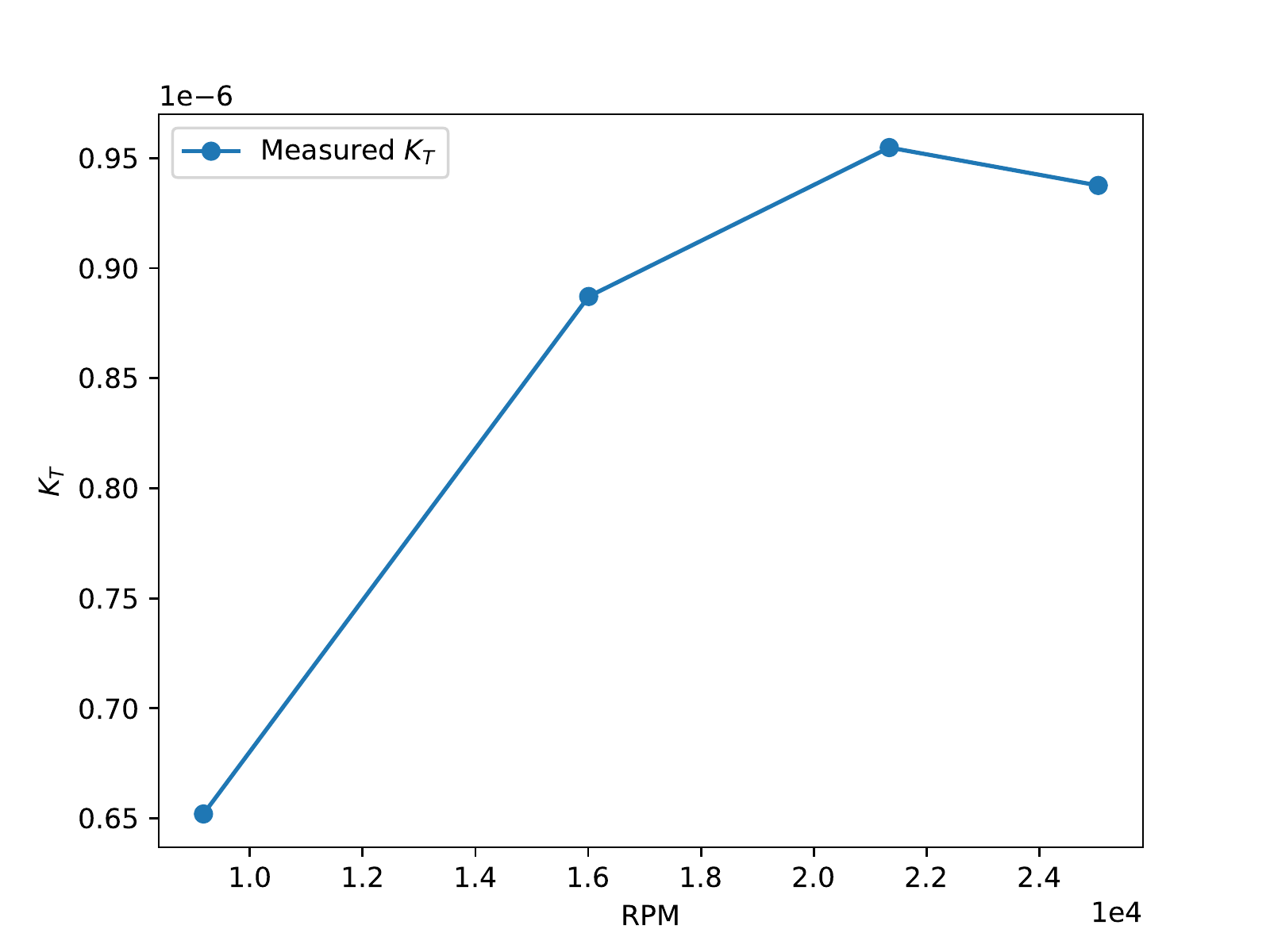}
        \caption{Thrust constant}
        %\label{fig:thrust}
    \end{subfigure}\\
    \begin{subfigure}{0.8\textwidth}
        \includegraphics[width=\textwidth]{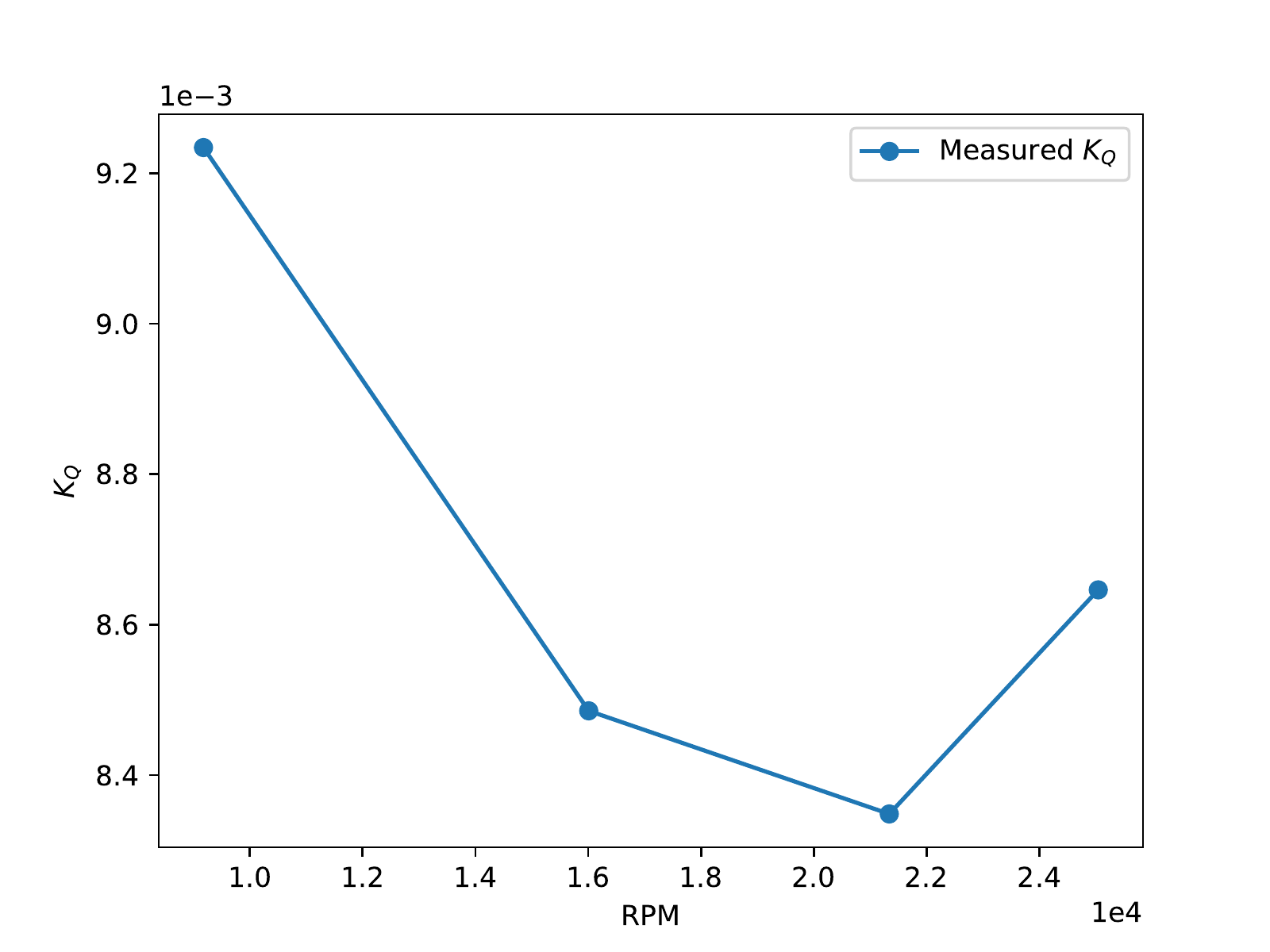}
        \caption{Torque constant}
        %\label{fig:torque}
    \end{subfigure}
    \caption{Motor model constants.}
    \label{fig:constants}
\end{figure}

Using the motor parameters found during experimentation we first derived the
thrust and torque coefficients and then use these to calculate the motor constants.
The thrust and torque coefficients in relation to the motor velocity is
displayed in Fig.~\ref{fig:coeff} while the thrust and torque motor constants
in relation to the motor velocity  
is displayed in Fig.~\ref{fig:constants}.

The motor model plugin configuration is completed with addition of the derived
motor constants providing the thrust and torque dynamics in simulation. 
With the completed model, we are able to validate the model using the dyno
simulator and compare the results to the experimental measured data. 
For thrust and torque these
results are
displayed in Fig.~\ref{fig:motor_plots}.
The results are comparable. 
We find the motor model to have an MAE of \torqueMAE for the torque output
compared to the real motor measurements, and
an MAE of \thrustMAE for the thrust output.
 
The real measurements do experience a greater
delay  
however this is likely attributed to the use of static motor constants where we
can visually see in Fig.~\ref{fig:constants} the constants, as a function of
the rotor velocity, are not only not static, but nonlinear.

\section{Simulation Stability Analysis}
\label{sec:stable}
Multirotors capable of achieving high angular velocities, which
induce large centripetal forces, are at risk of becoming unstable during
simulation. The problem is exaggerated as the number of links in a model
increase. The  root cause of the simulation instability is due to the 
type of coordinate solver used by the physics engine. Generally speaking, a
physics engine's coordinate solver can be
categorized as either a maximal coordinate solver or a generalized
coordinate solver~(also know as reduced coordinates)~\cite{coumans2014exploring}.  A maximal coordinate solver treats each body (link) as a
separate rigid body with 6 degrees of freedom (3 for position and 3 for
orientation). Constraints are then used to connect
bodies and enforce the intended degrees of freedom. Because the bodies are not
represented as a single entity this solver
is known to cause bodies to drift due to coordinate redundancies and
inaccuracies enforcing constraints. 
On the other hand, a generalized coordinate solver represents the bodies only
by the degrees of freedom.

The Gazebo simulator supports the
following physics engines: 
ODE~\cite{ode}, Bullet~\cite{bullet},
Simbody~\cite{sherman2011simbody} and
DART~\cite{lee2018dart}.
ODE, the default physics engine for
Gazebo, uses a maximal coordinate system while DART advertises its self as being
accurate and stable due
its use of generalized coordinate solver. 

In this section we evaluate the stability our model using both the ODE
and DART physics engines as a precursor for establishing which will be
necessary to use for flight controller
synthesis.

\subsection{Measuring Stability}
In this section we describe our algorithm Alg~\ref{alg:stable}  we developed for measuring the simulation stability
of our aircraft model. In summary, this algorithm measures the stability metric
$\delta$  defined as the sum of
the absolute value differences between the current distances of the bodies from their initial
state at each time step of the simulator~(defined at line 8).
The simulation is considered unstable is there occurs any drifting between the
bodies (\ie $\delta > 0$).

As the forces between the bodies becomes more complex, the simulation becomes
more likely to become unstable. Thus we must measure $\delta$ for a range of
angular velocities starting from still to its maximum achievable angular
velocity of the aircraft. To perform this measurement, we assume this is
a precursor to developing the flight controller therefore the idea is to
\textit{excite} each motor permutation to reach a variety of angular velocities.
More specifically we set each action $A$~(\ie control signal set) of the
total possible $2^\text{M}$ permutations ($\sigma$) where $M$ is the total actuator count of the aircraft and each motor control
signal can either be off, $0$, or full throttle $1$.  
After each time step $t$, the simulator receives the current state $S$  which
contains the aircraft's current angular velocity $\Omega$ for each roll, pitch and yaw
axis. We also obtain the set of all the aircraft's individual links relative
positions $V(t)$.
We can think of the links as a undirected weighted graph where each link position is a
vertex and the edge weight is the relative distance from one link to the
other. Using the set of vertices we calculate a Euclidean distance matrix
$D(t)_{i,j}$ for each of the $i$, $j$ link combinations. 
The stability metric is then calculated using this distance matrix and added to
the result vector $Y$.
One can then use $Y$ to find at which velocities the simulation is stable for.

\begin{algorithm}
\DontPrintSemicolon
\SetKwInOut{Input}{Inputs}\SetKwInOut{Output}{Returns}
\Input{A \gymfcTwo environment $E$ with the aircraft model to be measured.}
\Output{A vector $Y$ where each element is a tuple of the stability measurement
$\delta$ at the corresponding angular velocity $\Omega$.}

$Y \leftarrow \emptyset $\;

\For{ $A \in \sigma$} {
    \While{$t = 1,2,\dots$ }{
        $S \leftarrow E.Step(A)$\;
        $\Omega \leftarrow GetAngularVelocities(S)$\;
        V(t) = GetBodyPoses(t)\;
        $D(t)_{i,j}$ = EuclideanDistanceMatrix(V)\; 
        $\delta \leftarrow
        \displaystyle\sum_{i=0}^{N-1}\displaystyle\sum_{j=0}^{N-1}  |D(t)_{i,j}
        - D(0)_{i,j}|$\;
        $Y \leftarrow Y + \{(\delta,\Omega)\}$\;
    }
}
\Return{Y}

\caption{Model Stability Measurement}
\label{alg:stable}
\end{algorithm}

\subsection{Implementation}

To implement this algorithm we used  \gymfcTwo to issue the actions to the
aircraft and wrote a script (available in \gymfcTwo) using
\texttt{py3gazebo}\cite{py3gazebo} to interface with Gazebo's
messaging API. This interface is based on a publish-subscribe architecture
allowing the client to subscribe to a number of events. Our script implements
the \texttt{GetBodyPoses} function (Alg.~\ref{alg:stable} line 6) by subscribing to the 
 \texttt{poses\_stamped} messages which contains an array of the model
links and their corresponding positions $V(t)$. 
The stability metric results $Y$ are then used to generate 3D plots to
visualize the stability of the model.

\subsection{Stability Results}

\begin{figure*}
\centering
{\includegraphics[trim=0 0 0
0,clip,width=0.75\textwidth]{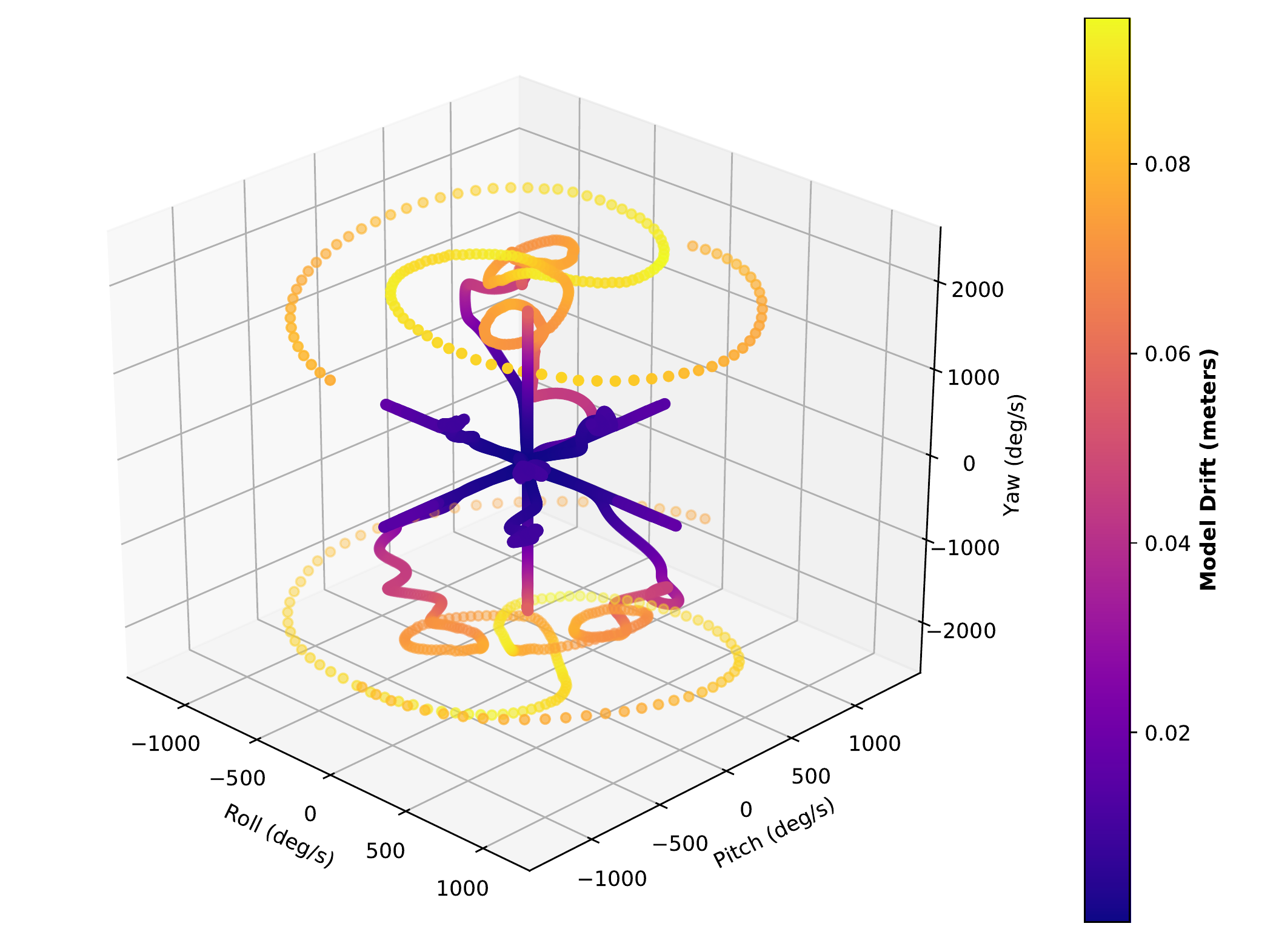}}
\caption{ODE physics engine with $2ms$ step size (500Hz).}
\label{fig:ode_002}
\end{figure*}

\begin{figure*}
\centering
{\includegraphics[trim=0 0 0
0,clip,width=0.75\textwidth]{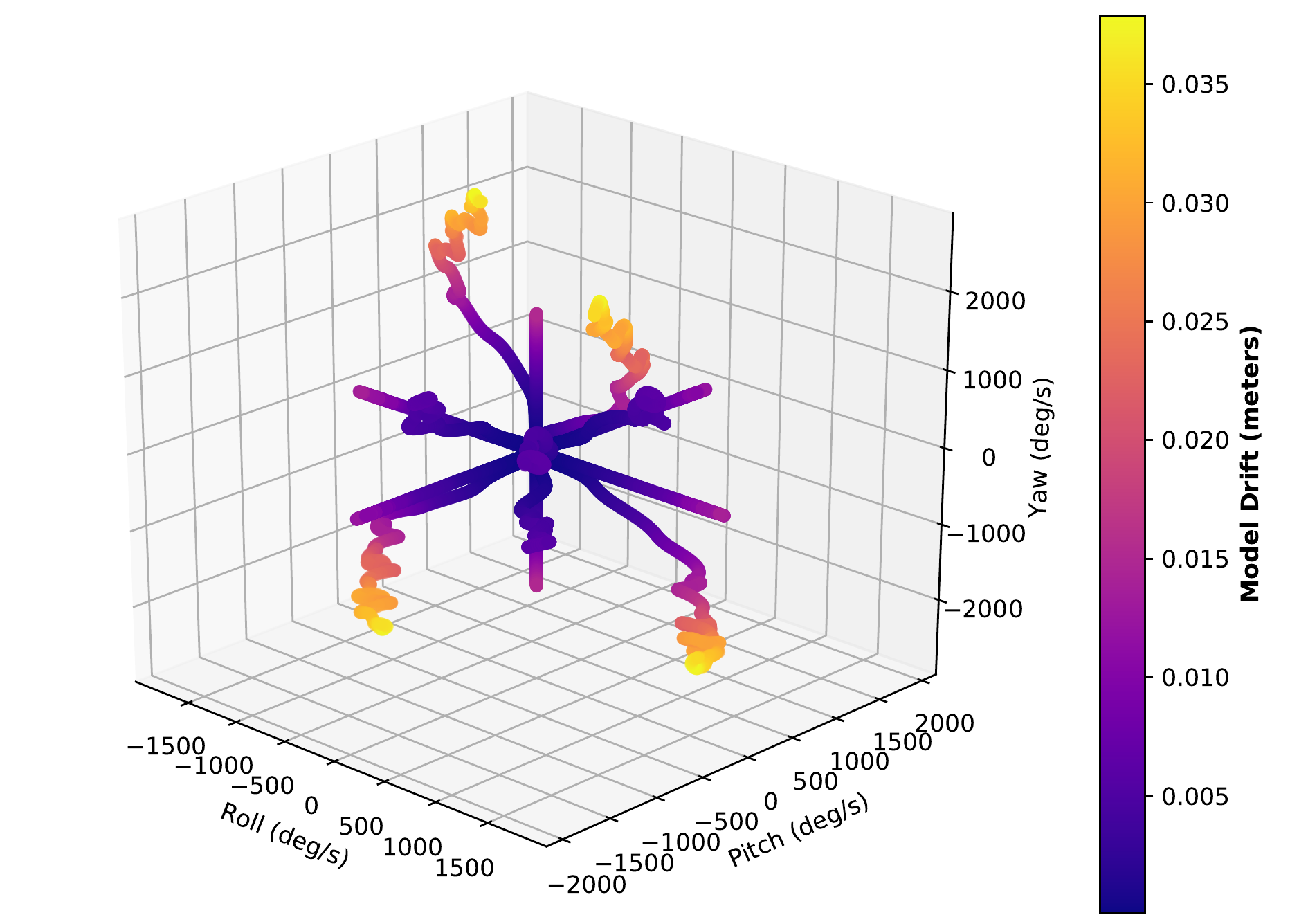}}
\caption{ODE physics engine with $1ms$ step size (1kHz).}
\label{fig:ode_001}
\end{figure*}

\begin{figure*}
\centering
{\includegraphics[trim=0 0 0
0,clip,width=0.75\textwidth]{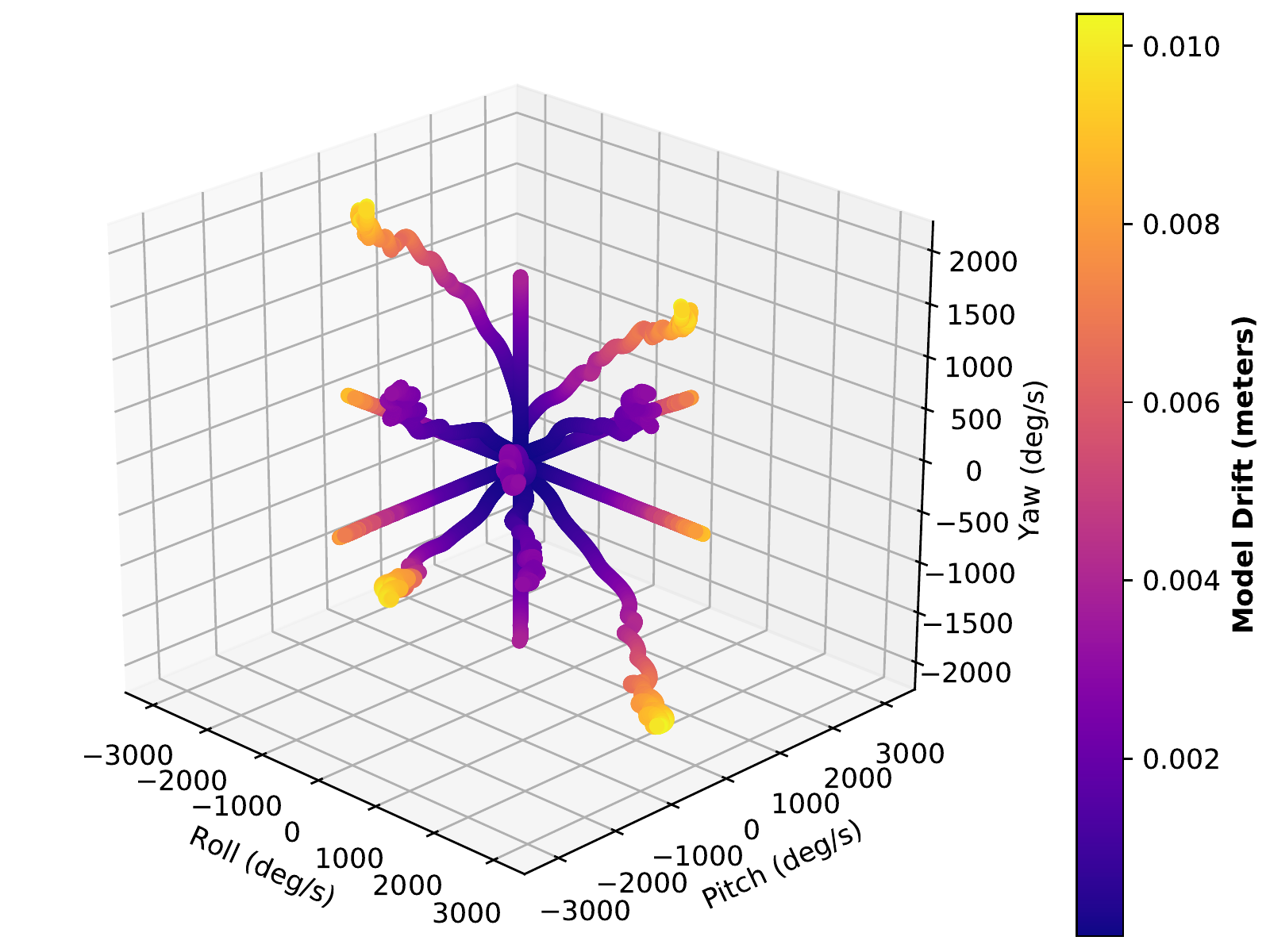}}
\caption{ODE physics engine with $500 \mu s$ step size (2kHz).}
\label{fig:ode_0005}
\end{figure*}

\begin{figure*}
\centering
{\includegraphics[trim=0 0 0
0,clip,width=0.75\textwidth]{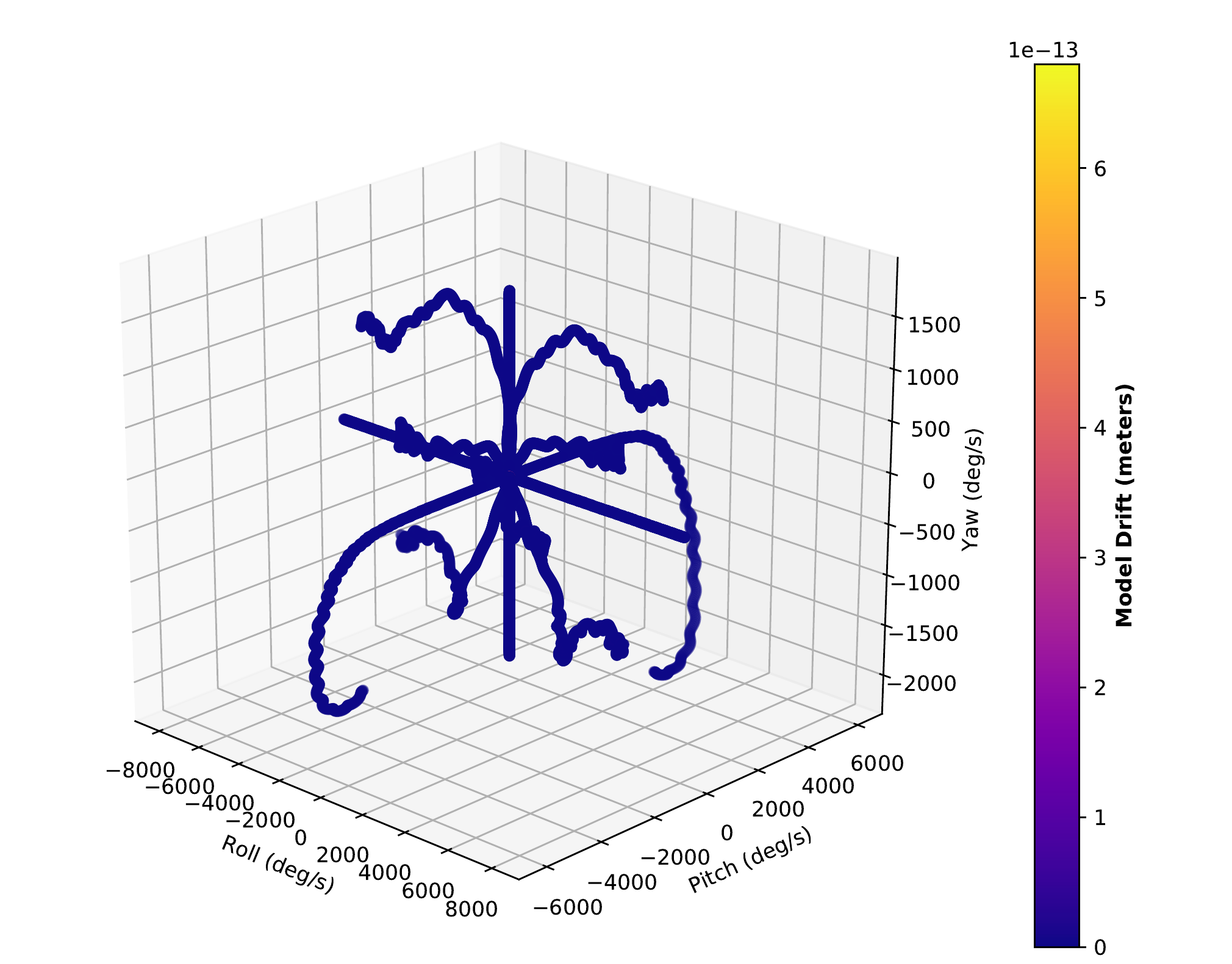}}
\caption{DART physics engine with $1ms$ step size (1kHz).}
\label{fig:dart}
\end{figure*}

We evaluated the stability of our model using  Gazebo's default physics engine
ODE with various simulation step sizes, and compared this to DART.
Our results for ODE using step sizes of 2ms~(500kHz), 1ms~(1kHz), and 500$\mu s$~(2kHz) 
are displayed in Fig.~\ref{fig:ode_002}, Fig.~\ref{fig:ode_001},
Fig.~\ref{fig:ode_0005} respectively. Results for DART are displayed in
Fig.~\ref{fig:dart}.

These simulation results show the execution of each of the motor permutations and
the angular velocity that is achieved. A heat map is used to indicate the value of
$\delta$ in meters for the corresponding angular velocity. 
As we can see the ODE physics
engine with the maximal coordinate solver results in a very unstable simulation
environment. 
For the largest step size of 2ms,  bodies start to separate at as low as
\instableODEtwo degrees per second with a max separation of \instableODETwoMax.
As the step size decreases (\ie simulation rate increases), stability
increases as the physics engine is
able to calculate the state more frequently. At a step size of 1ms, 
instability begin to occur at \instableODEone degrees per second with a max
separation of \instableODEoneMax and at a step size  of 500$\mu$s bodies begin
to separate at 
\instableODEfive degrees per second with a max separate of \instableODEFiveMax.
Now if we refer to Fig.~\ref{fig:dart} we can see that by using  a generalized
coordinate solver (\ie DART) zero drifting occurs.
Thus in summary, we find stability can be accomplished by two methods:
\begin{enumerate}
    \item  If using ODE or a maximal coordinate solver, decrease the step size
        until the minimum angular velocity in which body separate occurs is
        greater than the flight envelop of the aircraft.
    \item Use a physics engine with a generalized coordinate solver such as DART.
        It is recommended to use this option unless there is a specific reason
        in which this solver can not be used.
\end{enumerate}
Based on these findings \gymfcTwo has DART enabled by default.

\section{Neuro-flight Controller Training Implementation}
\label{sec:twin:impl}
In Section~\ref{sec:twin} we discussed in great detail our methodology for
creating a multicopter digital twin, one of the user provided modules. In this
section we will discuss our implementation of the remaining user supplied
modules to be used with \gymfcTwo for synthesizing a neuro-flight controller
via RL. 

\subsection{User Provided Modules}
\label{sec:impl:user}

\subsubsection{Control Algorithm and Tuner}
We use a neural network with the same architecture as in
Chapter~\ref{chapter:nf} with the only difference being the number of hidden
nodes has increased from 32 to 64.
Increasing the size of the network further did not
provide any additional performance benefits. 

Similar to Chapter~\ref{chapter:nf}  we trained using the PPO1 implementation
from OpenAI Baselines. 
We did put in a considerable amount of effort migrating to
Tensorforce~\cite{schaarschmidt2017tensorforce} in order to experiment with the beta
distribution and LSTM networks, however we could not reach even close to the
level of performance we could with OpenAI Baselines. 
This was even after using a hyperparameter tuning~\cite{falkner2018bohb}.
The primary challenge was
due to the lack of documentation and using hyperparameter definitions that
differ from the original PPO paper~\cite{schulman2017proximal}.  
An additional reason could be due to the differences in implementations of the
algorithm, which prior research has shown to greatly affect
performance~\cite{henderson2018deep}.
We did find that we needed to increase the step size to $1 \times 10^{-3}$
when using the beta distribution, yet this was still not enough to match the
performance provided by OpenAI Baselines.

For tuning~(\ie training) the \nn we used the hyperparameters defined in
Table~\ref{tab:twin:ppo}. The horizon and batch size were slightly increased
from Chapter~\ref{chapter:nf}. 
\begin{table}[]
    \centering
\begin{tabular}{l|c}
Hyperparameter            & Value          \\ \hline
Horizon (T)               & 512  \\
Adam stepsize             & \rPpoStepsize  \\
Num. epochs               & \rPpoEpochs    \\
Minibatch size            & 64 \\
Discount ($\gamma$)       & \rPpoDiscount  \\
GAE parameter ($\lambda$) & \rPpoGae      
\end{tabular}
\caption{PPO hyperparameters where $\rho$ is linearly annealed over the
course of training from 1 to 0.}
\label{tab:twin:ppo}
\end{table}

\subsubsection{Environment Interface}
To support RL training, our environment interface implements an OpenAI Gym, to
provide an interface for the PPO algorithm. The environment interface
implements the OpenAI Gym functions, \texttt{step}, and \texttt{reset}.
The \texttt{step} function makes a call to four important functions we have
implemented for our training environment:
\texttt{transform\_input},  \texttt{transform\_output}, \texttt{generate\_command} 
and \texttt{compute\_reward}. 
The functions \texttt{transform\_input} and  \texttt{transform\_output}  
support transforming  the aircraft state to the \nn input, and the \nn
output to the control signal, respectively. 
The function \texttt{generate\_command} generates the   
angular velocity setpoint for each axis of rotation  the agent must achieve for
the given time step. 
Lastly, the \texttt{compute\_reward} function 
calculates the reward for the agent at each time step. In the remainder of this
section we will discuss each function in detail.

\textbf{Transformation functions.} The \texttt{transform\_input} function takes as input the aircraft state, $S$
which contains the angular velocity~$\Omega$
and the desired angular
velocity~$\Omega^*$, and computes the network input as defined in
Eq.~\ref{eq:nf:input}.

The \texttt{transform\_output} functions scales and adds a bias to the \nn output $y$ to
derive 
the
control signals $u$ in the range $[0,1]$ required by the \nf firmware. 
Because the output of the \nn is the mean from the Gaussian distribution, the
output is first clipped to the action bounds $y_\text{low}=-1$ and
$y_\text{high}=1$. Next the scaling and bias is performed, where $u_\text{low}=0$,
and $u_\text{high}=1$,
\begin{align}
y &= \text{clip}(y, y_\text{low}, y_\text{high})\\
u &= \frac{(u_\text{high} - u_\text{low}) (y - y_\text{low}) }{ (y_\text{high} -
y_\text{low})} + u_\text{low}
\end{align}

\textbf{Command generation.} The \texttt{generate\_command} function computes
the angular velocity setpoint. The objective of
the agent is to reach this setpoint. 
From
Chapter~\ref{chapter:nf} we found that is was important to expose the agent to,
not only acceleration, but also deceleration for transferring the agent to the
real world. Thus \newgym  
continuously generates new commands until a predefined time out is reached.   
However, with such a long episode, analyzing the individual step response
caused by the change in the command input
increases in complexity as you have to slice the episode into the individual
pulses before analyzing.  
Additionally, during early stages of training,  the agent can get the aircraft
into extremely fast  angular velocities, well exceeding the target, which is
undesirable to allow this behavior to last the entire episode.

To address these concerns,
this command generator simplifies the environment to only a single pulse. We
begin by setting $\Omega^*=[0, 0, 0]$ for half a second. This allows the agent
to learn its idle or hover state. A command is then
randomly sampled  and held for two seconds which teaches the agent to
accelerate to a desired angular velocity, followed by a steady state. The command is then set back to $\Omega^*=[0, 0, 0]$ 
for an another additional two seconds to teach deceleration. 
The question becomes, what is the best distribution to sample the setpoints
from? In previous chapters we sampled from uniform random, however through our
experience, the agent will perform best through its sampled range. It is more
desirable to be accurate within the flight envelop than extreme cases.

\begin{figure}
\centering
\begin{subfigure}{0.5\textwidth}
    {\includegraphics[trim=0 0 35
		35,clip,width=\textwidth]{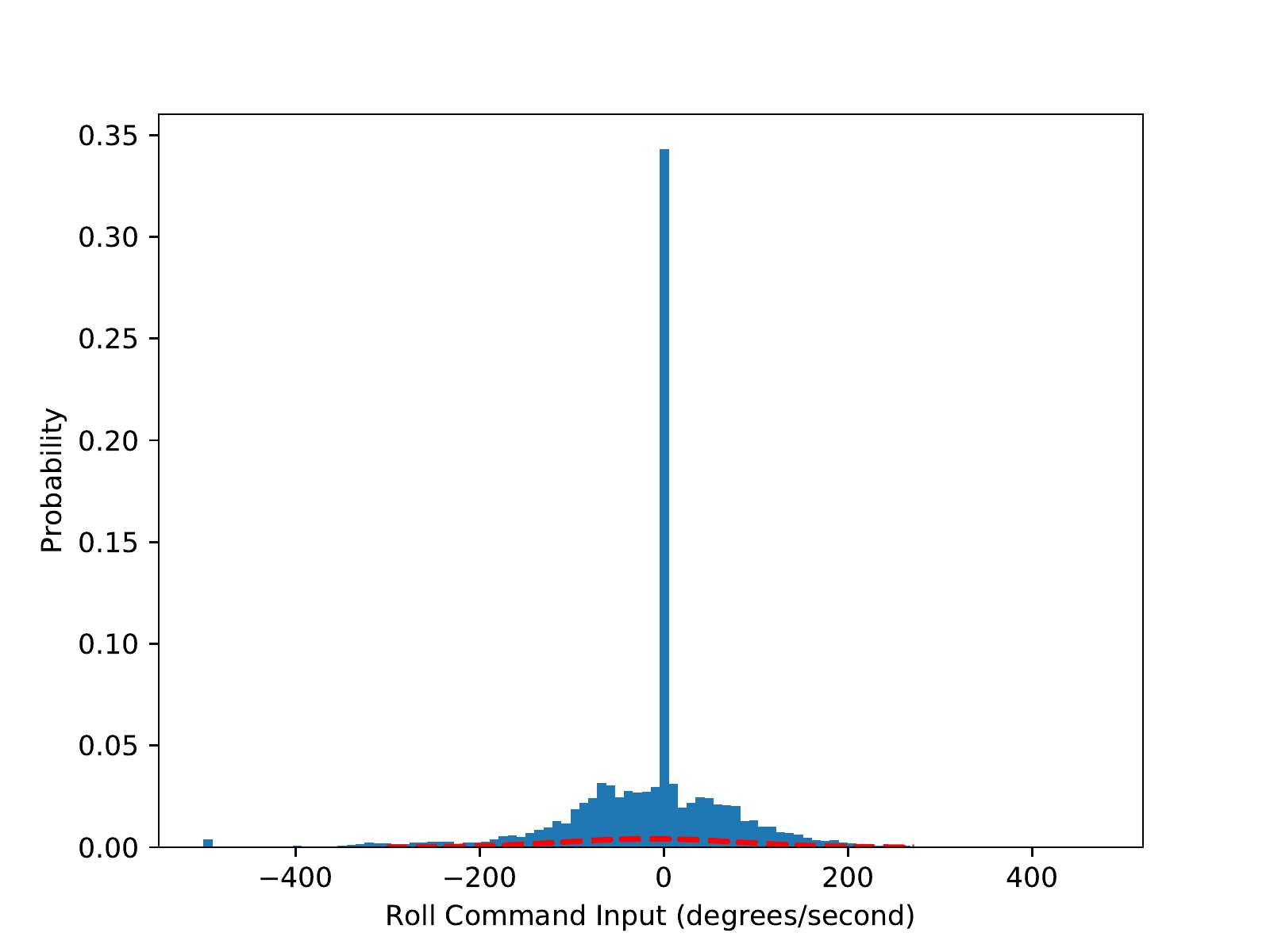}}
	\caption{Roll}
\end{subfigure}
\\
\begin{subfigure}{0.5\textwidth}
    {\includegraphics[trim=0 0 35
		35,clip,width=\textwidth]{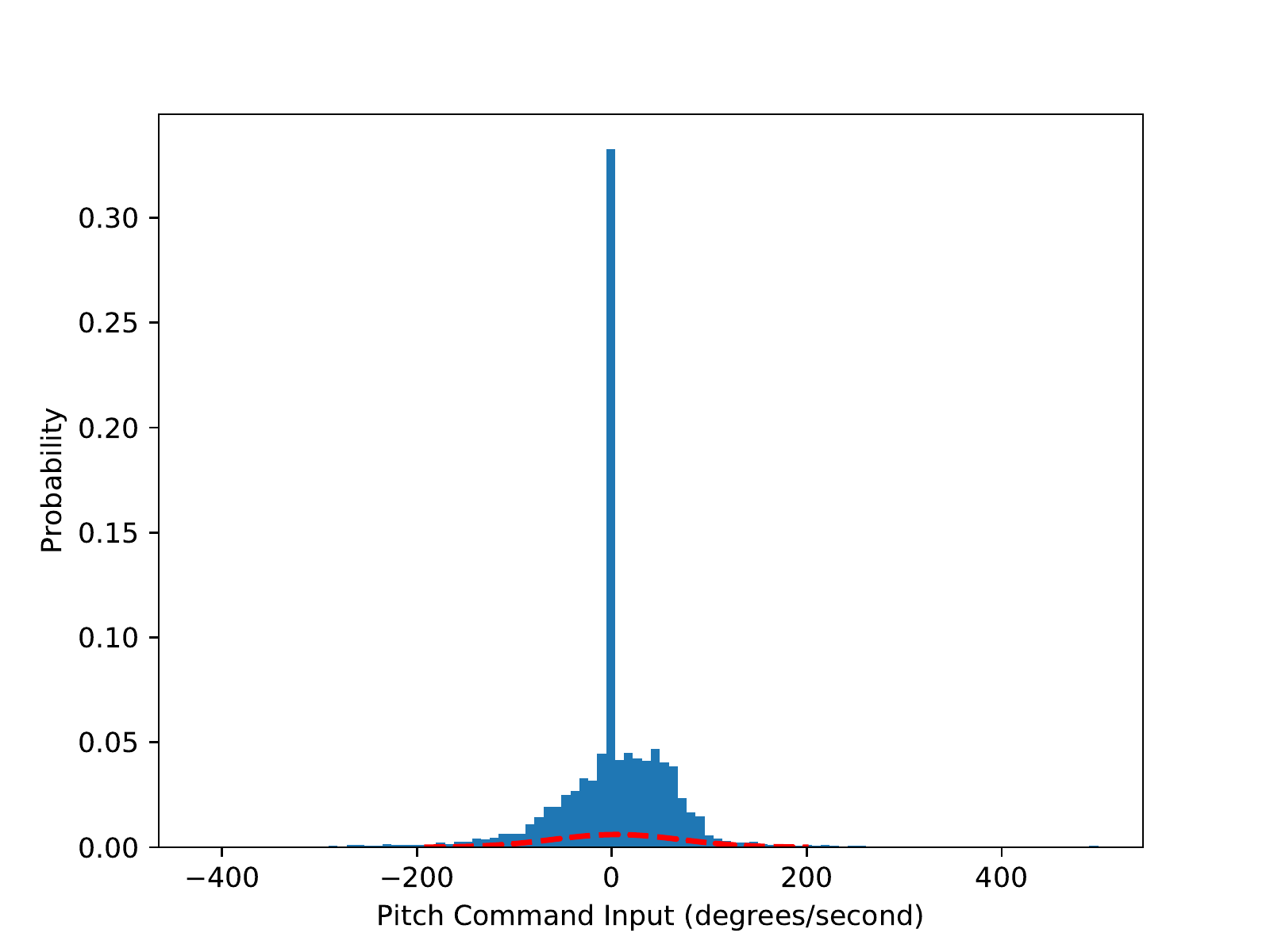}}
	\caption{Pitch}
\end{subfigure}
\\
\begin{subfigure}{0.5\textwidth}
    {\includegraphics[trim=0 0 35
		35,clip,width=\textwidth]{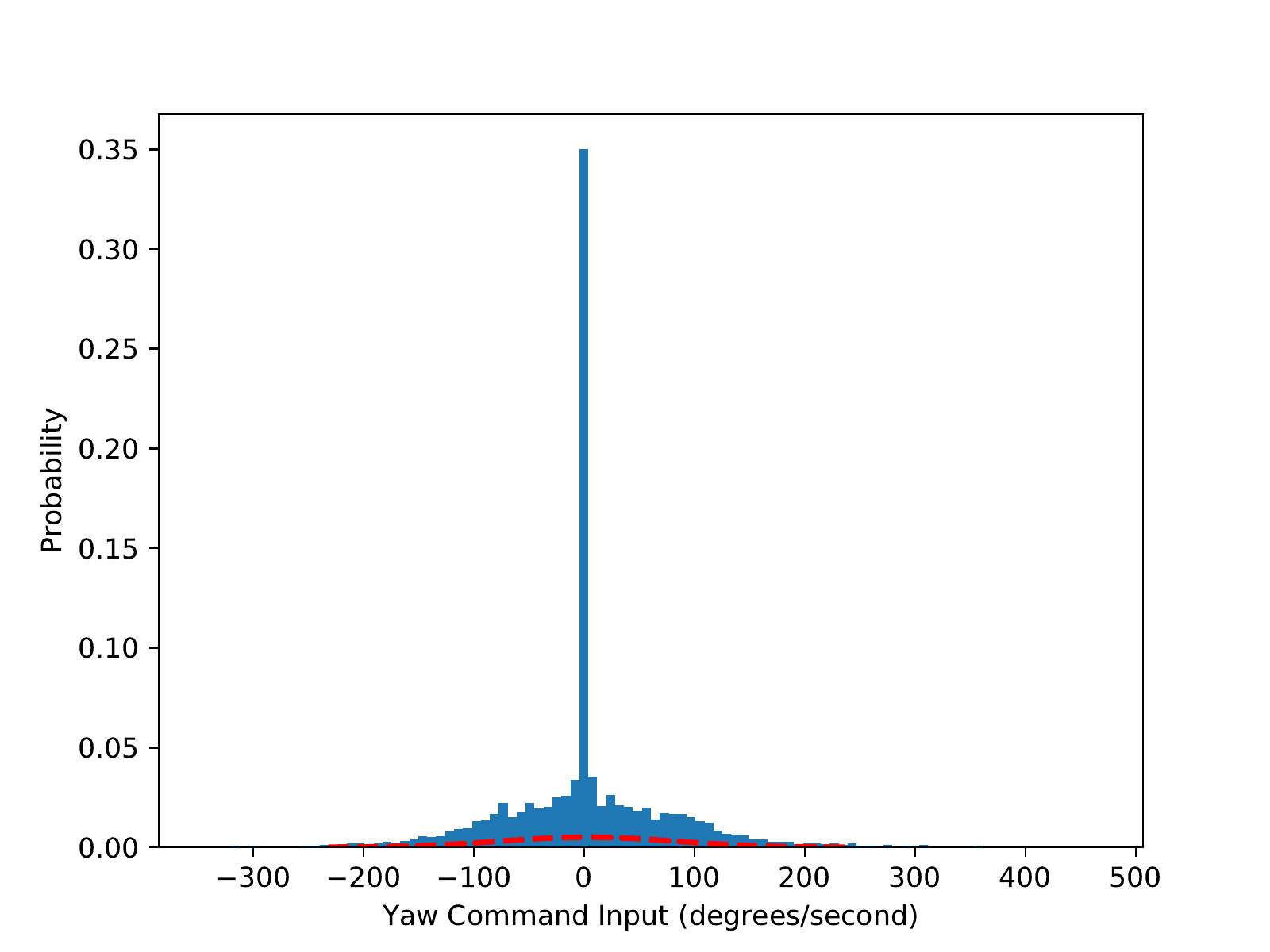}}
	\caption{Yaw}
\end{subfigure}
\caption{PDF of Pilot Command Inputs}
\label{fig:cmd_input}
\end{figure}

To discover the underlying command input distribution, we obtained a total of 
\cmdN  pilot input commands,  from real test flights, and created a histogram with 20 bins. Results are show in
Fig.~\ref{fig:cmd_input} for each axis, while a dashed red line is a fitted
 to a normal distribution PDF. As we can see the command inputs
roughly fit a normal distribution with an average control input of \cmdAve with
a standard deviation of \cmdStd. The average command input, centering around zero
degrees, was expected. This is because the majority of the time during flight
 a  heading is maintained  in which the angular velocity changes very little.
Minor adjustments may be made to compensate for external disturbances acting upon
the aircraft. The variance will be correlated to the type of
flying performed. For example, frequent aggressive aerobatic maneuvers  would
use a greater range of the flight envelope resulting in a wider variance, while more conservative tasks, such as aerial
photography and video would result in a narrower variance. 
  Based on these results, the command generation function samples  from a normal distribution
with $\mu=0$, however we increase the standard deviation to  $\sigma=100$ because we want to
evaluate the performance performing aggressive maneuvers on the edge of the
flight envelope.

\textbf{Reward function.} Our reward function is an improved iteration from Chapter~\ref{chapter:nf} with
some additional changes to increase stability. The reward function is defined
in Alg.~\ref{alg:reward} and is
called at each time step. In lines 1-4, a reward is given capturing the agents
progress to minimizing the error. We found this to provide more stability than the
sum of squared errors. At line 5,  the agent is penalized for the max changes
in the 
control signal to reduce output oscillations. This is scaled by the constant
$\beta > 0$.
At line 6 and 7, a reward is given to the agent for reducing their average control
signal output if they are in an error band defined
by the percent $\epsilon$  of the target angular velocity.
The remaining penalties, defined in lines 8 to 12, are to help stabilize the
learning process and consist of events that should never happen. 
We define a max penalty high enough such that the agent will not repeat the
behavior. We set \texttt{MAX\_PENALTY} = $1 \times 10^9$ however there is some
flexibility to this value. 
Line 8,
penalizes the agent for saturating the output. Recall the agents output~($a$),
for a stochastic policy,
is
the mean of a Gaussian distribution. The action is unbounded and thus can exceed the bounds of
the control signal. Although this value is clipped during the transformation
function, we found without this penalty, the angular velocity of the aircraft
would rapidly increase
and not come back down. We believe this to be primarily caused by the delayed motor
response as the control signal provided by the agent will not immediately
result in a change. Lines 9 and 10 penalize the agent if they have saturated
all the control outputs which should never happen. 
While lines 11 and 12
penalize the agent in cases where they  do nothing.
This penalty is derived from the basic quadcopter
dynamics such that at least two motors are required to preform one of desired
commands. If more than two motors are zero, and the target angular velocity is
not zeros, the penalty is applied.

\begin{algorithm}
\DontPrintSemicolon
\SetKwInOut{Input}{Inputs}\SetKwInOut{Output}{Returns}
\Output{Reward $r$ at time $t$}
$r \leftarrow 0$\;
$r_{e,t} \leftarrow -(e_{\phi}^2 + e_{\theta}^2 + e_{\psi}^2)$\;
$r \leftarrow r +  r_{e,t} - r_{e,t-1}$\;
$r_{e, t-1} \leftarrow r_{e, t}$\;
$r \leftarrow r - \beta \texttt{max}(|u_{t} - u_{t-1}|)$\;
\If {$|e| < \epsilon |\Omega^*|$}{
    $r \leftarrow \alpha \left( 1-\bar{u} \right)$\;
}
$r \leftarrow r - \texttt{MAX\_PENALTY} \sum \texttt{max}(a - 1, 0)$\;
\If {$\forall u_i \in u, u_i \equiv 1 $} {
    $r \leftarrow r -\texttt{MAX\_PENALTY}$\; 
}
\If {$ 2 < \sum_{u_i \in u: u_i \neq 0} 1 \texttt{~and~} \exists \Omega_i \in
    \Omega^*: \Omega_i
    > 0$} {
    $r \leftarrow r -\texttt{MAX\_PENALTY}$\; 
}
\Return{r}
\caption{Reward function}
\label{alg:reward}
\end{algorithm}

\section{Evaluation}
\label{sec:twin:eval}

In this section we synthesize a  neuro-controller via RL using the  \gymfcTwo
implementation. Most importantly, the evaluations differs from the other
evaluations such that the controllers are tuned and  evaluated in simulation
using our digital twin of \aircraft.

We evaluate the flight controller in simulation, and also in
the real world. As we have done in earlier chapters, we also provide a
simulation baseline using a PID controller. Using a PID tuning platform
implementation of \gymfcTwo, we tune our PID controller and compare the
performance to that of the neuro-flight controller in simulation. 

\subsection{Neuro-Controller Synthesis}
\label{sec:twin:train}

Before training, we disabled gravity in the simulation environment as we did in \newgym.
We did experiment with gravity enabled, and while the agent is able to minimize
the error without problem, minimizing the control output and oscillations were
more difficult. We believe this is partially explained by less exposure to certain
conditions that encourage our desired behavior. In other words, with gravity
disabled, we have no additional downward force acting on the aircraft,
therefore, in the simulation environment we do not need to care about how the
orientation will affect the control of the aircraft.  With gravity enabled, if a command puts the
aircraft in a state outside of its flight envelope, (\eg perpendicular to the
ground), it will negatively affect training. Thus the only time the agent
is exposed to a condition for idle, is at the beginning of episode when the setpoint is zeros. Intuitively  we thought adding a quaternion to the \nn input
would help the agent distinguish between these states however this did not help
reduce the control output and oscillations. In future work we will investigate
how we can build more stable training environments for when gravity is
enabled to create a more realistic training environment. 

Using our RL implementation of \gymfcTwo, we train our \nn for 10 million time
steps with the architecture and hyperparameters defined in
Section~\ref{sec:impl:user}. 
Training was conducted on a desktop computer
    running Ubuntu 18.04 with an eight-core i7-7700 CPU and an NVIDIA GeForce GT
    730 graphics card. 

During training, every 100,000 steps, a Tensorflow checkpoint of the policy is
saved. In parallel, a monitoring program watches for new checkpoints. The
monitoring program allows for the training progress to be monitored and
evaluation of the performance of the controller. This is helpful during reward
engineering to identify if the rewards are doing what we actually intend them
to do and identify trends.  
If we recall from Section~\ref{sec:reward}, during training the output of the
\nn is stochastic to aid in exploration. However when deployed, we use the
deterministic output of the \nn. The monitoring program evaluates each
checkpoint, deterministically, for a total of five episodes. 

\begin{figure*}
\centering
{\includegraphics[trim=0 0 0
    0,clip,width=0.7\textwidth]{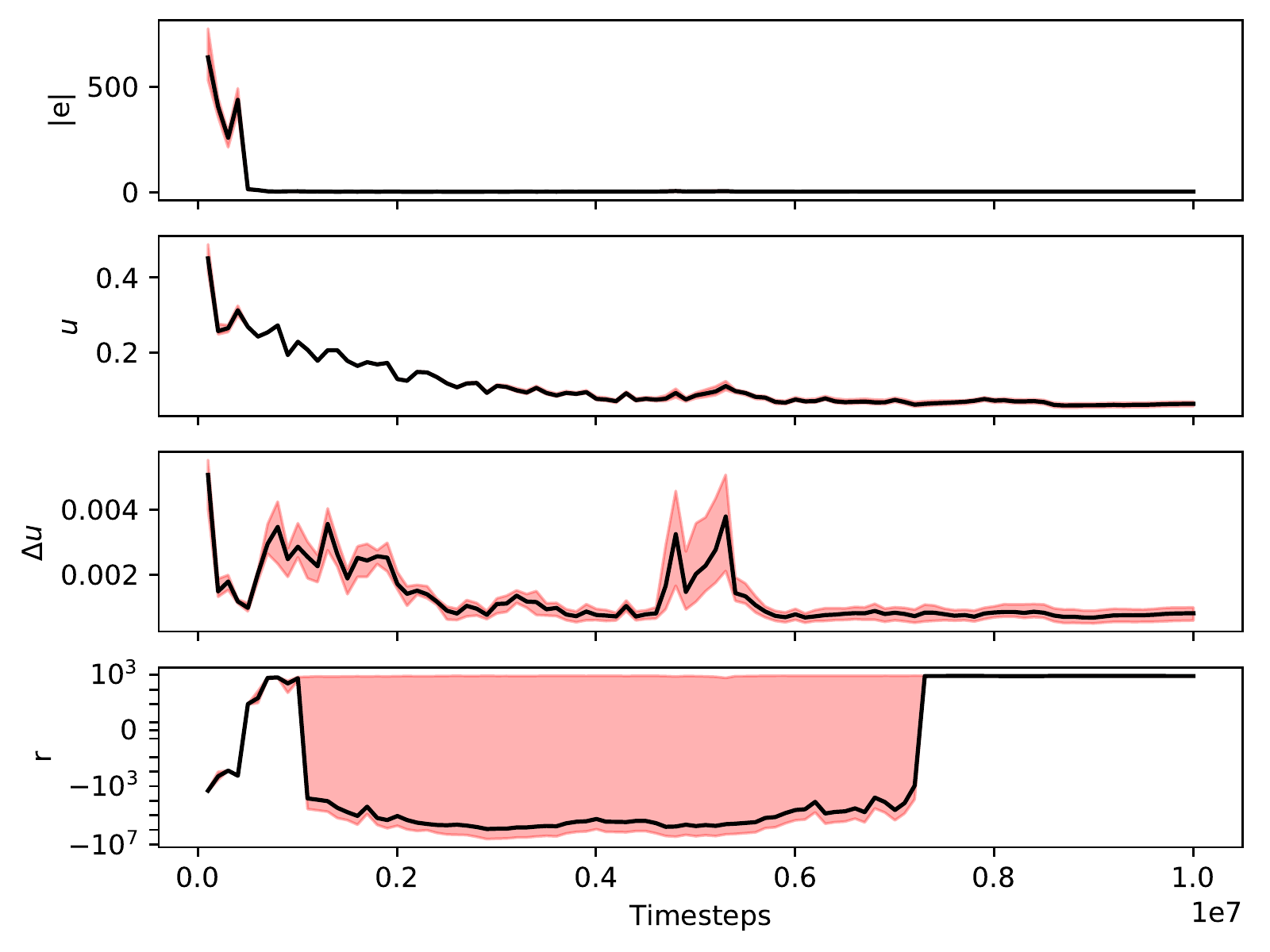}}
\caption{PPO training validation.}
\label{fig:twin:train:validate}
\end{figure*}

Fig.~\ref{fig:twin:train:validate} displays the results of the checkpoint 
validations throughout training for four metrics. The plots report the average metrics for each
checkpoint indicated by the black line, while the red regions define the min
and max values experienced for each metric. 
The first subplot reports the mean absolute error represented by $|e|$. The
second subplot is the average control output $u$, while the third subplot is
the average change in the control output $\Delta u$.
Last we have the average reward $r$ the controller would have received during
training (remember this is validation, not training). 

From the validation plot we can see the agent first minimizes the error which,
in turn, accumulates the majority of the reward. This happens very quickly and
consistently, within one million time steps. 
Once the error has been minimizes, and the agent is in the error
band, the agent begins to accumulate more reward for minimizing the control output. 
Minimizing the control output, also helps to reduce high amplitude oscillations
and reduce the output oscillations. 
As we can see by the increase in red, there is more variation in the change to
the control signal. Developing a reward system to balance the control output
effort
and oscillations has been one of the greater challenges.  
The last subplot, displaying the reward, uses a symmetrical log scale on the y
axis. The dip we see is due to the agent attempting to further decrease the error
by increasing acceleration, and in turn,  over saturating its control outputs. As a result it is hit with the maximum
penalty. Eventually toward the end of training the agent consistently acquires
the maximum rewards. 

Once training was complete, we select the checkpoint that provided the
most stable step responses, which occurred after 2,500,000 steps to use as our
flight controller policy.

\subsubsection{PID Baseline Evaluation}
\begin{figure}
\centering
{\includegraphics[trim=0 100 0
0,clip,width=\textwidth]{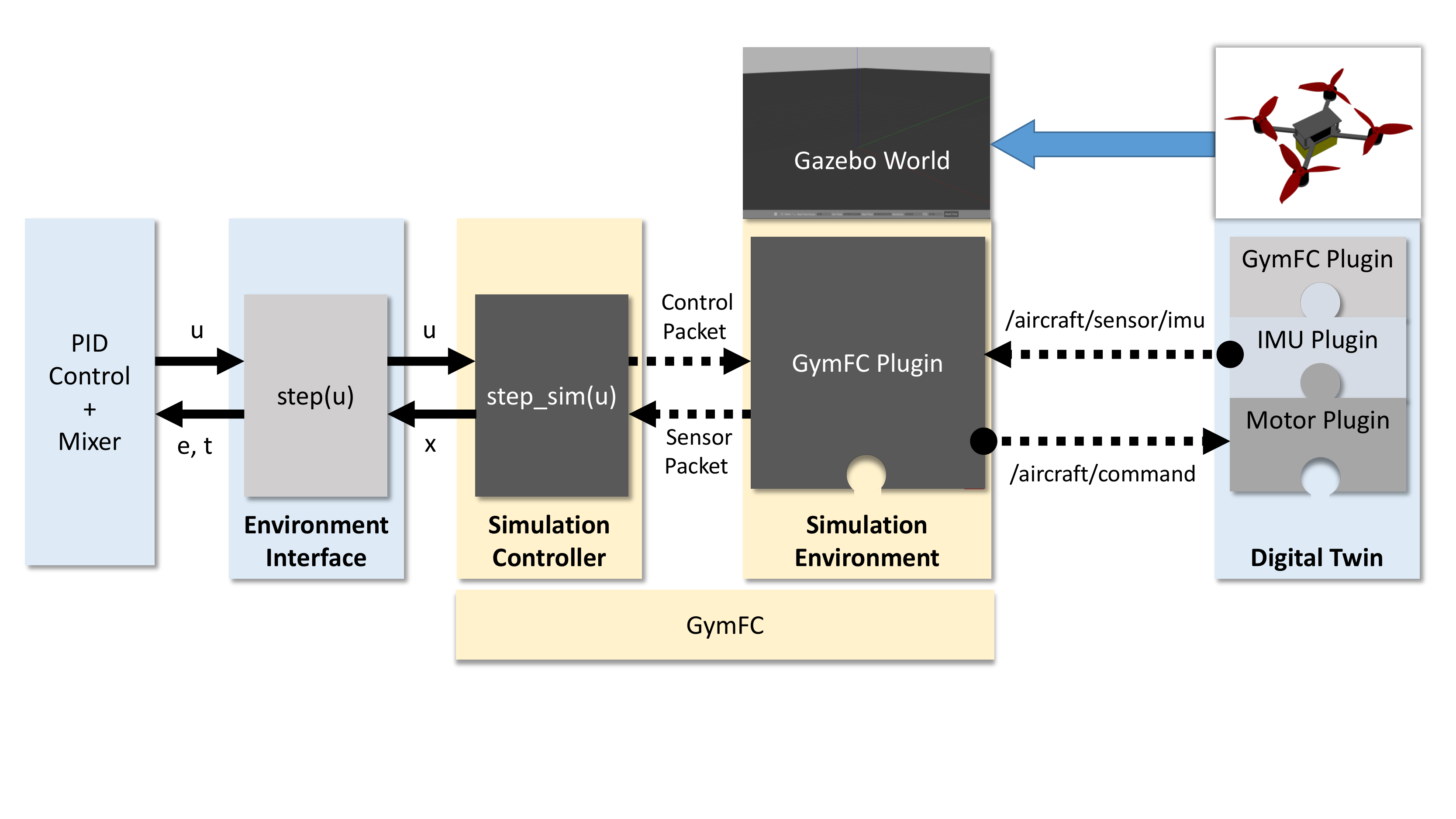}}
\caption{Implementation of \gymfcTwo for PID control tuning and SITL testing.}
\label{fig:gymfcpid}
\end{figure}

To provide a performance baseline
for our simulation evaluation, we use the traditional PID control algorithm.
However the PID attitude controller requires 9
static gains to be tuned specific for our new digital twin. To accomplish this
task  we implemented a tuning platform using \gymfcTwo. This architecture is displayed in
Fig.~\ref{fig:gymfcpid}. We will now discuss the user provided modules.

\textbf{Control algorithm and tuner.}  We use the open source Ivmech PID
Controller\cite{pidimpl} for the implementation of our attitude PID controller,
for each of the three axis. As we have
previously discussed in Section~\ref{sec:bg:pid}, the collective output of the
three PID controllers, must be mixed together to form the control signal. We
ported over the mixing implementation from Betaflight~\cite{betaflight} and with
a little glue code to create our PID controller.  
To tune the PID controller, we use the classical
Ziegler-Nichols method~\cite{ziegler1942optimum}.

\textbf{Environment Interface} We create an environment interface to provide
command generation, and 
transformation functions of the aircraft state. To support tuning using the
Ziegler-Nichols method, at $t=0$, we issue a command that is held for the
entire duration of the simulation to 
obtain the step response
from the controller.  
The environment can be provided with a
specific setpoint to allow each axis to be tuned independently, or if absent,
defaults to randomly sampling a setpoint so the performance can be randomly
evaluated.  
The environment interface also  transforms the aircraft state, into
the angular velocity error which is requires  as input to the PID controller. 

Using the \gymfcTwo PID tuning platform, we obtained the following gains for
each axis of rotation: $K_\phi = [2.4, 33.24, 0.033 ]$, $K_\theta =
[4.2, 64.33, 0.059]$, $K_\psi = [2, 5, 0]$, where
$K_\text{axis} = [K_P,K_I,K_D]$ for each proportional, integrative,
and derivative gains, respectively.

\subsection{Simulation Evaluation}
In this section we evaluate the neuro-flight controllers performance in
simulation, and compare it to the previously tuned PID controller.
We evaluated both controllers against 100 never before seen command inputs,
using the episode environment~(pulse control input) used during training of the \nn-based controller.
The average metrics are reported in Table~\ref{tab:twin:sim_valid_ppo} for
the \nn-based controller, while the PID controller metrics are reported in 
Table~\ref{tab:twin:sim_valid_pid}. Overall, results are consistent from our
previous findings. The \nn-based controller trained via PPO outperforms the PID
controller in all of our
error metrics. 
We additionally calculated the average control output produced by the controller,
as well as the average absolute change in the control output. These values are also
associated with their error, falling within a 95\% confidence interval. For the \nn controller,
the average control output and change in output was $\overline{u}=$\ppoUMu, and
$\overline{|\Delta u|}=$\ppoUDiffMu respectively.
While for the PID controller, the average control output and change in output was 
$\overline{u}=$\pidUMu and 
$\overline{|\Delta u|}=$\pidUDiffMu respectively. Although the PID controller
uses less effort, for the first time in this work, we have been able to synthesize a
controller that results in less change to the control output, and in effect,
less oscillations, than that produced by a PID controller.

\def\meta{3\xspace}
\def\metb{2\xspace}
\def\metc{2\xspace}
\def\metd{2\xspace}
\def\mete{148\xspace}
\def\metf{135\xspace}
\def\metg{66\xspace}
\def\meth{117\xspace}
\def\meti{3,311\xspace}
\def\metj{2,235\xspace}
\def\metk{2,075\xspace}
\def\metl{2,541\xspace}
\def\metm{148,804\xspace}
\def\metn{135,805\xspace}
\def\meto{66,807\xspace}
\def\metp{117,138\xspace}
\def\metq{6,233\xspace}
\def\metr{4,033\xspace}
\def\mets{3,435\xspace}
\def\mett{4,567\xspace}
\def\metu{237,168\xspace}
\def\metv{211,846\xspace}
\def\metw{95,983\xspace}
\def\metx{181,666\xspace}

\begin{table}[]
    \centering
{\setlength{\tabcolsep}{0.2em}
\def\arraystretch{1.15}%
\begin{tabular}{l|ccc|c|}
\cline{2-5}
                             & \multicolumn{4}{c|}{NN Controller (PPO)}                                                \\ \hline
\multicolumn{1}{|l|}{Metric} & Roll ($\phi$) & Pitch($\theta$) & Yaw ($\psi$)  & \cellcolor[HTML]{DAE8FC}Average       \\ \hline
\multicolumn{1}{|l|}{MAE}    & \meta         & \metb           & \metc         & \cellcolor[HTML]{DAE8FC}\metd         \\
\multicolumn{1}{|l|}{MSE}    & \mete      & \metf          & \metg        &
\cellcolor[HTML]{DAE8FC}\meth        \\
\multicolumn{1}{|l|}{IAE}    & \meti     & \metj       & \metk     &
\cellcolor[HTML]{DAE8FC}\metl     \\
\multicolumn{1}{|l|}{ISE}    & \metm  & \metn      & \meto    &
\cellcolor[HTML]{DAE8FC}\metp    \\
\multicolumn{1}{|l|}{ITAE}   
& \metq    & \metr    & \mets    & 
\cellcolor[HTML]{DAE8FC}\mett    \\
\multicolumn{1}{|l|}{ITSE}   & \metu & \metv   & \metw &
\cellcolor[HTML]{DAE8FC}\metx \\ \hline
\end{tabular}
}
\caption{\new{Simulation validation of  
        performance metrics of NN controller  
 trained with policy using digital twin. 
Metrics are reported for each individual axis, along with the average.
Lower values are better.}}
\label{tab:twin:sim_valid_ppo}
\end{table}

\def\meta{4\xspace}
\def\metb{4\xspace}
\def\metc{3\xspace}
\def\metd{4\xspace}
\def\mete{414\xspace}
\def\metf{492\xspace}
\def\metg{199\xspace}
\def\meth{368\xspace}
\def\meti{4,773\xspace}
\def\metj{4,941\xspace}
\def\metk{3,829\xspace}
\def\metl{4,514\xspace}
\def\metm{414,216\xspace}
\def\metn{493,033\xspace}
\def\meto{199,662\xspace}
\def\metp{368,970\xspace}
\def\metq{7,680\xspace}
\def\metr{7,937\xspace}
\def\mets{6,574\xspace}
\def\mett{7,397\xspace}
\def\metu{608,092\xspace}
\def\metv{712,863\xspace}
\def\metw{300,222\xspace}
\def\metx{540,392\xspace}

\begin{table}[]
    \centering
{\setlength{\tabcolsep}{0.2em}
\def\arraystretch{1.15}%
\begin{tabular}{l|ccc|c|}
\cline{2-5}
                             & \multicolumn{4}{c|}{PID Controller}                                                \\ \hline
\multicolumn{1}{|l|}{Metric} & Roll ($\phi$) & Pitch($\theta$) & Yaw ($\psi$)  & \cellcolor[HTML]{DAE8FC}Average       \\ \hline
\multicolumn{1}{|l|}{MAE}    & \meta         & \metb           & \metc         & \cellcolor[HTML]{DAE8FC}\metd         \\
\multicolumn{1}{|l|}{MSE}    & \mete      & \metf          & \metg        &
\cellcolor[HTML]{DAE8FC}\meth        \\
\multicolumn{1}{|l|}{IAE}    & \meti     & \metj       & \metk     &
\cellcolor[HTML]{DAE8FC}\metl     \\
\multicolumn{1}{|l|}{ISE}    & \metm  & \metn      & \meto    &
\cellcolor[HTML]{DAE8FC}\metp    \\
\multicolumn{1}{|l|}{ITAE}   
& \metq    & \metr    & \mets    & 
\cellcolor[HTML]{DAE8FC}\mett    \\
\multicolumn{1}{|l|}{ITSE}   & \metu & \metv   & \metw &
\cellcolor[HTML]{DAE8FC}\metx \\ \hline
\end{tabular}
}
\caption{\new{Simulation validation of  
        performance metrics of PID controller  
 trained with policy using digital twin. 
Metrics are reported for each individual axis, along with the average.
Lower values are better.}}
\label{tab:twin:sim_valid_pid}
\end{table}

\begin{figure}
\centering
{\includegraphics[trim=0 0 0
0,clip,width=\textwidth]{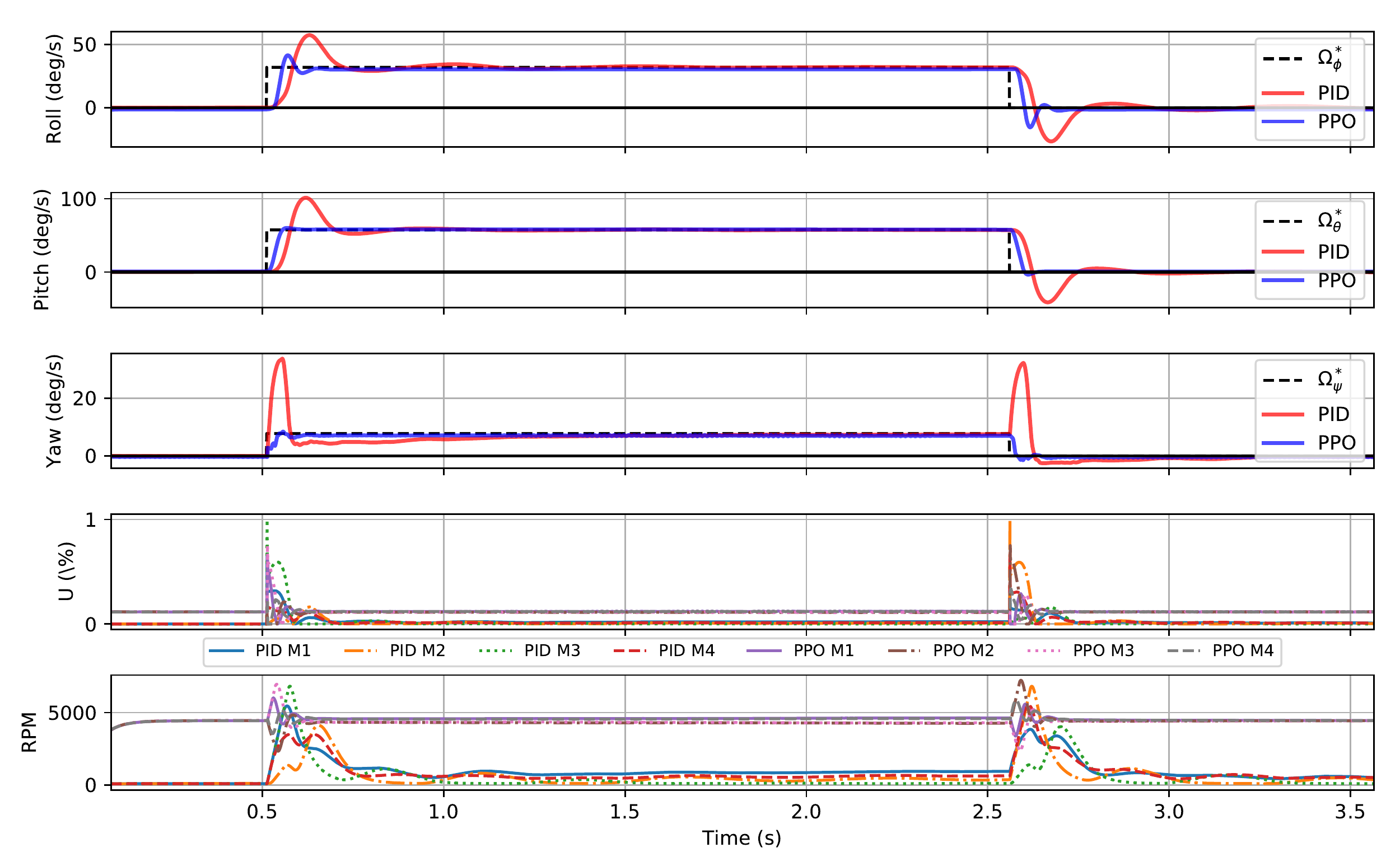}}
\caption{Step response comparison between PPO-based flight controller, and PID
flight controller.}
\label{fig:twin:pidvsppo}
\end{figure}

\begin{figure}
\centering
{\includegraphics[trim=0 0 0
0,clip,width=\textwidth]{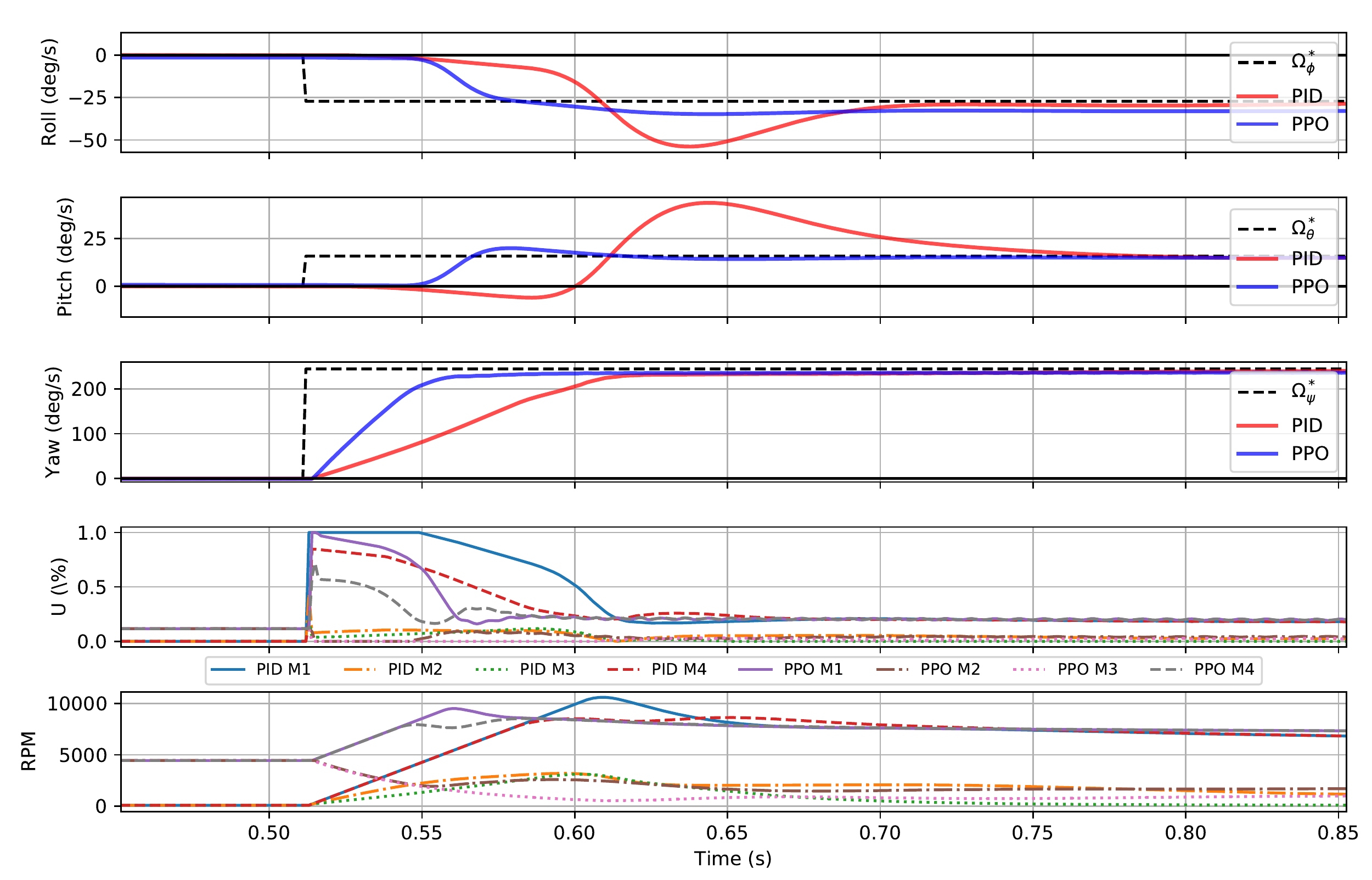}}
\caption{Zoomed in comparison between PPO-based flight controller, and PID
flight controller.}
\label{fig:twin:pidvsppoaccel}
\end{figure}

We visually  compare the performance between the PPO controller and the PID
controller in Fig.~\ref{fig:twin:pidvsppo}. In this example, the PID controller
suffers significant overshoot in the yaw axis. With the exception of minor
overshoot on the roll axis, the PPO controller tracks the setpoint quite well.
We sample another episode and zoom in on the step response to the command in
Fig.~\ref{fig:twin:pidvsppoaccel}. Here we can more clearly compare the control
signals between the two controllers. In this figure, the legend is shared
between the last two subplots which correspond to the control signal and motor
RPM respectively. The control signals generated by the two controllers are very
similar and follow similar responses. In the RPM plot, we can see the affect
each control signal has on each motor velocity.


\textbf{Flight envelope.} In the following experiment, we wish to characterize the flight envelope 
of the two controllers.
More specifically we would like to compare the operating regions of each
controller, in regards to how well the controller can maintain a desired
angular velocity. %

\begin{figure}
\centering
{\includegraphics[trim=35 0 0
0,clip,width=0.8\textwidth]{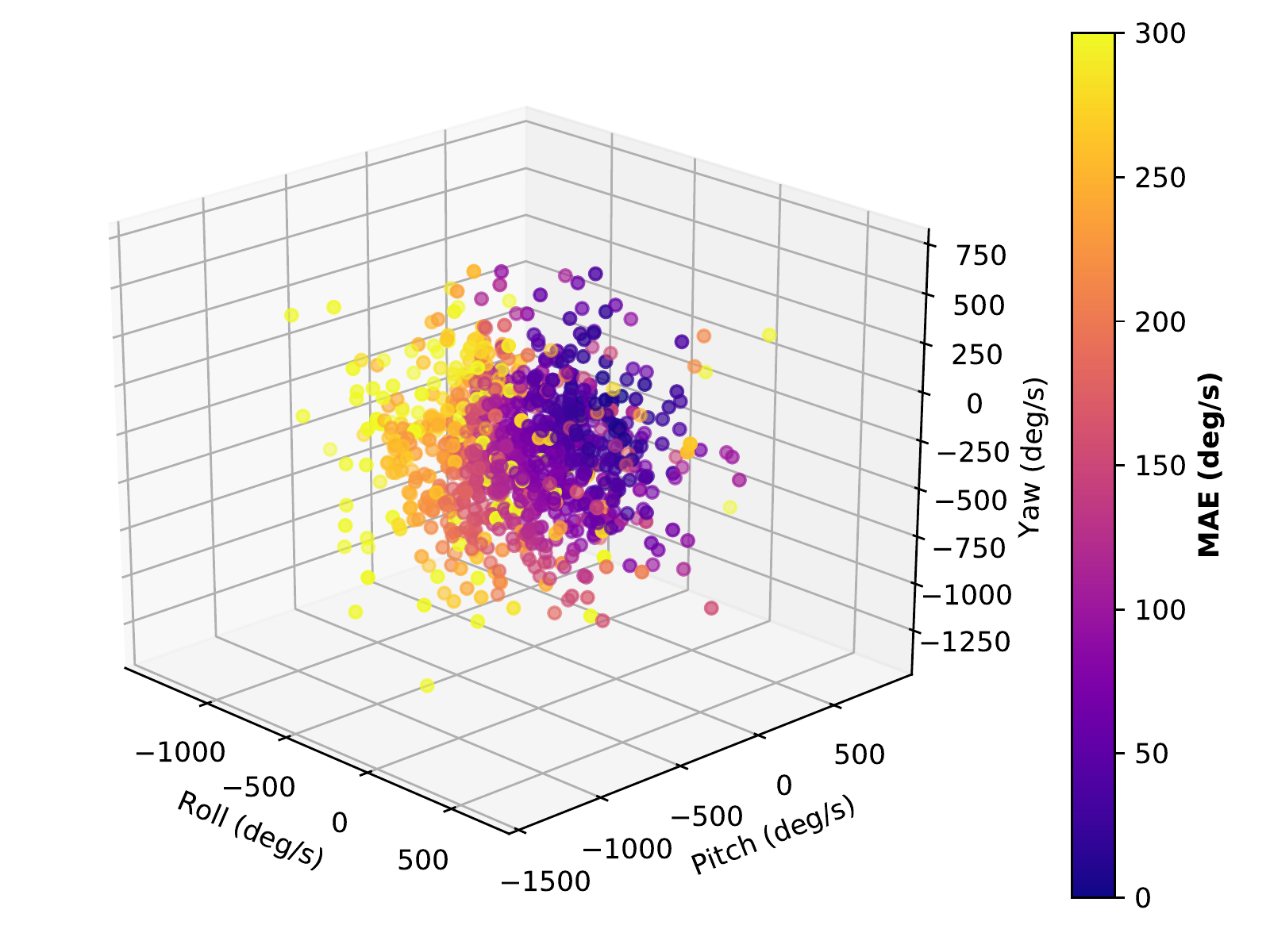}}
\caption{Flight envelope of PID flight controller.}
\label{fig:twin:pidenvelope}
\end{figure}

\begin{figure}
\centering
{\includegraphics[trim=35 0 0
0,clip,width=0.8\textwidth]{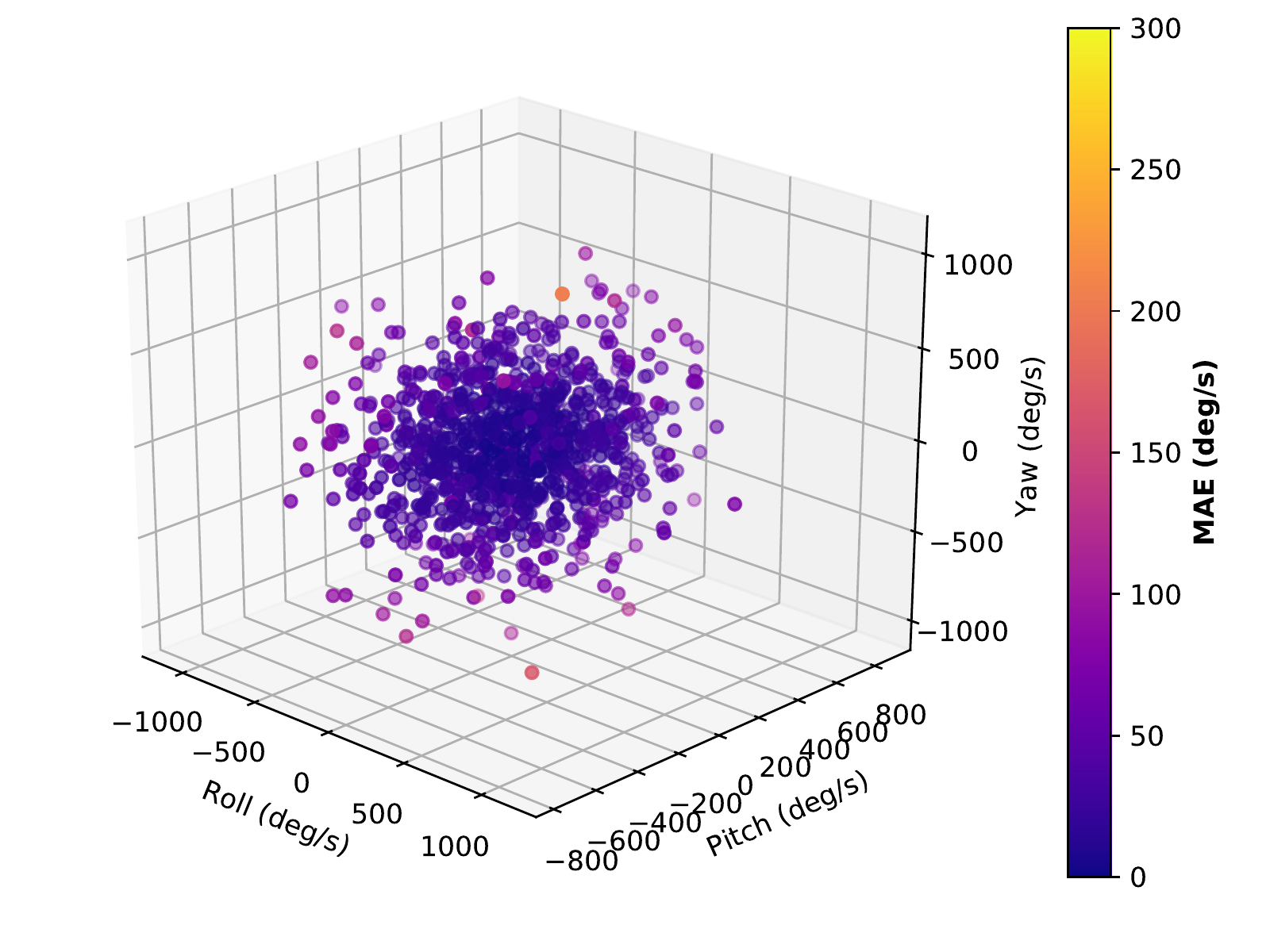}}
\caption{Flight envelope of neuro-flight controller.}
\label{fig:twin:ppoenvelope}
\end{figure}

To perform this measurement, we used the step input environment created for tuning the PID controller to
randomly sample an angular velocity from a Gaussian distributions with $\mu=0$
and $\sigma=300$. For each controller we evaluate 1,000 different setpoints.  For each
trial, the mean absolute error~(MAE) is calculated. We then created a 3D
scatter plot, where each point is a setpoint, and its color corresponds to the
MAE. Results for the PID controller and \nn controller are displayed in 
Fig.~\ref{fig:twin:pidenvelope} and 
Fig.~\ref{fig:twin:ppoenvelope} respectively. Looking closely at the scale of
the color bar, we can see that the \nn controller experiences almost three
times less error in the evaluation region. 
To measure stability, we counted the number of times each controller was able
to remain in a 10\% error band, in relation to the setpoint, after 500ms have lapsed. 
The \nn controller was able to stay within  the error band, 72\% of the time,
compared to PID controller only doing so 16\% of the time. We speculated the poor performance of the PID controller
could be due to the slower rise time, or overshoot. We increased the time
before we started measuring the error band
 till after 750ms  which only increased the PID controller to 29\%, however this also
 increased the \nn to 76\%.
 Manually inspecting the step response  it became clear that once the set
 points diverged greatly from its tuning region, its became very unstable with
 significant oscillations. On the other hand, the \nn controller was able to
 maintain stability upwards to angular velocities exceeding 1,000 degrees per
 second. These results showcase the robustness of the \nn controller, and the
 expanded flight envelope  in comparison to PID control.

\subsection{Neuroflight Flight Evaluations}
\label{sec:twin:flighteval}
In this section we perform real flight evaluation of the \nn policy.
Before conducting these test flights, we compile the policy into the
Neuroflight firmware and flash our flight
controller using  the Neuroflight toolchain described in
Section~\ref{sec:nf:sys}.

\begin{figure*}
\centering
{\includegraphics[trim=0 0 0
    0,clip,width=\textwidth]{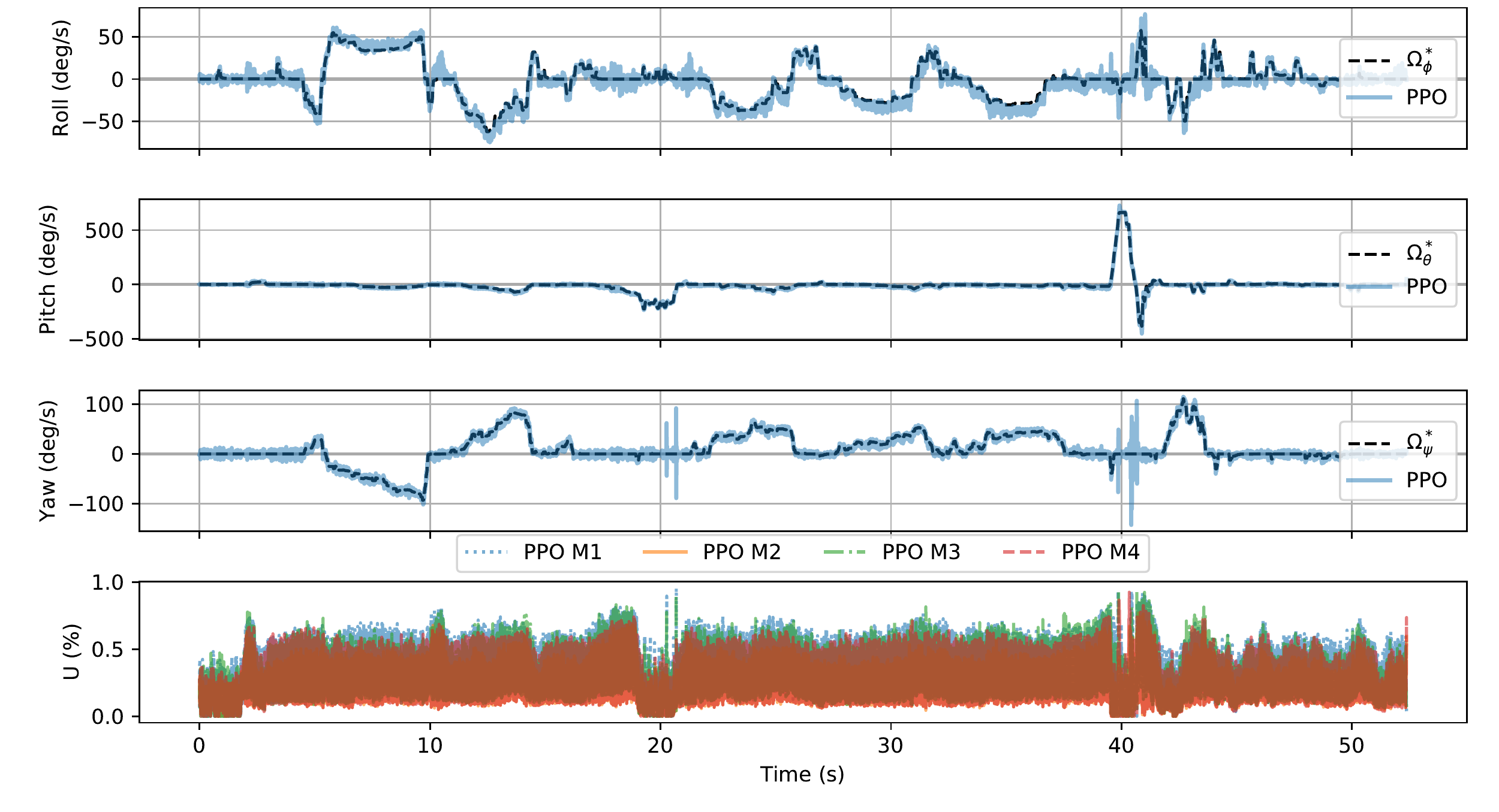}}
\caption{Flight test for neuro-flight controller synthesized with digital twin.}
\label{fig:twin:flight}
\end{figure*}

\begin{figure*}
\centering
{\includegraphics[trim=0 0 0
    0,clip,width=\textwidth]{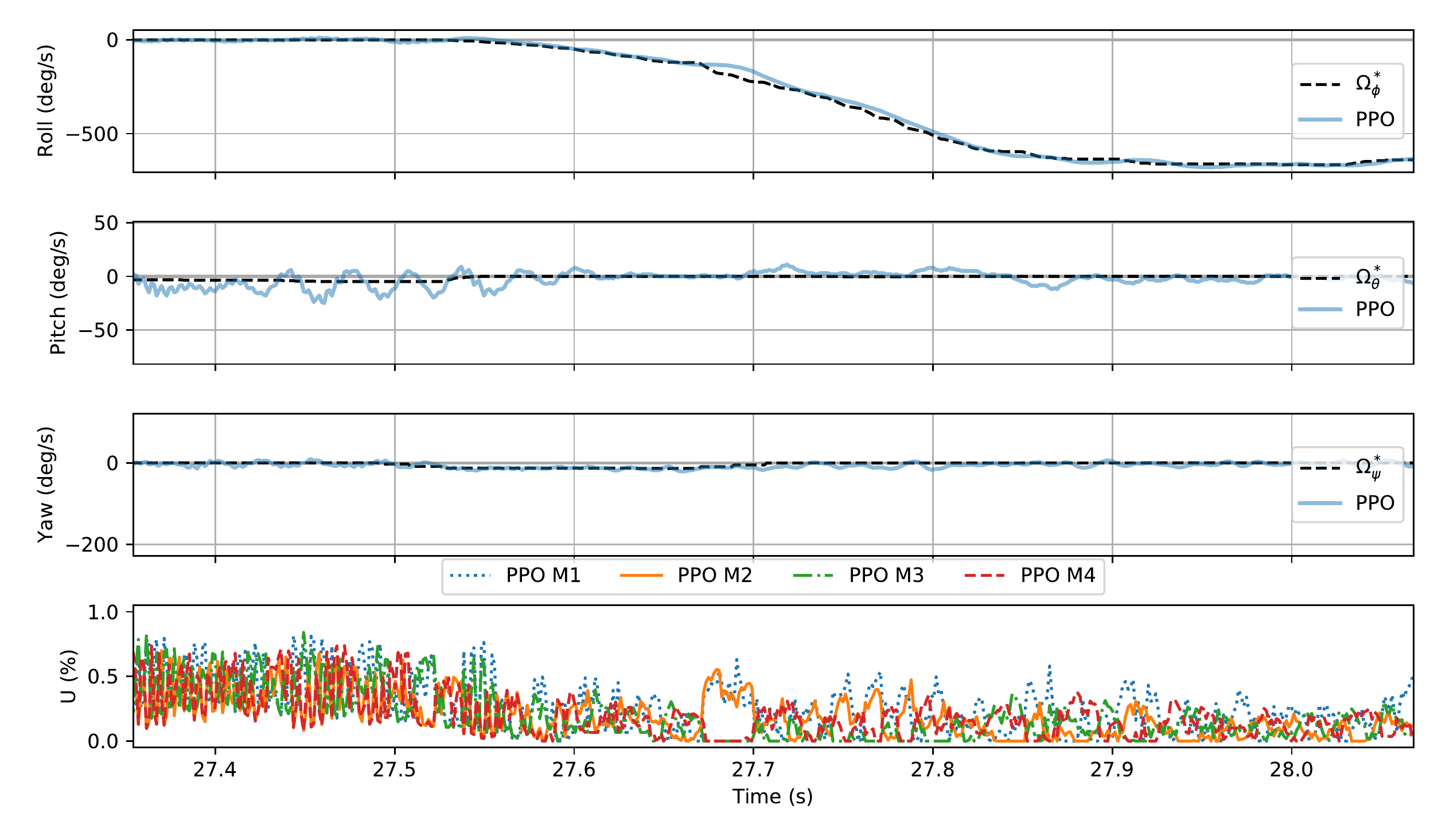}}
\caption{Zoomed in portion of a roll being executed.}
\label{fig:twin:flightzoom}
\end{figure*}

We conducted a total of \realFlightCount test flights executing a variety of
basic and advanced flight maneuvers while logging the angular velocity
reported by the gyro, the desired setpoint, and the motor control signals.
All FPV videos of the test flights can be viewed
at \videourl. The pilot reported precise and smooth handling. The FPV videos
do not show any signs of oscillations of other issues. Furthermore the pilot did not
experience  the drifting issues that were  reported in the policy trained with \newgym. 
Fig.~\ref{fig:twin:flight} shows the performance of the \nn controller tracking the
desired setpoint during one of the real flights. The controller is able to track the
pitch axis remarkably well. To inspect the tracking more closely, we zoomed in
on a roll being performed in Fig.~\ref{fig:twin:flightzoom}. With the exception
of some minor oscillations in the pitch axis, the tracking of the setpoints is
observed to be 
quite smooth.

Afterward the test flights were conducted, we analyzed the flight
logs and generated the performance metrics  reported in
Table~\ref{tab:twin:real_flight}. These error metrics are an average across all
test flights. 
Comparing these average errors to those from the controller trained with \newgym in Table~\ref{tab:real_flight}, we
can see the drastic reduction in error through the use of training the policy
using the digital twin. There is an 11 degrees per second decrease in the
MAE  as well as a significant drop is MSE indicating a decrease in large
fluctuations in the error. 

To measure the performance gap between the real world and simulation world, we
took the desired angular velocities recorded during the test flights and played
them back to the \nn controller in the \gymfcTwo simulation environment
controlling 
the digital twin. The same error metrics were generated and are reported in
Table~\ref{tab:twin:sim_valid_ppo_playback}.
From this comparison we can see the reality gap has been greatly reduced. The
average MAE for the simulation playback is  3 degrees per second which was also the same
measured in the
\newgym environment.
However we now only have a 2 degrees per second increase in
MAE in the real world compared to 13 degrees per second previously measured
when not using the digital twin (Table~\ref{tab:real_flight}).

One important observation we made during the test flights was the immense heat
being generated by the motors. 
This is usually a sign of rapid switching of the ESC. To prevent motor damage, 
we would 
allow the motors to cool between test flights. 
To quantify the switching in the control signals, 
we calculated %the average control output to be $\overline{u}=$\realUMu and 
the average
absolute change in the control output ($\overline{|\Delta u|}$) to be
\realUDiffMu in the real world, and 
\playbackTwoUDiffMu in the simulation world. 

The increase in the control signal output is problematic and
confirmed our suspicions while in the field conducting test flights. 
Further experimental tests need to be conducted to validate whether the heat
generated by the control signal oscillations are significant enough to cause
damage to the motor wires and permanent magnets. 
Visually we can confirm the aggressive oscillations in
Fig.~\ref{fig:twin:flightzoom} of one of the test flights.
What is most surprising is the significant gap in the performance between the
oscillations in simulation verse the real world. Although we found  
 the reward and environment described in Section~\ref{sec:twin:impl} to train 
 policies to transfer well to the real world, we
experimented with dozens of other policies, each of which contained such severe
visual oscillations, the test flights had to be abandoned. 
Through our experience, 
minimizing the control signal oscillations has
been the greatest challenge. 

Nevertheless, the accuracy of the \nn controller in the real world, when
trained using the digital twin, demonstrates remarkable tracking performance.
We have established a solid foundation for synthesizing accurate controllers
which can now be used to develop controllers with advanced control goals.

\def\meta{6\xspace}
\def\metb{5\xspace}
\def\metc{3\xspace}
\def\metd{5\xspace}
\def\mete{136\xspace}
\def\metf{64\xspace}
\def\metg{53\xspace}
\def\meth{84\xspace}
\def\meti{4,438\xspace}
\def\metj{3,846\xspace}
\def\metk{2,748\xspace}
\def\metl{3,677\xspace}
\def\metm{96,779\xspace}
\def\metn{46,009\xspace}
\def\meto{37,865\xspace}
\def\metp{60,218\xspace}
\def\metq{171,530\xspace}
\def\metr{145,893\xspace}
\def\mets{103,179\xspace}
\def\mett{140,201\xspace}
\def\metu{3,952,545\xspace}
\def\metv{1,847,723\xspace}
\def\metw{1,866,962\xspace}
\def\metx{2,555,743\xspace}

\begin{table}[]
    \centering
{\setlength{\tabcolsep}{0.2em}
\def\arraystretch{1.15}%
\begin{tabular}{l|ccc|c|}
\cline{2-5}
                             & \multicolumn{4}{c|}{NN Controller (PPO)}                                                \\ \hline
\multicolumn{1}{|l|}{Metric} & Roll ($\phi$) & Pitch($\theta$) & Yaw ($\psi$)  & \cellcolor[HTML]{DAE8FC}Average       \\ \hline
\multicolumn{1}{|l|}{MAE}    & \meta         & \metb           & \metc         & \cellcolor[HTML]{DAE8FC}\metd         \\
\multicolumn{1}{|l|}{MSE}    & \mete      & \metf          & \metg        &
\cellcolor[HTML]{DAE8FC}\meth        \\
\multicolumn{1}{|l|}{IAE}    & \meti     & \metj       & \metk     &
\cellcolor[HTML]{DAE8FC}\metl     \\
\multicolumn{1}{|l|}{ISE}    & \metm  & \metn      & \meto    &
\cellcolor[HTML]{DAE8FC}\metp    \\
\multicolumn{1}{|l|}{ITAE}   
& \metq    & \metr    & \mets    & 
\cellcolor[HTML]{DAE8FC}\mett    \\
\multicolumn{1}{|l|}{ITSE}   & \metu & \metv   & \metw &
\cellcolor[HTML]{DAE8FC}\metx \\ \hline
\end{tabular}
}
\caption{\new{Average error metrics of the NN controller from flights in the real
world trained with the digital twin. Metrics are reported for each individual axis, along with the average.
Lower values are better.}}
\label{tab:twin:real_flight}
\end{table}

\def\meta{3\xspace}
\def\metb{3\xspace}
\def\metc{4\xspace}
\def\metd{3\xspace}
\def\mete{35\xspace}
\def\metf{20\xspace}
\def\metg{26\xspace}
\def\meth{27\xspace}
\def\meti{3,879\xspace}
\def\metj{3,337\xspace}
\def\metk{4,091\xspace}
\def\metl{3,769\xspace}
\def\metm{35,144\xspace}
\def\metn{20,928\xspace}
\def\meto{26,893\xspace}
\def\metp{27,655\xspace}
\def\metq{101,586\xspace}
\def\metr{86,477\xspace}
\def\mets{106,814\xspace}
\def\mett{98,293\xspace}
\def\metu{955,123\xspace}
\def\metv{554,219\xspace}
\def\metw{708,319\xspace}
\def\metx{739,220\xspace}

\begin{table}[]
    \centering
{\setlength{\tabcolsep}{0.2em}
\def\arraystretch{1.15}%
\begin{tabular}{l|ccc|c|}
\cline{2-5}
                             & \multicolumn{4}{c|}{NN Controller (PPO)}                                                \\ \hline
\multicolumn{1}{|l|}{Metric} & Roll ($\phi$) & Pitch($\theta$) & Yaw ($\psi$)  & \cellcolor[HTML]{DAE8FC}Average       \\ \hline
\multicolumn{1}{|l|}{MAE}    & \meta         & \metb           & \metc         & \cellcolor[HTML]{DAE8FC}\metd         \\
\multicolumn{1}{|l|}{MSE}    & \mete      & \metf          & \metg        &
\cellcolor[HTML]{DAE8FC}\meth        \\
\multicolumn{1}{|l|}{IAE}    & \meti     & \metj       & \metk     &
\cellcolor[HTML]{DAE8FC}\metl     \\
\multicolumn{1}{|l|}{ISE}    & \metm  & \metn      & \meto    &
\cellcolor[HTML]{DAE8FC}\metp    \\
\multicolumn{1}{|l|}{ITAE}   
& \metq    & \metr    & \mets    & 
\cellcolor[HTML]{DAE8FC}\mett    \\
\multicolumn{1}{|l|}{ITSE}   & \metu & \metv   & \metw &
\cellcolor[HTML]{DAE8FC}\metx \\ \hline
\end{tabular}
}
\caption{\new{Error metrics of simulation playback NN controller  
 trained with policy using digital twin. 
Metrics are reported for each individual axis, along with the average.
Lower values are better.}}
\label{tab:twin:sim_valid_ppo_playback}
\end{table}

\subsection{Discussion}
Throughout this research, one of the most difficult challenges has been
managing the \nn control signal oscillations.  Through discussions with other
researchers, this appears to be a challenge, not only for flight control, but
neuro-control in general. 
Reducing oscillations has been discussed briefly in some of the literature. 
For  helicopter RL-based navigation controllers, \cite{bagnell2001autonomous} added a low pass digital filter  to the
control signal outputs, while in
\cite{kim2004autonomous}, a penalty 
based on the quadratic sum of actions is used to promote smooth and small changes
to the output. 

It is perplexing
that the only other work that discusses concerns with output oscillations for
quadcopter control is in
\cite{molchanov2019sim}. In this work they found removing the gyro low pass
filter in the CrazyFlie firmware, decrease delay, while also decreasing
physical oscillations. 
However we did not find this to help. This is most
likely due to us sampling the gryo at a considerably faster speed.
In this work we sample the gyro
at 4kHz and execute our control loop at 1kHz, while work in
\cite{molchanov2019sim} executes their control loop at 500Hz.
This work also reports the highest frequency found in the control signal output,
however without any relationship to performance (\ie causes visual oscillations,
increases motor temperature, etc.) these metrics are not  
meaningful as different propeller propulsion systems will be affected
differently by  the control signal.  

Other work related to \nn-based flight control~\cite{hwangbo2017control,palossi201964mw}
have not reported any details relating to the control signals generated by
their neuro-controller. Thus the questions arise, do the control signals
oscillate more than traditional control methods, such as PID? If not, what is
different about their approach such that this is not a concern? 
In \cite{hwangbo2017control} the authors combine the output of a PD controller
with the \nn, for attitude control, during training in order to stabilize the learning process. The authors mention it does not aid 
the controller after the training process however it is unclear if it is removed
from the controller when transferred to hardware. If it is removed, this work
does not discuss how the controller compensates for the absence of the PD
controller output.  
One possible reason high oscillations are not as prominent could be due to the differences in control goals. Our work
is concerned with low level control while the previously mentioned work 
is related to guidance and navigation tasks. Perhaps 
position estimation provided by motion capture systems and video results in  
decreased control signal oscillations. If this can be verified it would be
interesting future work to explore.

\section{Related Work}
\label{sec:twin:related}
In this section we will review simulators used for flight control
testing, the aircraft models they provide and motor models they use. 
Additionally we will review work related to UAV propulsion system modelling.

\subsection{Flight Simulators and Aircraft Models}
The Gazebo simulator provides  an Iris quadcopter, and
Zephr fixed wing UAV aircraft model. To achieve flight, an  aerodynamic plugin
is provided. For a multicopter, the aerodynamic plugin calculates the lift for
each blade and motor response from a PID controller.

RotorS~\cite{Furrer2016} is a micro air vehicle~(MAV) Gazebo simulator framework for software in the loop testing of flight control systems with a focus in navigation and guidance. The framework is tightly integrated with the Robotics Operating System~(ROS) and includes a number of multirotor models such as the AscTec Hummingbird, the AscTec Pelican, or the AscTec Firefly. Their documentation briefly describes how to assemble  
 your own MAV into their simulator however this does not describe methods for motor modelling. 
Additionally they acknowledge challenges transferring to a real MAV and share the same goals as this work to reduce effort transferring to real hardware however this also requires the target real aircraft to support ROS as well.

The PX4 project has extracted the motor models and dynamics from RotorS to
create a new project that is independent  of ROS for SITL and HITL testing~\cite{px4sitl}. Unfortunately there does not exist any
 documentation   
for deriving your own motor models.

AirSim~\cite{shah2018airsim} has similar goals to RotorS with a focus in
computer vision. To create realistic environments, this simulator uses the
Unreal engine which is difficult to achieve in Gazebo.
The Unreal engine uses PhysX~\cite{physx} as the physics engine backend which supports both
generalized and maximal coordinate solvers. 

Both RotoS and AirSim derive the motor forces and torques in a similar fashion
using element blade
element theory~\cite{mccormick1995aerodynamics} and model motor response using
first order filters. This method derives the force and torques from the entire
motor and propeller pair rather than Gazebo's aerodynamic which calculates
these forces from the individual blades.

Our work shares many of the same ambitions as the previous work primarily in
regards  of providing seamless transfer to hardware. 
However previous work is primarily focused in higher level tasks
while \gymfcGeneric's primary goal is to provide a tuning framework with a focus in low level attitude control. 

\subsection{Propeller Propulsion System Data}
A propeller database published by University of Illinois Urbana-Champaign
contains wind tunnel measurements for over 200 small-scale propellers~\cite{uiuc}.  
The database contains the advance ratios, thrust and torque coefficients.
Details of the experimental methodology and the test stand are presented
in~\cite{deters2014reynolds}. Follow up work \cite{deters2017static} performed 
static propeller testing for four popular quadcopters including the 3D Robotics
Solo and DJI Phantom 3. Thrust and power coefficients are also reported. 

In \cite{gong2018performance} a study of propeller propulsion systems, including
the ESC was conducted. Models were also developed for the ESC. The model was derived by fitting the efficiency data to a bi-linear equation as a function of the throttle and current.

Unfortunately of the previously discussed work, time behavior of
the propulsion system is not reported thus a motor response model can be not obtained.  

A large database of static propeller propulsion system performance data, commonly found on
multicopters has been published by
MiniquadTestBench~\cite{miniquad}. Thrust, torque, power and motor velocity
have been recorded for a number of different control inputs. Although thrust
and power coefficients are not provided, one could derive these values from the
raw data. Given the raw data one is also able to measure the motor response. %However accuracy may vary if a different ESC is used. 

\section{Conclusion and Future Work}
\label{sec:twin:con}
In this work we present a universal tuning framework, \gymfcTwo, as a means to 
synthesizing neuro-flight controllers unique to their digital twin aircraft. 
We introduce our methodology for creating a 
digital twin and demonstrate the approach producing a digital twin for the
\aircraft. Using our digital twin we analyze its stability 
in the Gazebo simulator using the default physics engine ODE and compare these results to DART.

We further showcase the flexibility of the \gymfcTwo framework through the
implementation 
of a 
 dynamometer for validating motor model thrust, torque and velocity
 performance, as well as a  platform for PID tuning.
Evaluating our synthesized  neuro-flight controller in simulation, we find
 this
 class of controllers has a larger flight envelope
 than a classical PID controller. Our real world flight evaluations provide
convincing evidence training using the digital twin  reduced the
reality gap. Nonetheless, the controller experienced high frequency motor
output oscillations that must be addressed in future work.  
In summary, our future work consists of making improvements to the
digital twin, and addressing control signal oscillations. 

We have identified three potential sources of error in the aircraft model that
need further attention in future work. 
\begin{enumerate}
\item \textbf{Moments of inertia.} In this work we compute the moments of
inertia using the measured mass of each individual aircraft part and the moments
of inertia matrix measured from each mesh model. However as we previously
discussed in Section~\ref{sec:twin:rigid} these methods assume a uniform mass
distribution of the object. In future work we will investigate methods for validating the accuracy of our
approach through experimental real world measurements. Possible approaches may
consist of building a torsional pendulum and an apparatus for the aircraft body to derive the inertia
measurements. 
 
\item \textbf{Motor model.} To model motor dynamics we have used 
the PX4 motor models as a foundation which embodies established models from element blade theory. However
as we have seen in our experimental measurements, the torque and thrust
coefficients greatly vary in relation to the motor RPM. Thus using static thrust and
torque coefficient will introduce errors. Based on these observations in the future   
we plan to develop more accurate models and investigate using \nn
to train an inverse plant model of the motor model.

\item \textbf{Aircraft attitude.} The challenge of developing the
digital twin is being able to model the individual components and then compose them
such that the resulting model is accurate. In regards to the motor performance,
we were able to validate  the thrust, torque and rotor velocity models in
simulation. In the future we plan to validate the angular velocity of the
aircraft in the real world. This will require the developed of an apparatus to fix the aircraft along each
axis with sensors to measure the angular velocity of the aircraft body. From
this data we can calculate other forces acting upon the aircraft such as drag
to further improve the accuracy of the simulation. 
\end{enumerate}

In regards to addressing oscillations in the control signal outputs experienced
in the real world, we plan to take the following approaches.
\begin{enumerate}
    \item \textbf{\nn state and architectures}  The ultimate goal is to  
        develop a neuro-controller which can make its decisions based on 
        the complete internal state of the aircraft.
        Thus we must work towards integrating addition sensors as input, while
        still maintaining a high level of performance. For one, we would
        like to perform experiments to identify if any correlation exists between the
        motor temperate and the ESC temperature. If so, we are able to access
        the ESC temperate through ESC telemetry  which can aid in building
        policies to prevent the aircraft getting into a state that could cause
        damage, for example shorting the motor wires.
    \item \textbf{Domain randomization} In the future we plan to integrating
        additional dynamics and forces, such wind, gravity, and other generic 
        force acting upon the aircraft body.  
        In this work we emulated gyro noise that was
        modelled from empirical data. However in future work we plan to investigate other domain randomization techniques such
action delays and noise to the set point.
\end{enumerate}

Work and results outlined in this chapter have helped progress the
state-in-the-art in intelligent flight control, bringing us one step closer to these
controllers being practical to be adopted in the real world.

\cleardoublepage

\chapter{Conclusions}
\label{chapter:conclusion}
\thispagestyle{myheadings}
The rapid advances in machine learning, big data, material sciences and manufacturing will
transform the aviation industry as we know it.  
The aircraft of the future will be self-aware providing remarkable levels of
performance, safety and reliability. This will be in part due to advanced
flight control systems, providing the abilities to learn, plan and adapt. For
example, the aircraft will be able to learn its current flight envelope to
determine what its current capabilities are. Furthermore, the aircraft will be able to
plan,
in real-time, for potential future system failures and mitigate them from
occurring before they happen.  
Lastly, the aircraft will be able to adapt to changes, such as shifts in
payload. 
To support these advanced control goals we require a new generation of 
intelligent control systems that will be capable of providing  high order
executive functions.  
To that end, this dissertation makes the following contributions.

\section{Summary of Contributions}
This dissertation investigates using the digital twinning paradigm for
synthesizing 
\nn based flight control systems. The resulting flight controller is
unique to the digital twin, providing optimal control for the specific
aircraft digital twin. Using \nns,  these controllers have the fundamental
 building blocks to support our future advanced control goals that are out of reach of traditional control methods. 
 This work has established a foundation
 for these next generation flight control systems, by developing software to
 synthesize 
 stable, precise and accurate \nn-based attitude controllers. 
Specifically, we developed a full solution stack for synthesizing  
\nn-based flight controllers via RL. 
This solution stack consists of a universal
tuning framework called \gymfcGeneric, a digital twin development methodology, and 
a \nn supported flight control firmware named Neuroflight.  
In summary this dissertation makes the following contributions in the study of
intelligent flight control systems.

\textbf{Tuning framework and training environment.} In this work we introduce
\gymfcGeneric, an open source universal flight control tuning framework.
We implement an RL training environment using \gymfcGeneric and benchmark  
a number of state-of-the-art RL algorithms, in simulation for quadcopter
attitude control, including DDPG, TRPO and PPO. We find PPO to out perform all
other RL algorithms, as well as traditional PID control. 

We introduce the reward function used to synthesize
attitude flight controllers via RL which achieves remarkable accuracy in the real world. 
We  further showcase the  modular design of the \gymfcGeneric
framework implementing a virtual dynamometer for motor modelling and an
environment for 
PID control tuning. The \gymfcGeneric architecture provides a platform for
researchers to develop tools and environments to aid in developing next generation flight
control.
Furthermore, \gymfcGeneric  opens up new possibilities for  performing SITL and HITL
sensitivity analysis of various environment, controller and aircraft
parameters to aid in controller and aircraft design.

\textbf{Digital twin development.} To reduce the reality gap we
have proposed our methodology for creating a digital twin of a multicopter.
This included the creation of the aircraft rigid bodies, and the construction of a
dynamometer to obtain measurements for deriving motor models. 
We have  developed  motor response  models  %and  integrated into existing motor models
to increase realism of the motor dynamics. 
Additionally, we have published software to perform a stability analysis of the
digital twin in simulation. 
Our evaluations show the digital twin has almost completely eliminated the reality
gap in terms of angular velocity error.

\textbf{Flight control firmware.} This dissertation introduced Neuroflight, the
world's first open source \nn supported flight control firmware. We have proposed  our
 toolchain for deploying a trained \nn policy to highly resource constrained
 off-the-shelf microcontrollers.
 %consisting of a toolchain to automatically compile Tensorflow models to an
 %object file targeting hardware with hard floating point arithmetic. 
%
%
Our timing analysis shows the \nn controller can execute faster than
2kHz allowing for faster digital ESC protocols to be utilized to support high precision flight.
Our real world flight evaluations demonstrate the \nn policies provide stable,
accurate flight and
are capable of performing aerobatic maneuvers.

\section{Open Challenges and Future Work}
The work proposed in this dissertation establishes a foundation for next
generation flight control systems however this is just the first stepping stone
and a number of opportunities lie ahead for future work.  

\begin{enumerate}

\item \textbf{Simulation  improvements.}
\gymfcGeneric is able to train attitude controllers independent of navigation tasks
through our approach of fixing the aircraft about its center of thrust to the
simulation world. Although autonomous flight control is currently more prominent in
literature than low level attitude control, manual override and be necessary
for these control systems to be adopted in the real world. Unfortunately the
majority of work related to autonomous flight do not address these issues.  
Our training strategy allows for the agent to learn the mapping of
the desired angular velocity setpoint to the  corresponding motor control
signal however there are side affects that we have previously discussed in
Section~\ref{sec:reward}
such as the agent using more power than needed. We are able to compensated for
this undesired behavior through
the reward functions however this is not ideal as %for transferability however this is not ideal as
increasing reward complexity can affect tracking accuracy.

The quadcopters we have trained in this work are agile racing drones, however one must be careful with
command generation  if the aircraft is not balanced. Multicopters
where the center of mass does not equal the center of thrust, for example
because of a gimbal, may not have the capability to perform full rotations in
the simulation environment. For
these type of aircraft one must make sure they will stay within their flight
envelope, which will result in additional logic for command generation during
training. 

To further increase realism of the environment additional environment dynamics
need to be modelled such as gravity, wind, aerodynamic affects of the aircraft and
other external disturbances acting upon the body during flight.

\item \textbf{Digital twin development.}
The modular design of the \gymfcGeneric framework opens up a number of possibilities for
increasing the fidelity of the digital twin. % and adding additional dynamics.
This could include the integration of power models to simulate discharging of a
battery, and  modelling material stress-strain analysis.   
Furthermore, in Chapter~\ref{sec:twin} we identified and discussed a number of
errors in the motor model, such as using a static torque and thrust
coefficient which does not accurately model the nonlinear motor dynamics that
are exaggerated for smaller multicopters such as our racing drone.

This work has been scoped to synthesizing flight controllers offline and as a
result we have not investigated methods for synchronizing the digital twin
with the real aircraft after it is deployed in the real world. To achieve the
true potential of the  digital twin, 
future work must develop methods for updating the digital twin with data
obtained from the real aircraft  so the controller, in the virtual environment, can
continually be improved. Essentially we need to create an inverse plant of the
aircraft  however current modelling depends on the thrust and torque of each
motor which can be difficult to obtain during flight.

\item \textbf{Continuous learning.} Our current approach trains \nns exclusively using offline learning. However, in order to 
reduce the performance gap between the simulated and {real world} we expect that a hybrid architecture involving online incremental
learning will be necessary. % to provide continuouslearning.
Online learning will allow the aircraft to adapt, in real-time, and
compensate for any modelling errors that existed  during offline training.
This presents interesting challenges for designing architectures to
hot-swap the \nn weights.  If we recall from Chapter~\ref{chapter:nf}
when the flight
controller is trained offline, the resulting NN graph is ``frozen'' and
AOT-compiled to execute on the quadcopters onboard controller.
The compiled \nn is a mix of arithmetic operations and hard-coded network
weights and is treated just as any other function. For resource constrained environments loading large networks into
memory may not be an option. Thus we will need to develop new software, and
hardware architectures that can support this functionality.  

Online learning will be complimentary to  training on the digital twin.
The digital twin can utilize the power of the cloud to  perform heavier
computation than the aircraft's onboard computer. For example, the digital twin
can be used to run through multiple different scenarios, and forecast 
system failures before they occur.

\item \textbf{\nn architecture.}
Several performance benefits can be realized from an optimal
network architecture for flight control including improved
accuracy and
faster execution. 
An extensive survey needs to be conducted investing the pros and
cons of various architectures. %Given our limited access to computational resources this was

Long short-term memory~(LSTM)
networks may help with time varying dynamics such as the motor response. 
Alternative
distributions such as
the beta function which is naturally
bounded~\cite{chou2017improving} may help with motor oscillation issues.
Furthermore the use of the rectified linear unit~(ReLU) activation functions
may reduce the execution
time of the \nn due to it being more computationally efficient than the hyperbolic tangent 
function.

Moving forward it will be important to develop modular networks. The current
research direction of RL based flight controllers for navigation are
 do not allow
for manual flight~\cite{hwangbo2017control,palossi201964mw}.
For these controllers to be deployed in the real world there must  be a way to
manually pilot the aircraft for maintenance and management purposes.
Using hierarchical network structures could be beneficial in creating modular
neuro-flight controllers.
\end{enumerate}

An exciting future lies ahead for developing next generation aircraft and their
corresponding flight control systems. As  embedded
computing platforms continue to reduce in size, it will allow for revolutionary advancements in flight
control, 
supporting sophisticated control goals such as the ability to learn, plan and
adapt.
The work presented in this thesis has provided a foundation for the community
to build upon, using our solution stack
to explore the full potential of \nn-based flight control systems.

\cleardoublepage

\newpage
\singlespace

\begingroup
\sloppy
\RaggedRight
\bibliographystyle{apalike}

\bibliography{biblio/survey,biblio/gymfc,biblio/nf,biblio/twin,biblio/thesis}

\begin{thebibliography}{}

\bibitem[lea, 2015]{learnrc}
 (2015).
\newblock {Brushless motor constant explained}.
\newblock \url{http://learningrc.com/motor-kv/}.
\newblock Accessed: 2018-12-07.

\bibitem[ard, 2018]{ardupilot}
 (2018).
\newblock {ArduPilot}.
\newblock \url{http://ardupilot.org/}.
\newblock Accessed: 2018-03-13.

\bibitem[baz, 2018]{bazel}
 (2018).
\newblock {Bazel - a fast, scalable, multi-language and extensible build
  system}.
\newblock \url{https://bazel.build/}.
\newblock Accessed: 2018-11-25.

\bibitem[bet, 2018]{betaflight}
 (2018).
\newblock {BetaFlight}.
\newblock \url{https://github.com/betaflight/betaflight}.
\newblock Accessed: 2017-10-18.

\bibitem[cle, 2018]{cleanflight}
 (2018).
\newblock {CleanFlight}.
\newblock Accessed: 2018-11-25.

\bibitem[gzb, 2018]{gzbug}
 (2018).
\newblock {gzserver doesn't close disconnected sockets}.
\newblock
  \url{https://bitbucket.org/osrf/gazebo/issues/2397/gzserver-doesnt-close-disconnected-sockets}.
\newblock Accessed: 2018-03-13.

\bibitem[iri, 2018]{iris}
 (2018).
\newblock {Iris Quadcopter}.
\newblock \url{http://www.arducopter.co.uk/iris-quadcopter-uav.html}.
\newblock Accessed: 2018-03-28.

\bibitem[min, 2018]{miniquad}
 (2018).
\newblock {Motor Data Explorer}.
\newblock \url{https://www.miniquadtestbench.com/motor-explorer.html}.
\newblock Accessed: 2018-11-25.

\bibitem[rot, 2018]{rotorbuild}
 (2018).
\newblock {NF1: Neuroflight Test Aircraft 1}.
\newblock \url{https://rotorbuilds.com/build/15163}.
\newblock Accessed: 2018-11-25.

\bibitem[pro, 2018]{protobuf}
 (2018).
\newblock {Protocol Buffers}.
\newblock \url{https://developers.google.com/protocol-buffers/}.
\newblock Accessed: 2018-03-13.

\bibitem[STM, 2018]{STM32F745VG}
 (2018).
\newblock {STM32F745VG}.
\newblock \url{https://www.st.com/en/microcontrollers/stm32f745vg.html}.
\newblock Accessed: 2018-11-25.

\bibitem[ten, 2018]{tensorflow}
 (2018).
\newblock {Tensorflow:An Open Source Machine Learning Framework for Everyone}.
\newblock \url{https://github.com/tensorflow/tensorflow/}.
\newblock Accessed: 2018-11-25.

\bibitem[erl, 2019]{erle}
 (2019).
\newblock Erle-copter.
\newblock \url{http://docs.erlerobotics.com/erle_robots/erle_copter}.
\newblock Accessed: 2019-07-26.

\bibitem[fre, 2019]{freecad}
 (2019).
\newblock {FreeCAD}.
\newblock \url{https://www.freecadweb.org/}.
\newblock Accessed June 17, 2019.

\bibitem[px4, 2019]{px4sitl}
 (2019).
\newblock {Gazebo Sim Plugin}.
\newblock \url{https://github.com/PX4/sitl_gazebo}.
\newblock Accessed June 17, 2019.

\bibitem[ine, 2019]{inertia}
 (2019).
\newblock {Inertial parameters of triangle meshes}.
\newblock \url{http://gazebosim.org/tutorials?tut=inertia&cat=build_robot}.
\newblock Accessed June 17, 2019.

\bibitem[int, 2019]{intelaero}
 (2019).
\newblock {Intel Aero}.
\newblock
  \url{https://click.intel.com/intel-aero-ready-to-fly-drone-2679.html}.
\newblock Accessed: 2019-07-26.

\bibitem[phy, 2019]{physx}
 (2019).
\newblock {PhysX}.
\newblock
  \url{https://gameworksdocs.nvidia.com/PhysX/4.0/documentation/PhysXGuide/Manual/Introduction.html}.
\newblock Accessed: 2019-07-20.

\bibitem[pid, 2019]{pidimpl}
 (2019).
\newblock {Python PID Controller}.
\newblock \url{https://github.com/ivmech/ivPID}.
\newblock Accessed: 2019-07-26.

\bibitem[rcb, 2019]{rcbenchmark}
 (2019).
\newblock {RCbenchmark}.
\newblock \url{https://www.rcbenchmark.com/}.
\newblock Accessed June 17, 2019.

\bibitem[sdf, 2019]{sdf}
 (2019).
\newblock {SDFormat}.
\newblock \url{http://sdformat.org/}.
\newblock Accessed June 17, 2019.

\bibitem[Abbeel et~al., 2007]{abbeel2007application}
Abbeel, P., Coates, A., Quigley, M., and Ng, A.~Y. (2007).
\newblock An application of reinforcement learning to aerobatic helicopter
  flight.
\newblock In {\em Advances in neural information processing systems}, pages
  1--8.

\bibitem[Abdulrahim et~al., 2019]{abdulrahim2019defining}
Abdulrahim, M., Bates, T., Nilson, T., Bloch, J., Nethery, D., and Smith, T.
  (2019).
\newblock Defining flight envelope requirements and handling qualities criteria
  for first-person-view quadrotor racing.
\newblock In {\em AIAA Scitech 2019 Forum}, page 0825.

\bibitem[Andrychowicz et~al., 2018]{andrychowicz2018learning}
Andrychowicz, M., Baker, B., Chociej, M., Jozefowicz, R., McGrew, B., Pachocki,
  J., Petron, A., Plappert, M., Powell, G., Ray, A., et~al. (2018).
\newblock Learning dexterous in-hand manipulation.
\newblock {\em arXiv preprint arXiv:1808.00177}.

\bibitem[{\AA}str{\"o}m and Wittenmark, 2013]{aastrom2013adaptive}
{\AA}str{\"o}m, K.~J. and Wittenmark, B. (2013).
\newblock {\em Adaptive control}.
\newblock Courier Corporation.

\bibitem[Bagnell and Schneider, 2001]{bagnell2001autonomous}
Bagnell, J.~A. and Schneider, J.~G. (2001).
\newblock Autonomous helicopter control using reinforcement learning policy
  search methods.
\newblock In {\em Robotics and Automation, 2001. Proceedings 2001 ICRA. IEEE
  International Conference on}, volume~2, pages 1615--1620. IEEE.

\bibitem[Black et~al., 2014]{black2014adaptive}
Black, W.~S., Haghi, P., and Ariyur, K.~B. (2014).
\newblock Adaptive systems: History, techniques, problems, and perspectives.
\newblock {\em Systems}, 2(4):606--660.

\bibitem[Blitzer et~al., 2008]{blitzer2008learning}
Blitzer, J., Crammer, K., Kulesza, A., Pereira, F., and Wortman, J. (2008).
\newblock Learning bounds for domain adaptation.
\newblock In {\em Advances in neural information processing systems}, pages
  129--136.

\bibitem[Bobtsov et~al., 2016]{bobtsov2016hybrid}
Bobtsov, A., Guirik, A., Budko, M., and Budko, M. (2016).
\newblock Hybrid parallel neuro-controller for multirotor unmanned aerial
  vehicle.
\newblock In {\em Ultra Modern Telecommunications and Control Systems and
  Workshops (ICUMT), 2016 8th International Congress on}, pages 1--4. IEEE.

\bibitem[Bouabdallah et~al., 2004]{bouabdallah2004design}
Bouabdallah, S., Murrieri, P., and Siegwart, R. (2004).
\newblock Design and control of an indoor micro quadrotor.
\newblock In {\em Robotics and Automation, 2004. Proceedings. ICRA'04. 2004
  IEEE International Conference on}, volume~5, pages 4393--4398. IEEE.

\bibitem[Brandt et~al., 2015]{uiuc}
Brandt, J.~B., Deters, R.~W., Ananda, G.~K., and Selig, M.~S. (2015).
\newblock \url{https://m-selig.ae.illinois.edu/props/propDB.html}.
\newblock Accessed: 2019-07-20.

\bibitem[Brockman et~al., 2016]{brockman2016openai}
Brockman, G., Cheung, V., Pettersson, L., Schneider, J., Schulman, J., Tang,
  J., and Zaremba, W. (2016).
\newblock {OpenAI Gym}.
\newblock {\em arXiv preprint arXiv:1606.01540}.

\bibitem[Brooks, 1992]{brooks1992artificial}
Brooks, R.~A. (1992).
\newblock Artificial life and real robots.
\newblock In {\em Proceedings of the First European Conference on artificial
  life}, pages 3--10.

\bibitem[Cheng et~al., 2018]{cheng2018end}
Cheng, Z., West, R., and Einstein, C. (2018).
\newblock End-to-end analysis and design of a drone flight controller.
\newblock {\em IEEE Transactions on Computer-Aided Design of Integrated
  Circuits and Systems}, 37(11):2404--2415.

\bibitem[Chou et~al., 2017]{chou2017improving}
Chou, P.-W., Maturana, D., and Scherer, S. (2017).
\newblock Improving stochastic policy gradients in continuous control with deep
  reinforcement learning using the beta distribution.
\newblock In {\em Proceedings of the 34th International Conference on Machine
  Learning-Volume 70}, pages 834--843. JMLR. org.

\bibitem[Cignoni et~al., 2008]{cignoni2008meshlab}
Cignoni, P., Callieri, M., Corsini, M., Dellepiane, M., Ganovelli, F., and
  Ranzuglia, G. (2008).
\newblock Meshlab: an open-source mesh processing tool.
\newblock In {\em Eurographics Italian chapter conference}, volume 2008, pages
  129--136.

\bibitem[Coumans, 2014]{coumans2014exploring}
Coumans, E. (2014).
\newblock Exploring mlcp solvers and featherstone.
\newblock In {\em Game Developers Conf}, pages 17--21.

\bibitem[Coumans, 2015]{bullet}
Coumans, E. (2015).
\newblock Bullet physics simulation.
\newblock In {\em ACM SIGGRAPH 2015 Courses}, SIGGRAPH '15, New York, NY, USA.
  ACM.

\bibitem[Cybenko, 1989]{cybenko1989approximation}
Cybenko, G. (1989).
\newblock Approximation by superpositions of a sigmoidal function.
\newblock {\em Mathematics of control, signals and systems}, 2(4):303--314.

\bibitem[Deters et~al., 2014]{deters2014reynolds}
Deters, R.~W., Ananda~Krishnan, G.~K., and Selig, M.~S. (2014).
\newblock Reynolds number effects on the performance of small-scale propellers.
\newblock In {\em 32nd AIAA applied aerodynamics conference}, page 2151.

\bibitem[Deters et~al., 2017]{deters2017static}
Deters, R.~W., Kleinke, S., and Selig, M.~S. (2017).
\newblock Static testing of propulsion elements for small multirotor unmanned
  aerial vehicles.
\newblock In {\em 35th AIAA Applied Aerodynamics Conference}, page 3743.

\bibitem[Dewey, 2014]{dewey2014reinforcement}
Dewey, D. (2014).
\newblock Reinforcement learning and the reward engineering principle.
\newblock In {\em 2014 AAAI Spring Symposium Series}.

\bibitem[Dhariwal et~al., 2017]{baselines}
Dhariwal, P., Hesse, C., Klimov, O., Nichol, A., Plappert, M., Radford, A.,
  Schulman, J., Sidor, S., and Wu, Y. (2017).
\newblock Openai baselines.
\newblock \url{https://github.com/openai/baselines}.

\bibitem[Dierks and Jagannathan, 2010]{dierks2010output}
Dierks, T. and Jagannathan, S. (2010).
\newblock Output feedback control of a quadrotor uav using neural networks.
\newblock {\em IEEE transactions on neural networks}, 21(1):50--66.

\bibitem[dos Santos et~al., 2012]{dos2012experimental}
dos Santos, S. R.~B., Givigi, S.~N., and J{\'u}nior, C. L.~N. (2012).
\newblock An experimental validation of reinforcement learning applied to the
  position control of uavs.
\newblock In {\em 2012 IEEE International Conference on Systems, Man, and
  Cybernetics (SMC)}, pages 2796--2802. IEEE.

\bibitem[Duan et~al., 2016]{duan2016benchmarking}
Duan, Y., Chen, X., Houthooft, R., Schulman, J., and Abbeel, P. (2016).
\newblock Benchmarking deep reinforcement learning for continuous control.
\newblock In {\em International Conference on Machine Learning}, pages
  1329--1338.

\bibitem[Ebeid et~al., 2018]{ebeid2018survey}
Ebeid, E., Skriver, M., Terkildsen, K.~H., Jensen, K., and Schultz, U.~P.
  (2018).
\newblock A survey of open-source uav flight controllers and flight simulators.
\newblock {\em Microprocessors and Microsystems}, 61:11--20.

\bibitem[Falkner et~al., 2018]{falkner2018bohb}
Falkner, S., Klein, A., and Hutter, F. (2018).
\newblock Bohb: Robust and efficient hyperparameter optimization at scale.
\newblock {\em arXiv preprint arXiv:1807.01774}.

\bibitem[Farha, 2016]{f16}
Farha, F. (2016).
\newblock {Hovakimyan's adaptive control to be tested on fighter jet}.
\newblock
  \url{https://mechanical.illinois.edu/news/hovakimyans-adaptive-control-be-tested-fighter-jet}.
\newblock Accessed: 2019-07-18.

\bibitem[Fatan et~al., 2013]{fatan2013adaptive}
Fatan, M., Sefidgari, B.~L., and Barenji, A.~V. (2013).
\newblock An adaptive neuro pid for controlling the altitude of quadcopter
  robot.
\newblock In {\em Methods and models in automation and robotics (mmar), 2013
  18th international conference on}, pages 662--665. IEEE.

\bibitem[Furrer et~al., 2016]{Furrer2016}
Furrer, F., Burri, M., Achtelik, M., and Siegwart, R. (2016).
\newblock {\em Robot Operating System (ROS): The Complete Reference (Volume
  1)}, chapter RotorS---A Modular Gazebo MAV Simulator Framework, pages
  595--625.
\newblock Springer International Publishing, Cham.

\bibitem[Gabor et~al., 2016]{gabor2016simulation}
Gabor, T., Belzner, L., Kiermeier, M., Beck, M.~T., and Neitz, A. (2016).
\newblock A simulation-based architecture for smart cyber-physical systems.
\newblock In {\em Autonomic Computing (ICAC), 2016 IEEE International
  Conference on}, pages 374--379. IEEE.

\bibitem[Glaessgen and Stargel, 2012]{glaessgen2012digital}
Glaessgen, E.~H. and Stargel, D. (2012).
\newblock The digital twin paradigm for future nasa and us air force vehicles.
\newblock In {\em 53rd Struct. Dyn. Mater. Conf. Special Session: Digital Twin,
  Honolulu, HI, US}, pages 1--14.

\bibitem[Gong et~al., 2018]{gong2018performance}
Gong, A., MacNeill, R., and Verstraete, D. (2018).
\newblock Performance testing and modeling of a brushless dc motor, electronic
  speed controller and propeller for a small uav application.
\newblock In {\em 2018 Joint Propulsion Conference}, page 4584.

\bibitem[Grieves, 2014]{grieves2014digital}
Grieves, M. (2014).
\newblock Digital twin: Manufacturing--excellence through virtual factory
  replication (white paper).
\newblock {\em Michael Grieves (University of Michigan) LLC (http://innovate.
  fit. edu/plm/documents/doc\_mgr/912/1411.0
  \_Digital\_Twin\_White\_Paper\_Dr\_Grieves. pdf)}.

\bibitem[Hagan and Demuth, 1999]{hagan1999neural}
Hagan, M.~T. and Demuth, H.~B. (1999).
\newblock Neural networks for control.
\newblock In {\em American Control Conference, 1999. Proceedings of the 1999},
  volume~3, pages 1642--1656. IEEE.

\bibitem[Hattem, 2019]{numpystl}
Hattem, R. (2019).
\newblock {numpy-stl}.
\newblock \url{https://numpy-stl.readthedocs.io/en/latest/}.
\newblock Accessed June 17, 2019.

\bibitem[Henderson et~al., 2018]{henderson2018deep}
Henderson, P., Islam, R., Bachman, P., Pineau, J., Precup, D., and Meger, D.
  (2018).
\newblock Deep reinforcement learning that matters.
\newblock In {\em Thirty-Second AAAI Conference on Artificial Intelligence}.

\bibitem[Hill et~al., 2018]{stable-baselines}
Hill, A., Raffin, A., Traore, R., Dhariwal, P., Hesse, C., Klimov, O., Nichol,
  A., Plappert, M., Radford, A., Schulman, J., Sidor, S., and Wu, Y. (2018).
\newblock Stable baselines.
\newblock \url{https://github.com/hill-a/stable-baselines}.

\bibitem[Hovakimyan et~al., 2011]{hovakimyan2011mathcal}
Hovakimyan, N., Cao, C., Kharisov, E., Xargay, E., and Gregory, I.~M. (2011).
\newblock L1 adaptive control for safety-critical systems.
\newblock {\em IEEE Control Systems}, 31(5):54--104.

\bibitem[Hunt et~al., 1992]{hunt1992neural}
Hunt, K.~J., Sbarbaro, D., {\.Z}bikowski, R., and Gawthrop, P.~J. (1992).
\newblock Neural networks for control systems—a survey.
\newblock {\em Automatica}, 28(6):1083--1112.

\bibitem[Husbands and Harvey, 1992]{husbands1992evolution}
Husbands, P. and Harvey, I. (1992).
\newblock Evolution versus design: Controlling autonomous robots.
\newblock In {\em AI, Simulation and Planning in High Autonomy Systems, 1992.
  Integrating Perception, Planning and Action., Proceedings of the Third Annual
  Conference of}, pages 139--146. IEEE.

\bibitem[Hwangbo et~al., 2017]{hwangbo2017control}
Hwangbo, J., Sa, I., Siegwart, R., and Hutter, M. (2017).
\newblock Control of a quadrotor with reinforcement learning.
\newblock {\em IEEE Robotics and Automation Letters}, 2(4):2096--2103.

\bibitem[Jakobi et~al., 1995]{jakobi1995noise}
Jakobi, N., Husbands, P., and Harvey, I. (1995).
\newblock Noise and the reality gap: The use of simulation in evolutionary
  robotics.
\newblock {\em Advances in artificial life}, pages 704--720.

\bibitem[Karpathy, 2018]{normalize}
Karpathy, A. (2018).
\newblock {Deep Reinforcement Learning: Pong from Pixels}.
\newblock Accessed: 2018-03-29.

\bibitem[Kim et~al., 1993]{kim1993nonlinear}
Kim, B.~S., Calise, A., and Kam, M. (1993).
\newblock Nonlinear flight control using neural networks and feedback
  linearization.
\newblock In {\em Proceedings. The First IEEE Regional Conference on Aerospace
  Control Systems,}, pages 176--181. IEEE.

\bibitem[Kim et~al., 2004]{kim2004autonomous}
Kim, H.~J., Jordan, M.~I., Sastry, S., and Ng, A.~Y. (2004).
\newblock Autonomous helicopter flight via reinforcement learning.
\newblock In {\em Advances in neural information processing systems}, pages
  799--806.

\bibitem[Koch, 2018a]{gymfccode}
Koch, W. (2018a).
\newblock {GymFC}.
\newblock \url{https://github.com/wil3/gymfc}.

\bibitem[Koch, 2018b]{neuroflight}
Koch, W. (2018b).
\newblock {Neuroflight Github Repository}.
\newblock \url{https://github.com/wil3/neuroflight}.

\bibitem[Koch, 2018c]{neuroflightproject}
Koch, W. (2018c).
\newblock {Neuroflight Project Page}.
\newblock \url{https://wfk.io/neuroflight}.

\bibitem[Koch, 2019a]{gymfcplugins}
Koch, W. (2019a).
\newblock {GymFC Aircraft Plugins}.
\newblock Accessed June 17, 2019.

\bibitem[Koch, 2019b]{py3gazebo}
Koch, W. (2019b).
\newblock {Py3Gazebo: Python3 bindings for the Gazebo multi-robot simulator.}
\newblock \url{https://github.com/wil3/py3gazebo}.
\newblock Accessed June 17, 2019.

\bibitem[Koch et~al., 2019a]{koch2019neuroflight}
Koch, W., Mancuso, R., and Bestavros, A. (2019a).
\newblock Neuroflight: Next generation flight control firmware.
\newblock {\em arXiv preprint arXiv:1901.06553}.

\bibitem[Koch et~al., 2019b]{gymfc}
Koch, W., Mancuso, R., West, R., and Bestavros, A. (2019b).
\newblock Reinforcement learning for uav attitude control.
\newblock {\em ACM Transactions on Cyber-Physical Systems}, 3(2):22.

\bibitem[Koenig and Howard, 2004]{koenig2004design}
Koenig, N. and Howard, A. (2004).
\newblock Design and use paradigms for gazebo, an open-source multi-robot
  simulator.
\newblock In {\em Intelligent Robots and Systems, 2004.(IROS 2004).
  Proceedings. 2004 IEEE/RSJ International Conference on}, volume~3, pages
  2149--2154. IEEE.

\bibitem[Krishnan et~al., 2019]{krishnan2019air}
Krishnan, S., Borojerdian, B., Fu, W., Faust, A., and Reddi, V.~J. (2019).
\newblock Air learning: An ai research platform for algorithm-hardware
  benchmarking of autonomous aerial robots.
\newblock {\em arXiv preprint arXiv:1906.00421}.

\bibitem[Lee et~al., 2018]{lee2018dart}
Lee, J., Grey, M.~X., Ha, S., Kunz, T., Jain, S., Ye, Y., Srinivasa, S.~S.,
  Stilman, M., and Liu, C.~K. (2018).
\newblock Dart: Dynamic animation and robotics toolkit.
\newblock {\em The Journal of Open Source Software}, 3(22):500.

\bibitem[Leith and Leithead, 2000]{leith2000survey}
Leith, D.~J. and Leithead, W.~E. (2000).
\newblock Survey of gain-scheduling analysis and design.
\newblock {\em International journal of control}, 73(11):1001--1025.

\bibitem[Li and Song, 2012]{li2012survey}
Li, Y. and Song, S. (2012).
\newblock A survey of control algorithms for quadrotor unmanned helicopter.
\newblock In {\em Advanced Computational Intelligence (ICACI), 2012 IEEE Fifth
  International Conference on}, pages 365--369. IEEE.

\bibitem[Liang, 2018]{looptime}
Liang, O. (2018).
\newblock {Looptime and Flight Controller}.
\newblock \url{https://oscarliang.com/best-looptime-flight-controller/}.
\newblock Accessed: 2018-11-25.

\bibitem[Lillicrap et~al., 2015]{lillicrap2015continuous}
Lillicrap, T.~P., Hunt, J.~J., Pritzel, A., Heess, N., Erez, T., Tassa, Y.,
  Silver, D., and Wierstra, D. (2015).
\newblock Continuous control with deep reinforcement learning.
\newblock {\em arXiv preprint arXiv:1509.02971}.

\bibitem[Luukkonen, 2011]{luukkonen2011modelling}
Luukkonen, T. (2011).
\newblock Modelling and control of quadcopter.
\newblock {\em Independent research project in applied mathematics, Espoo}.

\bibitem[Maleki et~al., 2016]{maleki2016reliable}
Maleki, K.~N., Ashenayi, K., Hook, L.~R., Fuller, J.~G., and Hutchins, N.
  (2016).
\newblock A reliable system design for nondeterministic adaptive controllers in
  small uav autopilots.
\newblock In {\em Digital Avionics Systems Conference (DASC), 2016 IEEE/AIAA
  35th}, pages 1--5. IEEE.

\bibitem[Maxwell, 1868]{maxwell1868governors}
Maxwell, J.~C. (1868).
\newblock I. on governors.
\newblock {\em Proceedings of the Royal Society of London}, (16):270--283.

\bibitem[McCormick, 1995]{mccormick1995aerodynamics}
McCormick, B. (1995).
\newblock Aerodynamics aeronautics and flight mechanics john wiley \& sons inc.

\bibitem[Meier et~al., 2015]{meier2015px4}
Meier, L., Honegger, D., and Pollefeys, M. (2015).
\newblock Px4: A node-based multithreaded open source robotics framework for
  deeply embedded platforms.
\newblock In {\em 2015 IEEE international conference on robotics and automation
  (ICRA)}, pages 6235--6240. IEEE.

\bibitem[Miglino et~al., 1995]{miglino1995evolving}
Miglino, O., Lund, H.~H., and Nolfi, S. (1995).
\newblock Evolving mobile robots in simulated and real environments.
\newblock {\em Artificial life}, 2(4):417--434.

\bibitem[Minh and Ha, 2010]{minh2010modeling}
Minh, L.~D. and Ha, C. (2010).
\newblock Modeling and control of quadrotor mav using vision-based measurement.
\newblock In {\em Strategic Technology (IFOST), 2010 International Forum on},
  pages 70--75. IEEE.

\bibitem[Mnih et~al., 2013]{mnih2013playing}
Mnih, V., Kavukcuoglu, K., Silver, D., Graves, A., Antonoglou, I., Wierstra,
  D., and Riedmiller, M. (2013).
\newblock Playing atari with deep reinforcement learning.
\newblock {\em arXiv preprint arXiv:1312.5602}.

\bibitem[Molchanov et~al., 2019]{molchanov2019sim}
Molchanov, A., Chen, T., H{\"o}nig, W., Preiss, J.~A., Ayanian, N., and
  Sukhatme, G.~S. (2019).
\newblock Sim-to-(multi)-real: Transfer of low-level robust control policies to
  multiple quadrotors.
\newblock {\em arXiv preprint arXiv:1903.04628}.

\bibitem[Nicol et~al., 2008]{nicol2008robust}
Nicol, C., Macnab, C., and Ramirez-Serrano, A. (2008).
\newblock Robust neural network control of a quadrotor helicopter.
\newblock In {\em Electrical and Computer Engineering, 2008. CCECE 2008.
  Canadian Conference on}, pages 001233--001238. IEEE.

\bibitem[Palossi et~al., 2019]{palossi201964mw}
Palossi, D., Loquercio, A., Conti, F., Flamand, E., Scaramuzza, D., and Benini,
  L. (2019).
\newblock A 64mw dnn-based visual navigation engine for autonomous nano-drones.
\newblock {\em IEEE Internet of Things Journal}.

\bibitem[Palunko and Fierro, 2011]{palunko2011adaptive}
Palunko, I. and Fierro, R. (2011).
\newblock Adaptive control of a quadrotor with dynamic changes in the center of
  gravity.
\newblock {\em IFAC Proceedings Volumes}, 44(1):2626--2631.

\bibitem[Persopolo, 2019]{gemfan}
Persopolo (2019).
\newblock {Gemfan 5" propeller 5152R}.
\newblock \url{https://grabcad.com/library/gemfan-5-propeller-5152r-1}.
\newblock Accessed June 17, 2019.

\bibitem[Plappert, 2016]{plappert2016kerasrl}
Plappert, M. (2016).
\newblock keras-rl.
\newblock \url{https://github.com/keras-rl/keras-rl}.

\bibitem[Ringegni et~al., 2001]{ringegni2001experimental}
Ringegni, P., Actis, M., and Patanella, A. (2001).
\newblock An experimental technique for determining mass inertial properties of
  irregular shape bodies and mechanical assemblies.
\newblock {\em Measurement}, 29(1):63--75.

\bibitem[Santoso et~al., 2017]{santoso2017state}
Santoso, F., Garratt, M.~A., and Anavatti, S.~G. (2017).
\newblock State-of-the-art intelligent flight control systems in unmanned
  aerial vehicles.
\newblock {\em IEEE Transactions on Automation Science and Engineering}.

\bibitem[Schaarschmidt et~al., 2017]{schaarschmidt2017tensorforce}
Schaarschmidt, M., Kuhnle, A., and Fricke, K. (2017).
\newblock Tensorforce: A tensorflow library for applied reinforcement learning.
\newblock Web page.
\newblock Accessed: 2018-12-07.

\bibitem[Schulman et~al., 2015]{schulman2015trust}
Schulman, J., Levine, S., Abbeel, P., Jordan, M., and Moritz, P. (2015).
\newblock Trust region policy optimization.
\newblock In {\em International Conference on Machine Learning}, pages
  1889--1897.

\bibitem[Schulman et~al., 2017]{schulman2017proximal}
Schulman, J., Wolski, F., Dhariwal, P., Radford, A., and Klimov, O. (2017).
\newblock Proximal policy optimization algorithms.
\newblock {\em arXiv preprint arXiv:1707.06347}.

\bibitem[{Sergio Guadarrama, Anoop Korattikara, Oscar Ramirez, Pablo Castro,
  Ethan Holly, Sam Fishman, Ke Wang, Ekaterina Gonina, Neal Wu, Chris Harris,
  Vincent Vanhoucke, Eugene Brevdo}, 2018]{TFAgents}
{Sergio Guadarrama, Anoop Korattikara, Oscar Ramirez, Pablo Castro, Ethan
  Holly, Sam Fishman, Ke Wang, Ekaterina Gonina, Neal Wu, Chris Harris, Vincent
  Vanhoucke, Eugene Brevdo} (2018).
\newblock {TF-Agents}: A library for reinforcement learning in tensorflow.
\newblock \url{https://github.com/tensorflow/agents}.
\newblock Accessed: 2019-07-26.

\bibitem[Shah et~al., 2018]{shah2018airsim}
Shah, S., Dey, D., Lovett, C., and Kapoor, A. (2018).
\newblock Airsim: High-fidelity visual and physical simulation for autonomous
  vehicles.
\newblock In {\em Field and service robotics}, pages 621--635. Springer.

\bibitem[Shepherd~III and Tumer, 2010]{shepherd2010robust}
Shepherd~III, J.~F. and Tumer, K. (2010).
\newblock Robust neuro-control for a micro quadrotor.
\newblock In {\em Proceedings of the 12th annual conference on Genetic and
  evolutionary computation}, pages 1131--1138. ACM.

\bibitem[Sherman et~al., 2011]{sherman2011simbody}
Sherman, M.~A., Seth, A., and Delp, S.~L. (2011).
\newblock Simbody: multibody dynamics for biomedical research.
\newblock {\em Procedia Iutam}, 2:241--261.

\bibitem[Smith et~al., 2010]{smith2010design}
Smith, T., Barhorst, J., and Urnes, J.~M. (2010).
\newblock Design and flight test of an intelligent flight control system.
\newblock In {\em Applications of Neural Networks in High Assurance Systems},
  pages 57--76. Springer.

\bibitem[{Smith, Russel}, 2006]{ode}
{Smith, Russel} ({2006}).
\newblock {Open Dynamics Engine}.

\bibitem[Sutton and Barto, 1998]{sutton1998reinforcement}
Sutton, R.~S. and Barto, A.~G. (1998).
\newblock {\em Reinforcement learning: An introduction}, volume~1.
\newblock MIT press Cambridge.

\bibitem[Tobin et~al., 2017]{tobin2017domain}
Tobin, J., Fong, R., Ray, A., Schneider, J., Zaremba, W., and Abbeel, P.
  (2017).
\newblock Domain randomization for transferring deep neural networks from
  simulation to the real world.
\newblock In {\em 2017 IEEE/RSJ International Conference on Intelligent Robots
  and Systems (IROS)}, pages 23--30. IEEE.

\bibitem[Todorov et~al., 2012]{todorov2012mujoco}
Todorov, E., Erez, T., and Tassa, Y. (2012).
\newblock Mujoco: A physics engine for model-based control.
\newblock In {\em 2012 IEEE/RSJ International Conference on Intelligent Robots
  and Systems}, pages 5026--5033. IEEE.

\bibitem[Tuegel et~al., 2011]{tuegel2011reengineering}
Tuegel, E.~J., Ingraffea, A.~R., Eason, T.~G., and Spottswood, S.~M. (2011).
\newblock Reengineering aircraft structural life prediction using a digital
  twin.
\newblock {\em International Journal of Aerospace Engineering}, 2011.

\bibitem[Wang and Zhang, 2001]{wang2001fundamental}
Wang, L.~Y. and Zhang, J.-F. (2001).
\newblock Fundamental limitations and differences of robust and adaptive
  control.
\newblock In {\em American Control Conference, 2001. Proceedings of the 2001},
  volume~6, pages 4802--4807. IEEE.

\bibitem[Waslander et~al., 2005]{waslander2005multi}
Waslander, S.~L., Hoffmann, G.~M., Jang, J.~S., and Tomlin, C.~J. (2005).
\newblock Multi-agent quadrotor testbed control design: Integral sliding mode
  vs. reinforcement learning.
\newblock In {\em Intelligent Robots and Systems, 2005.(IROS 2005). 2005
  IEEE/RSJ International Conference on}, pages 3712--3717. IEEE.

\bibitem[Whitaker et~al., 1958]{whitaker1958design}
Whitaker, H.~P., Yamron, J., and Kezer, A. (1958).
\newblock {\em Design of model-reference adaptive control systems for
  aircraft}.
\newblock Massachusetts Institute of Technology, Instrumentation Laboratory.

\bibitem[Williams-Hayes, 2005]{williams2005flight}
Williams-Hayes, P.~S. (2005).
\newblock Flight test implementation of a second generation intelligent flight
  control system.
\newblock {\em infotech@ Aerospace, AIAA-2005-6995}, pages 26--29.

\bibitem[Zames, 1966]{zames1966input}
Zames, G. (1966).
\newblock On the input-output stability of time-varying nonlinear feedback
  systems part one: Conditions derived using concepts of loop gain, conicity,
  and positivity.
\newblock {\em IEEE transactions on automatic control}, 11(2):228--238.

\bibitem[Zamora et~al., 2016]{zamora2016extending}
Zamora, I., Lopez, N.~G., Vilches, V.~M., and Cordero, A.~H. (2016).
\newblock Extending the openai gym for robotics: a toolkit for reinforcement
  learning using ros and gazebo.
\newblock {\em arXiv preprint arXiv:1608.05742}.

\bibitem[Ziegler and Nichols, 1942]{ziegler1942optimum}
Ziegler, J.~G. and Nichols, N.~B. (1942).
\newblock Optimum settings for automatic controllers.
\newblock {\em trans. ASME}, 64(11).

\bibitem[Zulu and John, 2014]{zulu2014review}
Zulu, A. and John, S. (2014).
\newblock A review of control algorithms for autonomous quadrotors.
\newblock {\em Open Journal of Applied Sciences}, 4(14):547.

\end{thebibliography}
\endgroup

\cleardoublepage

\addcontentsline{toc}{chapter}{Curriculum Vitae}

\begin{center}
{\LARGE {\bf CURRICULUM VITAE}}\\
\vspace{0.5in}
{\large {\bf William Frederick Koch III}}\\
\vspace{0.25in}
wfkoch@bu.edu \hspace{1cm} https://wfk.io\\

\medskip
Boston University\\
Department of Computer Science\\
111 Cummington Mall, Boston MA 02215

\end{center}
{
\setlist[itemize]{ topsep=0pt}
\section*{Education}

\begin{itemize}
  \item
      \textbf{Boston University}{\hfill Boston, MA\\} % Location
    {PhD in Computer Science}{\hfill Sept. 2014 - Sept. 2019}\\ % Date(s)
     {Thesis title: Flight Controller Synthesis via Deep Reinforcement
         Learning}\\
     {GPA: 3.7/4.0}

  \item
      \textbf{Stevens Institute of Technology}{\hfill Hoboken, NJ\\} % Location
    {M.S. in Computer Engineering}{\hfill Jan. 2012 - Dec. 2013}\\ % Date(s)
    {Thesis title: A framework for assisting learners by incorporating
        knowledge to aid in predicting nerve guidance conduit performance}\\
    {GPA: 3.8/4.0}

  \item
      \textbf{University of Rhode Island} {\hfill Kingston, RI\\} % Location
    {B.S. in Computer Engineering, Minor in Mathematics}{\hfill Sept. 2003 - May 2008}\\ % Date(s)
    {GPA: 3.2/4.0}

\end{itemize}

\section*{Research Experience}

\begin{itemize}

  \item
      \textbf{Boston University} {\hfill Boston, MA}\\ % Location
    {Research Assistant}{\hfill Jan. 2017 - Present} % Date(s)
    {
      \begin{itemize} % Description(s) of tasks/responsibilities
        \item {Developing next generation flight control systems through the use of
        machine learning including the worlds first open-source neural network powered flight control
        firmware, Neuroflight.}
              \item {Conducted research in wide area of cyber security including static
        and dynamic malware analysis, vulnerability analysis, cyber defense and attacks
        and mobile security.}
      \end{itemize}
    }

  \item
      \textbf{MIT Lincoln Laboratory}{\hfill Lexington, MA}\\ % Location
    {Cyber Security Research Intern }{\hfill Jan. 2016 - June 2016} % Date(s)
    {
      \begin{itemize} % Description(s) of tasks/responsibilities
      \item {Developed novel SDN attack called Persona Hijacking which has been published in USENIX Security Symposium.}
      \end{itemize}
    }

  \item
      \textbf{Stevens Institute of Technology}{\hfill Hoboken, NJ}\\ % Location
    {Research Assistant}{\hfill Jan. 2012 - Dec. 2013} % Date(s)
    {
      \begin{itemize} % Description(s) of tasks/responsibilities
      \item {Worked on multi-discipline team consisting of biomedical and computer
engineers to advance nerve guidance conduit performance.}
       \item {Developed novel machine learning algorithms to predict nerve guidance conduit performance. }
      \end{itemize}
    }

\end{itemize}

\section*{Teaching Experience}

\begin{itemize}
  \item
  \textbf{Boston University} {\hfill Boston, MA}\\ % Location
    {Teaching Fellow}{\hfill Fall 2017, Spring 2019} % Date(s)
    {
      \begin{itemize} % Description(s) of tasks/responsibilities
\item {Designed lesson plans, taught discussion sections, developed written and programming assignments for class
Fundamentals of Computing Systems.}
      \end{itemize}
    }

  \item
      \textbf{Internal Drive Tech Camps} {\hfill Princeton, NJ}\\ % Location
    {Programming Instructor}{\hfill June 2016 - Aug. 2018} % Date(s)
    {
      \begin{itemize} % Description(s) of tasks/responsibilities
      \item {Created lesson plans for wide range of skill levels including object oriented fundamentals, polymorphism, exception handling and third-party library integration.}
\item {Emphasized lessons on coding style and best practices not taught and enforced in academia.}
\item{Advised students through final projects ranging from web crawlers to video games.}
      \end{itemize}
    }

  \item
      \textbf{Stevens Institute of Technology}{\hfill Hoboken, NJ}\\ % Location
    {Teachers Assistant}{\hfill Jan. 2012 - Dec. 2013} % Date(s)
    {
      \begin{itemize} % Description(s) of tasks/responsibilities
\item {Grader for graduate class CPE-555 Real-Time and Embedded Systems and
undergraduate class EE-250 Mathematics for Electrical Engineers.}
      \end{itemize}
    }

\end{itemize}

\section*{Additional Experience}

\begin{itemize}

  \item
      \textbf{Boston Drone Racing}{\hfill Boston, MA}\\ % Location
    {Founder} % Job title
    {\hfill Jan. 2017 - Present} % Date(s)
    {
      \begin{itemize} % Description(s) of tasks/responsibilities
      \item {Created website and designed logo. Established communication channels. Manage social
              media networks. }
      \item {Secured funding for racing track and supplies. }
      \item {Organize weekly races and monthly hack nights.}
      \end{itemize}
    }

  \item
    \textbf{Capsules, LLC} % Organization
    {\hfill Madison, CT}\\ % Location
    {Co-Founder/CEO} % Job title
    {\hfill June 2013 - Aug. 2014} % Date(s)
    {
      \begin{itemize} % Description(s) of tasks/responsibilities
      \item {Managed team to create a geo-location based augmented reality
              mobile app.}
      \item {Lead mobile developer responsible for overall architecture, design and implementation.}
      \end{itemize}
    }

  \item
      \textbf{Sikorsky Aircraft {\scriptsize\textnormal{(subcontracted through AIS Consulting
                and Sila SG)}}} % Organization
    {\hfill Shelton, CT}\\ % Location
    {Software Engineer} % Job title
    {\hfill Jun. 2006 - Jan. 2012} % Date(s)
    {
      \begin{itemize} % Description(s) of tasks/responsibilities
      \item {Lead software engineer on seven software applications for the Sikorsky CH-53K Aircraft's Integrated Support System (ISS). }
        \item {Designed and implemented continuous integration environment.}
        \item {Responsible for integration between third-party vendors.}
      \end{itemize}
    }

  \item
    \textbf{CT Hackerspace} % Organization
    {\hfill Watertown, CT}\\ % Location
    {Co-founder/Chairman} % Job title
    {\hfill Aug. 2010 - Aug. 2011} % Date(s)
    {
      \begin{itemize} % Description(s) of tasks/responsibilities
      \item {Established organization through the development of bylaws,
              identity, physical and web presence.}
      \item {Ran monthly board meetings to facilitate in the growth and direction of the hackerspace.}
      \end{itemize}
    }

\end{itemize}

\section*{Select Publications}

\begin{itemize}

  \item
    {Neuroflight: Next Generation Flight Control Firmware} % Position
    {\textbf{William Koch}, Renato Mancuso, and Azer Bestavros, \textit{In submission}} % Committee
    {} % Location
    {2019} % Date(s)
  \item
    {Reinforcement Learning for UAV Attitude Control} % Position
    {\textbf{William Koch}, Renato Mancuso, Richard West, and Azer Bestavros, \textit{ACM Transactions on Cyber-Physical Systems}} % Committee
    {} % Location
    {2019} % Date(s)
  \item
    {S3B: Software-Defined Secure Server Bindings} % Position
    {\textbf{William Koch}, and Azer Bestavros, \textit{IEEE International Conference on Distributed Computing Systems (ICDCS)}} % Committee
    {} % Location
    {2018} % Date(s)
  \item
    {Semi-automated discovery of server-based information oversharing vulnerabilities in Android applications} % Position
    {\textbf{William Koch}, Abdelberi Chaabane, Manuel Egele, William Robertson, and Engin Kirda, \textit{ACM SIGSOFT International Symposium on Software Testing and Analysis}} % Committee
    {} % Location
    {2017} % Date(s)
  \item
    {PayBreak: defense against cryptographic ransomware.} % Position
    {Eugene Kolodenker, \textbf{William Koch}, Gianluca Stringhini, and Manuel Egele, \textit{ACM on Asia Conference on Computer and Communications Security}} % Committee
    {} % Location
    {2017} % Date(s)
  \item
    {Identifier Binding Attacks and Defenses in Software-Defined Networks} % Position
    {Samuel Jero, \textbf{William Koch}, Richard Skowyra, Hamed Okhravi, Cristina Nita-Rotaru, and David Bigelow, \textit{USENIX Security Symposium}} % Committee
    {} % Location
    {2017} % Date(s)

  \item
    {Markov modeling of moving target defense games} % Position
    {Hoda Maleki,  Saeed Valizadeh, \textbf{William Koch}, Azer Bestavros, and Marten
        van Dijk, \textit{In Proceedings of the 2016 ACM Workshop on Moving Target Defense}} % Committee
    {} % Location
    {2016} % Date(s)

  \item
    {Provide: Hiding from automated network scans with proofs of identity} % Position
    {\textbf{William Koch}, and Azer Bestavros, \textit{IEEE Workshop on Hot Topics in Web Systems and Technologies (HotWeb)}} % Committee
    {} % Location
    {2016} % Date(s)

\end{itemize}

\section*{Projects}

\begin{itemize}
    \item
        \textbf{Neuroflight} % Organization
    {} % Location
    {} % Date(s)
    {
Neuroflight is the first open-source neuro-flight controller software (firmware) for remotely piloting multi-rotors and fixed wing aircraft. Neuroflight's primary focus is to provide optimal flight performance.
    }\\
    {\href{https://github.com/wil3/neuroflight}{https://github.com/wil3/neuroflight}} % Job title

\item
    \textbf{GymFC} % Organization
    {
GymFC is an OpenAI Gym environment designed for synthesizing intelligent flight control systems using reinforcement learning. This environment is meant to serve as a tool for researchers to benchmark controllers to progress the state-of-the art of intelligent flight control.
    }\\
    {\href{https://github.com/wil3/gymfc}{https://github.com/wil3/gymfc}} % Job title

\end{itemize}

}

\end{document}